\documentclass[lettersize,journal]{IEEEtran}
\usepackage{amsmath,amsfonts}
\usepackage{algorithmic}
\usepackage{algorithm}
\usepackage{array}
\usepackage[caption=false,font=normalsize,labelfont=sf,textfont=sf]{subfig}
\usepackage{textcomp}
\usepackage{stfloats}
\usepackage{url}
\usepackage{verbatim}
\usepackage{graphicx}
\usepackage{cite}
\usepackage{amsmath}
\usepackage{amsthm}
\usepackage{fancyhdr}
\usepackage{booktabs}
\usepackage{algorithm}
\usepackage{algorithmic}
\usepackage{bm}
\usepackage{amssymb}
\usepackage{color}
\usepackage{multirow}
\usepackage{makecell}
\usepackage{wrapfig}
\usepackage{threeparttable}
\usepackage[switch]{lineno}
\usepackage{pifont}
\usepackage{bbding}
\usepackage[dvipsnames]{xcolor}
\definecolor{c1}{HTML}{00B050}
\newcommand{\INPUT}{\item[\textbf{Input:}]}
\newcommand{\OUTPUT}{\item[\textbf{Output:}]}

\begin{document}

 \title{Striving for Faster and Better: A One-Layer Architecture with Auto Re-parameterization for Low-Light Image Enhancement} 
\author{Nan An, Long Ma,~\IEEEmembership{Member,~IEEE}, Guangchao Han, Xin Fan,~\IEEEmembership{Senior Member,~IEEE}, Risheng Liu,~\IEEEmembership{Member,~IEEE}
\thanks{This work was supported in part by the National Natural Science Foundation of China under Grant 62450072, U22B2052 and 62027826, in part by Distinguished Youth Fund Program of Liaoning Natural Science Foundation under Grant 2025JH6/101100001, in part by Innovation Supporting Plan of Distinguished Youth Scientific and Technological Talents of Dalian under Grant 2024RJ002, in part by Fundamental Research Funds for Dalian University of Technology under Grant DUT24LAB125, in part by China Postdoctoral Science Foundation under Grant 2023M740491, in part by Postdoctoral Fellowship Program of CPSF under Grant GZB20240098, and in part by the Dalian Science and Technology Innovation Fund-Young Tech Star under Grant 2023RQ017. 
	
	The authors are with the School of Software Technology, Dalian University of Technology, Dalian, 116024, China. (e-mail: nanan0527@hotmail.com; malone94319@gmail.com; guangchaohan66@gmail.com; xin.fan@dlut.edu.cn; rsliu@dlut.edu.cn).
%
%
%
	
	
	(Corresponding author: Risheng Liu)}

\thanks{}}

\markboth{Journal of \LaTeX\ Class Files,~Vol.~14, No.~8, August~2021}%
{Shell \MakeLowercase{\textit{et al.}}: A Sample Article Using IEEEtran.cls for IEEE Journals}


\maketitle

\IEEEpubid{\begin{minipage}{\textwidth}\ \\[35pt] \centering
		Copyright \copyright 2025 IEEE. Personal use of this material is permitted. However, permission to use this material for any other purposes \\ must be obtained from the IEEE by sending an email to pubs-permissions@ieee.org.
\end{minipage}}

\begin{abstract}
 Deep learning-based low-light image enhancers have made significant progress in recent years, with a trend towards achieving satisfactory visual quality while gradually reducing the number of parameters and improving computational efficiency. In this work, we aim to delving into the limits of image enhancers both from visual quality and computational efficiency, while striving for both better performance and faster processing. To be concrete, by rethinking the task demands, we build an explicit connection, \textit{i.e.,} visual quality and computational efficiency are corresponding to model learning and structure design, respectively. Around this connection, we enlarge parameter space by introducing the re-parameterization for ample model learning of a pre-defined minimalist network (\textit{e.g.,} just one layer), to avoid falling into a local solution. To strengthen the structural representation, we define a hierarchical search scheme for discovering a task-oriented re-parameterized structure, which also provides powerful support for efficiency. Ultimately, this achieves efficient low-light image enhancement using only a single convolutional layer, while maintaining excellent visual quality. Experimental results show our sensible superiority both in quality and efficiency against recently-proposed methods. Especially, our running time on various platforms (\textit{e.g.,} CPU, GPU, NPU, DSP) consistently moves beyond the existing fastest scheme. The source code will be released at \url{https://github.com/vis-opt-group/AR-LLIE}. 
\end{abstract}


\section{Introduction}
\label{sec:intro}
\IEEEPARstart{B}RIGHTENING images captured in low-light environments  (\textit{e.g.,} nighttime) to make more valuable information visible assumes a vital role in many actual applications, such as automatic driving and traffic surveillance. However, due to many unavoidable challenging shooting conditions, low-light images often suffer from various degradation factors (\textit{e.g.,} diminished contrast, color aberration, and significant noise), substantially increasing the difficulty of this task. The reduced visual quality of low-light images also hinders the development of some high-level semantic understanding tasks, \textit{e.g.,} detection~\cite{liang2021recurrent,liu2024infrared,liu2024task}, segmentation~\cite{sakaridis2019guided,liu2023multi,jiao2023pearl}, and tracking~\cite{lu2023cascaded,liu2020location,jin2014robust}. Hence, investigations into the enhancement of low-light images is of paramount importance and remains a longstanding topic in computer vision. 

Similar to the development trajectory of other classical image processing tasks, low-light image enhancement has also evolved from traditional methods to deep learning approaches. Early traditional methods typically designed algorithms based on the Retinex theory~\cite{land1971lightness}, which characterizes the physical imaging principles of low-light images and laid the foundation for model-based optimization techniques in low-light enhancement. These approaches can be categorized into two main types based on the optimization objectives: simultaneous optimization of multiple variables~\cite{gu2019novel,xu2020star,liu2018learning} and optimization of a core component (\textit{i.e.,} illumination)~\cite{guo2017lime,ma2022low}. The former suffers from low computational efficiency due to the simultaneous modeling and optimization of multiple variables. The latter addresses this challenge by simplifying the modeling process to optimize only a single core variable. Although this approach reduces computational complexity, it struggles to handle non-uniform low-light scenarios effectively.

Along with the vigorous development of deep learning, learning-based low-light image enhancers has experienced significant strides forward. In the early period of the developing deep learning-based low-light enhancers, researchers focused on designing complicated architectures to provide efficient enough mapping. Inspired by traditional model-based optimization approaches, representative deep learning methods primarily construct decomposition architectures grounded in physical imaging principles~\cite{wang2019underexposed}. These methods typically decompose low-light images into illumination and reflectance components, employing a dual-branch architecture and introducing various constraints to optimize these components separately. However, akin to the drawbacks of traditional methods, the complexity of the designed structure often results in low computational efficiency. Additionally, some methods based on optimization-inspired unrolling flows~\cite{liu2023low} and heuristically designed structures~\cite{he2023low} have also made certain progress. These methods mainly enhancing performance through meticulously designed modules tailored for the target scene according to prior knowledge. Although these approach can achieve excellent performance in specific scenarios, unfortunately, such methods often require significant human labor, rely heavily on the designer's engineering skills, and also face the issue of low computational efficiency.

\begin{figure*}[ht!]
	\scriptsize 
	\centering
	\begin{tabular}{c@{\extracolsep{1em}}c}
		\includegraphics[height=0.24\linewidth]{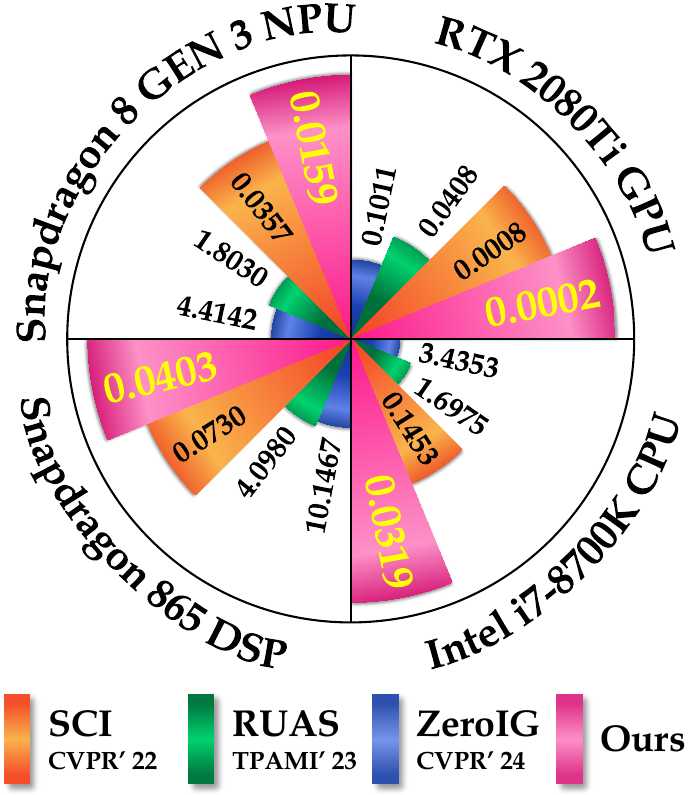}&
		\includegraphics[height=0.24\linewidth]{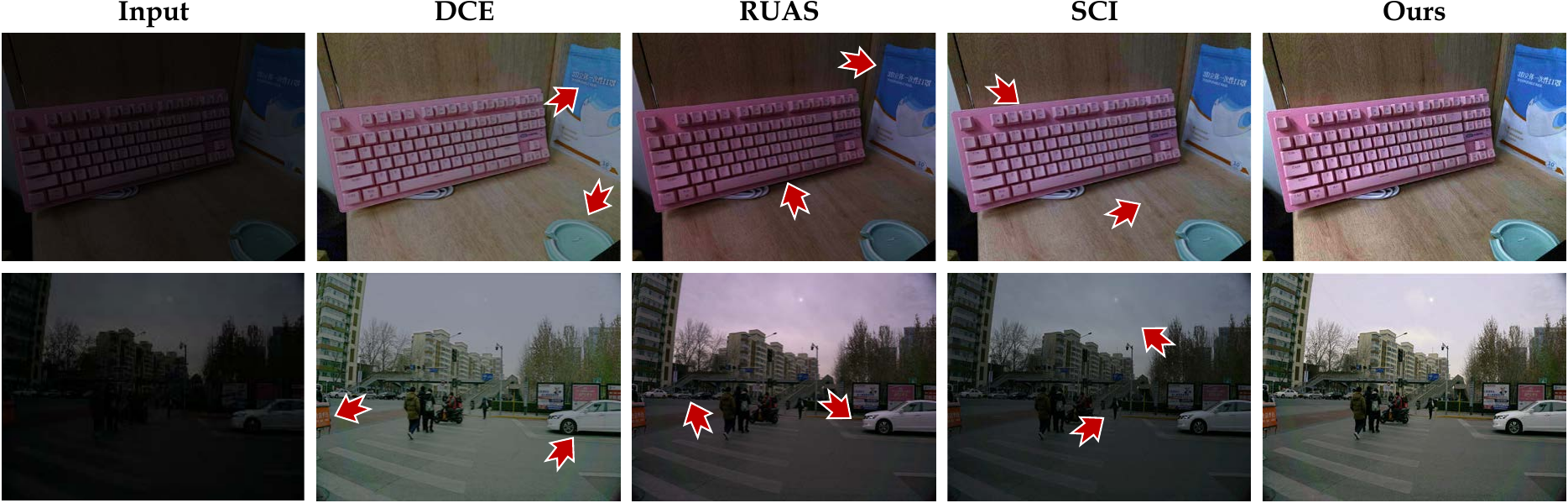}\\
		(a) Running time on various platforms&(b) Visual comparison on two challenging low-light scenarios\\
	\end{tabular}
	\caption{Performance evaluations among our AR-LLIE and three recent state-of-the-art representative low-light image enhancement approaches (ZeroIG~\cite{shi2024zero}, RUAS~\cite{liu2022learning}, and SCI~\cite{ma2022toward}). By evaluating the average running time of images with the size of 1920$\times$1080 on various computational platforms (CPU and GPU for PC, NPU and DSP for mobile) in (a), our AR-LLIE shows the consistently fastest time on all platforms. The visual comparisons of different scenarios in (b) further show the superior visual quality generated by our AR-LLIE.}
	\label{fig:FirstFigure}
\end{figure*}

{In recent years, with the success of learning strategy design in other fields~\cite{kandula2023illumination} and the increasing demand for real-time applications, the research focus of the realm of low-light image enhancement has gradually shifted to how to develop more effective learning strategies to enhance performance while ensuring efficiency. A straightforward approach is to apply neural architecture search (NAS) technology to overcome the limitations of heuristic design~\cite{liu2021retinex}. These approaches meticulously design structural constraints to enable the network to automatically discover optimal structures for specific tasks, aiming to construct lightweight networks that address low-light image enhancement. Additionally, another class of methods strives for efficiency while also addressing the over-reliance on paired data prevalent in existing works. These methods include approaches that frame the image enhancement problem as an image-specific curve estimation task~\cite{li2023lees} or introduce additional constraints based on the input image's own information for optimization~\cite{ma2022toward}. Such strategies aim to build unsupervised low-light image enhancement methods. Recently, with the increasing demand for real-time inference in practical applications, some works have attempted to achieve fast inference by utilizing re-parameterization techniques~\cite{Ding_2019_ICCV,ding2021diverse,zhang2021edge}. However, most existing methods~\cite{zhang2023learning} directly integrate pre-existing re-parameterized structures into the original architecture, neglecting the relevance between the chosen re-parameterization structure and the specific task requirements. As a result, these methods often require additional constraints or post-processing to achieve good performance.}

{To sum up, the research focuses for the above-mentioned deep enhancers progressively converts to be both better and faster from the single goal of better visual quality. Along with this trend, a straight issue is ``what is the ultimate limits of the deep enhancer for both visual quality and computational efficiency?''. In addition, a dramatic drawback for existing techniques is that the called high computational efficiency (\textit{e.g.,} DCE~\cite{Zero-DCE++}) is just verified in the PC devices, its actual practicability cannot be fully validated. Based on the above considerations and aiming to delve the limits of deep low-light image enhancers, we first rethink the illumination-based deep enhancer to explicitly bridge the sub-target for task (\textit{i.e.,} visual quality and computational efficiency) and sub-step for deep model design. Then we study these separated sub-tasks one by one to acquire our method. We introduce reparameterization to ensure structural simplicity and combine it with NAS to maximize the performance-enhancing effect of re-parameterization. The combination of these two techniques allows for flexible adjustment of the re-parameterization process based on actual needs, enabling us to achieve robust low-light image enhancement with just a single convolutional layer. In Fig.~\ref{fig:FirstFigure} (a), we report inference time on different computational platforms towards different devices including PC (CPU and GPU) and mobile (NPU and DSP). Obviously, our proposed method reaches a peak against other state-of-the-art methods, which reveals our practicability. The visual results shown in Fig.~\ref{fig:FirstFigure} (b) further indicate our superiority in visual quality. Our main contributions can be concluded as}
\begin{itemize}
	\item We design an auto re-parameterization mechanism to enlarge parameter space of a minimalist one-layer network, enabling sufficient parameter learning, avoiding local optima, and automatically discovering the optimal structure to ensure the model's capability. To the best of our knowledge, we are the first to combine NAS and re-parameterization for low-light image enhancement. 
	\item Considering that existing re-parameterization patterns are universally-designed and task-irrelevant, we define a hierarchical architecture search scheme for automatically discovering a task-oriented re-parameterized structure to significantly ameliorate the visual quality under guaranteeing computational efficiency. 
	\item Plenty of experiments are performed to verify our superiority in both visual quality and computational efficiency against other state-of-the-art methods. Remarkably, the running times on multiple platforms (\textit{e.g.,} CPU, GPU, NPU, DSP) in terms of different image sizes are all less than 30\% than the existing fastest method. 
\end{itemize}

\section{Related Work}
\label{sec:relate}

\subsection{Model-based Traditional Methods}
In the era of model-based traditional methods, prevailing approaches endeavored to guide the design of image enhancement through the incorporation of physical imaging models, with Retinex theory-based schemes standing out as the most emblematic and widely employed. The Retinex theory~\cite{land1971lightness} posits that low-light images can be decomposed into illumination and reflectance components, with the former closely tied to scene factors and the latter representing a well-exposed clear image. This theory plays a pivotal role in the field, spurring traditional methods to develop efficient statistical priors centered around Retinex theory.

A direct approach for Retinex theory-based traditional methods is to design regularization priors separately for the illumination and reflectance components. The work in~\cite{fu2015probabilistic} innovatively introduced the Total Variation (TV) regularization and applied it to the optimization of both illumination and reflectance. Gu~\emph{et al.}~\cite{gu2019novel} further refined integer-order TV regularization by employing fractional calculus to enhance the effectiveness of illumination decomposition. Conversely, the work in~\cite{xu2020star} designed a structure and texture-aware Retinex model, using an exponential local derivative to devise an exponential filter for regularizing the illumination and reflectance variables obtained from Retinex decomposition. Cai~\emph{et al.}\cite{cai2017joint} enhanced the illumination constraints by incorporating piece-wise shape prior and bright channel prior, establishing a unified intrinsic-extrinsic model.
Subsequently, to counteract the detrimental influence of potential noise, researchers turned their attention to formulating noise-suppressed reflectance priors. Fu~\emph{et al.}\cite{fu2016weighted} introduced weighted sparsity regularization to enforce piece-wise smoothing of the estimated reflectance. And the work in~\cite{li2018structure} proposed a robust Retinex model incorporating a noise map to achieve structure-revealed reflectance under a gradient-based fidelity term.



Despite the progress made by the aforementioned approaches that estimate and optimize dual variables using a single model, a notable drawback of these methods is their low computational efficiency, as the multi-variable setting significantly increases the inference burden. To address this issue, a series of methods focused on optimizing the core component (illumination) emerged and have become the mainstream in traditional approaches. LIME~\cite{guo2017lime} represents a seminal approach in this class of methods, employing Relative Total Variation (RTV) for direct illumination estimation, which initially devised for image smoothing~\cite{tsmoothing2012}. This method is notable for its simple single-variable optimization process, enabling it to achieve rapid inference while preserving impressive visual quality. Building on the success of single-variable optimization, subsequent works~\cite{ ma2022low} sought to address the over-exposure issues inherent in LIME. 
Zhang~\emph{et al.}~\cite{zhang2018high} introduced a perceptual bidirectional similarity criterion and combined it with Retinex theory to frame the exposure correction problem as an illumination optimization task, resulting in enhanced images with more uniform exposure. The work in~\cite{ma2022low} identified the over-exposure problem stemming from the use of image smoothing models for illumination estimation and proposed a novel fidelity measure to resolve it.


\subsection{Data-driven Deep Networks}
In recent years, spurred by the rapid advancements in deep learning and the exponential growth of low-light datasets, the research focus in low-light image enhancement has shifted from designing statistical priors to learning deep models from extensive datasets~\cite{liu2022learning,ma2022toward}. Data-driven deep network approaches now hold the central position in contemporary low-light image enhancement technology.


{Since the development of data-driven deep networks, designing network architectures~\cite{wei2018deep,jin2021bridging,zhang2019kindling,yang2021sparse,ma2021learning} has been a foundational and highly focused endeavor. Researchers have continuously experimented with various network structures to enhance the performance of low-light image enhancement effectively. RetinexNet~\cite{wei2018deep}, a milestone in this field, is inspired by Retinex theory and designs a decomposition network to simultaneously estimate illumination and reflectance. This classical decomposition framework has also inspired further exploration into more efficient modeling mechanisms. Building on this, KinD~\cite{zhang2019kindling} introduced additional modules to refine illumination and reflectance, strengthening the connections between different components. Kandula~\textit{et al.}~\cite{kandula2023illumination} proposed a method that primarily focuses on utilizing contextual information to guide and optimize illumination for unsupervised image enhancement. The work in~\cite{liu2024efficient} compared to the previous work, further integrated computational efficiency into the task objectives by constructing an efficient neural network without the need for additional decomposition networks or regularization functions. Moreover, with the recent surge in popularity of Transformers and diffusion models, a series of related novel approaches have emerged. LLFormer~\cite{wang2023ultra} introduces a Transformer-based low-light image enhancement algorithm designed for ultra-high-definition images, utilizing an axial multi-head self-attention mechanism and cross-layer attention fusion blocks to achieve low linear complexity. The work in ~\cite{wang2023lldiffusion} proposes a degradation-aware learning scheme using diffusion models, which considers latent degradation representations to guide the diffusion process, enabling effective brightness enhancement for low-light images. Recently, Hou~\emph{et al.}~\cite{hou2024global} introduced a global structure-aware regularization method, which helps preserve intricate details and enhance contrast during the diffusion process, mitigating noise generation.}



{Compared to methods centered on network architecture design, recent advancements in low-light image enhancement have increasingly focused on leveraging carefully designed learning strategies to improve computational efficiency, leading to the proposal of a series of efficiency-oriented approaches~\cite{Zero-DCE++,ma2022toward,liu2022learning,fu2023learning,shi2024zero,zhang2023learning}. Guo~\emph{et al.}~\cite{Zero-DCE++} introduced a lightweight unsupervised approach, meticulously tailoring a series of loss functions to capture pixel-level high-order curves. Ma~\emph{et al.}~\cite{ma2022toward} crafted a weight-sharing cascaded process for illumination estimation, accelerating the algorithm through self-calibration learning strategies that constrain each stage. The work in~\cite{liu2022learning} designed a cooperative learning framework, exploring the intrinsic connections between various visual tasks and features in low-light scenes. Fu~\emph{et al.}~\cite{fu2023learning} proposed establishing connections between images with identical content within the reflectance domain, employing a self-supervised mechanism to further eliminate the interference of redundant features in RAW images. ZeroIG~\cite{shi2024zero} developed a zero-shot image enhancement method, introducing denoising-related constraints in the illumination-guided enhancement process to better handle noise interference. SCLM~\cite{zhang2023learning} aims to simplify the network architecture to achieve lightweight image enhancement. They first build a single convolutional layer model for coarse enhancement
using re-parameterization, and then introduce a curve adjustment-based local adaptive module to optimize the enhancement results. However, due to the difficulty in ensuring the correlation between the predefined re-parameterization structure and task requirements, SCLM experiences certain limitations in performance enhancement.}




\section{Learning via Auto Re-parameterization}\label{sec:auto}
Currently, the predominant approach for image enhancement involves stacking network layers (denoted as $\bm{\alpha}^{\mathtt{t}}$) to form the model space (denoted as $\Theta(\bm{\alpha}^{\mathtt{t}})$) and derive parameters (denoted as $\bm{\theta}\in\Theta(\bm{\alpha}^{\mathtt{t}})$) using available data. From the task demand, an effective image enhancer is expected to meet two fundamental requirements: high visual quality and efficient inference. Notably, there exists a direct relationship between the aforementioned learning mechanism and the desired outcomes: the quality of the visual output is closely tied to the learning of model parameters, while the computational efficiency of the network is inherently linked to the complexity of its structure.


Notably, the provided viewpoints condense a target-oriented solving paradigm on how to design deep enhancers, \textit{i.e.,} from structure design (determines efficiency) to model learning (influences quality). In practice, most current deep enhancers prioritize visual quality as the primary task objective, often overlooking the need for efficient network architectures. Very recently, there appears a few studies~\cite{Zero-DCE++,ma2022toward,liu2022learning} that concentrate on designing highly efficient models. For instance, the work in~\cite{ma2022toward} achieves image enhancement using only three convolution blocks, which enters a new high-speed and high-quality era for low-light image enhancement.

Learning a mapping to infer the illumination from observations, as presented in several previous works~\cite{liu2022learning,ma2022toward}, and then deriving the reflectance (\textit{i.e.,} the desired objective) according to the Retinex theory, is a commonly adopted paradigm. This process can be defined as 
\begin{equation}
	\mathbf{x}=\mathbf{y}\oslash\mathbf{u},	
	\mathbf{u}=\mathcal{G}_{\bm{\alpha}}(\mathbf{y};\bm{\theta}),
\end{equation}
where $\mathbf{y}, \mathbf{u}, \mathbf{x}$ are the low-light observation, illumination, and reflectance, respectively. The mapping $\mathcal{G}$ contains the architecture $\bm{\alpha}$ that needs to be defined and the model parameters $\bm{\theta}$ (acquired by optimizing $\min_{\bm{\theta}}\mathcal{L}_{}(\bm{\theta})$ on available data). The symbol $\oslash$ denotes the element-wise division operation. 


Based on the above considerations, to explore the limits of an enhancer that can balance visual quality and efficiency, we consider an extreme scenario where the architecture $\bm{\alpha}$ is defined as a one-layer basic 3$\times$3 convolutional block with 3 channels (including a LeakyReLU nonlinearity layer). We also introduce a parameter-free residual mechanism to accelerate the learning process. The primary challenge associated with one-layer learning is the limited representation capacity, which intuitively leads to easily getting stuck in local optimal solutions. To counter this, in the upcoming sections, we will initially introduce the re-parameterization measure to expand the model space. Then, we'll further broaden the structural space for the re-parameterized operator through tiered NAS.

 \begin{figure}[!htb]
	\centering
	\footnotesize
	\begin{tabular}{c}
		\includegraphics[width=0.98\linewidth]{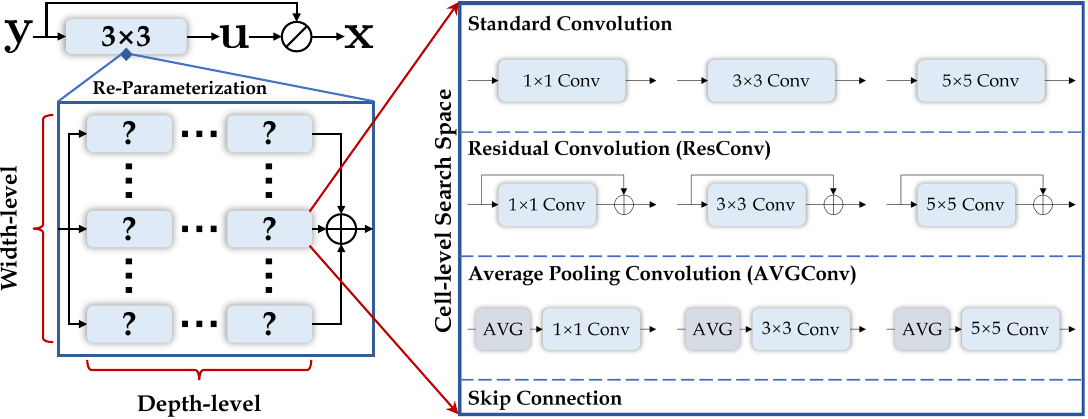}\\
	\end{tabular}
	\caption{{The overall framework of the our AR-LLIE.}}
	\vspace{-3mm}
	\label{fig: flowchart}
\end{figure}

\subsection{Re-parameterization in 1-Layer CNN}
{As shown in Fig.~\ref{fig: flowchart}, in our defined one-layer architecture, the model space is confined to only $3^4$ parameters that require learning. To enrich the representation capability during the learning phase and avoid increasing computational burden in the inference phase, we introduce the re-parameterization to expand the parameter space for learning,  according to the homogeneity and additivity for the convolution, we have}
\begin{equation}
	\mathcal{G}_{\bm{\alpha}}(\mathbf{y},\bm{\theta})=\mathcal{H}_{\bm{\alpha}}\Big(\mathbf{y},\sum_{s=1}^{S}\bm{\pi}_{s}\Big),
\end{equation}
where $\mathcal{H}_{\bm{\alpha}}$ represents the re-parameterized multi-branch structure $\bm{\alpha}$ with the learnable parameters $\sum_{s=1}^{S}\bm{\pi}_{s}$. The variable $S$ represents numbers of the basic unit in the re-parameterization. Moreover, the convolutional parameters that are composed of kernel weights and bias all satisfy the above permutation process. 


According to the above equivalence transformation, we can subsequently establish the following optimization goal
\begin{equation*}
	\min_{\bm{\pi}_{s}}\mathcal{L}_{}\Big(\sum_{s=1}^{S}\bm{\pi}_{s}\Big),\;\;
	s.t.,\;\sum_{s=1}^{S}\bm{\pi}_{s}=\bm{\theta},
\end{equation*}

It says that we adopt multi-branch architecture in the learning phase, and then merge parameters from multi-branch architecture into the desired parameters from one-layer network in the inference phase. 

Regrettably, the prevailing re-parameterized structures~\cite{ding2021repvgg} are typically fashioned for general representation, lacking a task-oriented mechanism, making it challenging to ensure the capture of crucial information for our proposed illumination estimation\footnote{Please refer to Sec.~\ref{sec:exp} for experimental verification}. Consequently, the next challenge lies in how to define the topology for the re-parameterized structure.


\begin{algorithm}[h] 
	\caption{Tiered NAS for Re-parameterization.}\label{alg:TN}
	\begin{algorithmic}[1] 
		\INPUT The search space $\mathcal{A}$, the training and validation datasets $\mathcal{D}_\mathtt{tr}$ and $\mathcal{D}_\mathtt{val}$ and other necessary hyper-parameters.
		\OUTPUT The searched architecture.
		
		\STATE Initialize $\bm{\alpha}=\{\bm{\alpha}_{w},\bm{\alpha}_{d},\bm{\alpha}_{c}\}$ and $\bm{\omega}$.
		\STATE // Search for $\bm{\alpha}_{w}$ at width-level.
		\WHILE {not converged}
		\STATE $\bm{\omega}^* \leftarrow\bm{\omega}-\nabla_{\bm{\omega}}\mathcal{L}_{}^{\mathtt{tr}}(\bm{\omega};\bm{\alpha})$.
		\STATE ${\bm{\alpha}_{w}^*} \leftarrow\bm{\alpha}_{w}-\nabla_{\bm{\alpha}_{w}}\mathcal{L}_{}^{\mathtt{val}}(\bm{\alpha}_{w}(\bm{\alpha}_{d}(\bm{\alpha}_{c}));\bm{\omega}^*(\bm{\alpha}))$.
		\ENDWHILE
		
		\STATE // Search for $\bm{\alpha}_{d}$ at depth-level with fixed $\bm{\alpha}_{w}^*$.
		\WHILE {not converged}
		\STATE $\bm{\omega}^* \leftarrow\bm{\omega}-\nabla_{\bm{\omega}}\mathcal{L}_{}^{\mathtt{tr}}(\bm{\omega};\bm{\alpha})$.
		\STATE ${\bm{\alpha}_{d}^*} \leftarrow\bm{\alpha}_{d}-\nabla_{\bm{\alpha}_{d}}\mathcal{L}_{}^{\mathtt{val}}({\bm{\alpha}_{w}^*}(\bm{\alpha}_{d}(\bm{\alpha}_{c}));\bm{\omega}^*(\bm{\alpha}))$.
		\ENDWHILE
		
		\STATE // Search for $\bm{\alpha}_{c}$ at cell-level with fixed $\bm{\alpha}_{w}^*$ and $\bm{\alpha}_{d}^*$.
		\WHILE {not converged}
		\STATE $\bm{\omega}^* \leftarrow\bm{\omega}-\nabla_{\bm{\omega}}\mathcal{L}_{}^{\mathtt{tr}}(\bm{\omega};\bm{\alpha})$.
		\STATE ${\bm{\alpha}_{c}^*} \leftarrow\bm{\alpha}_{c}-\nabla_{\bm{\alpha}_{c}}\mathcal{L}_{}^{\mathtt{val}}({\bm{\alpha}_{w}^*}({\bm{\alpha}_{d}^*}(\bm{\alpha}_{c}));\bm{\omega}^*(\bm{\alpha}))$.
		\ENDWHILE

		\RETURN Multi-branch architecture for re-parameterization derived based on ${\bm{\alpha}_{w}^*}, {\bm{\alpha}_{d}^*}$ and ${\bm{\alpha}_{c}^*}$.
	\end{algorithmic}
\end{algorithm}
\subsection{Auto Re-parameterization via Tiered NAS}
Although we have enlarged parameter space by re-parameterization, the limited structural representation still cannot realize the performance gain to the maximum extent. Therefore, it is essential to expand structure space for the re-parameterized model for better learning. 


\subsubsection{Multi-level structure modeling}
To construct a re-parameterized structure related to the demand of low-light image enhancement, we introduce the neural architecture search technique to automatically discover a	completely new task-oriented re-parameterization structure. In the previous section, we introduce the basic unit for the re-parameterization, it contains three hyper-parameters that need to be determined on different levels. 
Here we defined the whole search process to fully perform the characteristics of the re-parameterized structure into three levels, including width-level, depth-level and cell-level, presented as $\bm{\alpha}=\{\bm{\alpha}_{w},\bm{\alpha}_{d},\bm{\alpha}_{c}\}$. 
\begin{equation*}
	\begin{aligned}
		&\min_{\bm{\alpha}}\mathcal{L}_{}^{\mathtt{val}}\Big(\bm{\alpha}_{w}\big(\bm{\alpha}_{d}(\bm{\alpha}_{c})\big);\bm{\omega}^{*}\Big),\\
		&s.t.,\bm{\omega}^{*}=\arg\min_{\bm{\omega}}\mathcal{L}_{}^{\mathtt{tr}}(\bm{\omega};\bm{\alpha}),\\
	\end{aligned}
\end{equation*}
where $\mathcal{L}_{}^{\mathtt{val}}$ and $\mathcal{L}_{}^{\mathtt{tr}}$ respectively represent the loss for validation and training, and $\bm{\omega}$ refers to all learnable parameters of the pre-constructed multi-branch architecture. Based on the above modeling, directly applying existing neural architecture search strategies to jointly optimize the structure across all levels seems challenging due to complex coupling relationships between different levels (\textit{i.e.,} the maximum of the parameters space is directly influenced by the number and specific structure of cells which are also intimately connected with the width and depth of the network architecture). To address the intricacies of the constructed multi-level structure modeling from a clearer perspective, we propose a tiered NAS. In the following, we will provide detailed insights into our designed search procedure.

\subsubsection{Tiered NAS}
In which, as for width-level, it focuses on the number of branches in re-parameterized blocks and the maximum number of branch is 8. As for depth-level, it aims at determining the topology of every branch, specifically, the number of cells, which is up to 2. As for the minimum cell-level, it decides each specific operation in every branch from search space. We establish a lightweight search space with some simple operations including standard convolution, residual convolution and the convolution followed by average pooling. All kinds of the convolutions above contain three different kernel situation, including $1\times 1$, $3\times 3$ and $5 \times 5$. For the three aforementioned levels, we start the search by considering the width-level, which directly affects the upper limit of the parameter count. Building upon the initial multi-branch structure obtained in the previous stage, we further determine the depth of each branch. After the first two stages, the overall re-parameterization structure framework has been preliminarily established. Within this framework, based on the constructed lightweight search space, we conduct a search for the specific network structure at the minimum cell-level. This final structure is then used for subsequent re-parameterization. 

Alg.~\ref{alg:TN} summarizes the overall search process. Specifically, the entire search strategy can be divided into three stages. First, we update the width-level $\bm{\alpha}_{w}$ (Steps 3-6). Subsequently, we update the depth-level $\bm{\alpha}_{d}$ (Steps 8-10) while keeping the width settings (\textit{i.e.,} ${\bm{\alpha}_{w}}^*$) fixed. Finally, we update the specific structure of each cell $\bm{\alpha}_{c}$ (Steps 13-16), keeping both the width and depth (\textit{i.e.,} ${\bm{\alpha}_{w}}^*$ and ${\bm{\alpha}_{d}}^*$) fixed. Note that during the entire search process, $\bm{\omega}$ represents all learnable network parameters at the current stage. For each stage, we employ the widely used one-step finite difference technique to approximately calculate the gradient of the upper-level variables. For each level of the search (from width to depth to cells), we adopt an alternating update approach for  $\bm{\alpha}$ and  $\bm{\omega}$.

\subsection{Loss Functions}~\label{sec:loss}
We adopt a classical unsupervised loss~\cite{ma2022toward,liu2022learning} in search and training phases, described as 
\begin{equation*}
	\mathcal{L}_{total} = \mathcal{L}_{fidelity} + \lambda\mathcal{L}_{smooth},
\end{equation*}
where $\mathcal{L}_{fidelity}$ and $\mathcal{L}_{smooth}$ represent the fidelity loss and smoothing loss respectively. And $\lambda$ is the positive balancing parameter. The fidelity aims to maintain the pixel-level consistency between the low-light input and estimated illumination, formulated as $\mathcal{L}_{fidelity} = {\lVert \mathbf{u} - \mathbf{y} \rVert}^2$. As for smoothing loss~\cite{Zero-DCE++,zhang2021beyond}, we apply the spatial-variant $l_{1}$ norm as the smoothness term, formulated as $\mathcal{L}_{smooth} = \sum_{i=1}^{N} \sum_{j \in T\left(i\right)} \omega_{i,j}\left| \mathbf{u}_{i} - \mathbf{u}_{j} \right |$, where $i$ represent the $i$-th pixel, $N$ is the total number of pixels and $T\left(i\right)$ denotes the adjacent pixels of $i$. $\omega_{i,j}$  is formulated as $\omega_{i,j} = \exp \left(-\frac{\sum_{c}\left(\mathbf{y}_{i,c}-\mathbf{y}_{j,c}\right)^{2}}{2\sigma^{2}}\right)$, where $c$ denotes the image channel in RGB color space. As the standard deviations in Gaussian kernel, $\sigma$ is set to $0.1$.

\begin{table*}[ht]
	\footnotesize
	\setlength{\tabcolsep}{1.4mm}
	\renewcommand\arraystretch{1.5}	
	\centering
	\caption{{Quantitative comparison with advanced methods of low-light image enhancement on different benchmarks. Note that we adopt SCUNet as a post-processing denoiser for all methods to ensure fairness.}}
	\begin{tabular}{|c|c||ccccccccccc||c|}
		\hline
		\multicolumn{2}{|c||}{Method} &FIDE &KinD &UTVNet &RetinexNet &DCE &SCI &RUAS &RQ-LLIE &PairLIE &NeRCo &ZeroIG &\multirow{2}{*}{Ours}\\
		\cline{1-13}
		\multicolumn{2}{|c||}{Reference} &{\scriptsize \textit{CVPR' 20}} &{\scriptsize\textit{IJCV' 21}} &{\scriptsize \textit{ICCV' 21}} &{\scriptsize \textit{BMVC' 18}} &{\scriptsize \textit{TPAMI' 22}} &{\scriptsize \textit{CVPR' 22}} &{\scriptsize \textit{TPAMI' 23}} &{\scriptsize \textit{ICCV' 23}} &{\scriptsize \textit{CVPR' 23}} &{\scriptsize \textit{ICCV' 23}} &{\scriptsize \textit{CVPR' 24}} &\\
		\hline
		\multirow{3}{*}{MIT} &PSNR$\uparrow$ &17.0726 &16.1455 &18.0849 &12.7642 &16.2504 &14.9042 &\textbf{\textcolor{blue}{19.9250}} &18.7499 &12.9039 &18.3654 &15.9966 &\textbf{\textcolor{red}{24.0157}}\\
		~ &SSIM$\uparrow$ &0.6808 &0.7153 &0.6076 &0.6318 &0.7225 &0.6914 &\textbf{\textcolor{blue}{0.8078}} &0.7682 &0.6267 &0.6137 &0.6831 &\textbf{\textcolor{red}{0.8571}}\\
		~ &NIQE$\downarrow$ &4.5114 &3.8948 &3.6749 &4.5496 &\textbf{\textcolor{red}{3.4991}} &3.9512 &3.5355 &3.6841 &5.0973 &3.8496 &5.8804 &\textbf{\textcolor{blue}{3.5183}}\\
		\hline
		\multirow{3}{*}{LOL} &PSNR$\uparrow$ &18.3487 &16.6315 &16.1642 &15.6684 &14.8827 &15.7702 &16.2703 &10.0308 &16.9672 &\textbf{\textcolor{red}{18.9078}} &14.8360 &\textbf{\textcolor{blue}{18.4800}}\\
		~ &SSIM$\uparrow$ &0.5863 &0.5556 &0.5675 &0.5979 &\textbf{\textcolor{blue}{0.6153}} &0.6112 &0.5292 &0.3265 &0.5890 &0.5910 &0.4817 &\textbf{\textcolor{red}{0.6694}}\\
		~ &NIQE$\downarrow$ &5.1584 &5.3289 &5.1979 &4.8798 &4.3855 &4.6570 &4.8565 &5.8851 &\textbf{\textcolor{blue}{4.2763}} &4.2871 &5.2221 &\textbf{\textcolor{red}{4.1518}}\\
		\hline
	\end{tabular}
	
	\label{table:quan}
\end{table*}

\begin{figure*}[htb!]
	\centering
	\begin{tabular}{c@{\extracolsep{0.2em}}c@{\extracolsep{0.2em}}c@{\extracolsep{0.2em}}c@{\extracolsep{0.2em}}c}
		\includegraphics[width=0.195\linewidth]{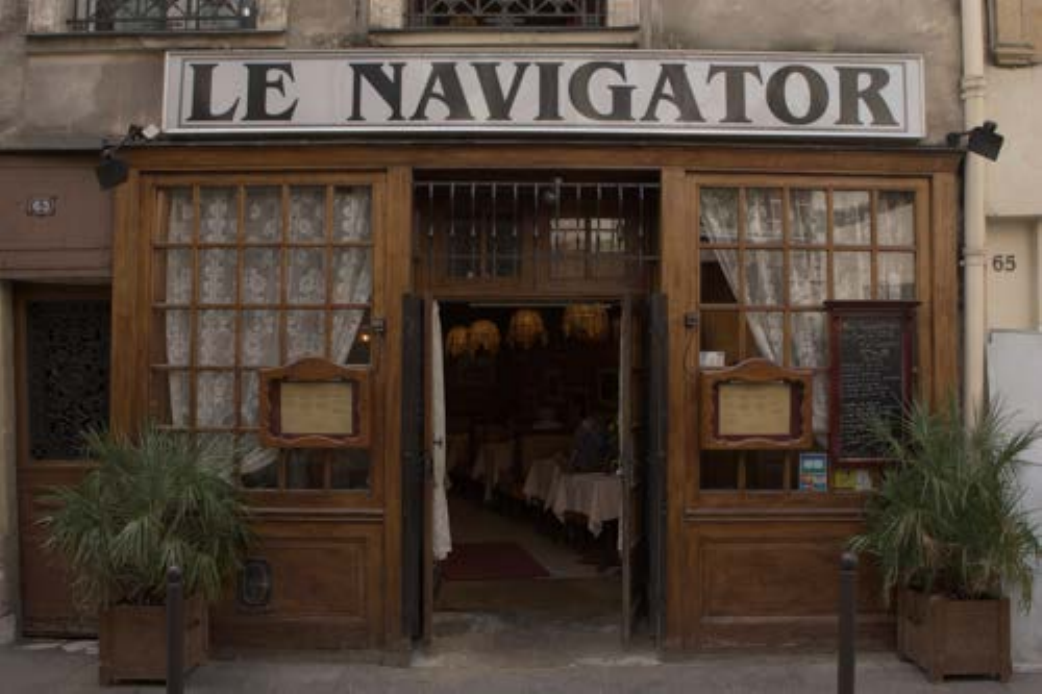}&
		\includegraphics[width=0.195\linewidth]{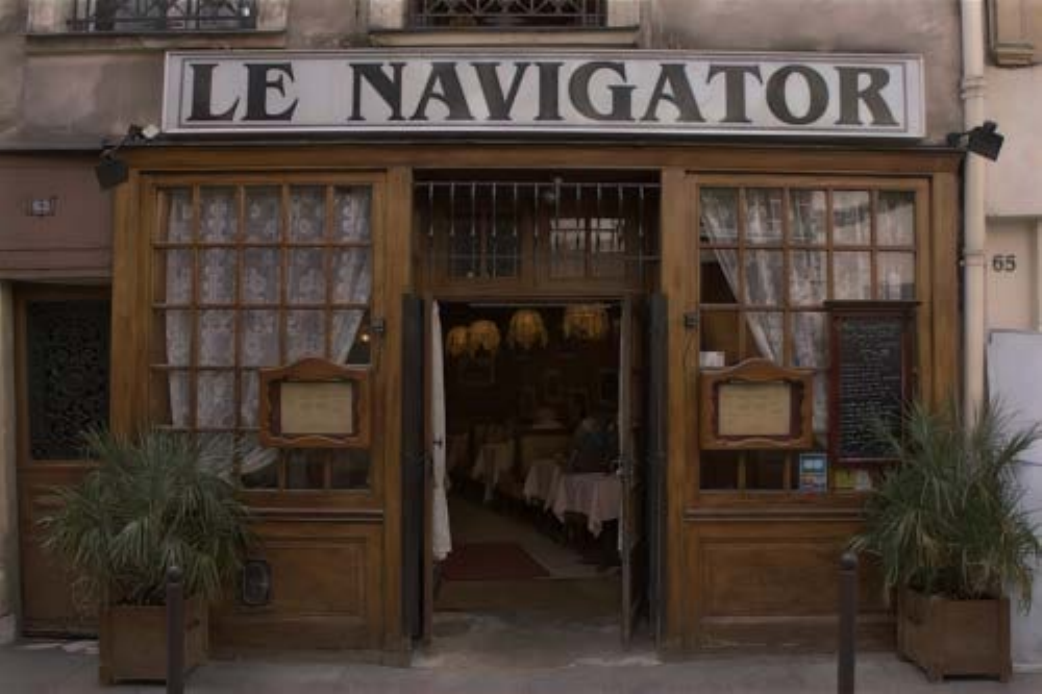}&
		\includegraphics[width=0.195\linewidth]{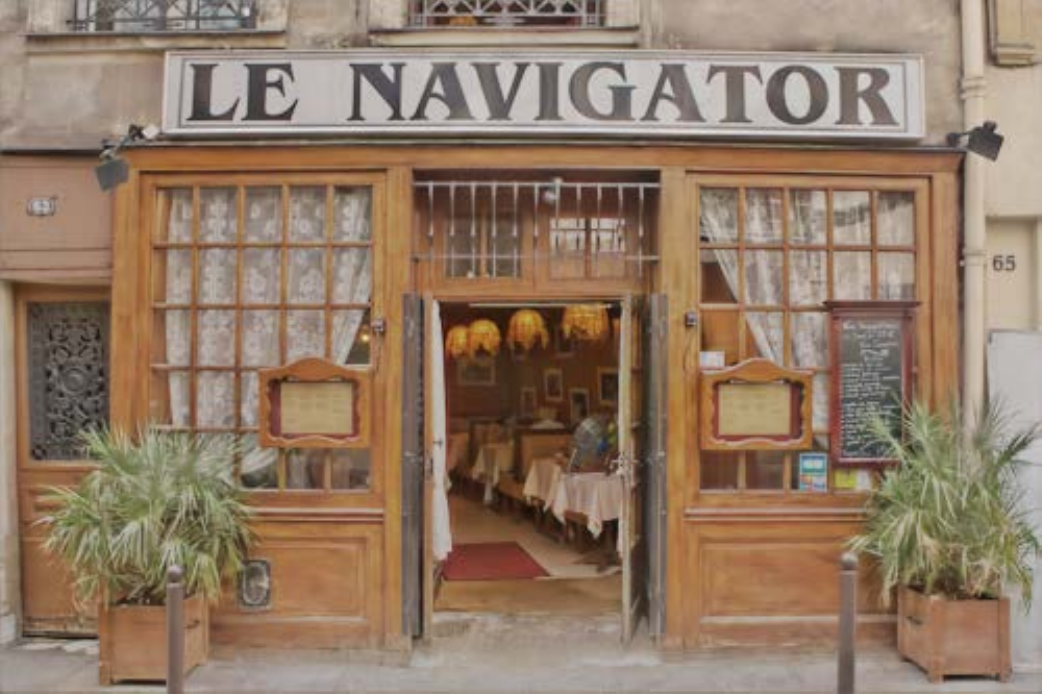}&
		\includegraphics[width=0.195\linewidth]{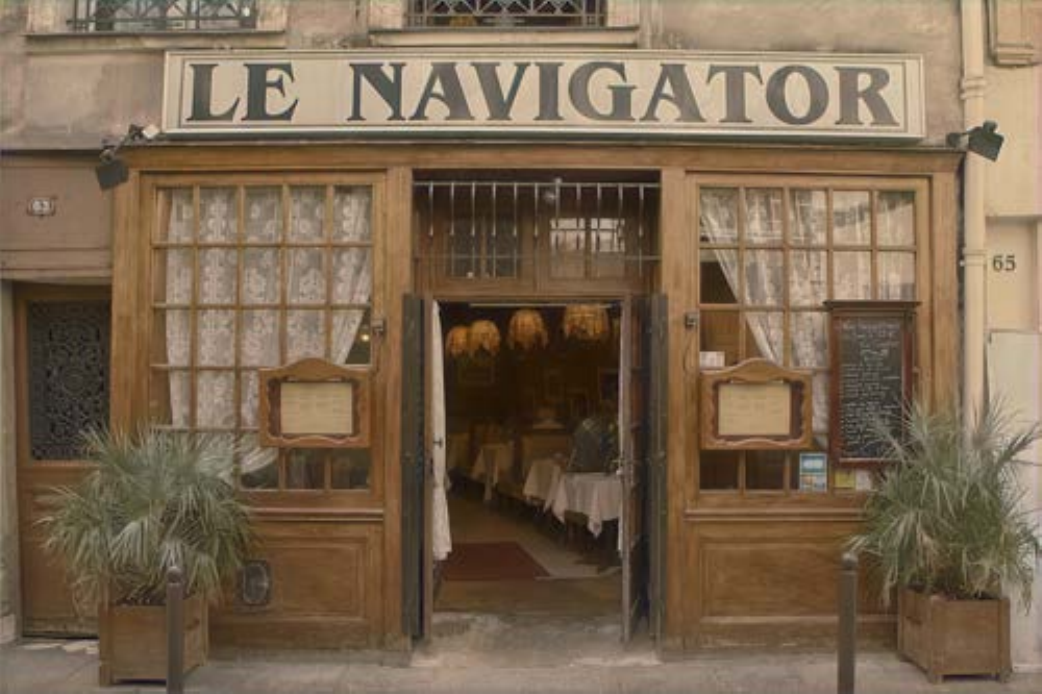}&
		\includegraphics[width=0.195\linewidth]{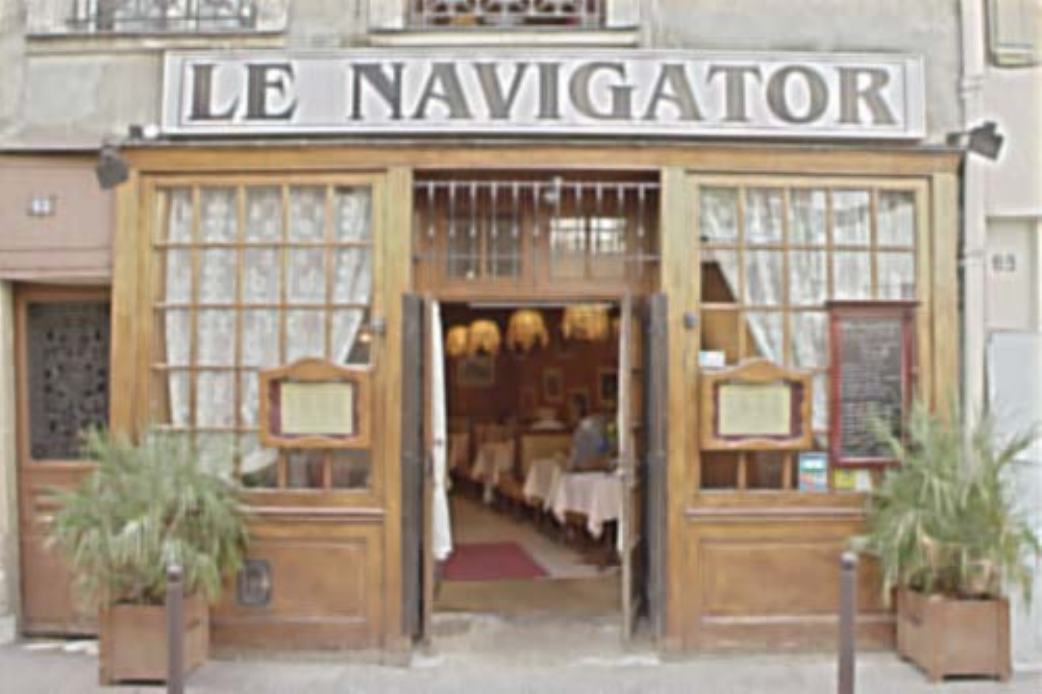}\\
		\includegraphics[width=0.195\linewidth]{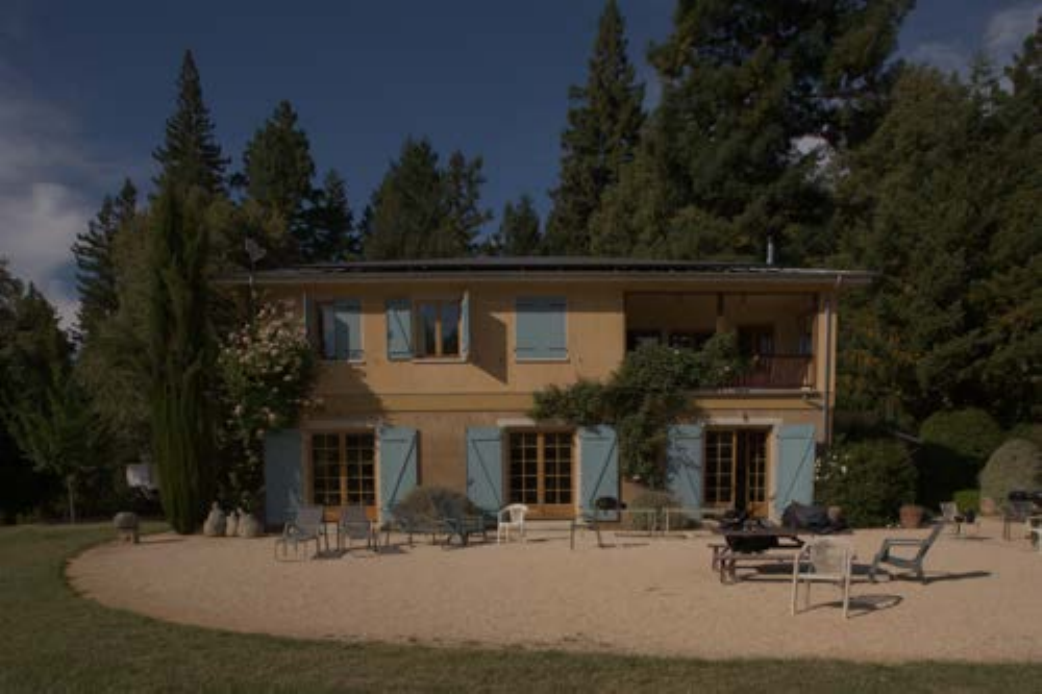}&
		\includegraphics[width=0.195\linewidth]{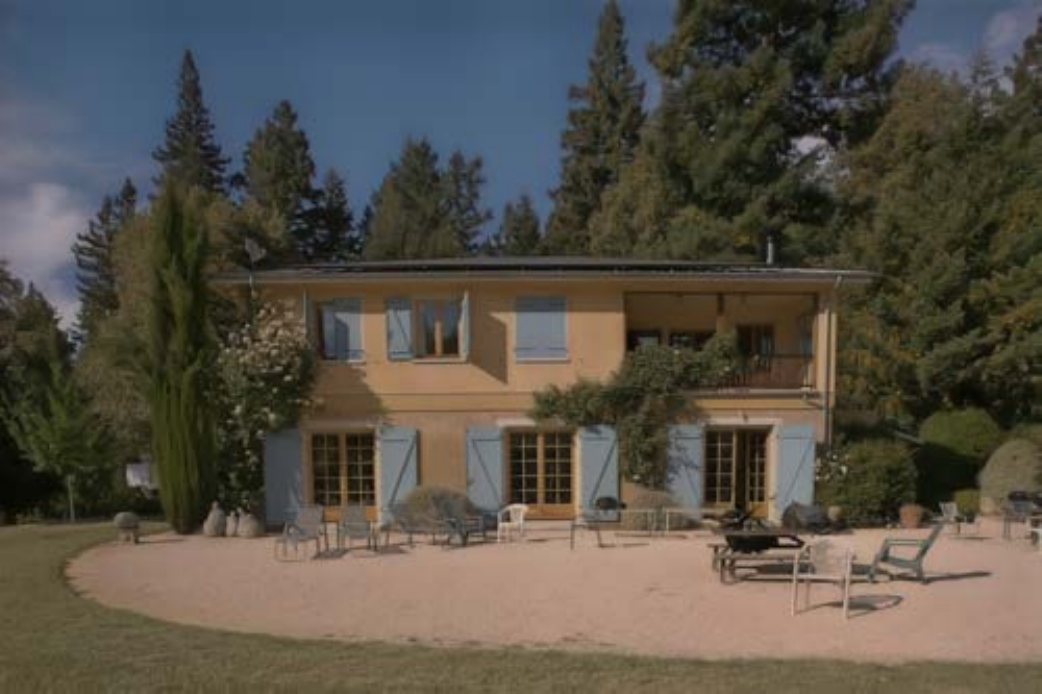}&
		\includegraphics[width=0.195\linewidth]{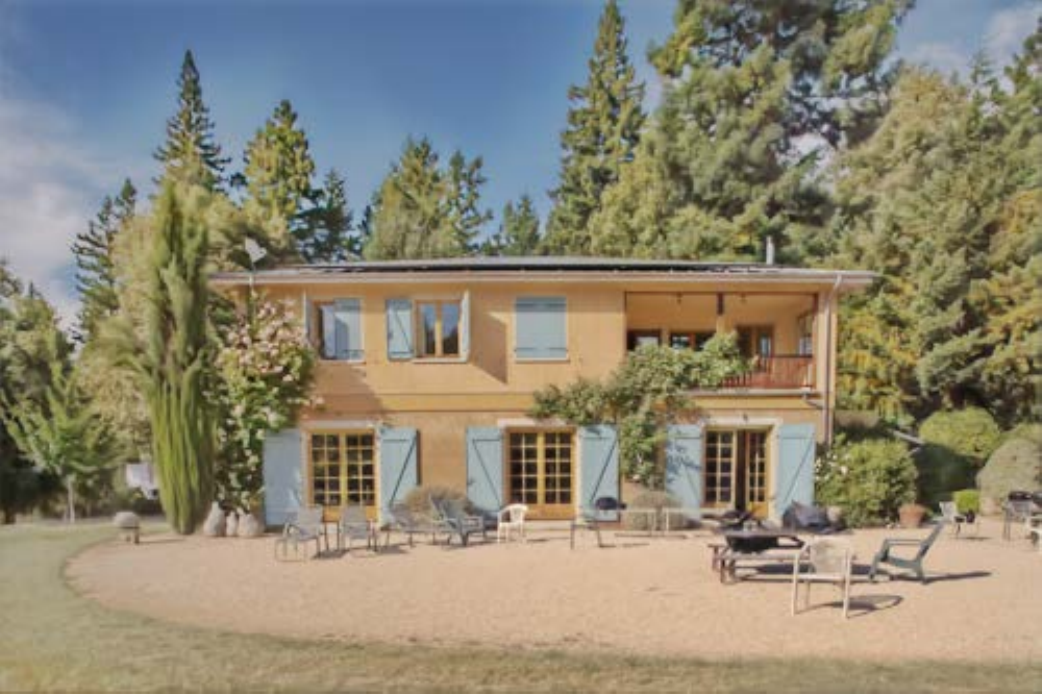}&
		\includegraphics[width=0.195\linewidth]{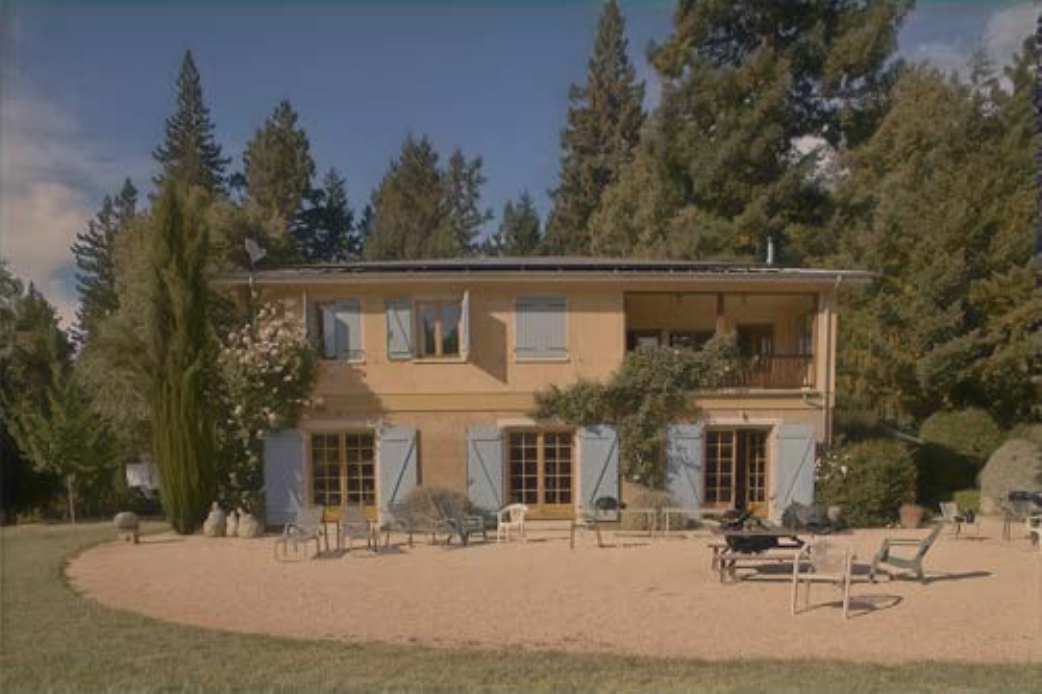}&
		\includegraphics[width=0.195\linewidth]{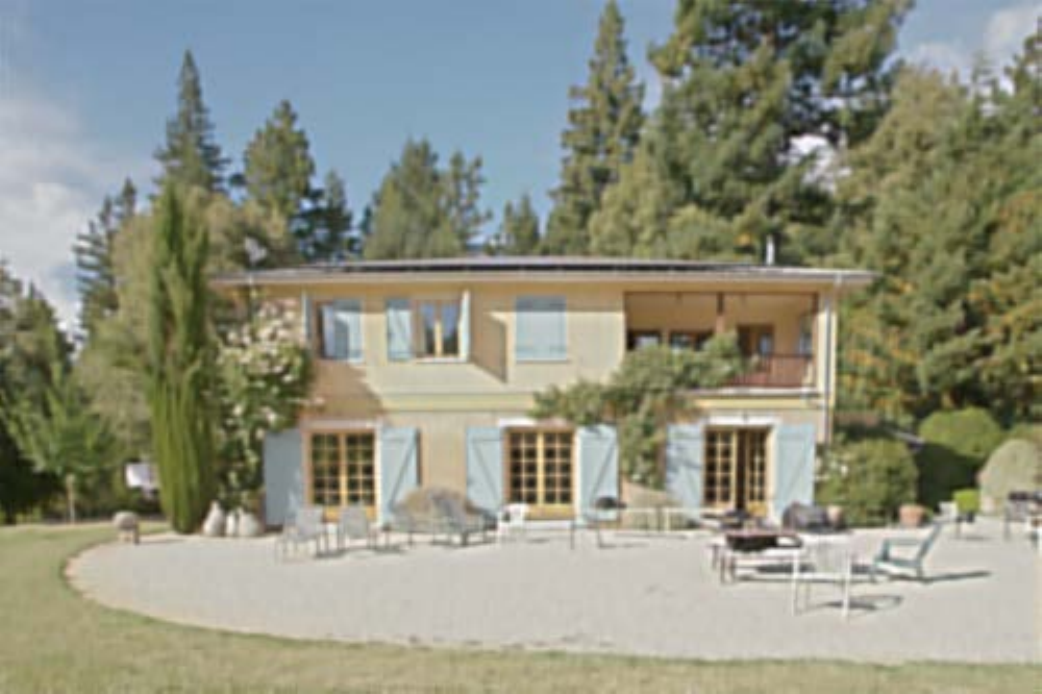}\\
		\footnotesize Input&\footnotesize FIDE&\footnotesize KinD&\footnotesize UTVNet&\footnotesize ZeroIG\\
		\includegraphics[width=0.195\linewidth]{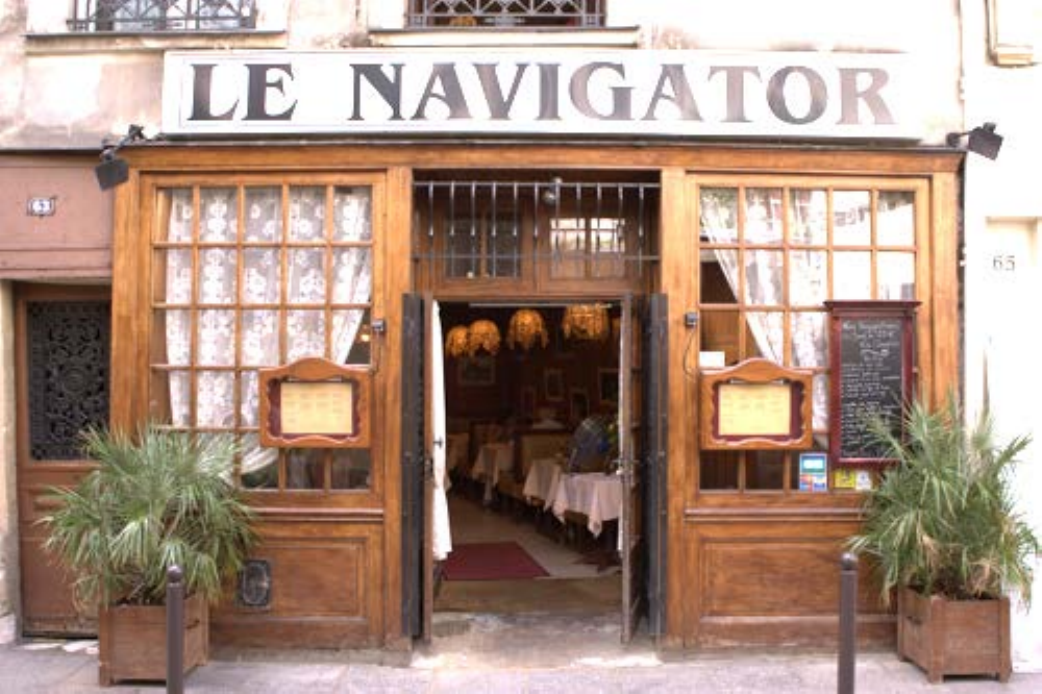}&
		\includegraphics[width=0.195\linewidth]{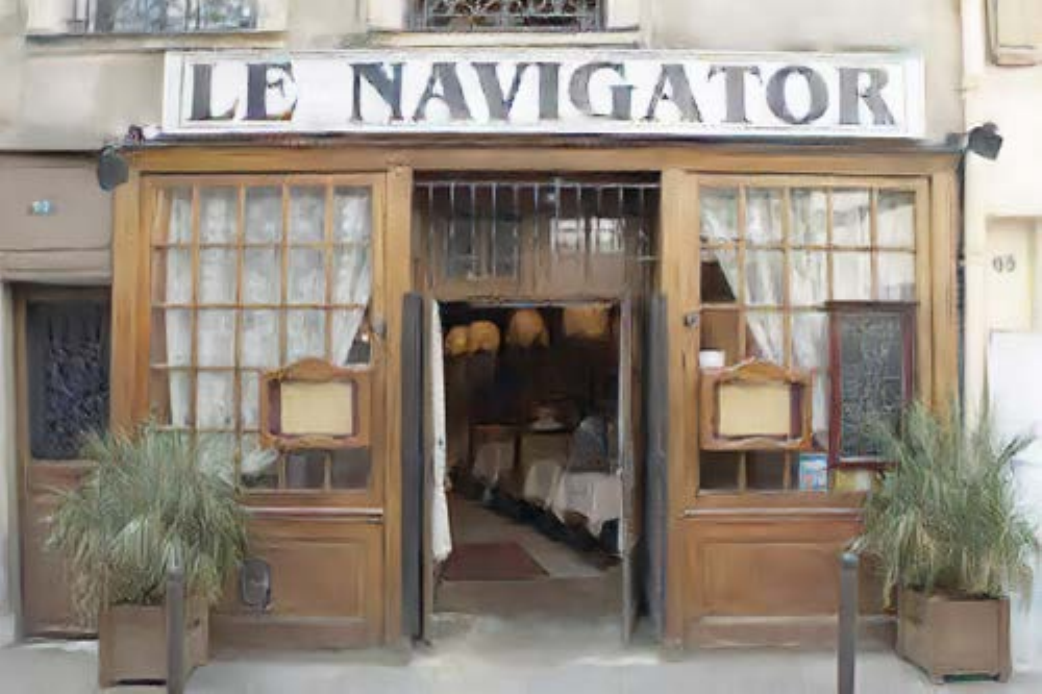}&
		\includegraphics[width=0.195\linewidth]{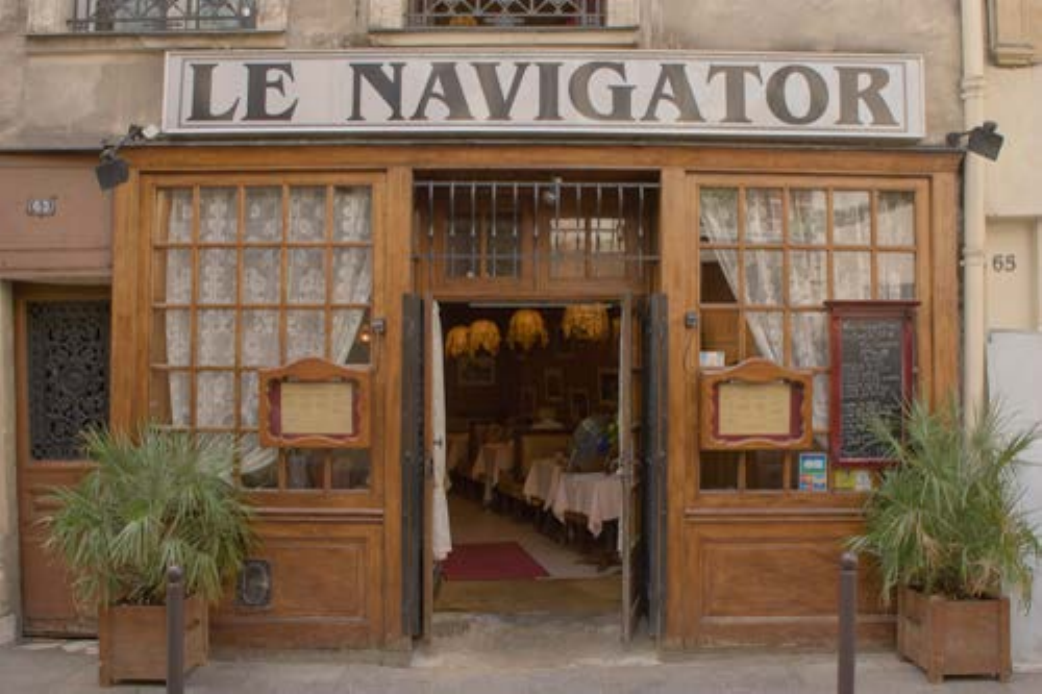}&
		\includegraphics[width=0.195\linewidth]{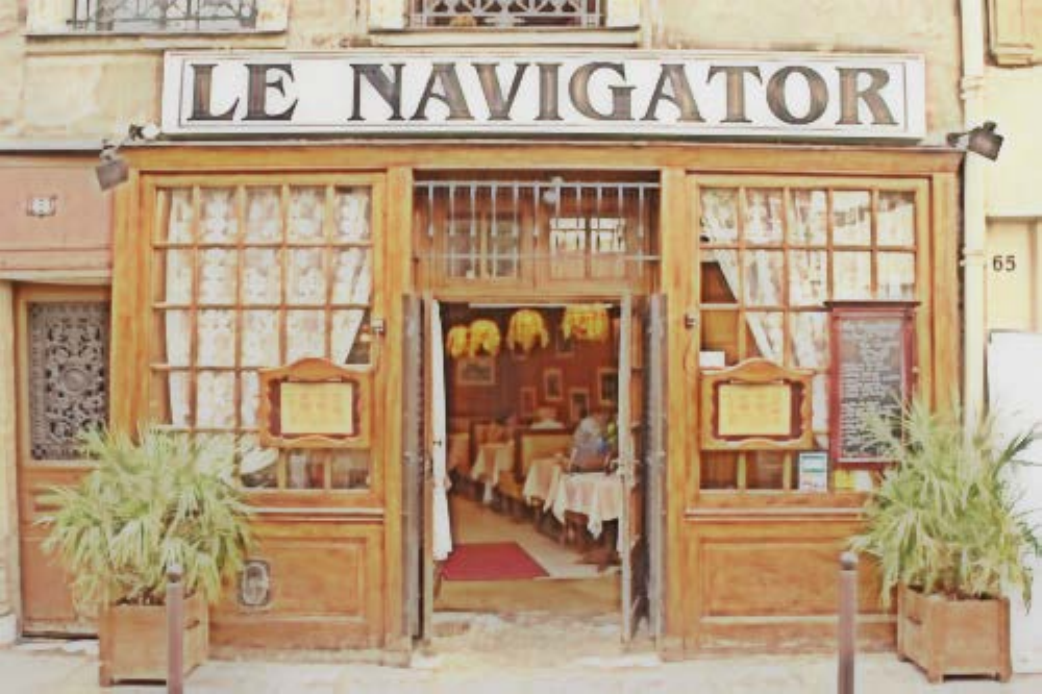}&
		\includegraphics[width=0.195\linewidth]{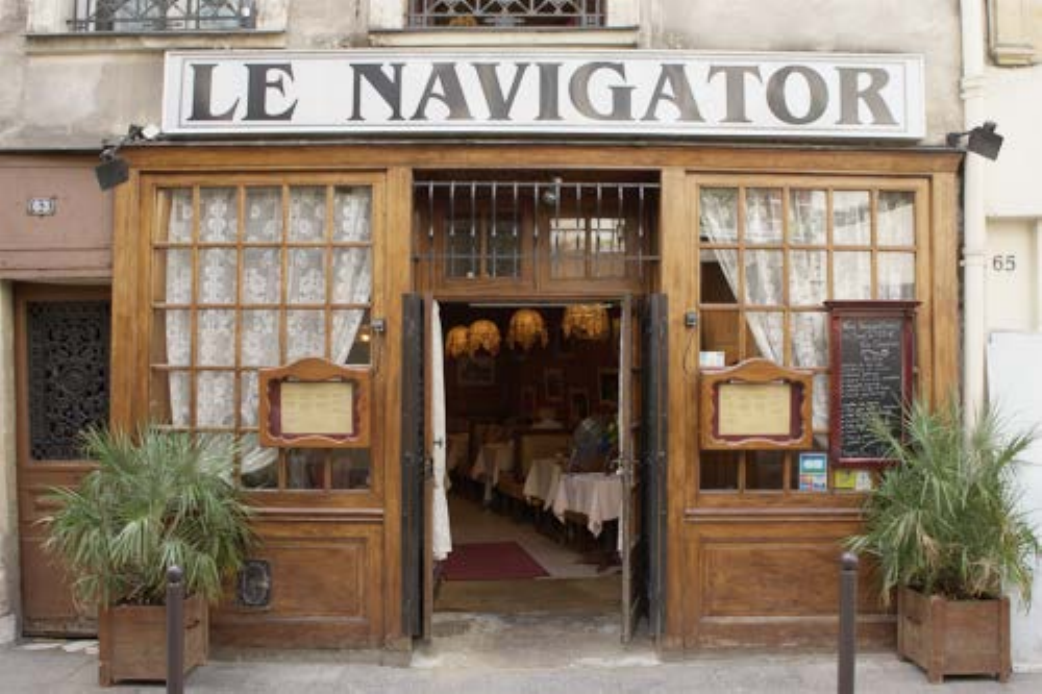}\\
		\includegraphics[width=0.195\linewidth]{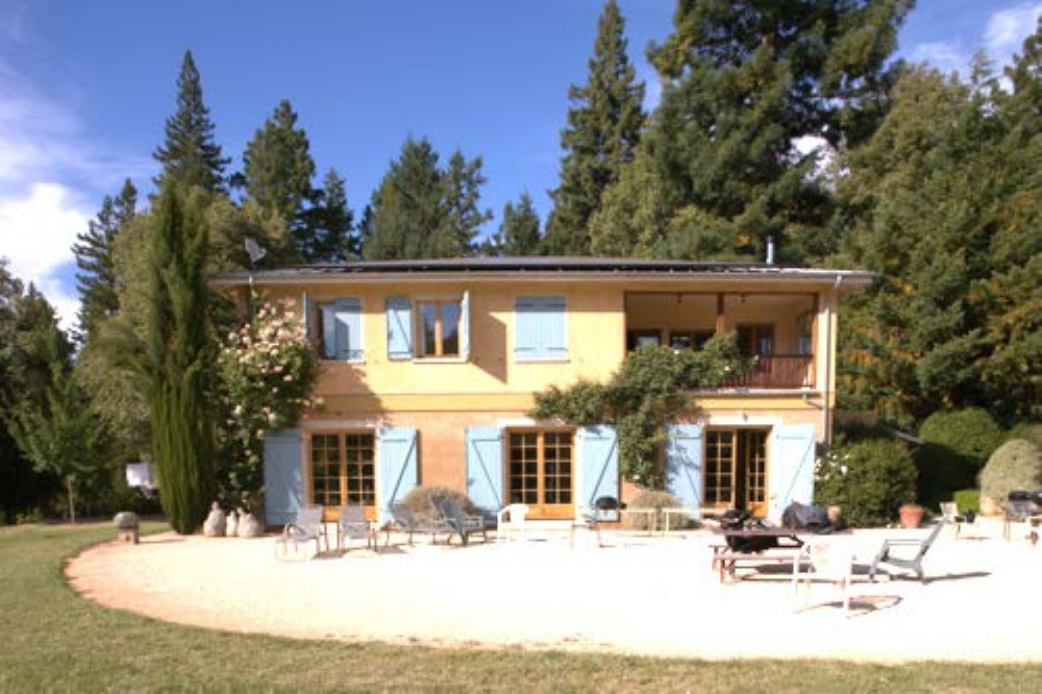}&
		\includegraphics[width=0.195\linewidth]{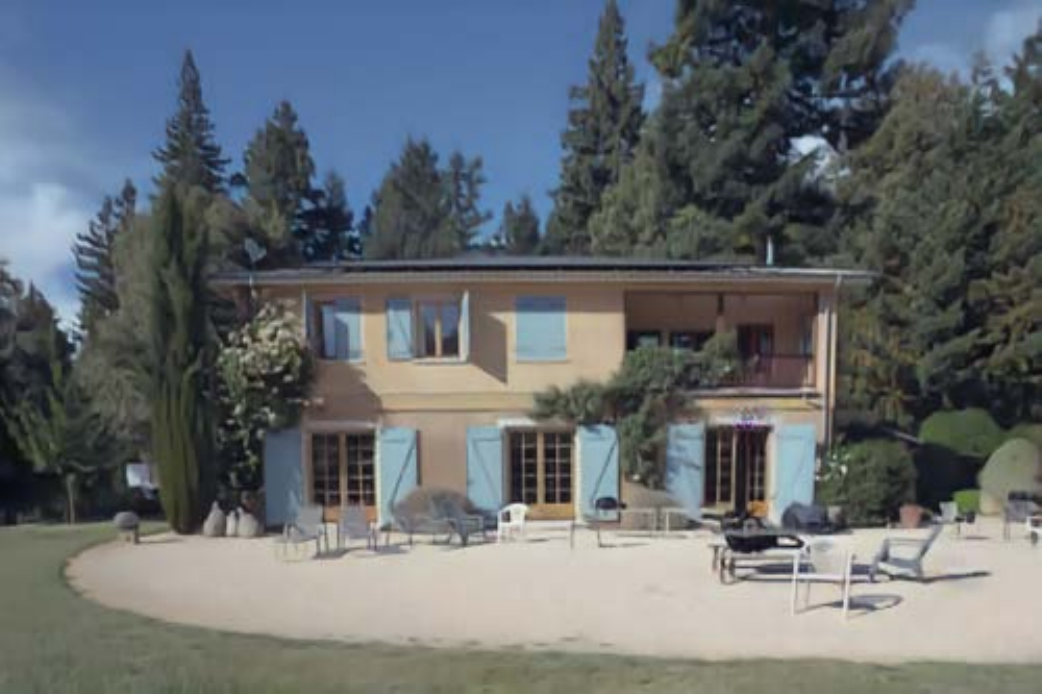}&
		\includegraphics[width=0.195\linewidth]{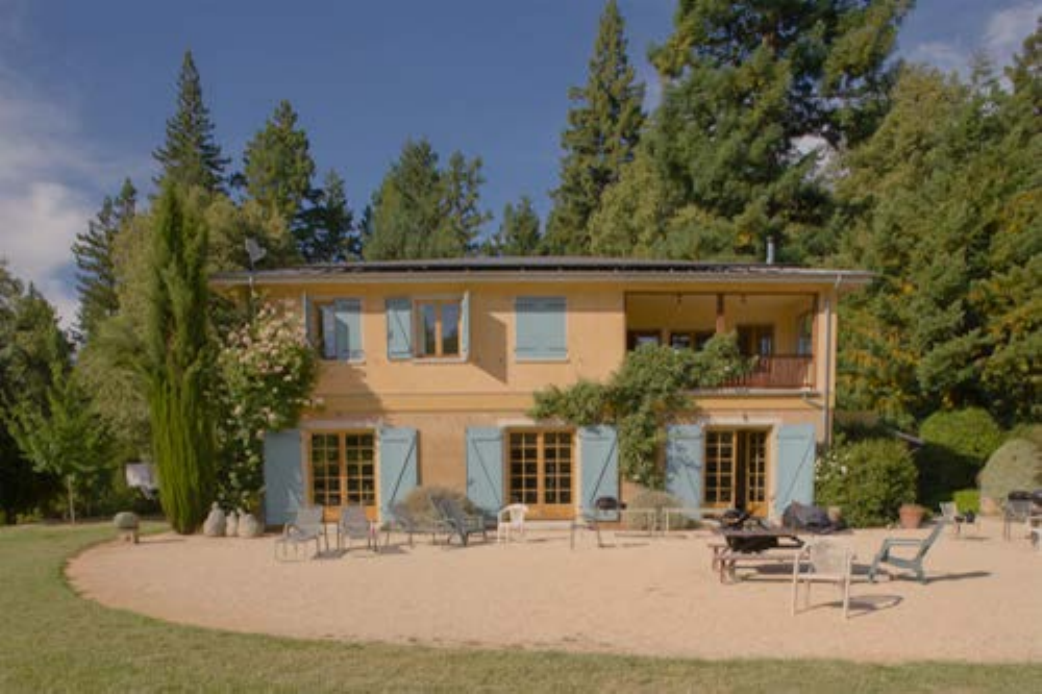}&
		\includegraphics[width=0.195\linewidth]{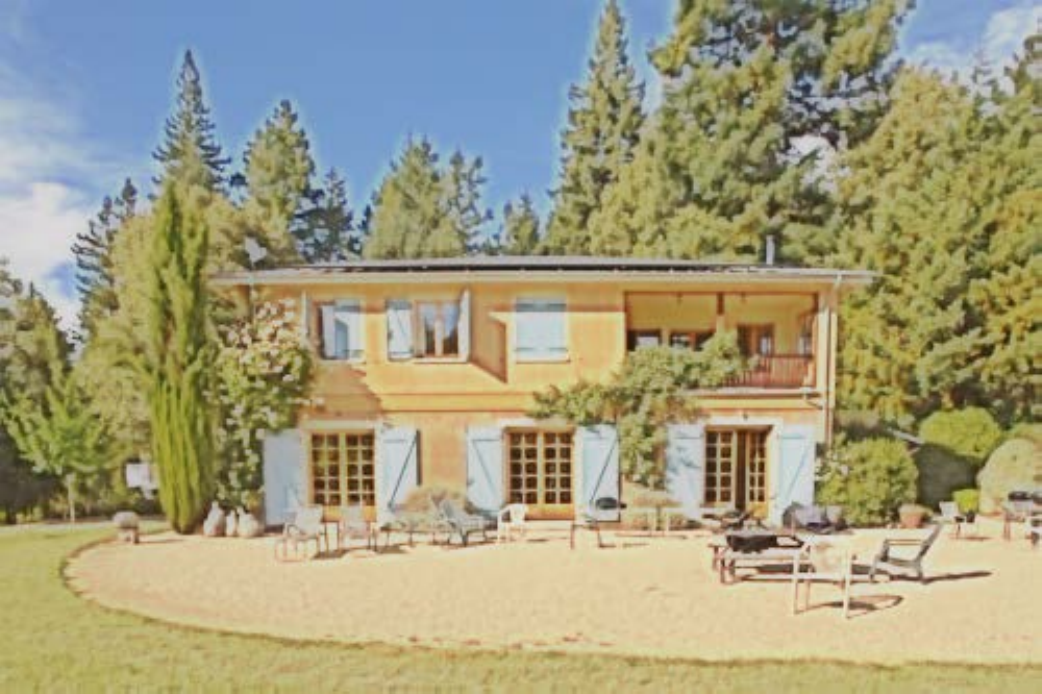}&
		\includegraphics[width=0.195\linewidth]{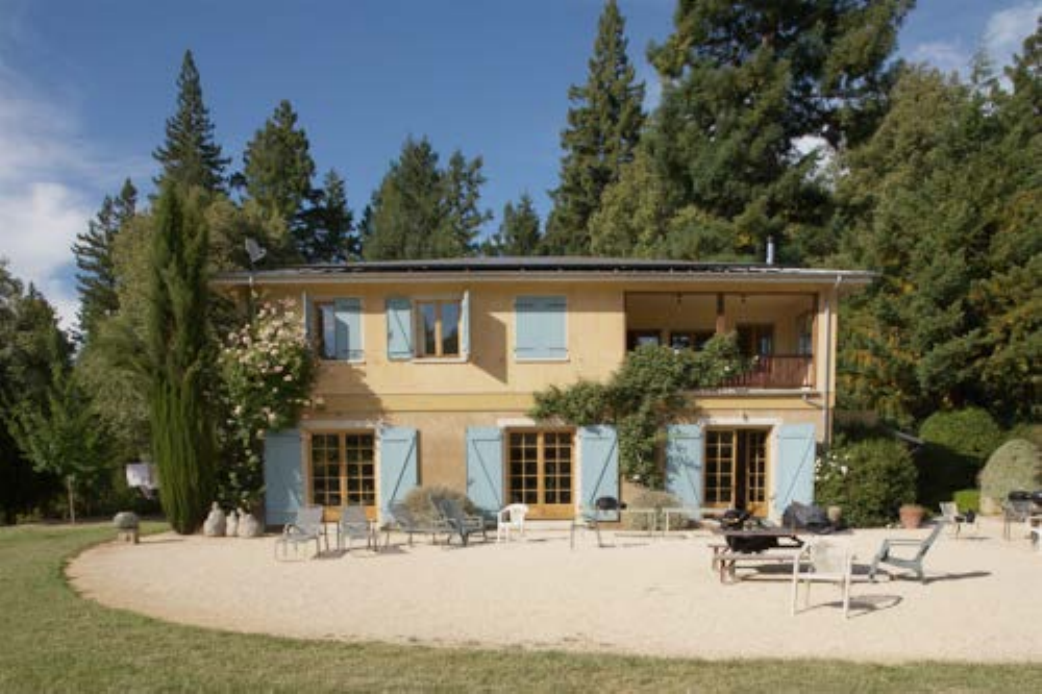}\\
		\footnotesize RUAS&\footnotesize NeRCo&\footnotesize RQ-LLIE&\footnotesize PairLIE&\footnotesize Ours\\
	\end{tabular}
	\caption{{Qualitative comparison with advanced methods of low-light image enhancement on the MIT dataset.}}
	\label{fig:mit}
\end{figure*}

\begin{table}[h]
	\centering
	\footnotesize
	\renewcommand{\arraystretch}{1.5}
	\setlength{\tabcolsep}{1pt}
	\caption{{Quantitative comparison results using LIQE and UNIQUE as metrics on the MIT dataset.}}

	\begin{tabular}{|c|ccccccc|c|}
		\hline
		{Method} &DCE&SCI&RUAS&RQ-LLIE&PairLIE&NeRCo&ZeroIG&Ours\\ \hline
		LIQE$\uparrow$&3.9134&\textcolor{blue}{\textbf{3.9483}}&3.9289&3.8134&2.8536&3.1103&2.2725&\textcolor{red}{\textbf{4.0273}}\\ 
		UNIQUE$\uparrow$&1.0974&1.1590&\textcolor{blue}{\textbf{1.2491}}&1.1086&1.1133&0.8682&0.4901&\textcolor{red}{\textbf{1.3187}}\\ \hline
	\end{tabular}
	\vspace{-1mm}
	\label{table: LIQE}
\end{table}

	 \begin{table}[h]
	\centering
	\footnotesize
	\setlength{\tabcolsep}{1.2mm}
	\renewcommand\arraystretch{1.5}	
	\caption{{Quantitative comparison with representative low-light image enhancement methods on the MIT dataset.}}
	\begin{tabular}{|c||cccc||c|}
		\hline
		Method &LLFormer&LightenDiffusion&GSAD&DiffLL&{Ours}\\ \hline
		PSNR$\uparrow$&16.7634&\textcolor{blue}{\textbf{19.8570}}&12.9437&16.9133&\textcolor{red}{\textbf{24.0157}}\\ 
		SSIM$\uparrow$&0.7120&\textcolor{blue}{\textbf{0.7571}}&0.6093&0.7064&\textcolor{red}{\textbf{0.8571}}\\
		LOE$\downarrow$&290.1542&\textcolor{blue}{\textbf{267.6065}}&244.1709&421.5476&\textcolor{red}{\textbf{67.897}}\\ \hline
	\end{tabular}
	\vspace{-3mm}
	\label{table: new}
\end{table}

\section{Experimental Results}\label{sec:exp}

In this section, we first introduce the experimental implementation details. We then compare the performance of AR-LLIE and other advanced  low-light enhancers across various low-light scenes of differing difficulty levels for a comprehensive evaluation. To provide a more thorough comparison, we further present visual results and user surveys on several other small-scale low-light datasets. Additionally, we compare the computational efficiency of each method on personal computers and mobile devices with different computing capabilities, which are often overlooked by most methods. Finally, we conduct a series of analytical experiments on the proposed automatic re-parameterization algorithm to demonstrate the effectiveness of our approach.

\subsection{Implementation Details}

For delving into the limits, we define a simple convolution (with a LeakyReLU activation function) as the default architecture (after re-parameterization) for our AR-LLIE. In the searching process, we calculated the gradients of the architecture and weight parameters following the proposed tiered NAS methods. As for the data setup, we conducted searches on the MIT, LOL, and DARKFACE datasets. We randomly selected 100 and 200 images from each dataset for the search phase, and then selected 300 and 400 images from the remaining data for subsequent training and testing. During the search and training processes, the images were kept at their original sizes, with a batch size of 2. The number of epochs for both search and training was set to 100. We adopt the ADAM optimizer with parameters $\beta_1=0.9$, $\beta_2=0.999$, and $\epsilon=10^{-8}$ for both searching and training. Note that all the experiments are implemented with PyTorch framework on a NVIDIA RTX 2080Ti GPU.

\subsection{Performance Evaluation}
In this section, we conduct a series of quantitative and qualitative comparisons between the proposed AR-LLIE and 10 representative advanced low-light image enhancement methods, including FIDE~\cite{Xu_2020_CVPR}, KinD~\cite{zhang2019kindling}, UTVNet~\cite{Zheng_2021_ICCV}, RetinexNet~\cite{wei2018deep}, DCE~\cite{Zero-DCE++}, SCI~\cite{ma2022toward}, RUAS~\cite{liu2022learning}, RQ-LLIE~\cite{liu2023low}, PairLIE~\cite{fu2023learning}, NeRCo~\cite{yang2023implicit}, and ZeroIG~\cite{shi2024zero}. {For the benchmarks, we consider three widely used datasets with varying difficulty levels: {the MIT dataset~\cite{bychkovsky2011learning} for regular difficulty (moderate darkness with no noise interference), the challenging LOL dataset~\cite{wei2018deep} (deeper darkness with significant noise) , DARKFACE dataset~\cite{yang2020advancing} (extremely severe darkness), and BAID dataset~\cite{lv2022backlitnet} (contains images with regions of varying brightness levels).} Regarding metrics, we utilize two full-reference metrics: PSNR (Peak Signal-to-Noise Ratio), SSIM (Structural Similarity Index), one no-reference metric: NIQE (Naturalness Image Quality Evaluator), and LOE (Lightness-Order-Error).}

	\begin{figure*}[htb!]
		\centering
		\begin{tabular}{c@{\extracolsep{0.2em}}c@{\extracolsep{0.2em}}c@{\extracolsep{0.2em}}c@{\extracolsep{0.2em}}c}
			\includegraphics[width=0.195\linewidth]{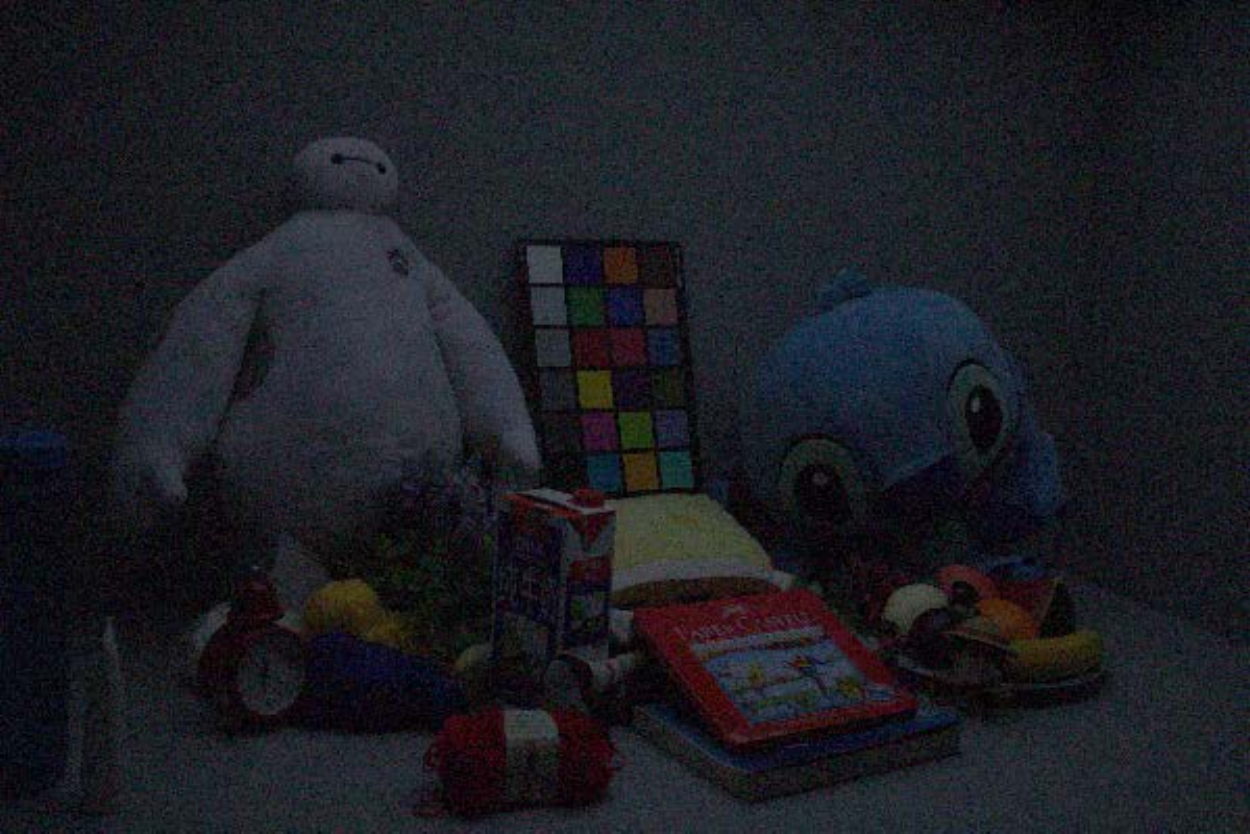}&
			\includegraphics[width=0.195\linewidth]{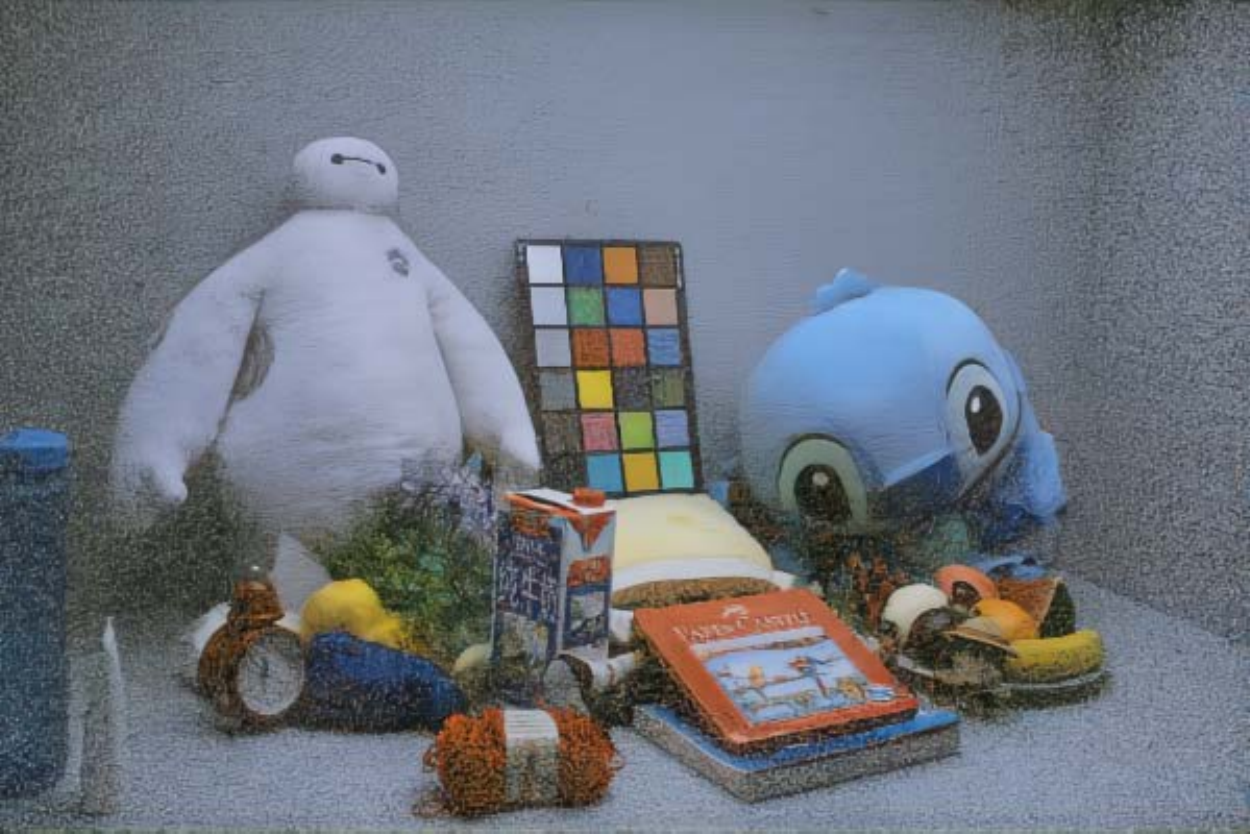}&
			\includegraphics[width=0.195\linewidth]{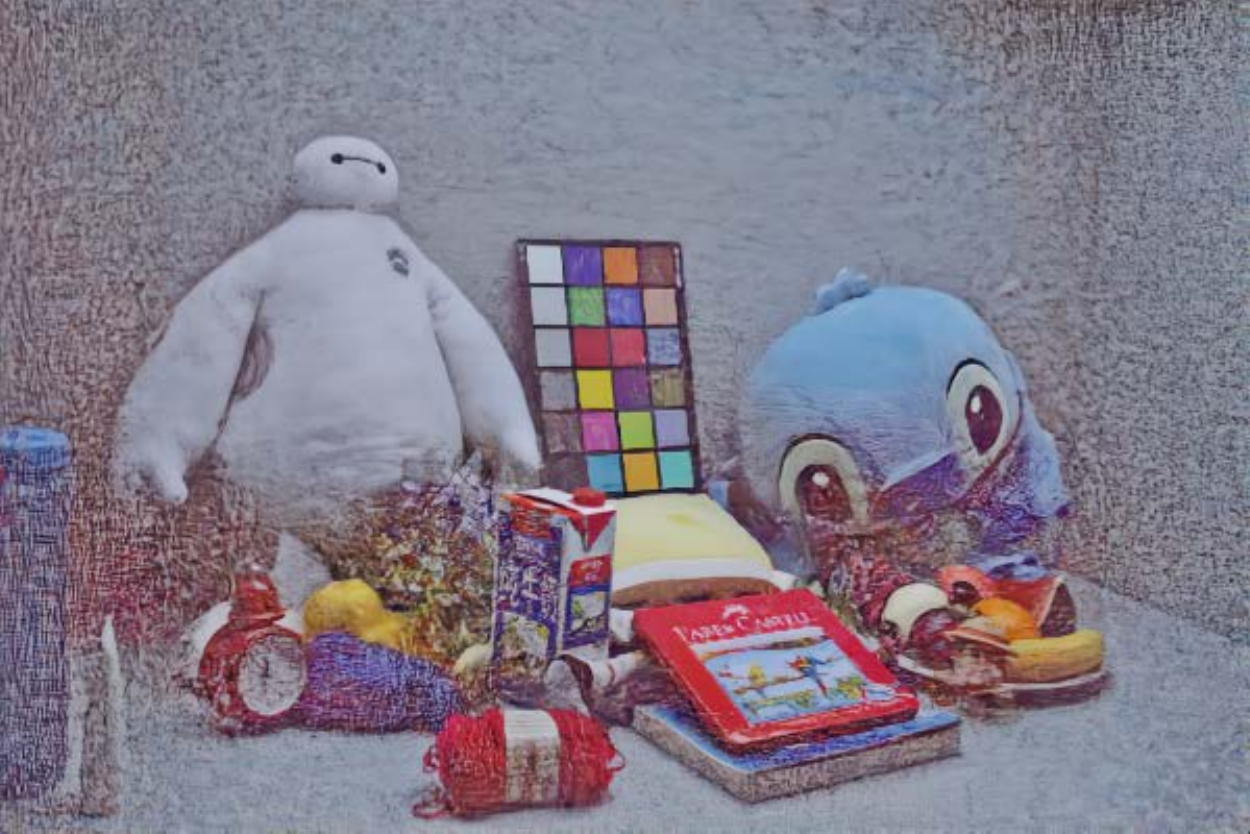}&
			\includegraphics[width=0.195\linewidth]{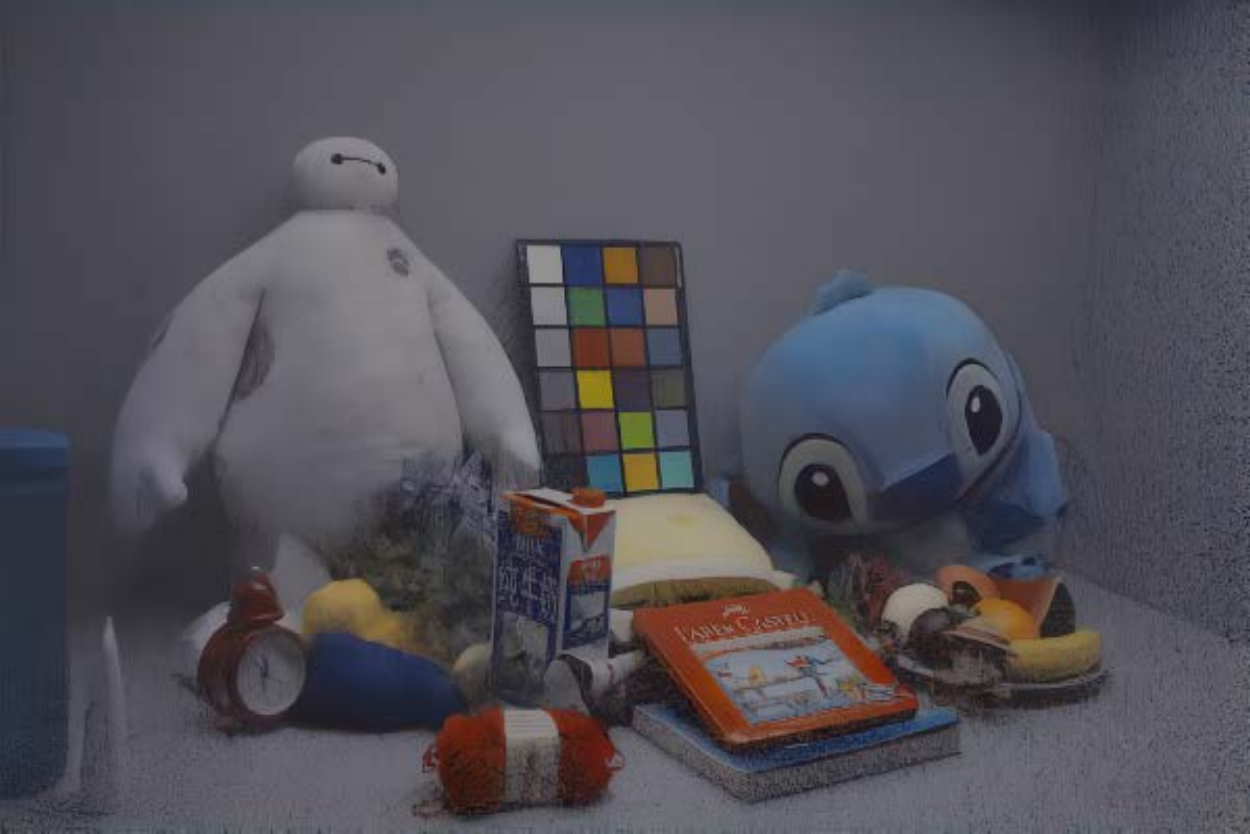}&
			\includegraphics[width=0.195\linewidth]{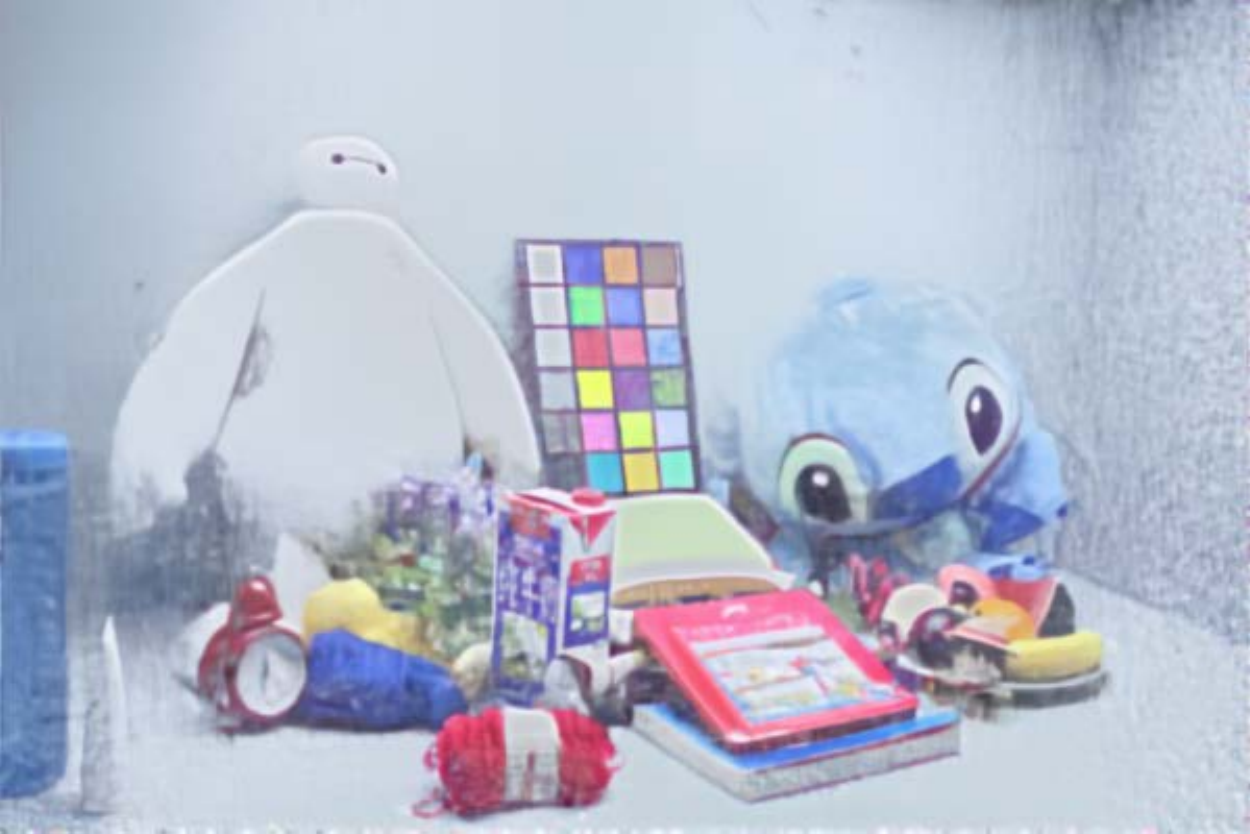}\\
			\includegraphics[width=0.195\linewidth]{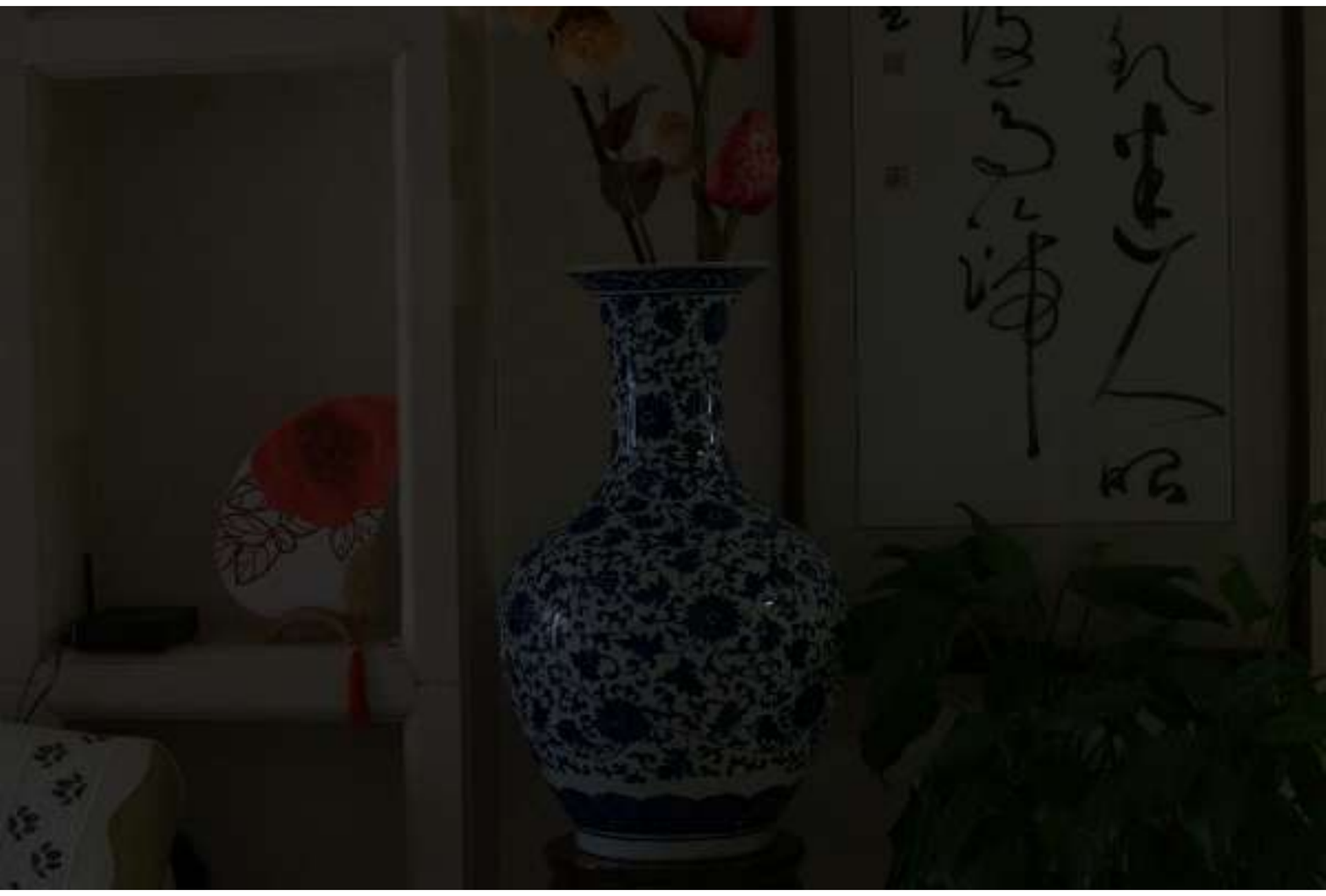}&
			\includegraphics[width=0.195\linewidth]{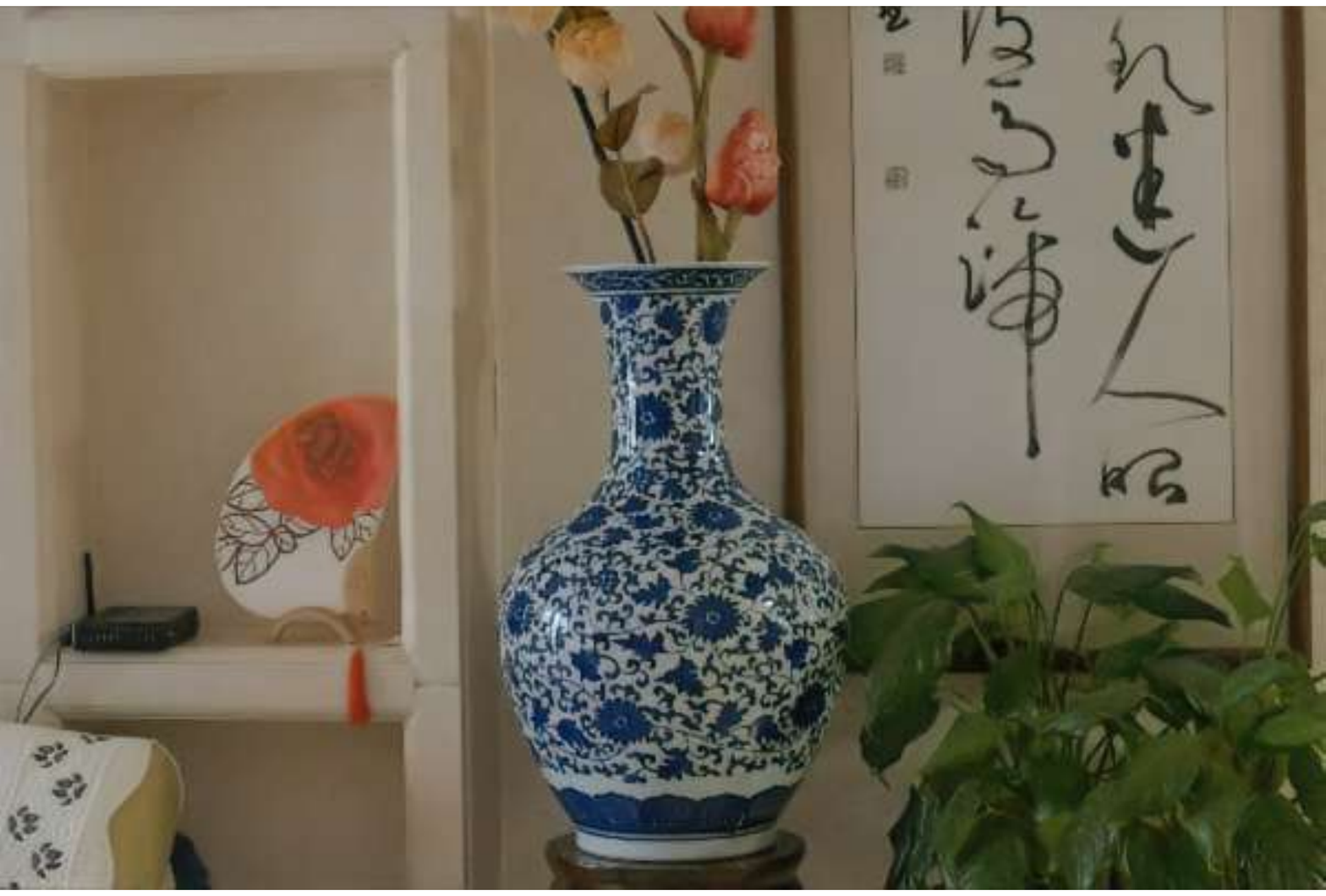}&
			\includegraphics[width=0.195\linewidth]{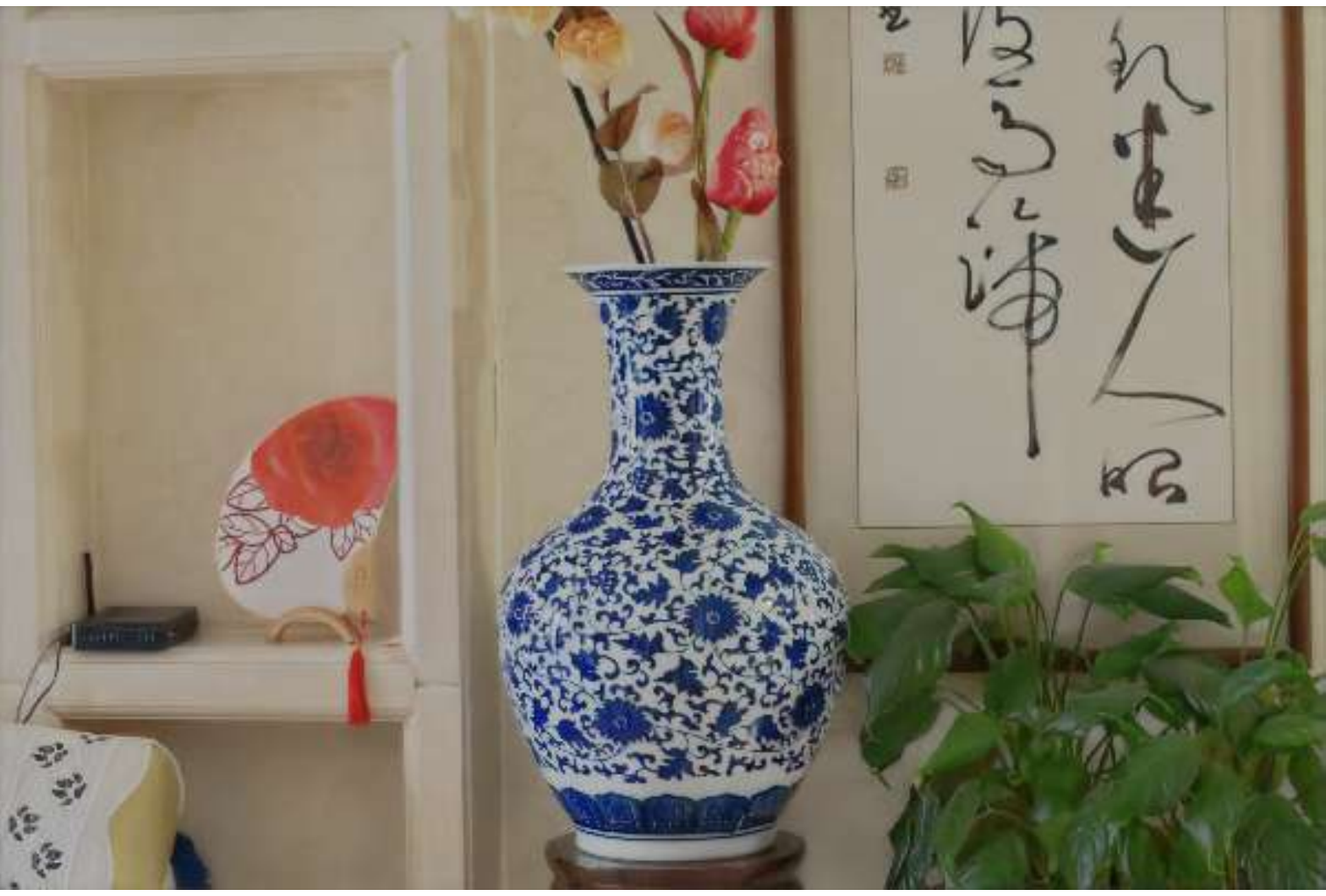}&
			\includegraphics[width=0.195\linewidth]{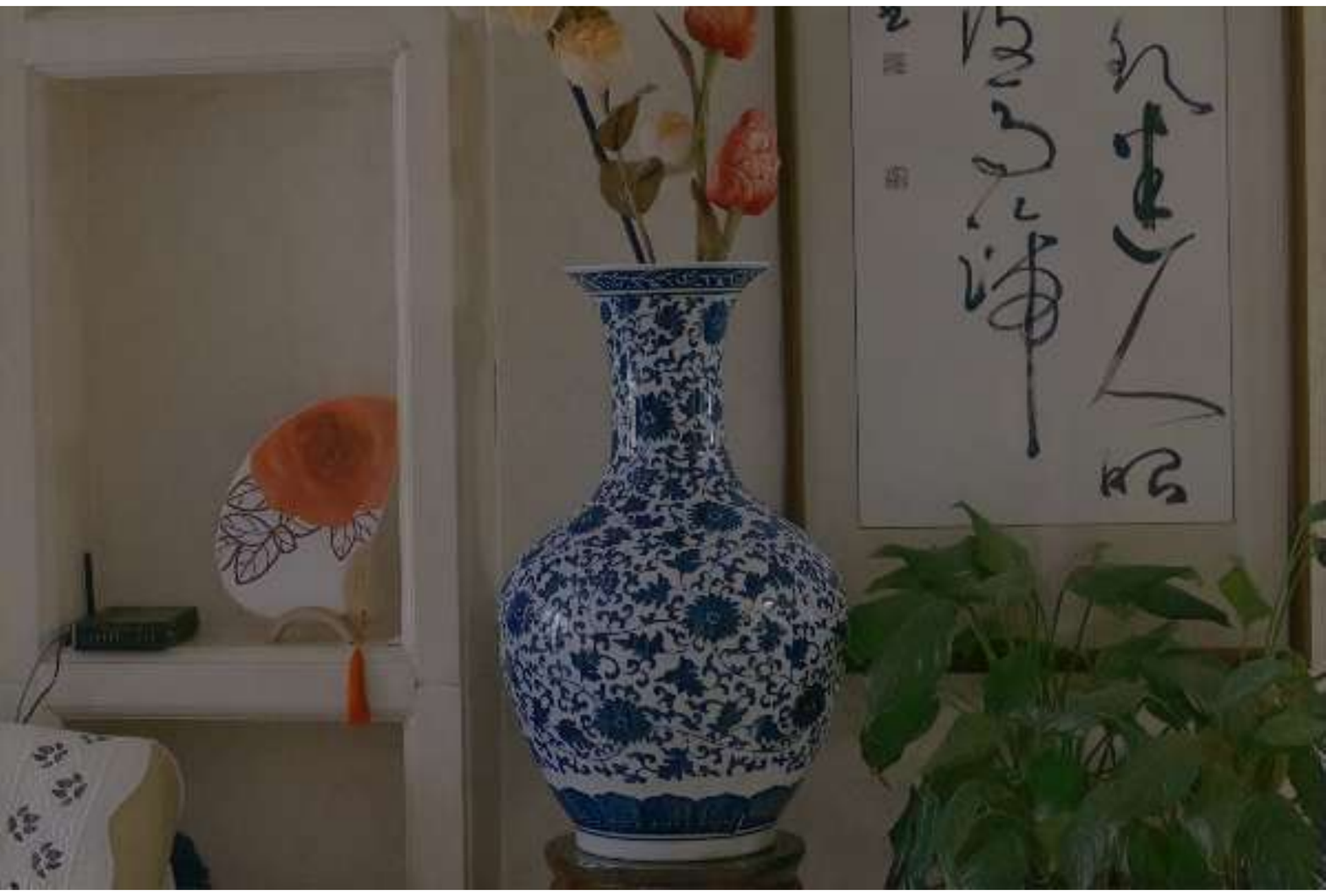}&
			\includegraphics[width=0.195\linewidth]{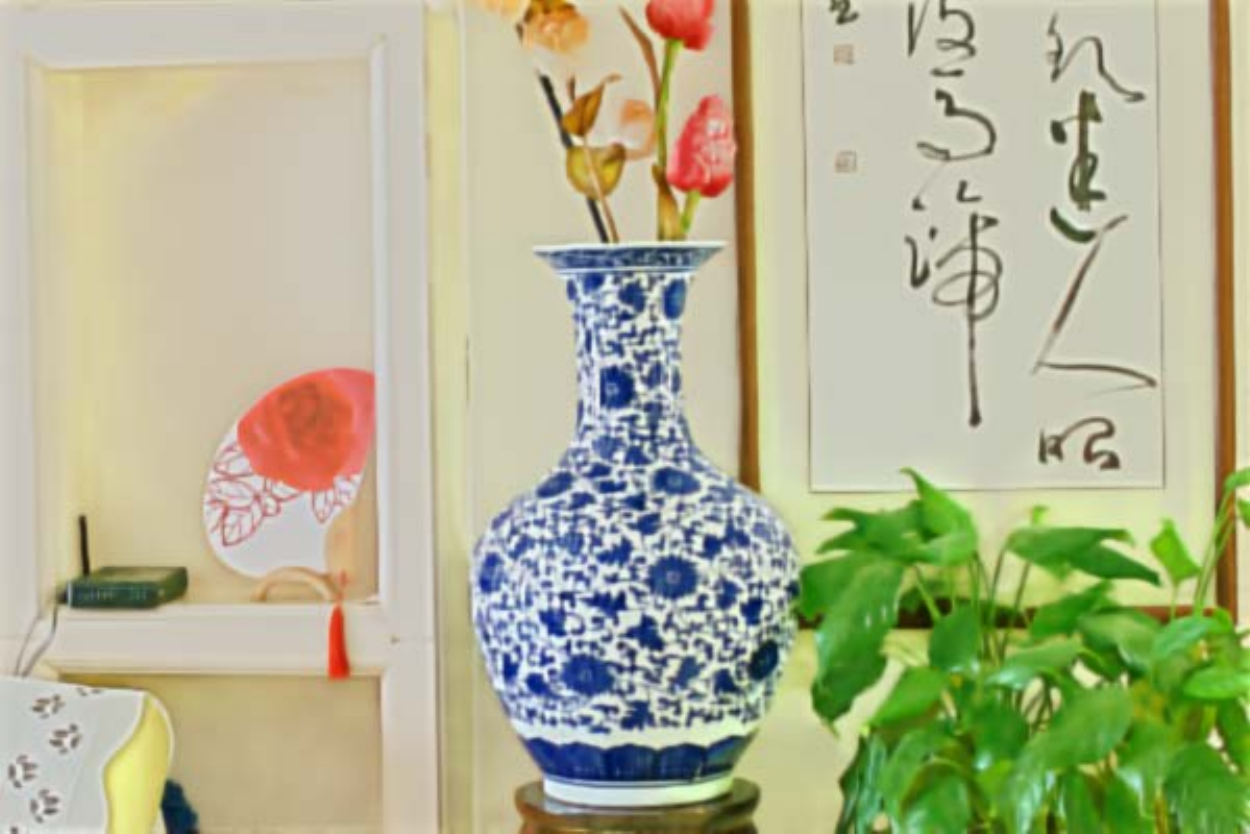}\\
			\footnotesize Input&\footnotesize FIDE&\footnotesize KinD&\footnotesize UTVNet&\footnotesize ZeroIG\\
			\includegraphics[width=0.195\linewidth]{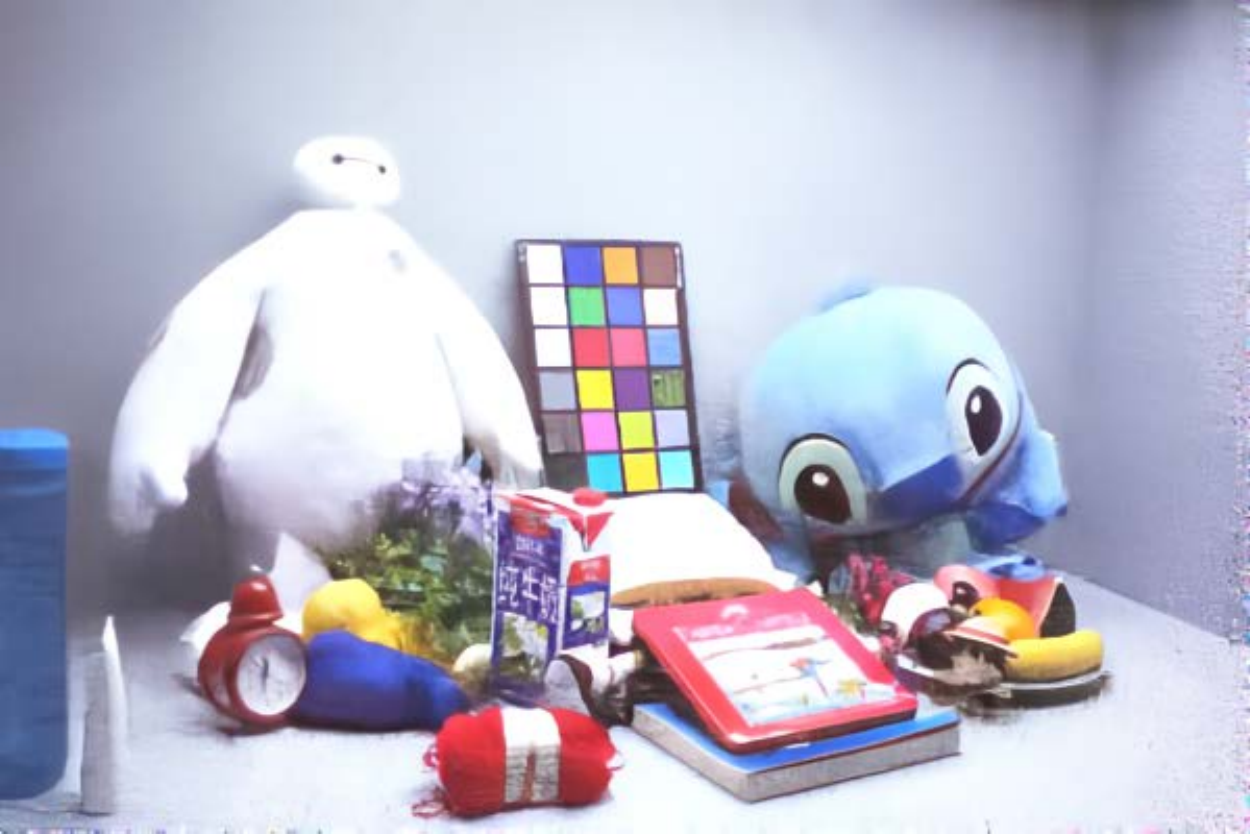}&
			\includegraphics[width=0.195\linewidth]{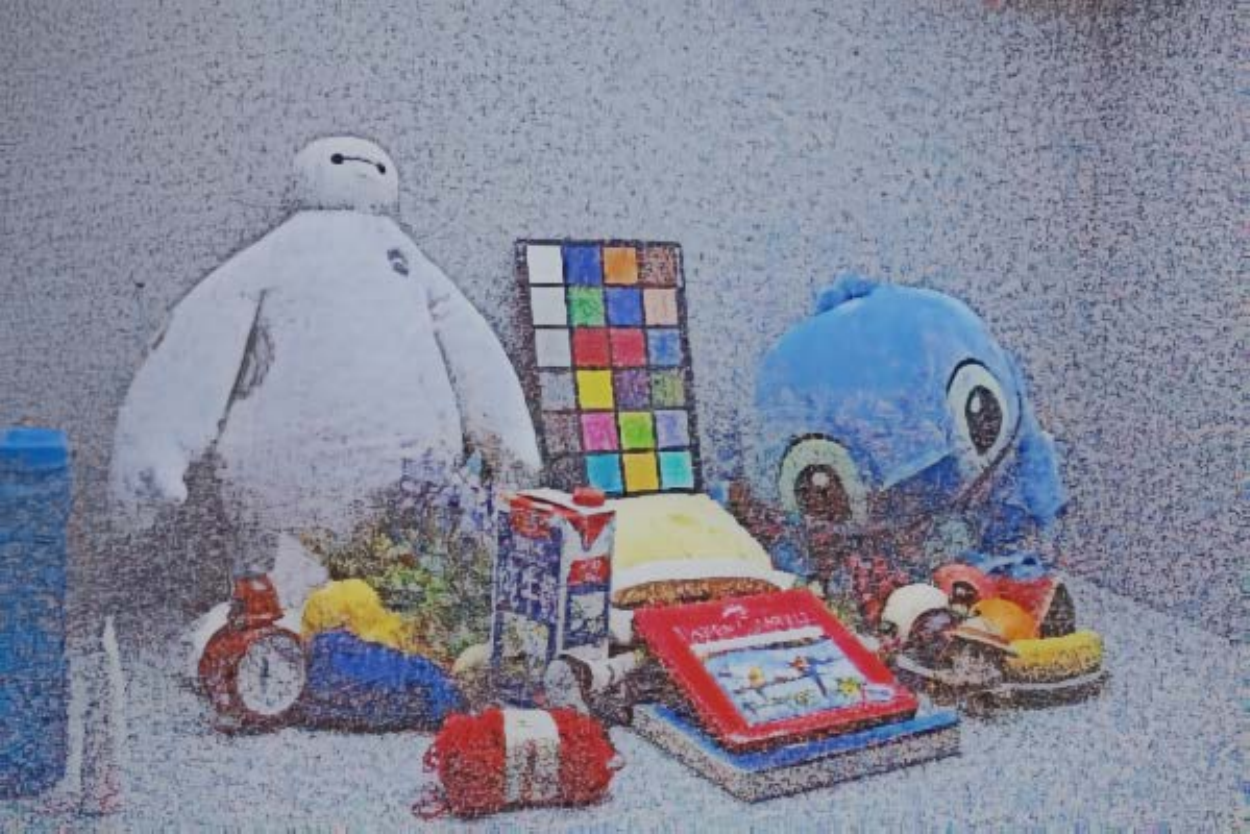}&
			\includegraphics[width=0.195\linewidth]{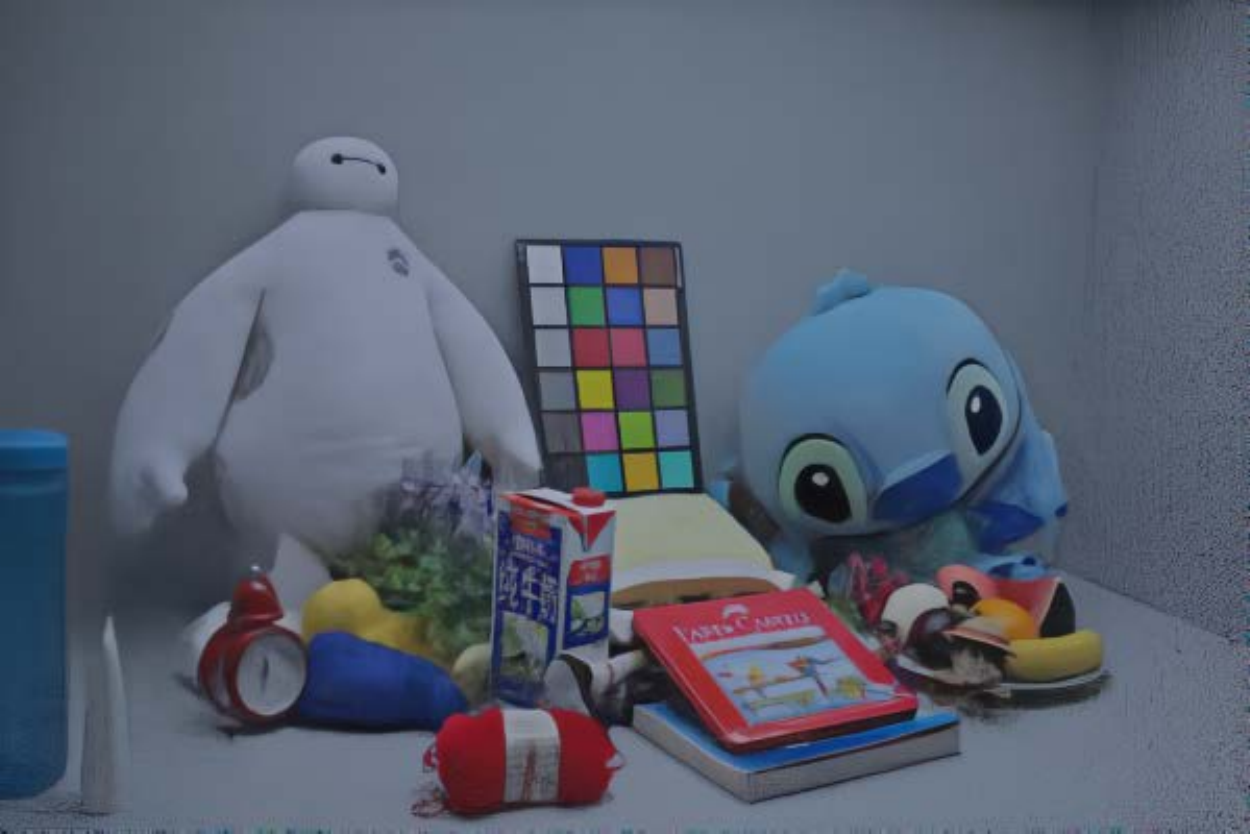}&
			\includegraphics[width=0.195\linewidth]{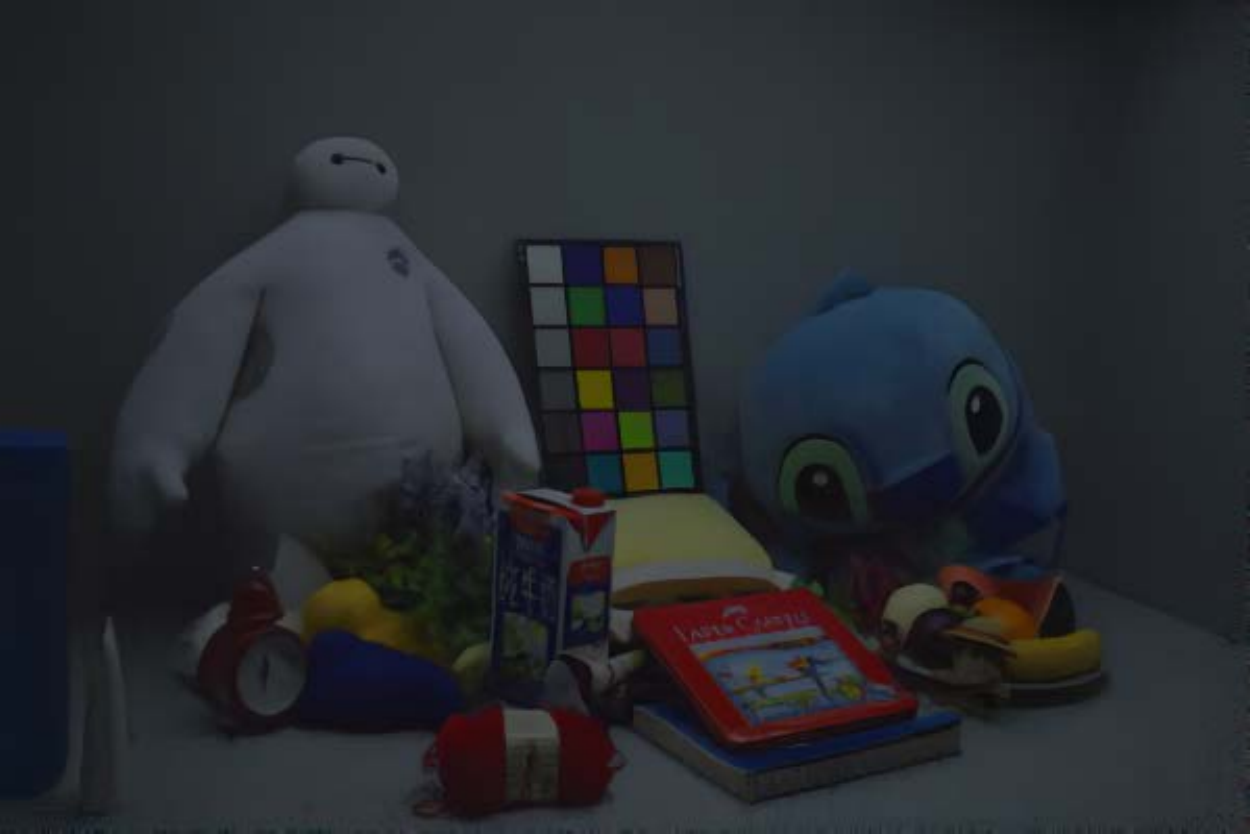}&
			\includegraphics[width=0.195\linewidth]{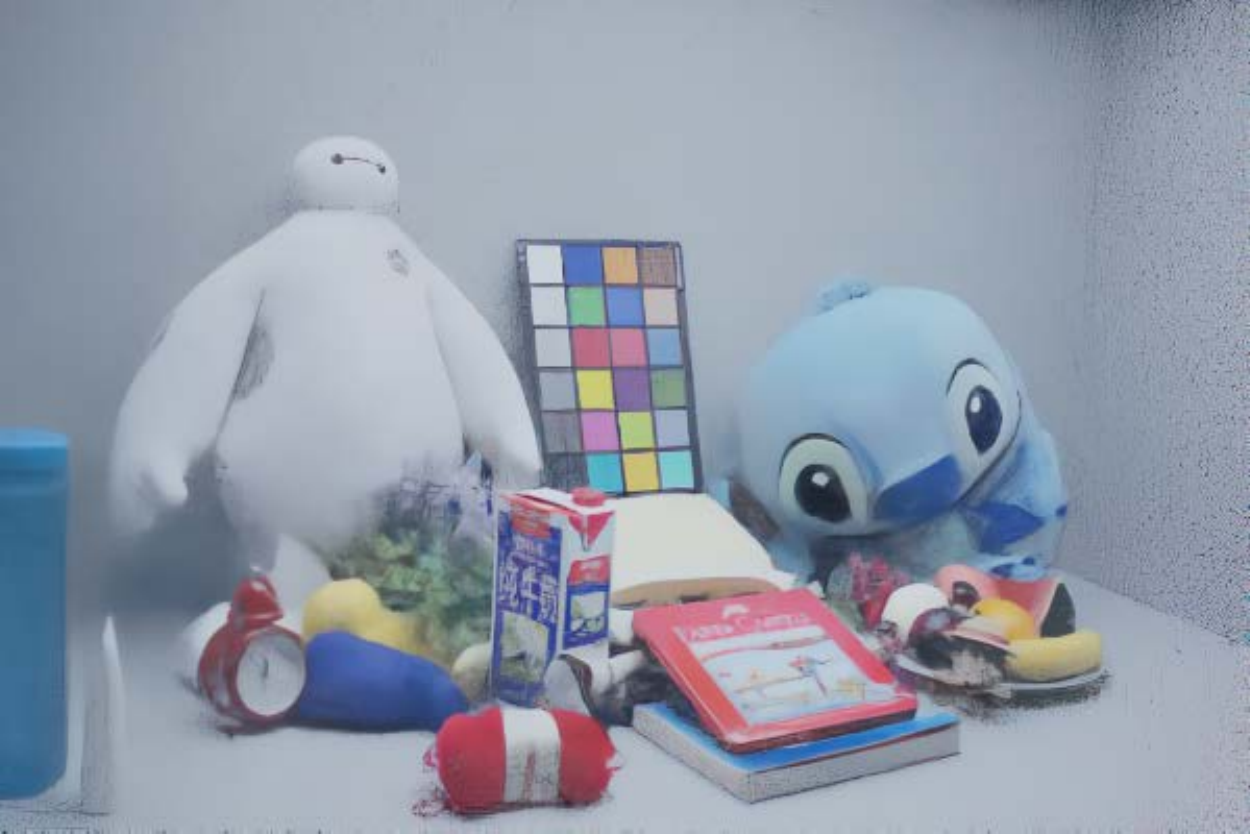}\\
			\includegraphics[width=0.195\linewidth]{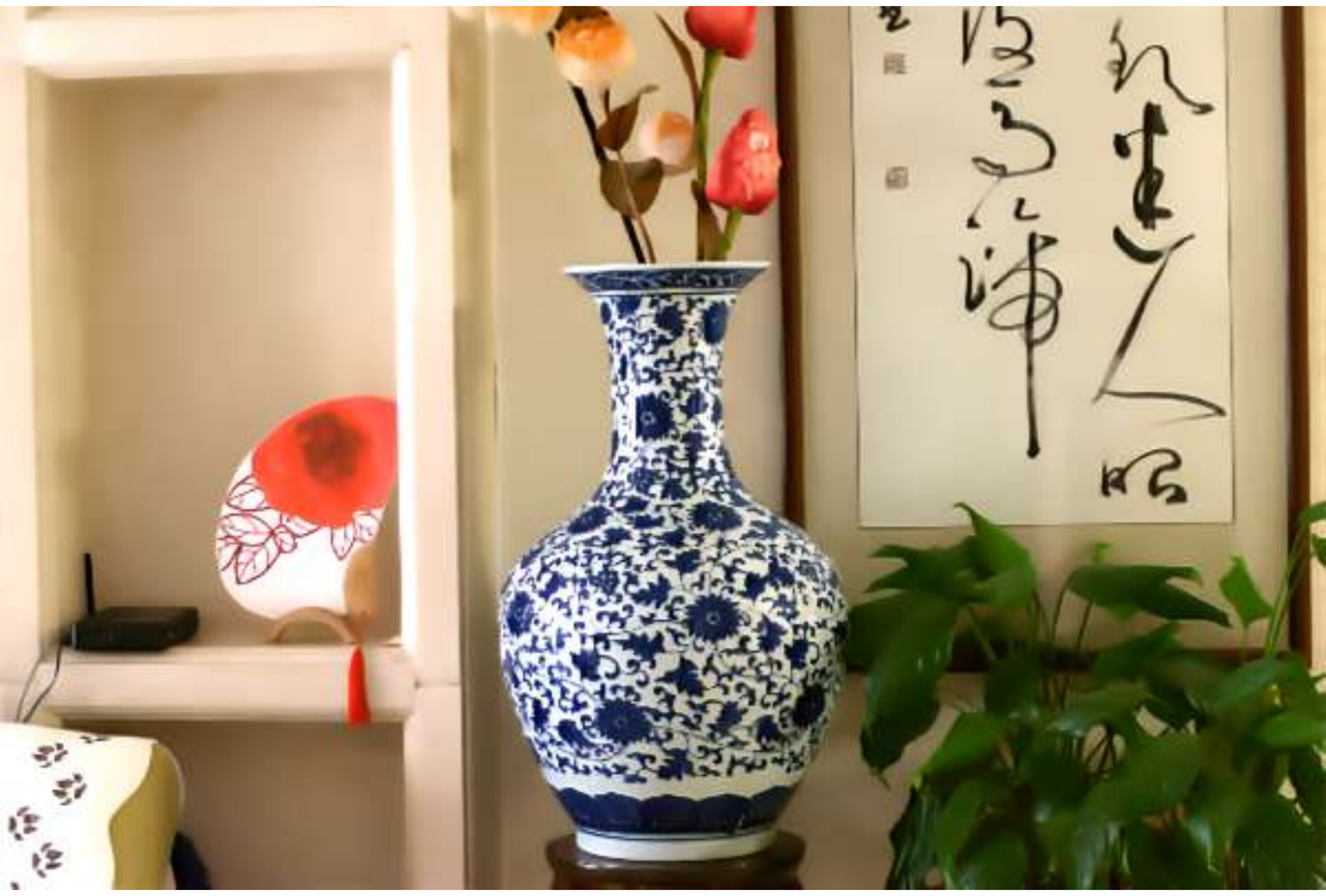}&
			\includegraphics[width=0.195\linewidth]{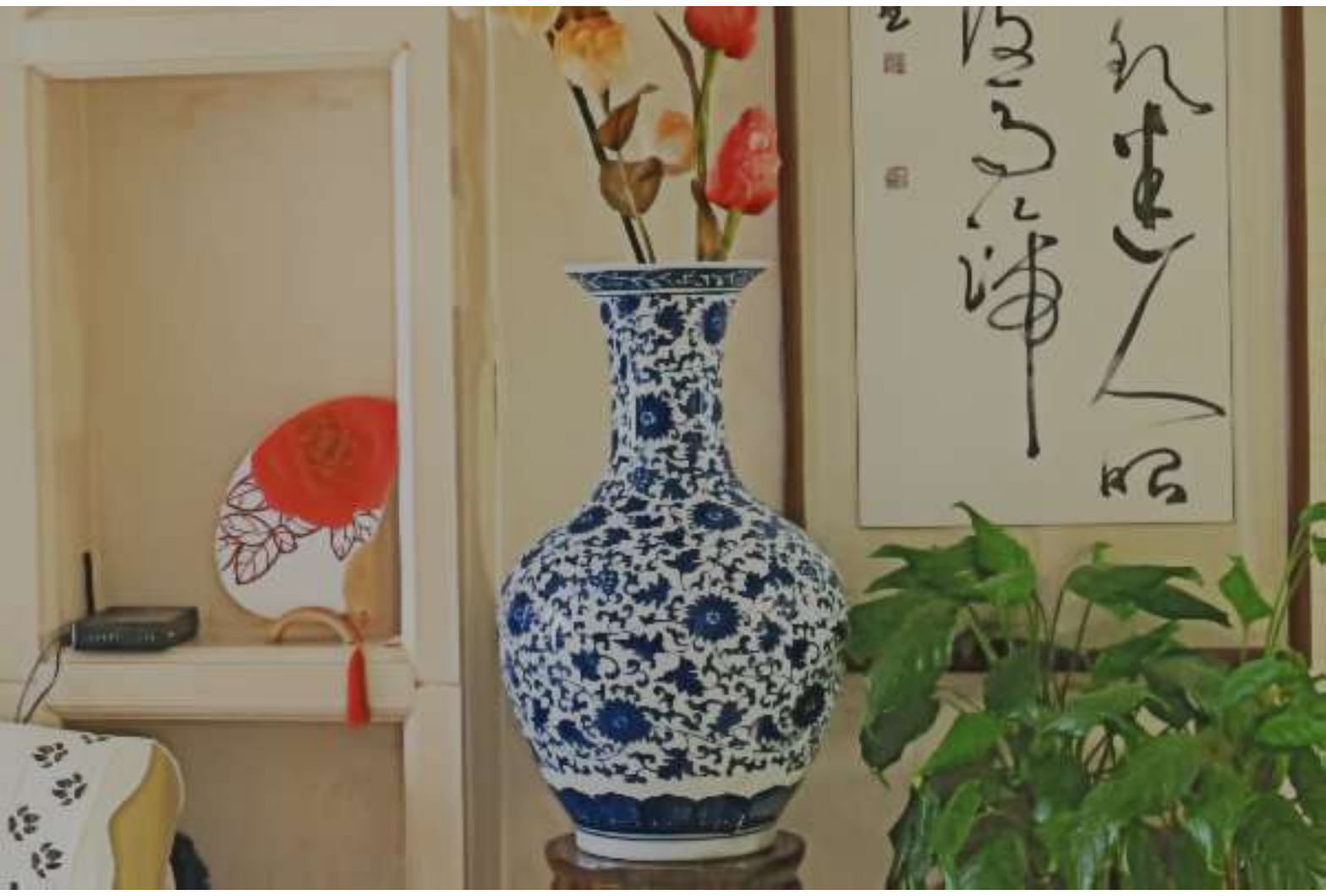}&
			\includegraphics[width=0.195\linewidth]{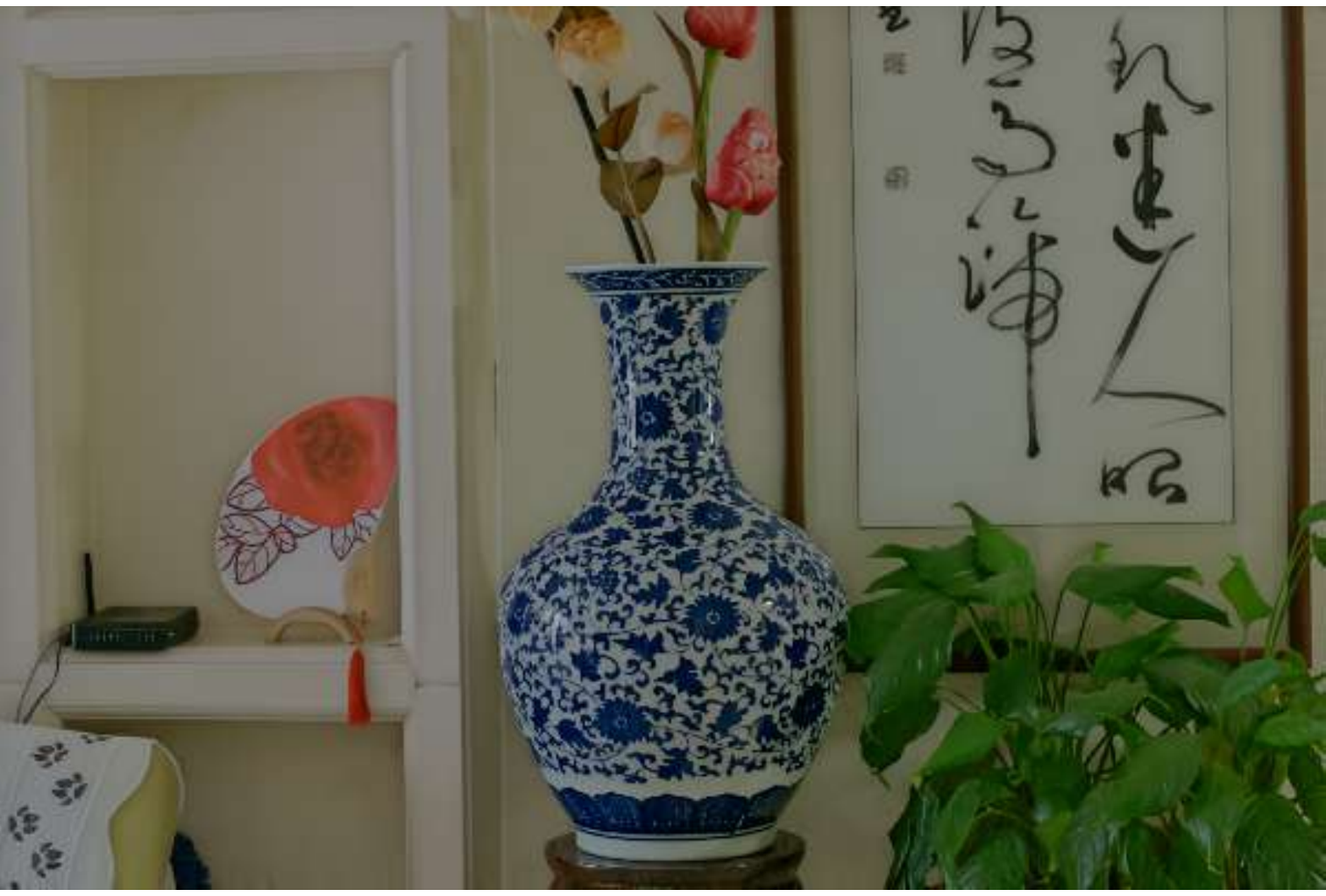}&
			\includegraphics[width=0.195\linewidth]{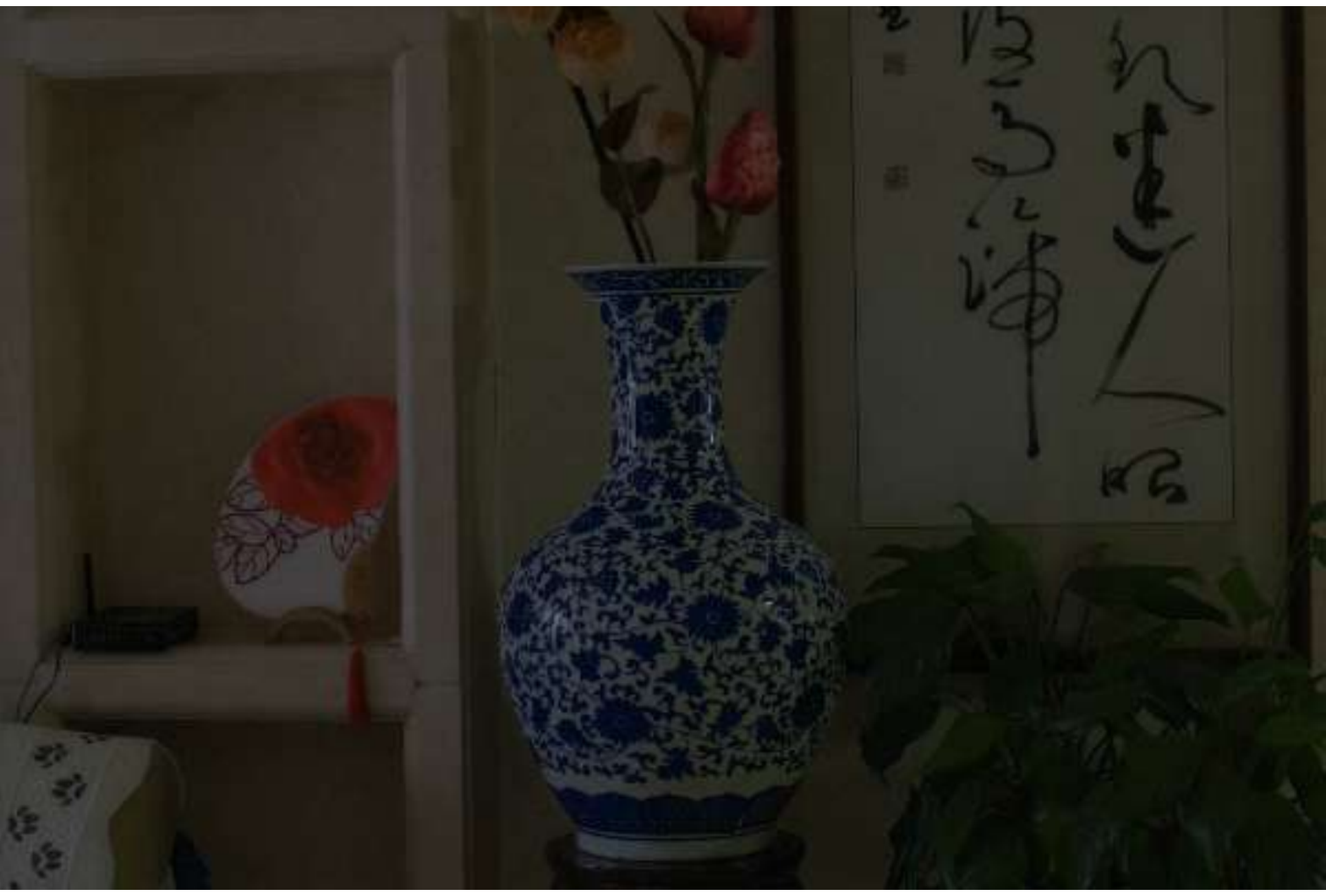}&
			\includegraphics[width=0.195\linewidth]{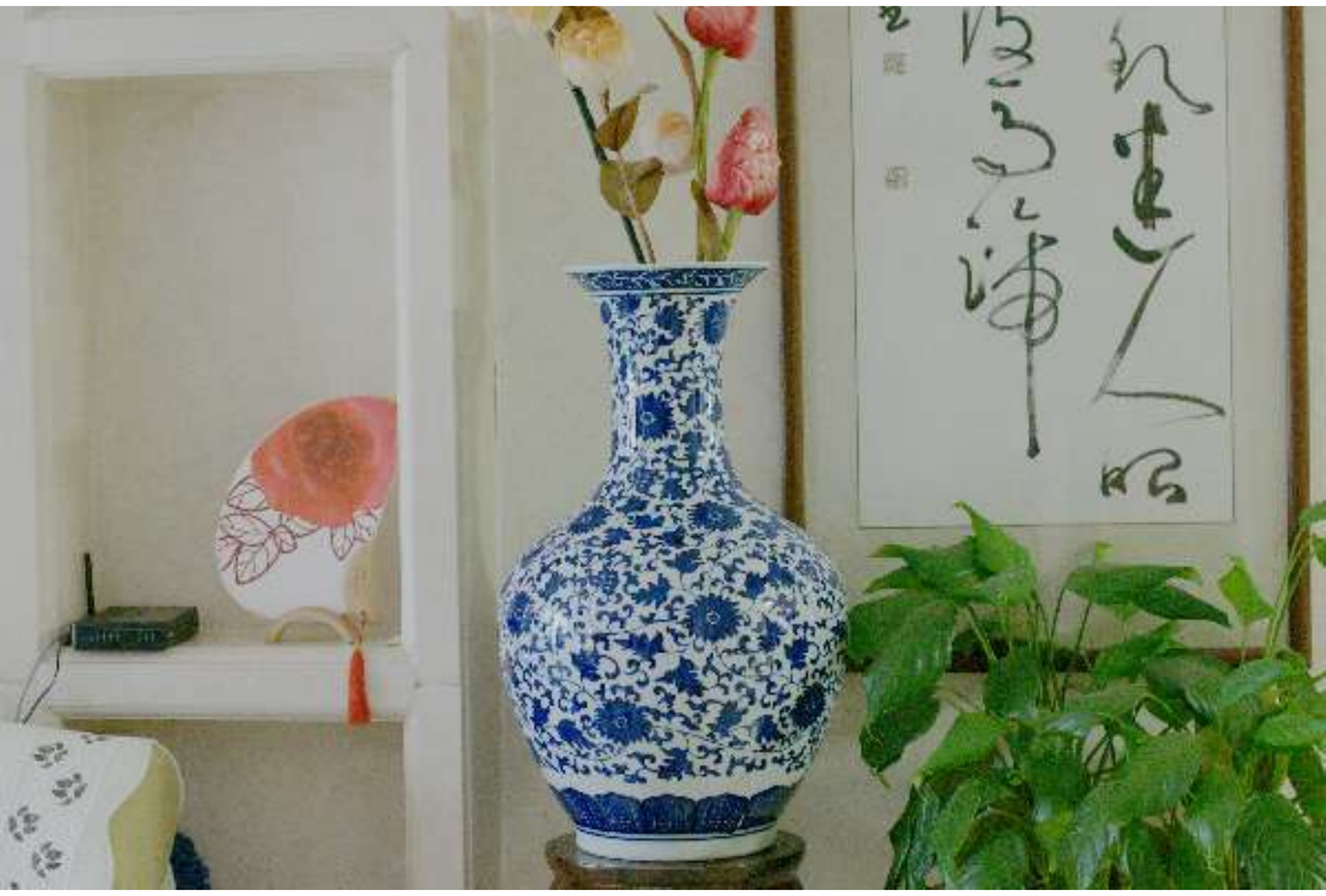}\\
			\footnotesize RUAS&\footnotesize PairLIE&\footnotesize DCE&\footnotesize RQ-LLIE&\footnotesize Ours\\
			
		\end{tabular}
		\caption{{Qualitative comparison with advanced methods of low-light image enhancement on samples with significant noise in LOL dataset.}}
		\label{fig:lol2}
	\end{figure*}
	\begin{figure*}[htb!]
		\centering
		\begin{tabular}{c@{\extracolsep{0.2em}}c@{\extracolsep{0.2em}}c@{\extracolsep{0.2em}}c@{\extracolsep{0.2em}}c@{\extracolsep{0.2em}}c}
			\includegraphics[width=0.155\linewidth]{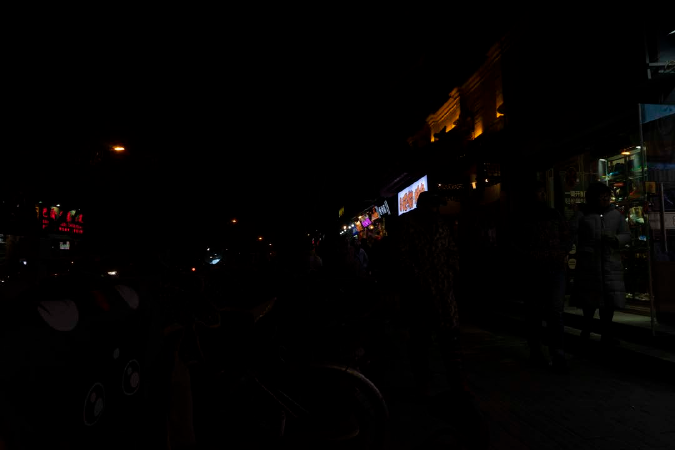}&
			\includegraphics[width=0.155\linewidth]{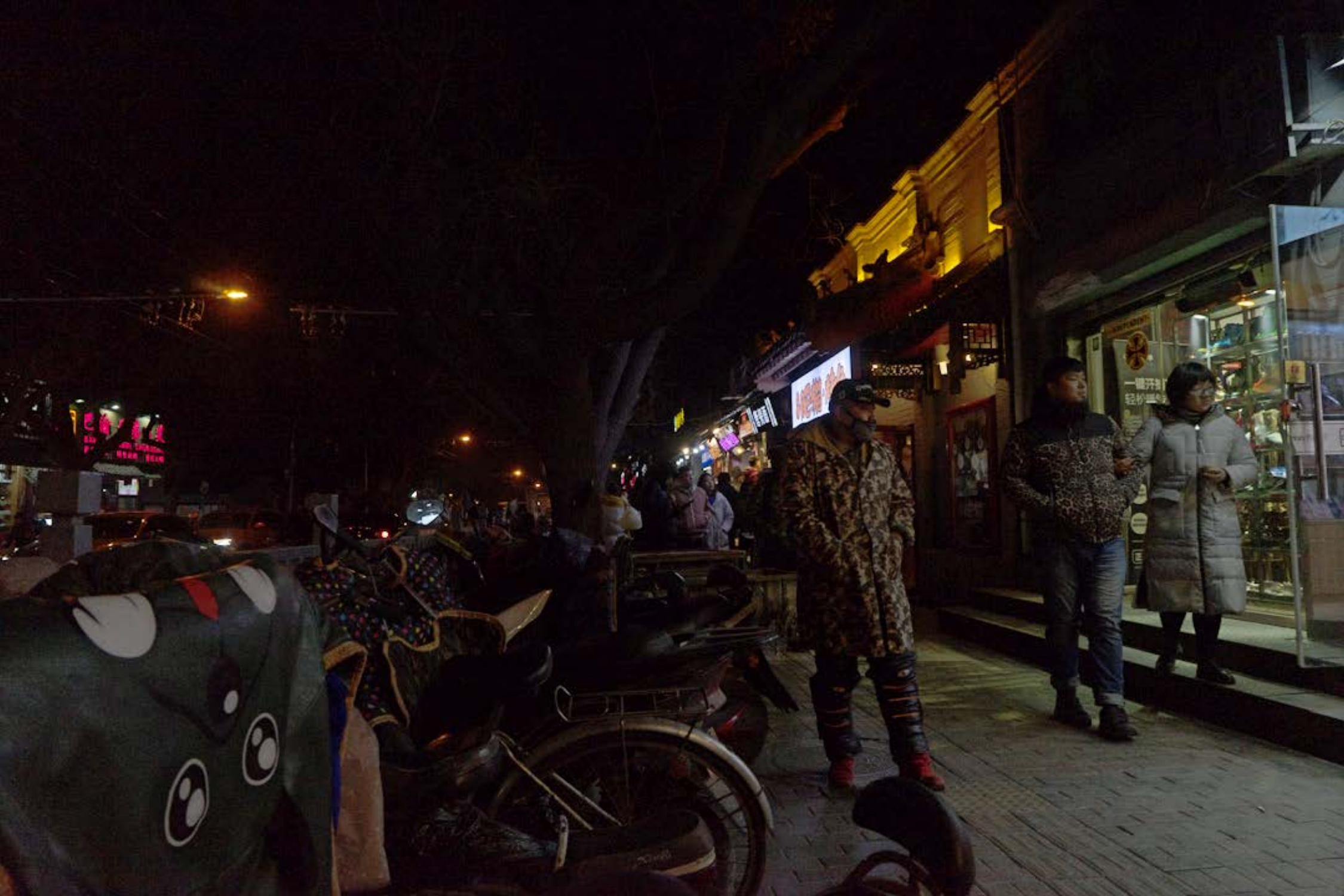}&
			\includegraphics[width=0.155\linewidth]{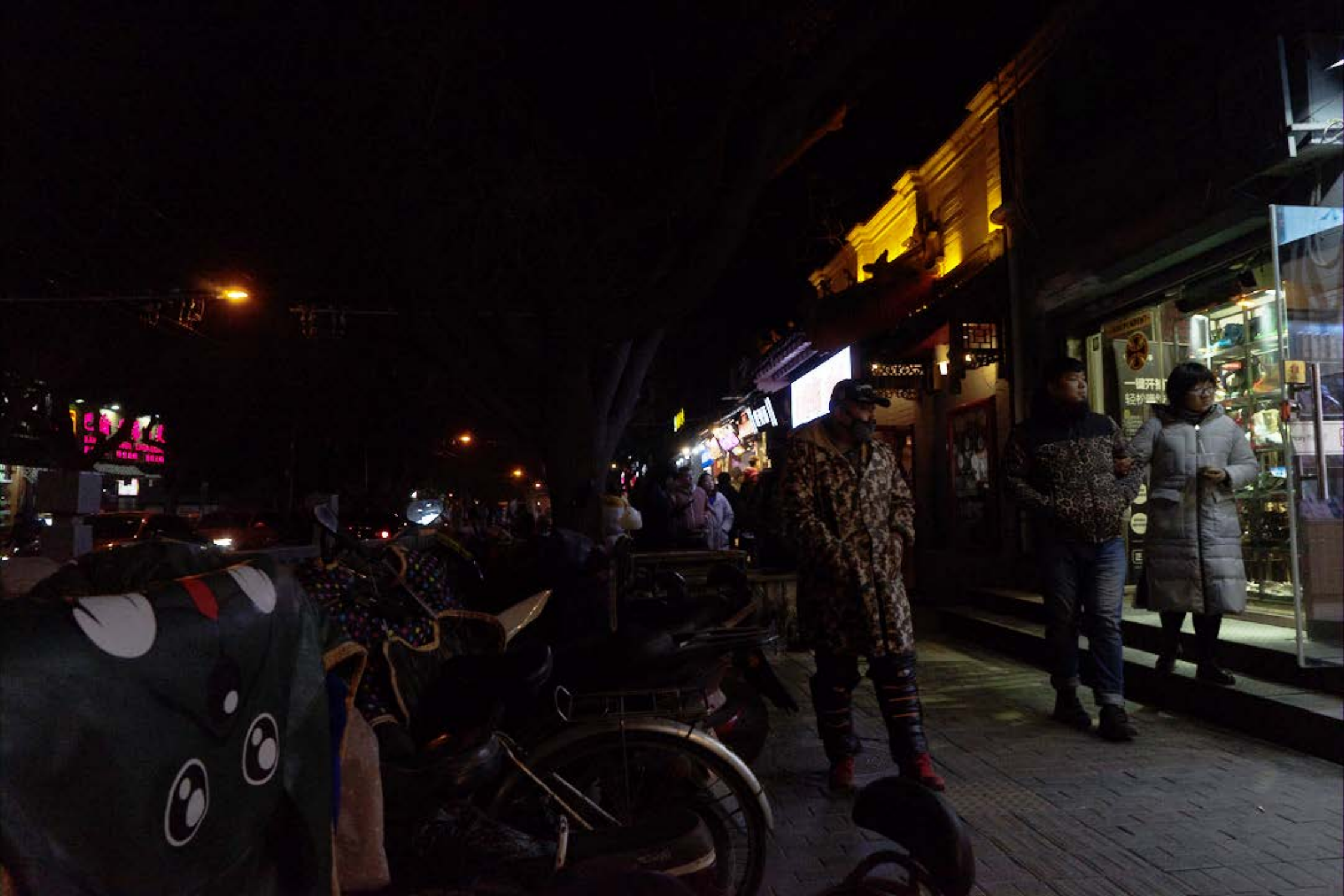}&
			\includegraphics[width=0.155\linewidth]{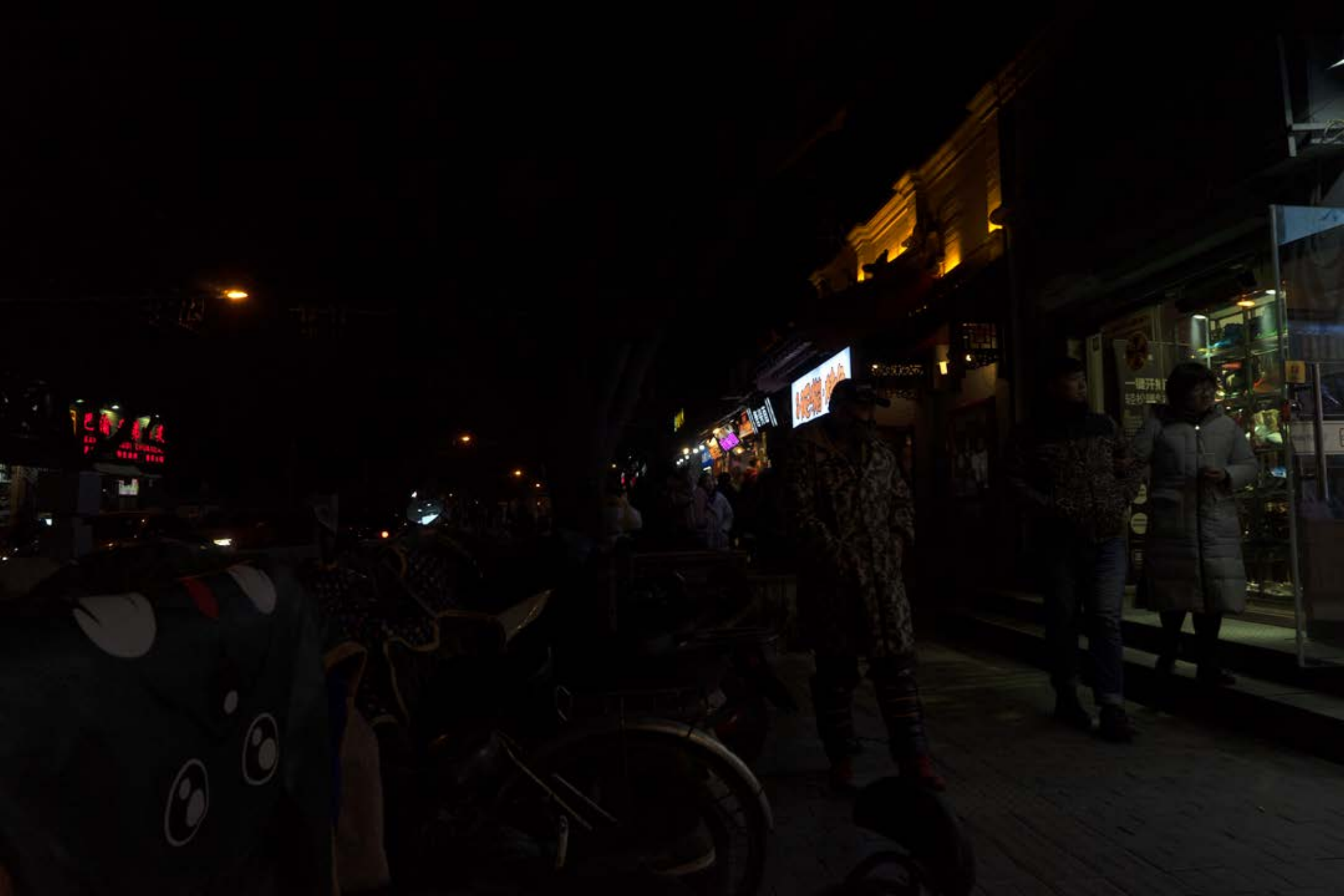}&
			\includegraphics[width=0.155\linewidth]{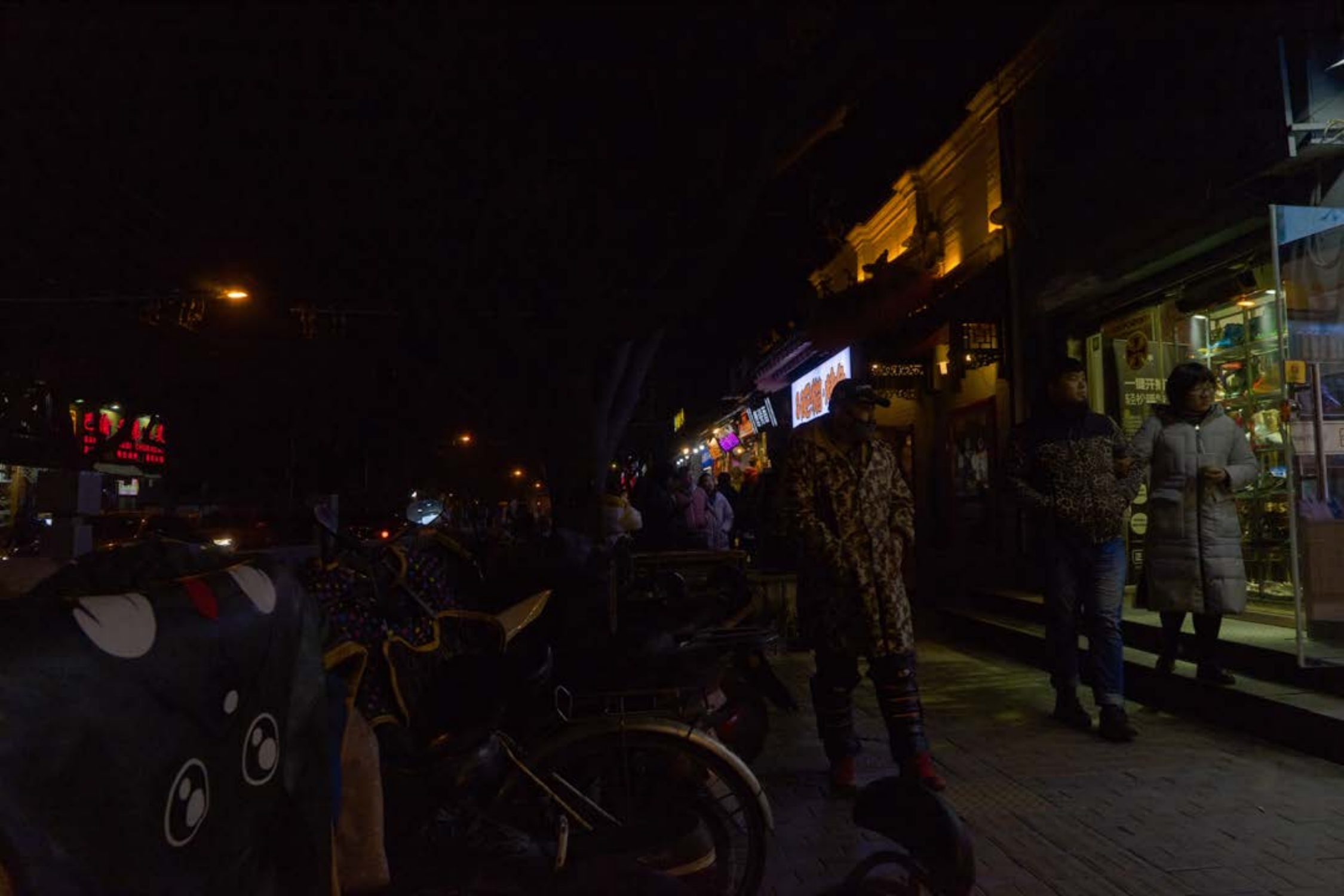}&
			\includegraphics[width=0.155\linewidth]{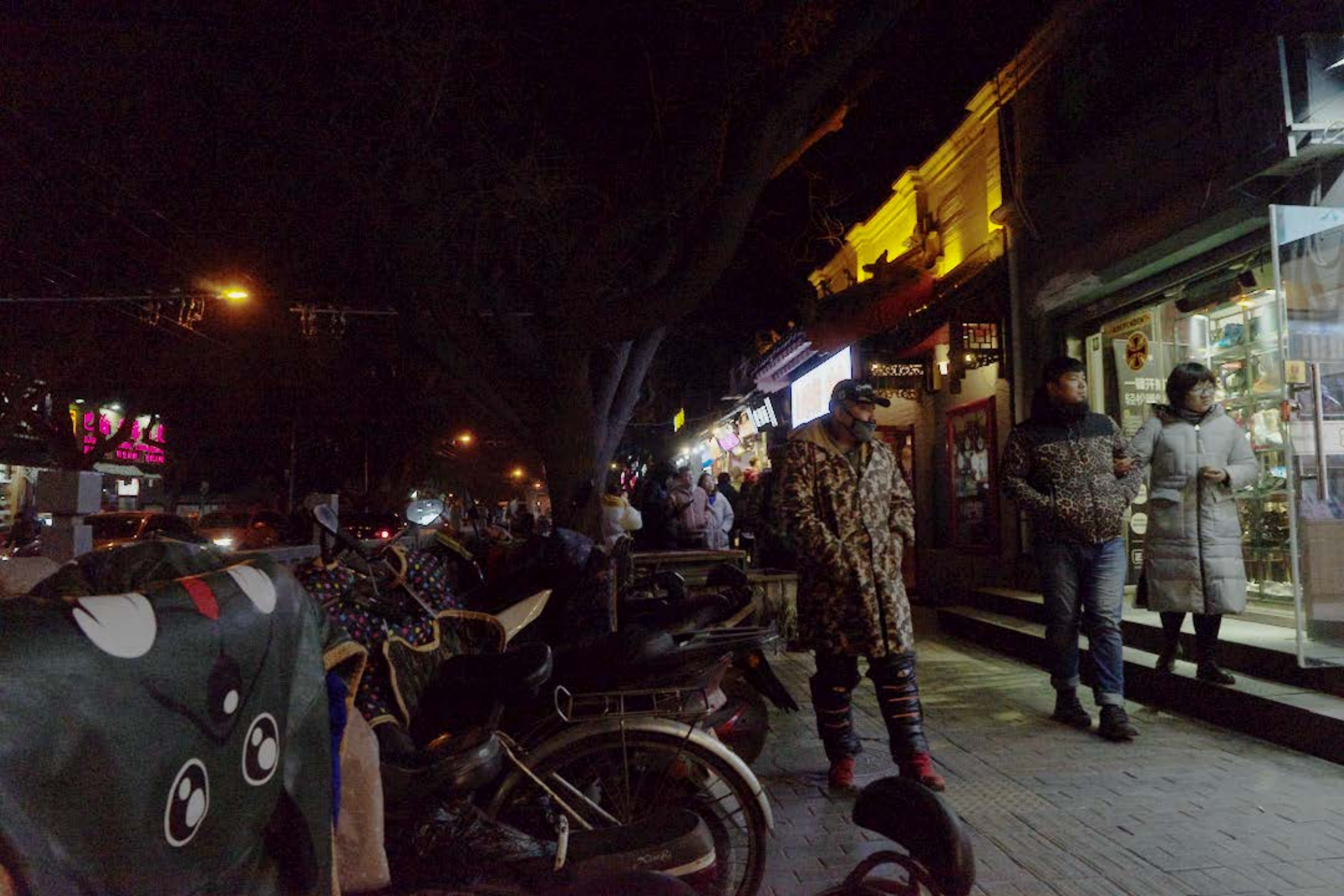}\\
			\includegraphics[width=0.155\linewidth]{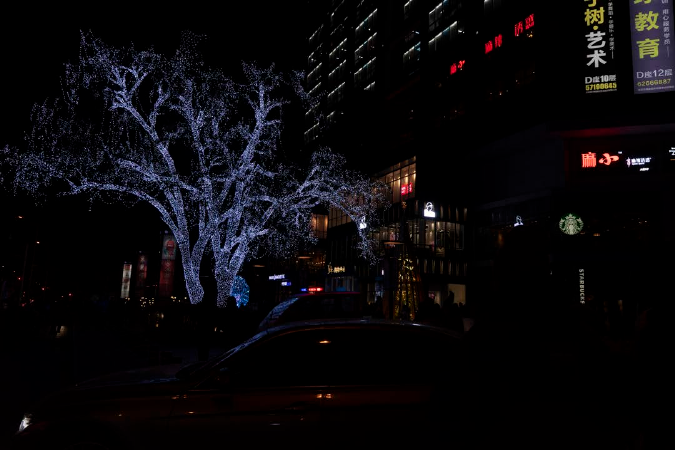}&
			\includegraphics[width=0.155\linewidth]{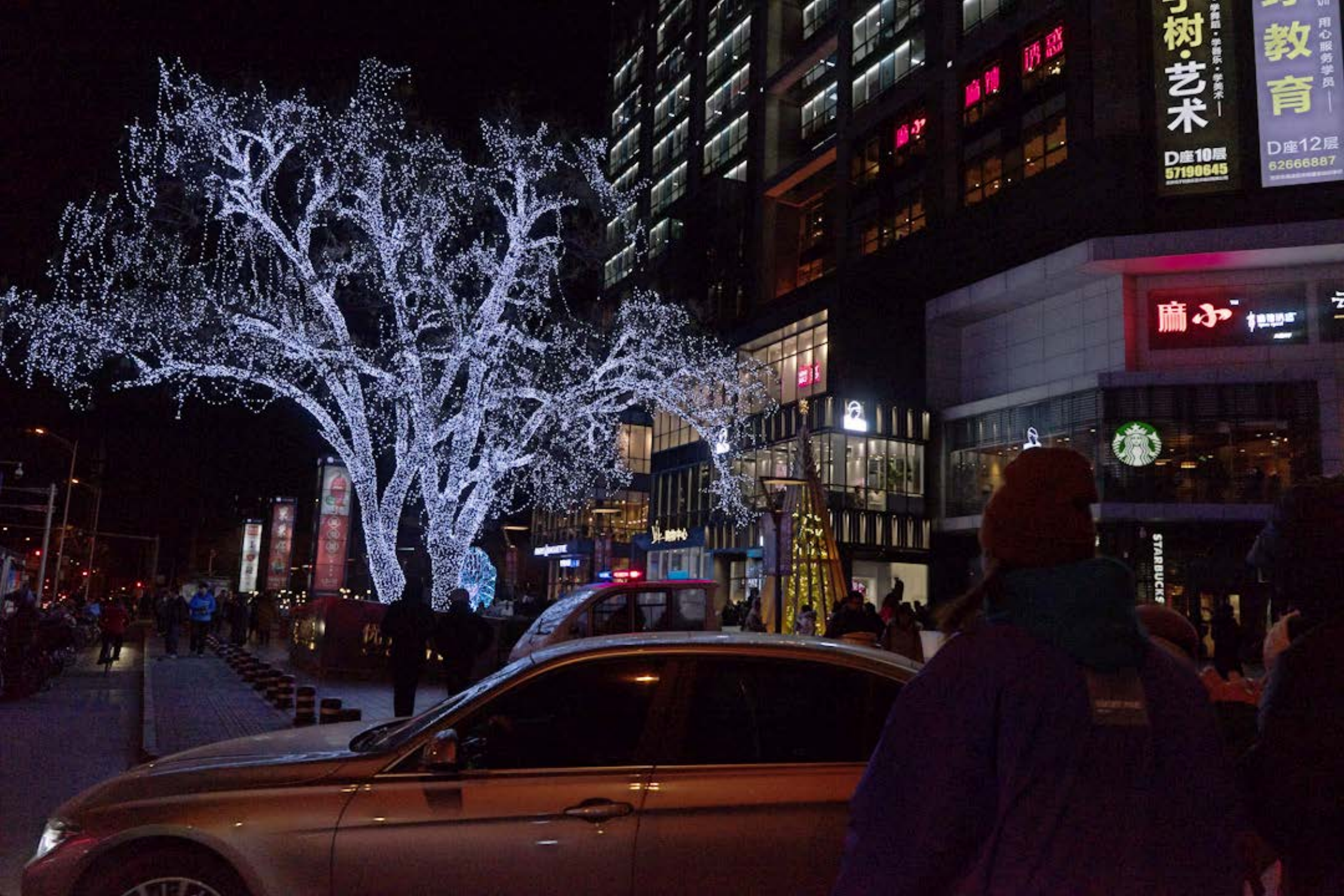}&
			\includegraphics[width=0.155\linewidth]{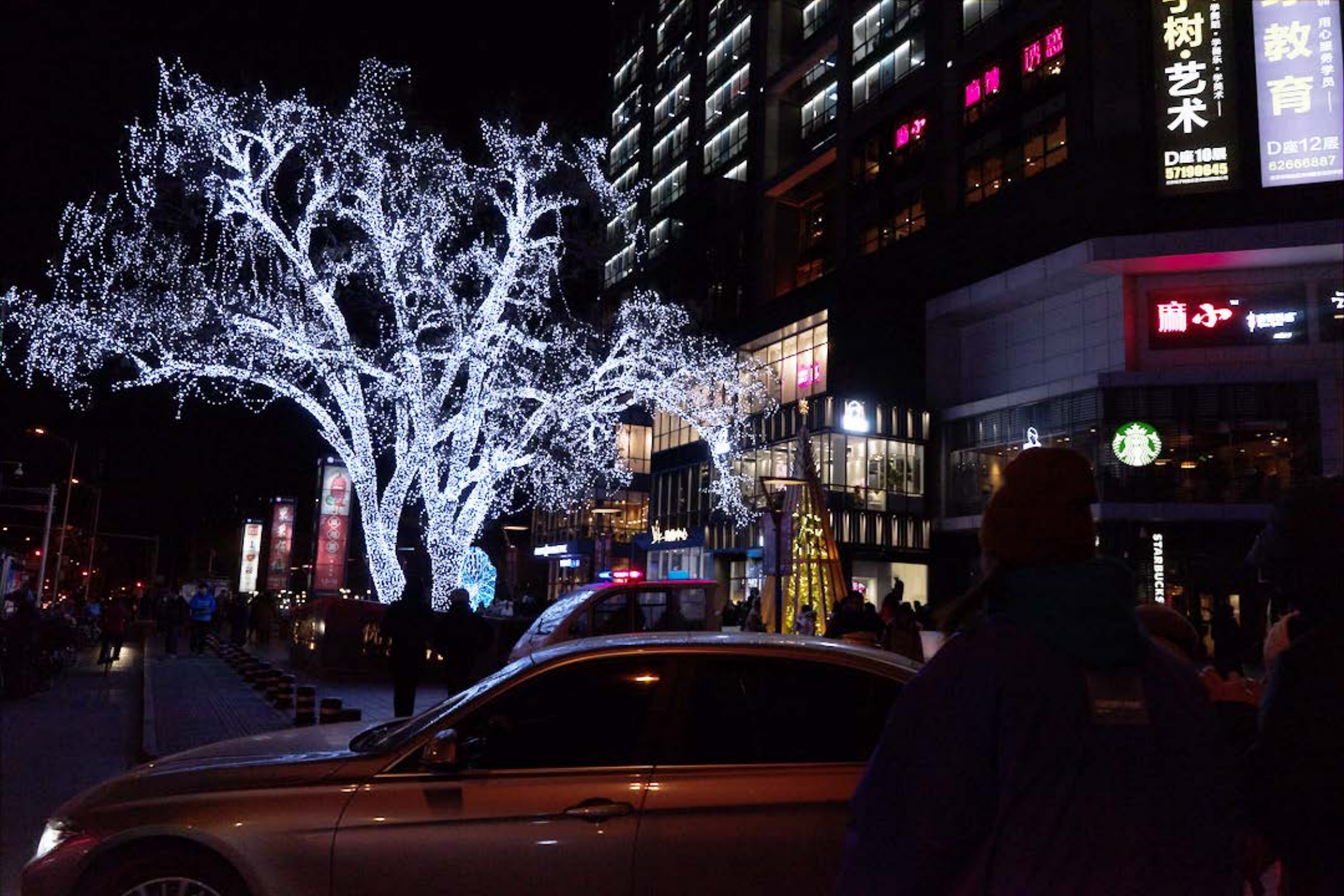}&
			\includegraphics[width=0.155\linewidth]{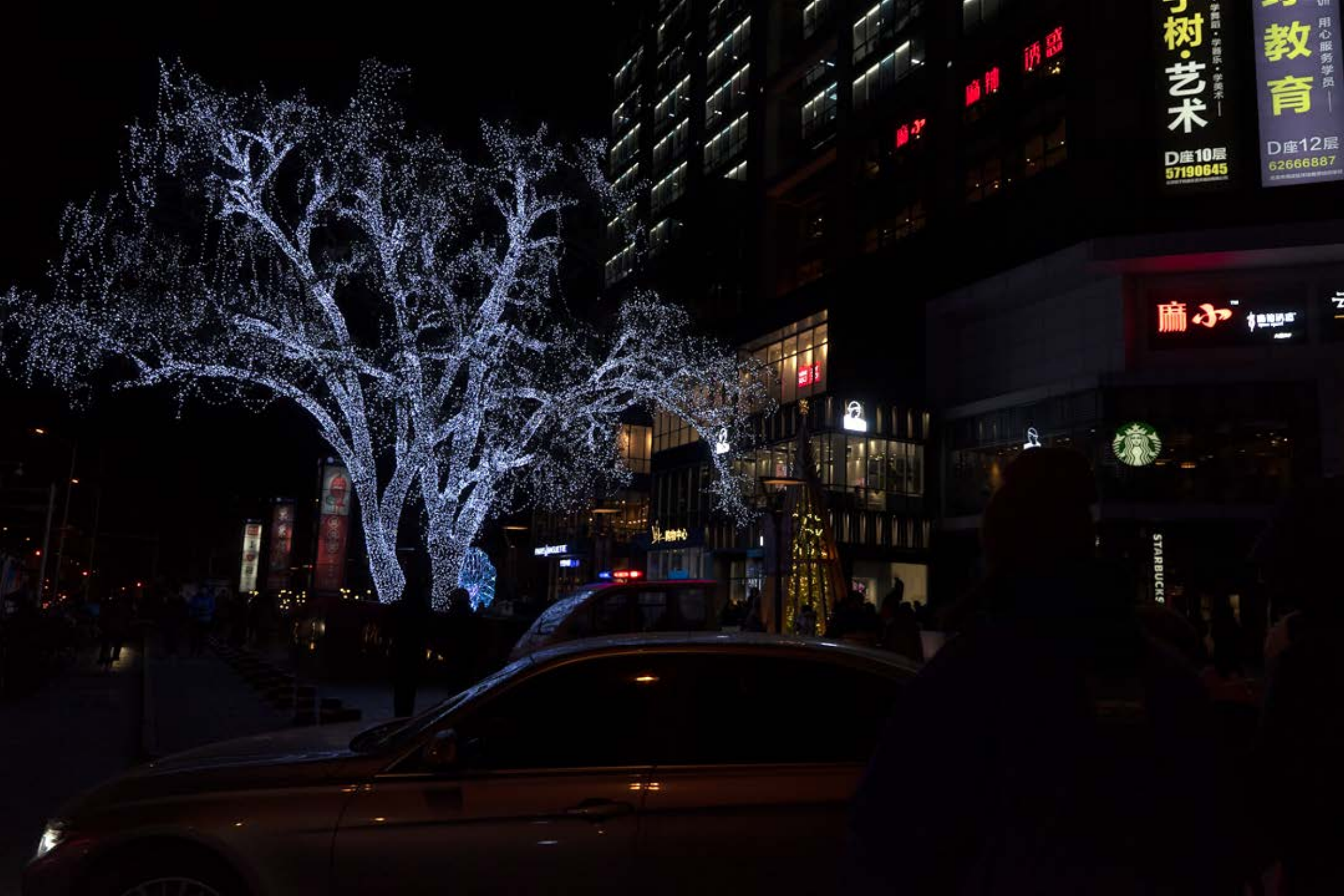}&
			\includegraphics[width=0.155\linewidth]{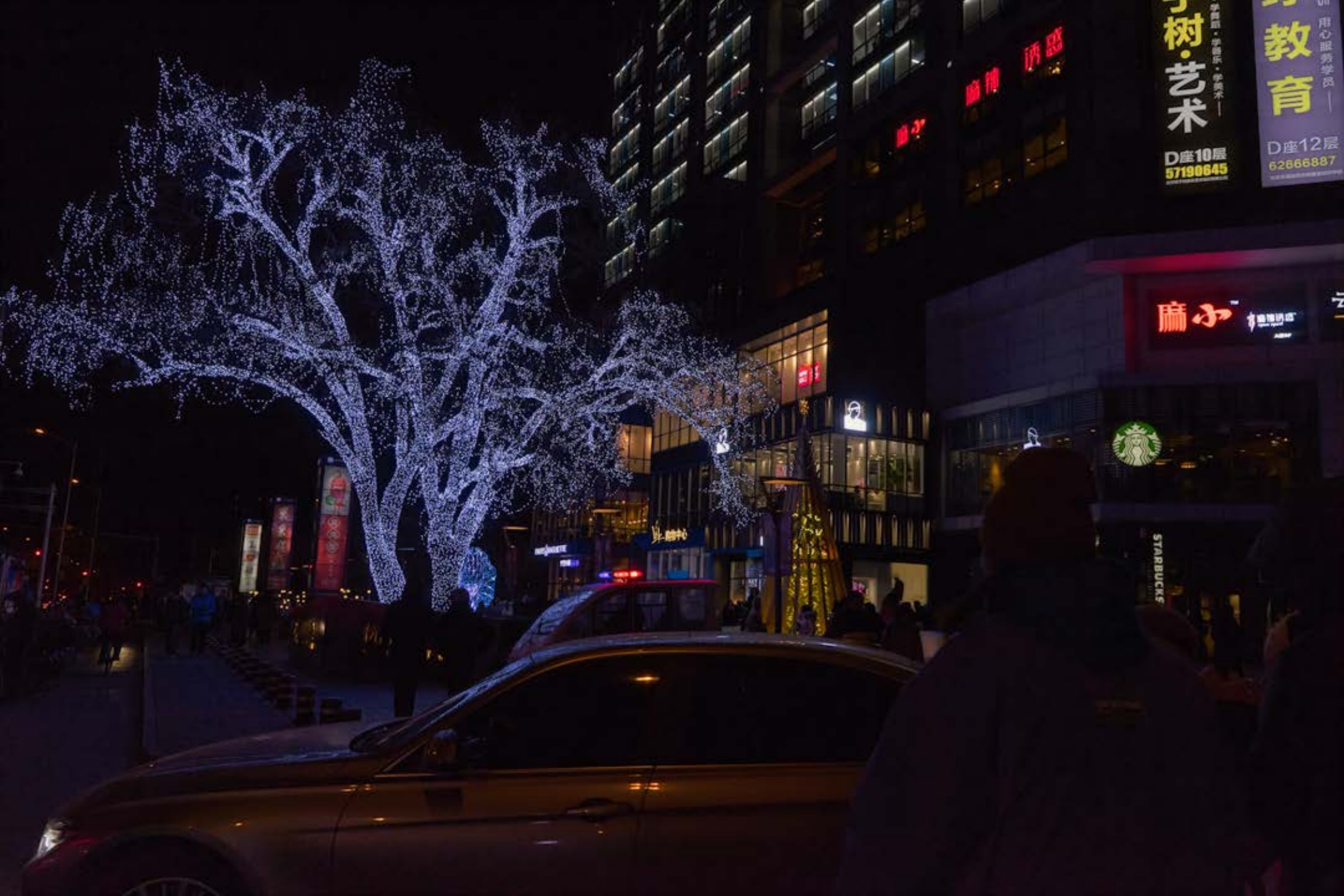}&
			\includegraphics[width=0.155\linewidth]{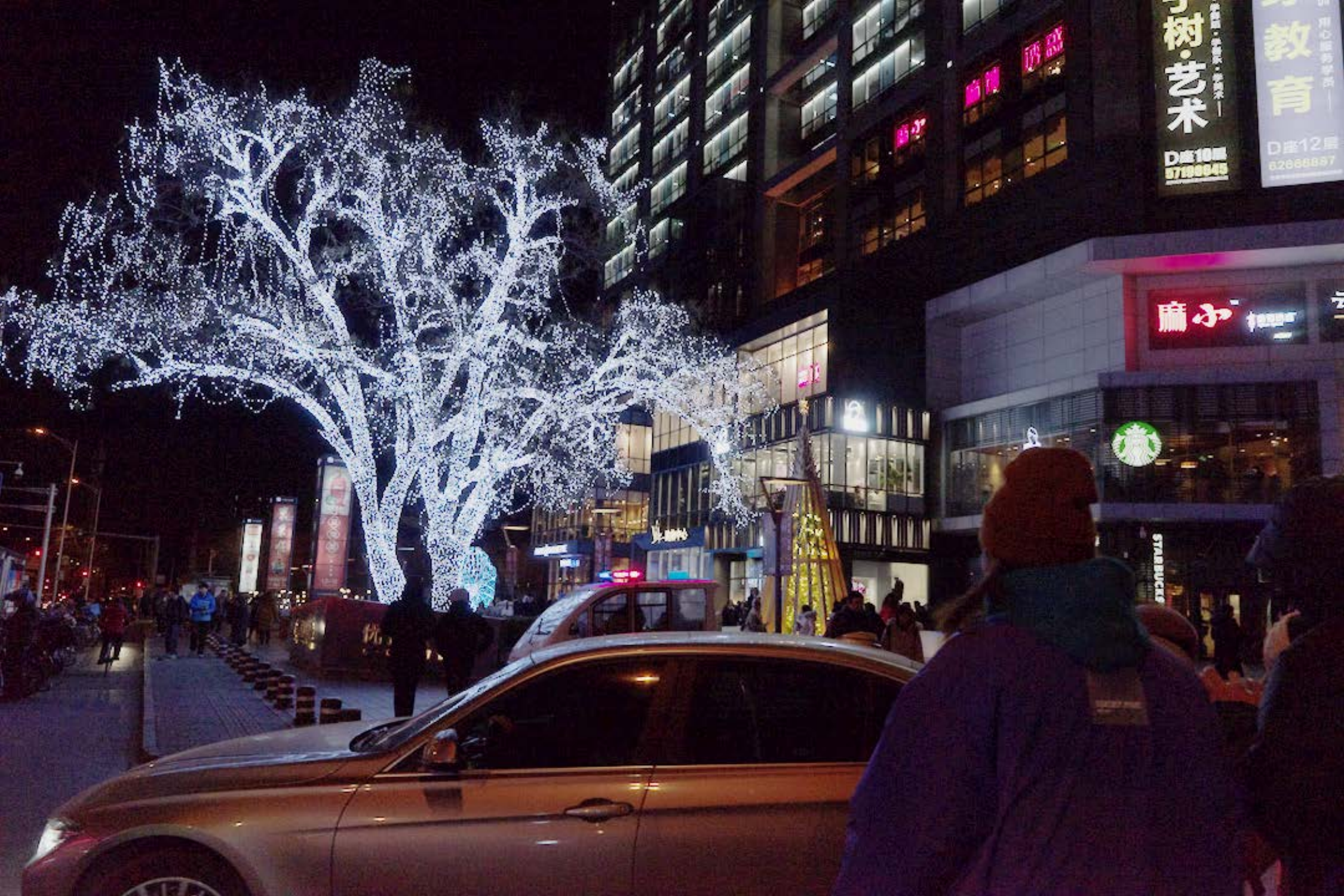}\\
			\includegraphics[width=0.155\linewidth]{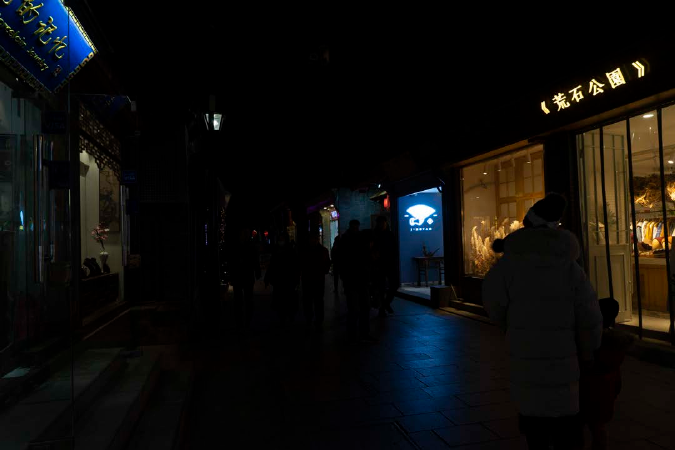}&
			\includegraphics[width=0.155\linewidth]{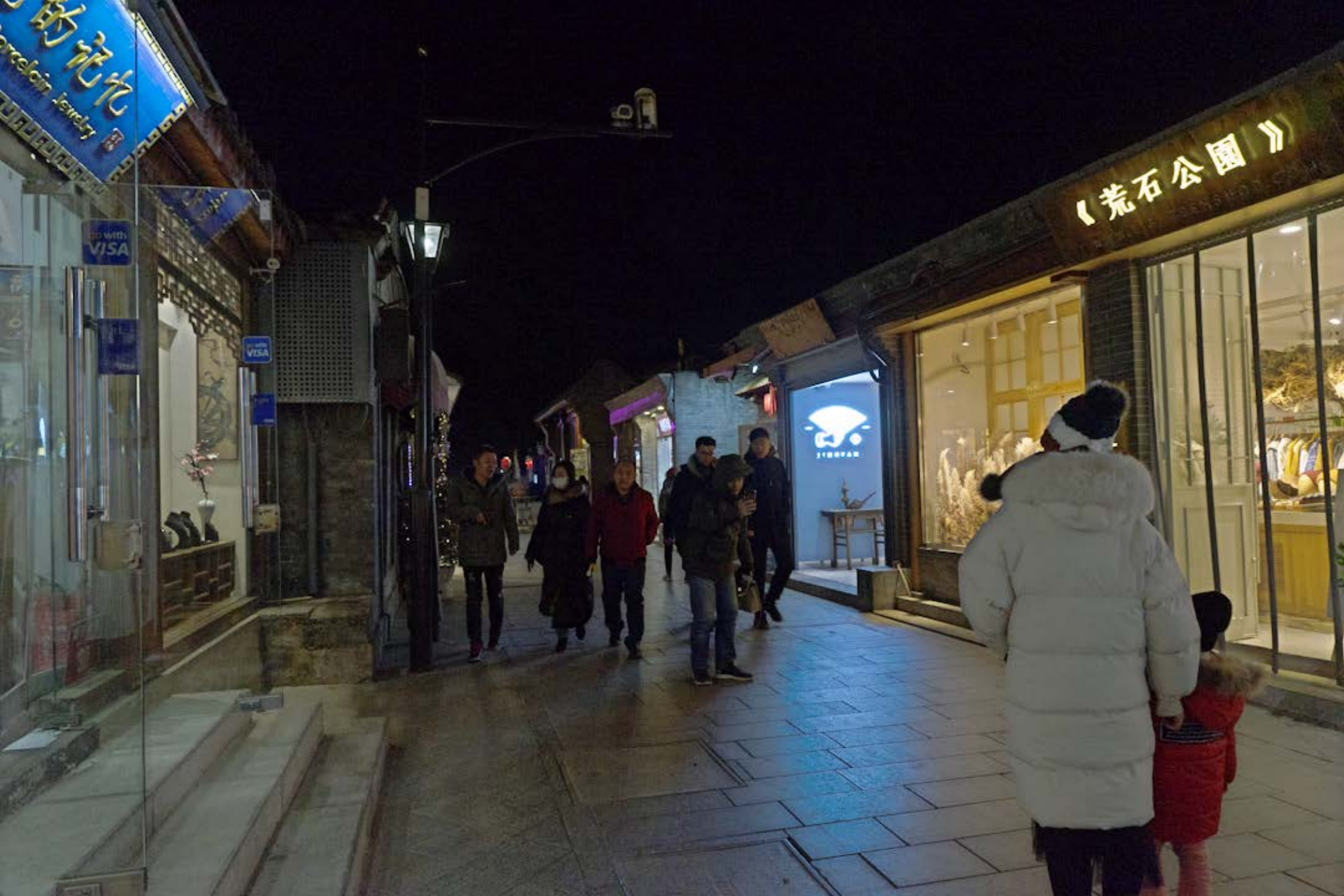}&
			\includegraphics[width=0.155\linewidth]{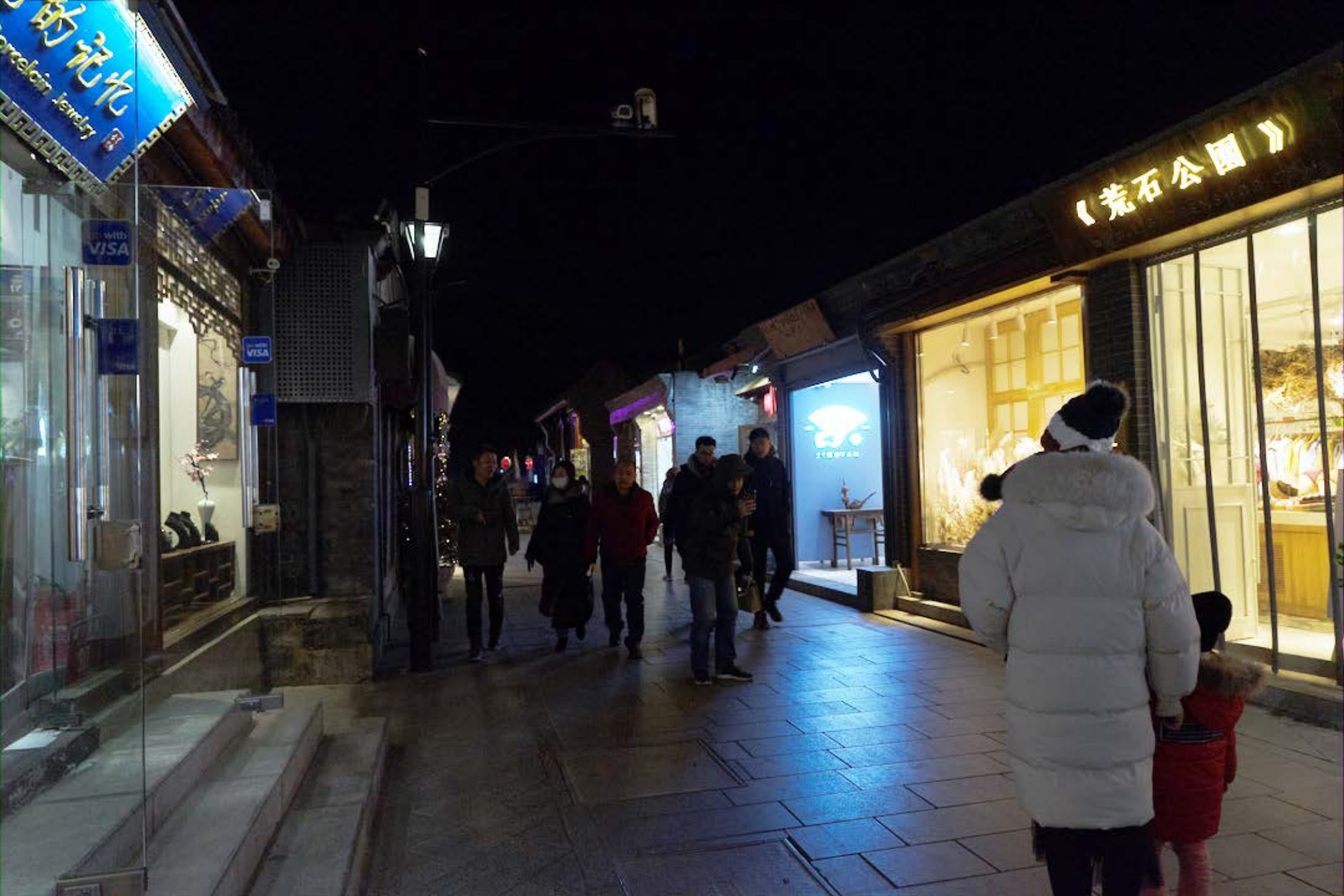}&
			\includegraphics[width=0.155\linewidth]{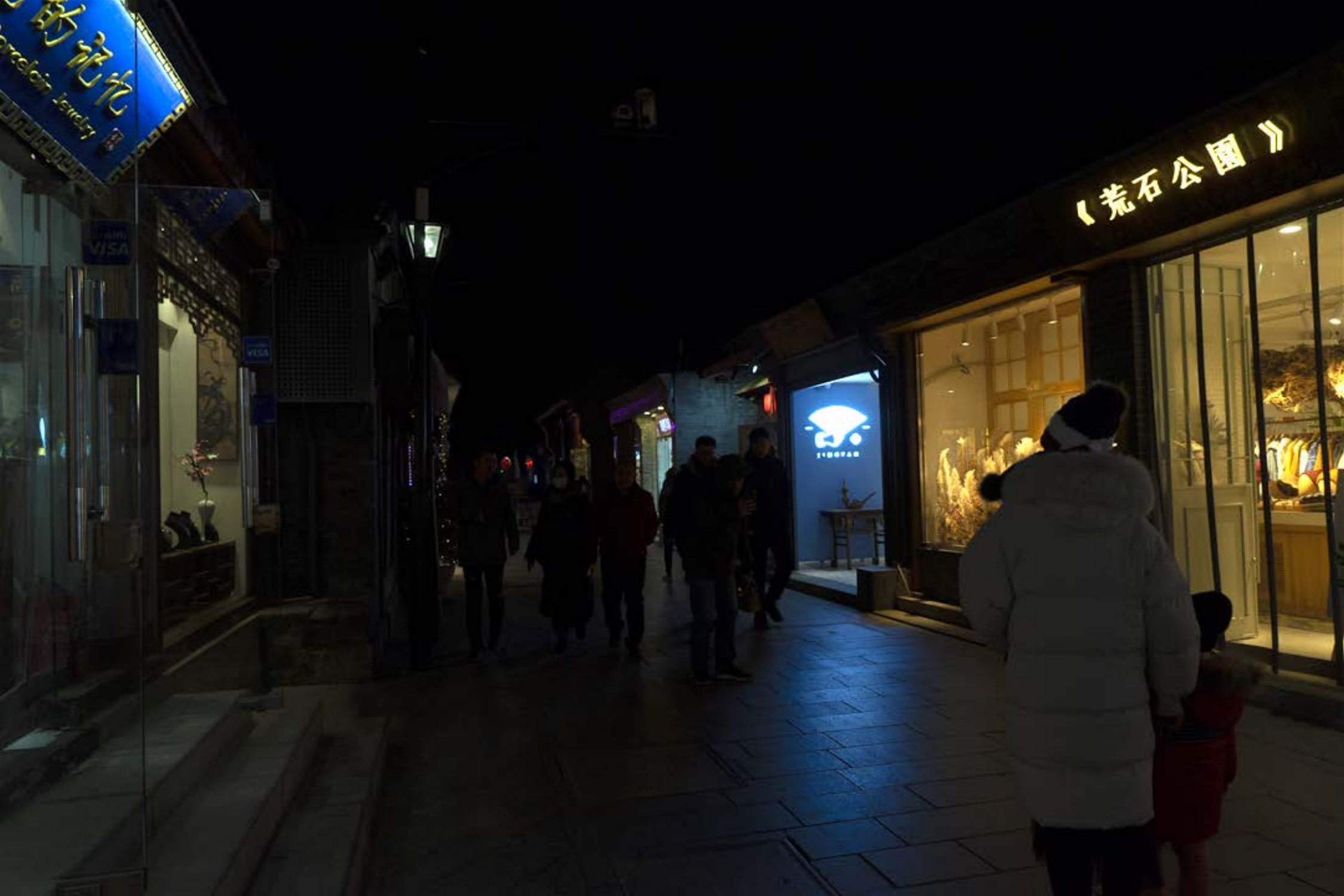}&
			\includegraphics[width=0.155\linewidth]{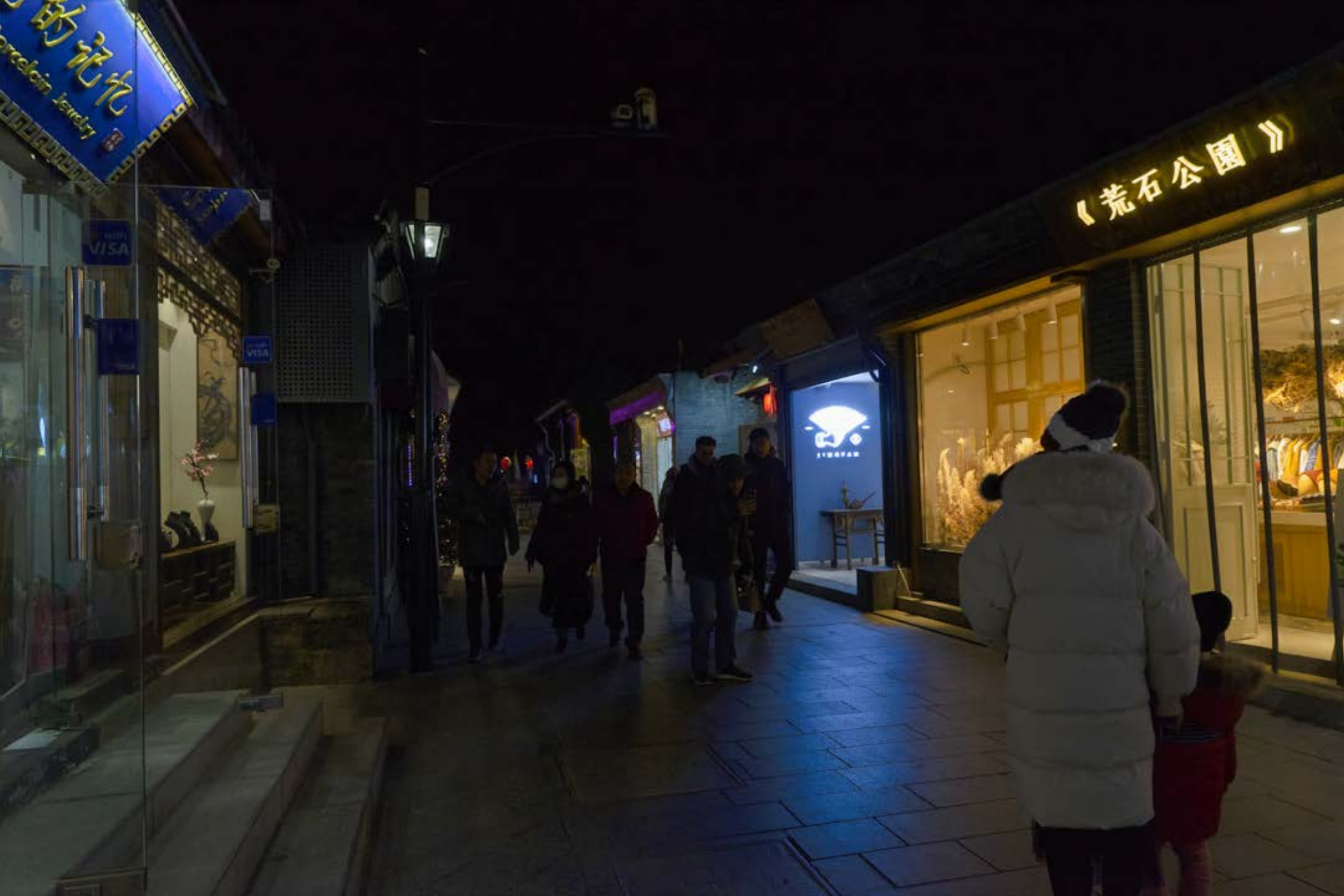}&
			\includegraphics[width=0.155\linewidth]{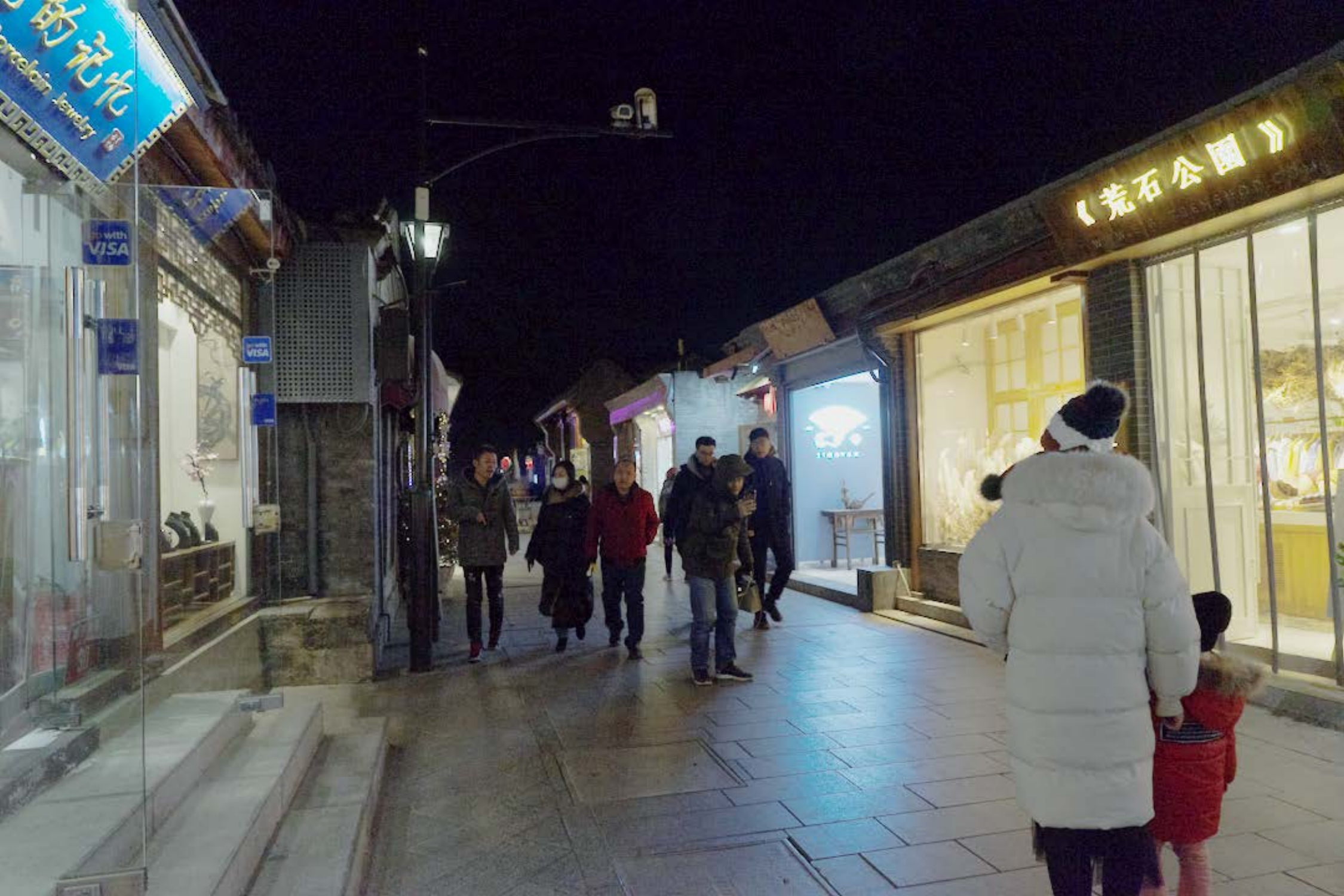}\\
			\footnotesize Input&\footnotesize DCE&\footnotesize SCI&\footnotesize RUAS&\footnotesize RQ-LLIE&\footnotesize Ours\\
		\end{tabular}
		\caption{{Qualitative comparison with advanced methods of low-light image enhancement on the DARKFACE dataset.}}
		\label{fig:darkface}
	\end{figure*}
	
	\begin{figure*}[htb!]
		\centering
		\begin{tabular}{c}
			\includegraphics[width=0.95\linewidth]{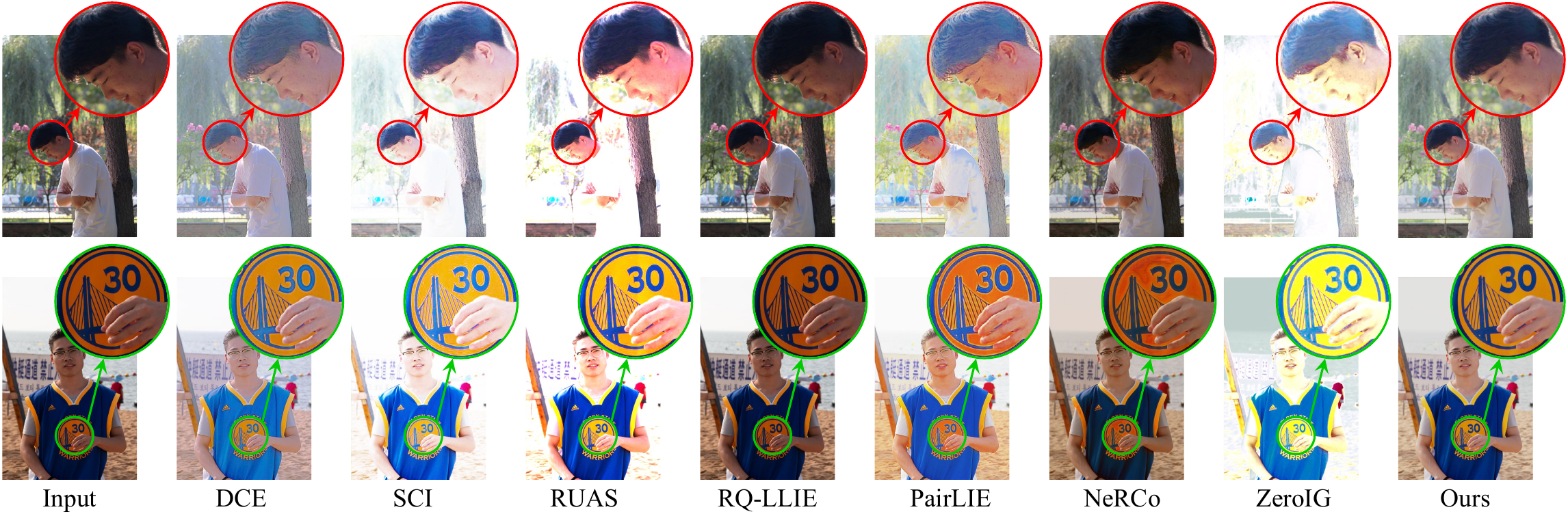}\\
		\end{tabular}
		\caption{{Visual comparison with state-of-the-art methods on the BAID dataset.}}
		\label{fig:BAID}
	\end{figure*}
	
	\begin{figure*}[!htb]
		\centering
		\footnotesize
		\begin{tabular}{c}
			\includegraphics[width=0.95\linewidth]{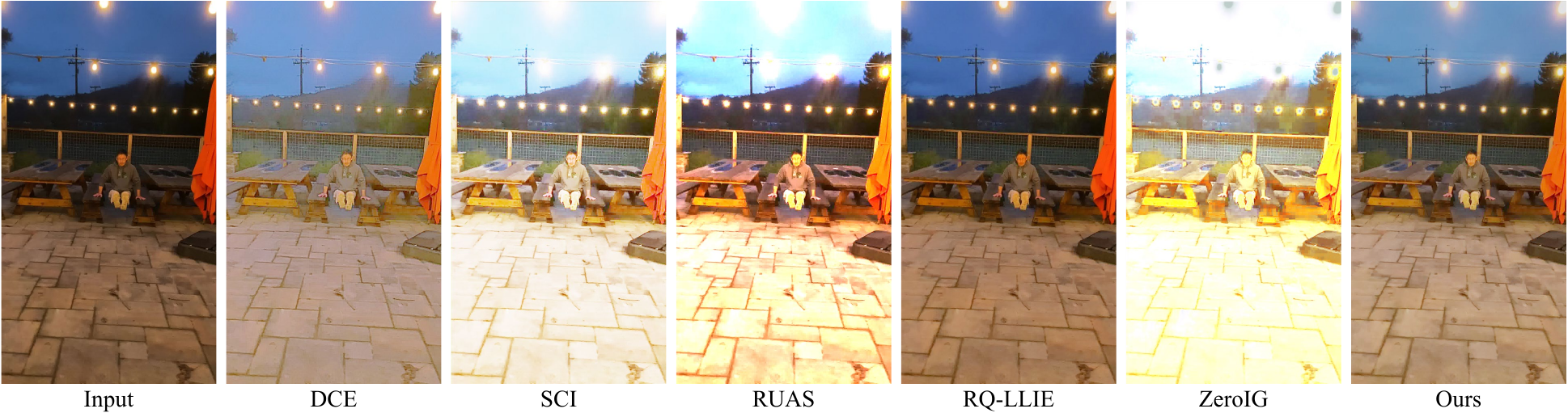}\\
		\end{tabular}
		\caption{{Visual comparison of results on the LOLi-Phone dataset.}}
		\label{fig: LOLi}
	\end{figure*}

	\subsubsection{Conventional Low-Light Scenes}
	The upper section of Table~\ref{table:quan} reports the numerical results on the MIT dataset. Clearly, our method ranks top in all metrics, demonstrating significant advantages. Thanks to the constructed auto re-parameterization mechanism, which expands the parameter space while keeping the predefined structure unchanged, AR-LLIE achieves competitive results compared to other advanced methods. Furthermore, compared to the fastest existing deep learning-based low-light enhancer SCI, AR-LLIE significantly reduces computation time, validating the superiority of our approach. Fig.~\ref{fig:mit} presents the qualitative results on the MIT dataset. It can be observed that all other comparative methods fail to effectively improve visual quality and often have unexpected degradation effects in their enhancement results. For example, FIDE and UTVNet exhibit obvious underexposure issues, while the results from KinD and ZeroIG show significant color shifts. RUAS and PairLIE suffer from varying degrees of overexposure. In contrast, our AR-LLIE method achieves accurate brightness control, natural colors, and clearer details. {To further evaluate the subjective quality of our results, we provide quantitative comparisons between our method and representative approaches on the widely used NR-IQA metrics LIQE~\cite{zhang2023blind} and UNIQUE~\cite{zhang2021uncertainty} in Table~\ref{table: LIQE}. Clearly, our method demonstrates significant advantages in both metrics, indicating that our approach achieves better visual quality, with superior preservation of structural information, and is closer to human perceptual evaluation standards compared to other methods. Furthermore, with the recent surge in popularity of Transformers and diffusion models, a series of related new approaches have emerged. In Table 1, we compare the proposed method with recently introduced Transformer- and diffusion-based methods, including LLFormer~\cite{wang2023ultra}, LightenDiffusion~\cite{jiang2025lightendiffusion}, GSAD~\cite{hou2024global}, and DiffLL~\cite{jiang2023low}. It can be observed that our AR-LLIE remains competitive and demonstrates certain advantages over these state-of-the-art approaches.}

	\begin{figure*}[htb!]
		\centering
		\begin{tabular}{c@{\extracolsep{0.15em}}c@{\extracolsep{0.15em}}c@{\extracolsep{0.15em}}c@{\extracolsep{0.15em}}c}
			\includegraphics[width=0.19\linewidth,height=0.13\linewidth]{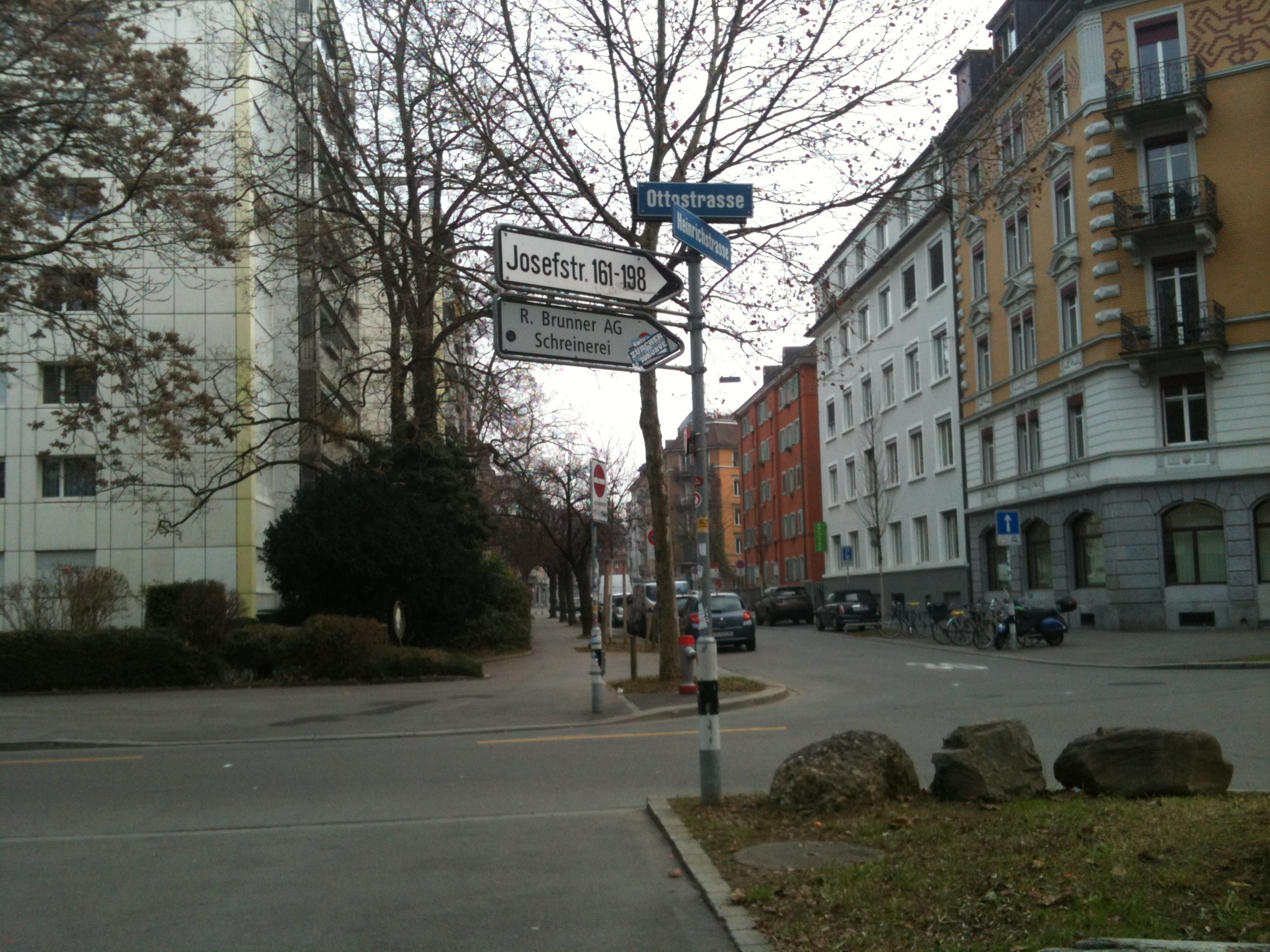}&
			\includegraphics[width=0.19\linewidth,height=0.13\linewidth]{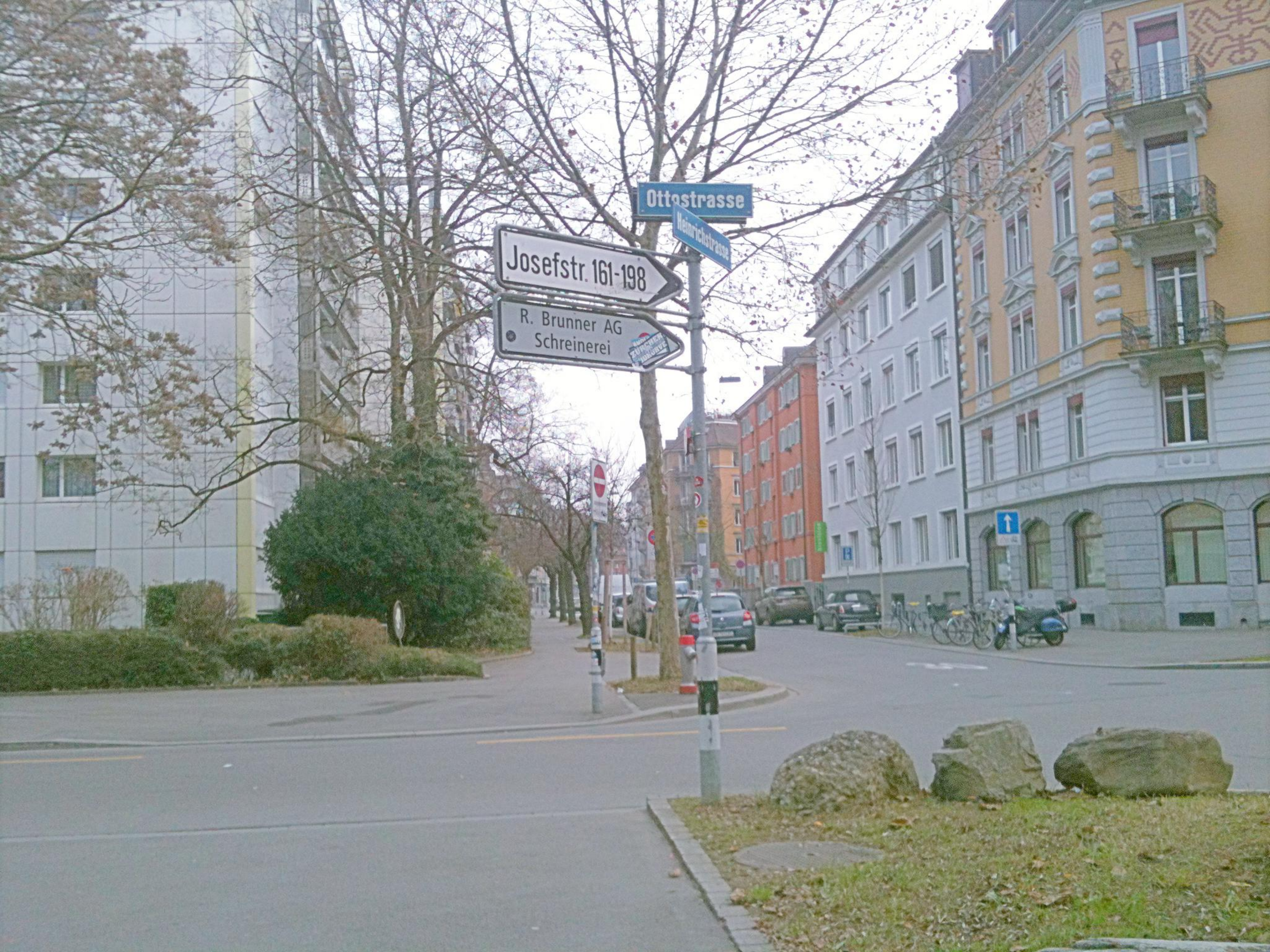}&
			\includegraphics[width=0.19\linewidth,height=0.13\linewidth]{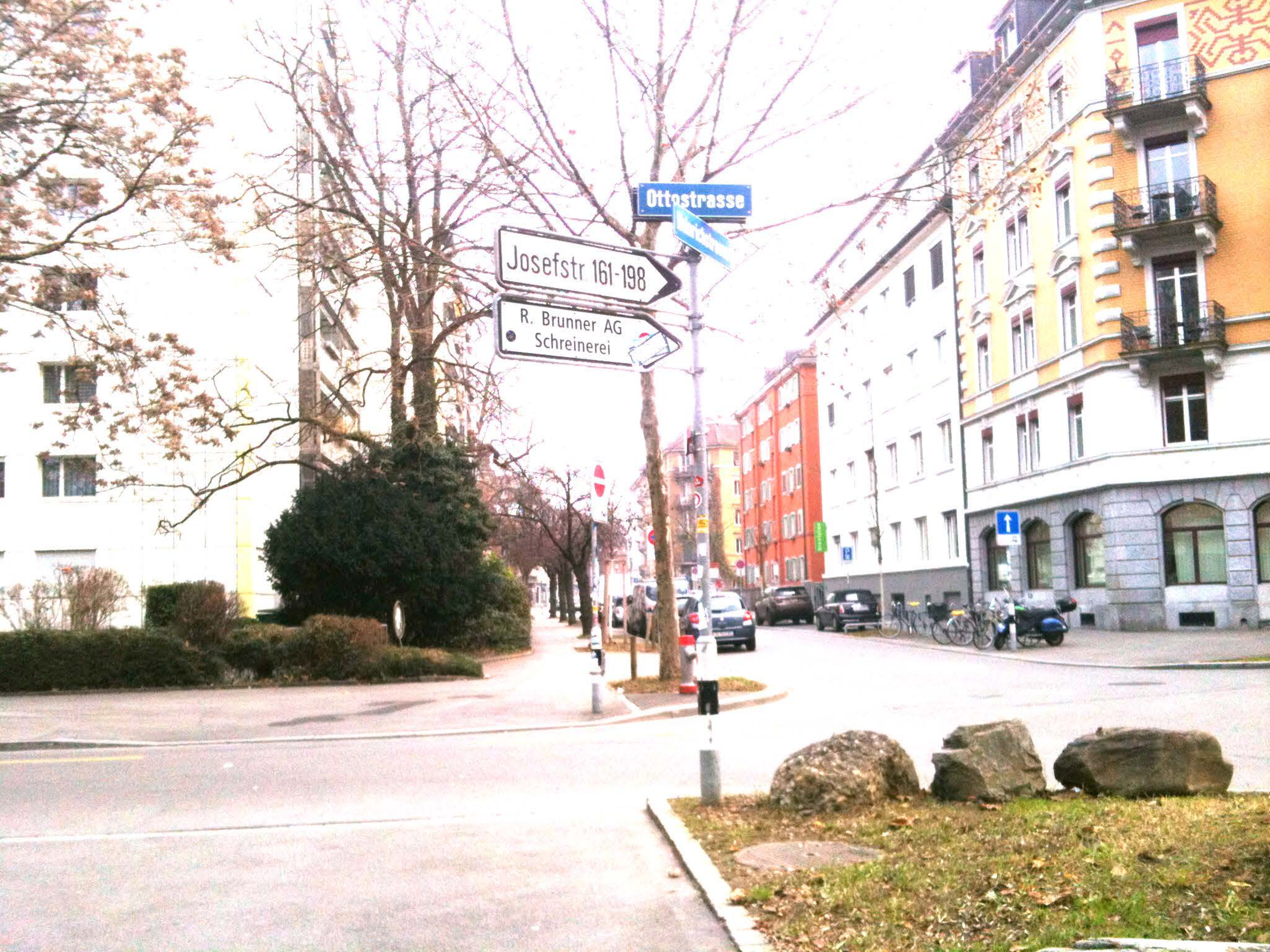}&
			\includegraphics[width=0.19\linewidth,height=0.13\linewidth]{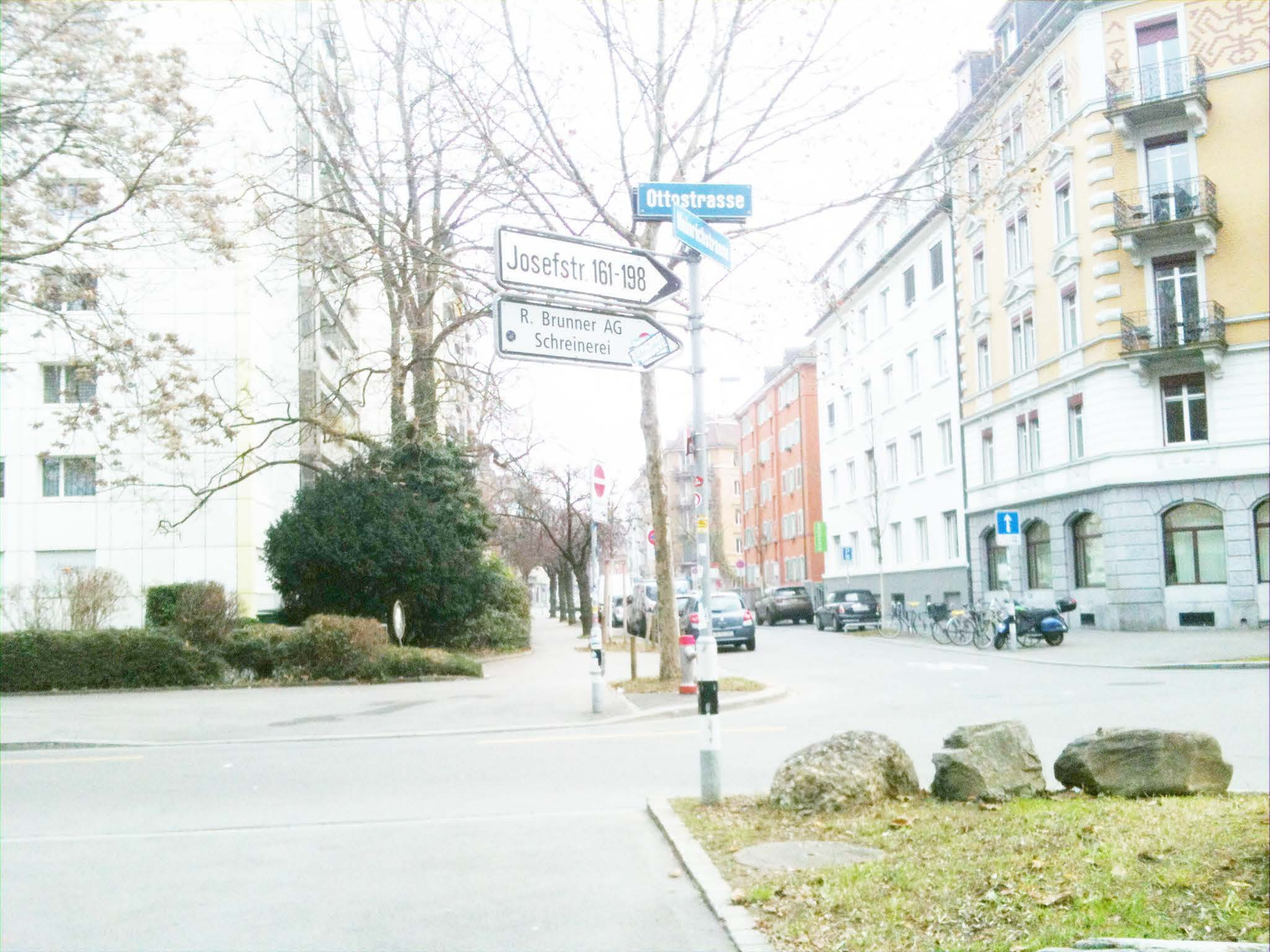}&
			\includegraphics[width=0.19\linewidth,height=0.13\linewidth]{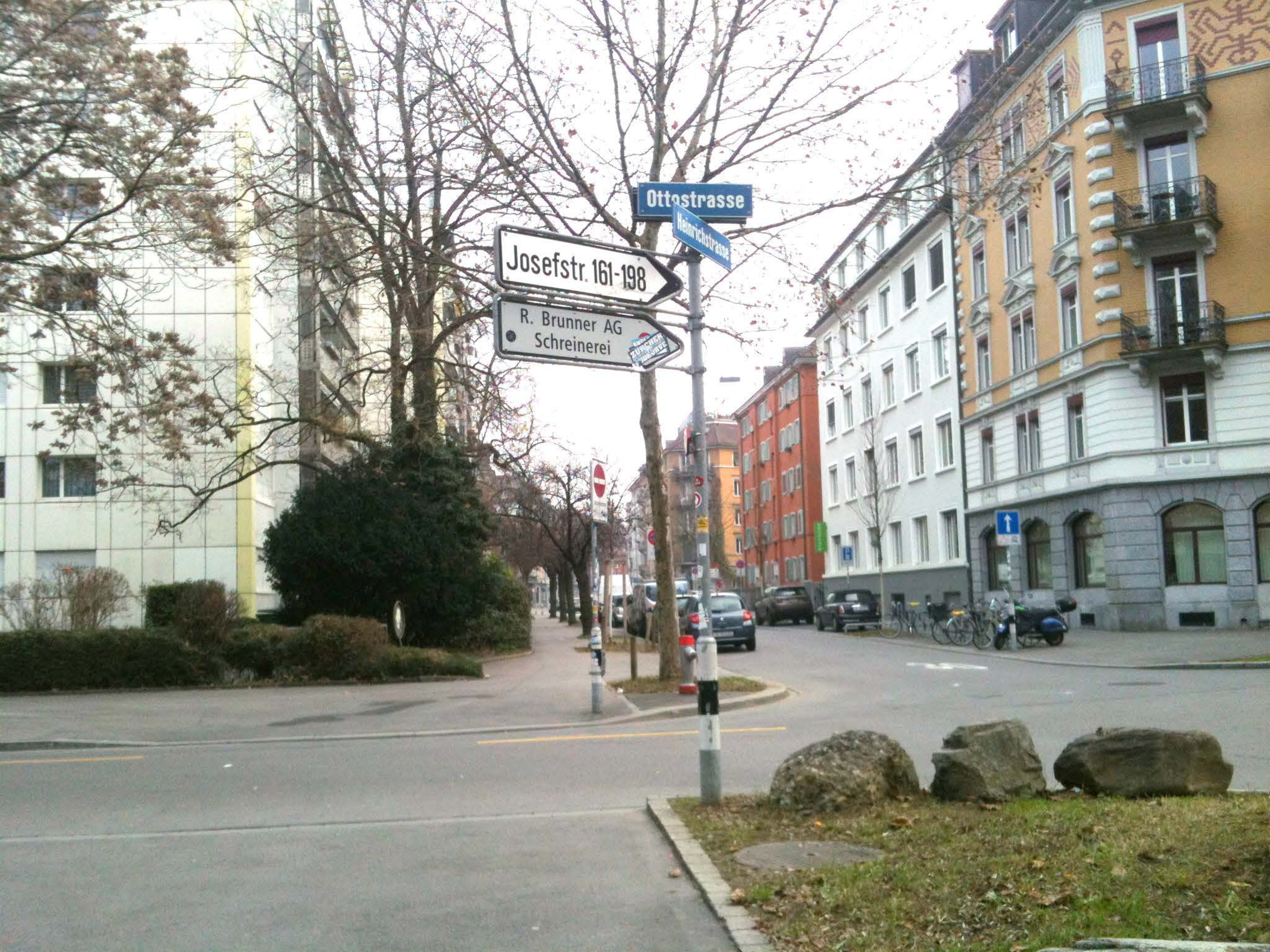}\\
			\includegraphics[width=0.19\linewidth,height=0.13\linewidth]{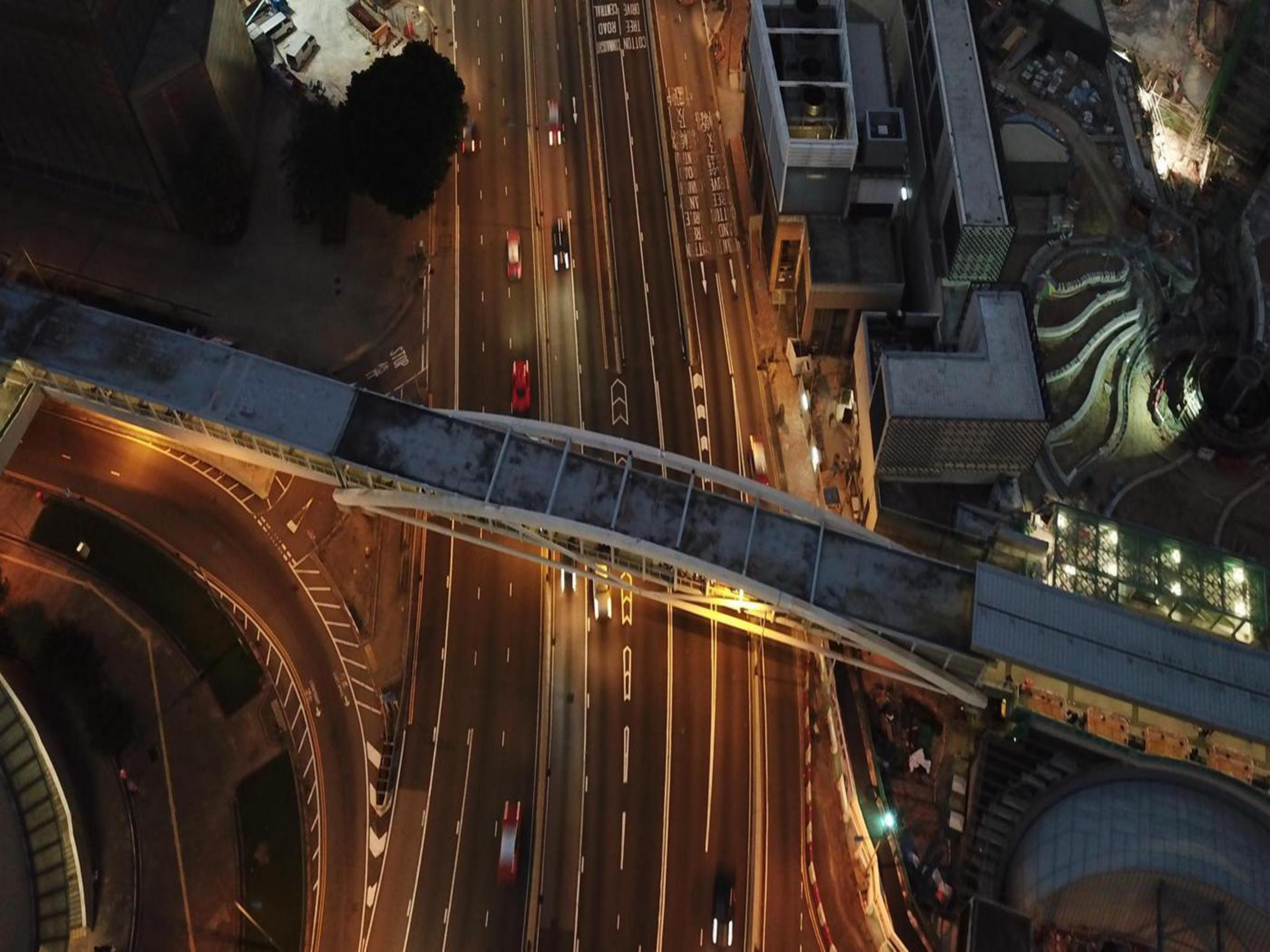}&
			\includegraphics[width=0.19\linewidth,height=0.13\linewidth]{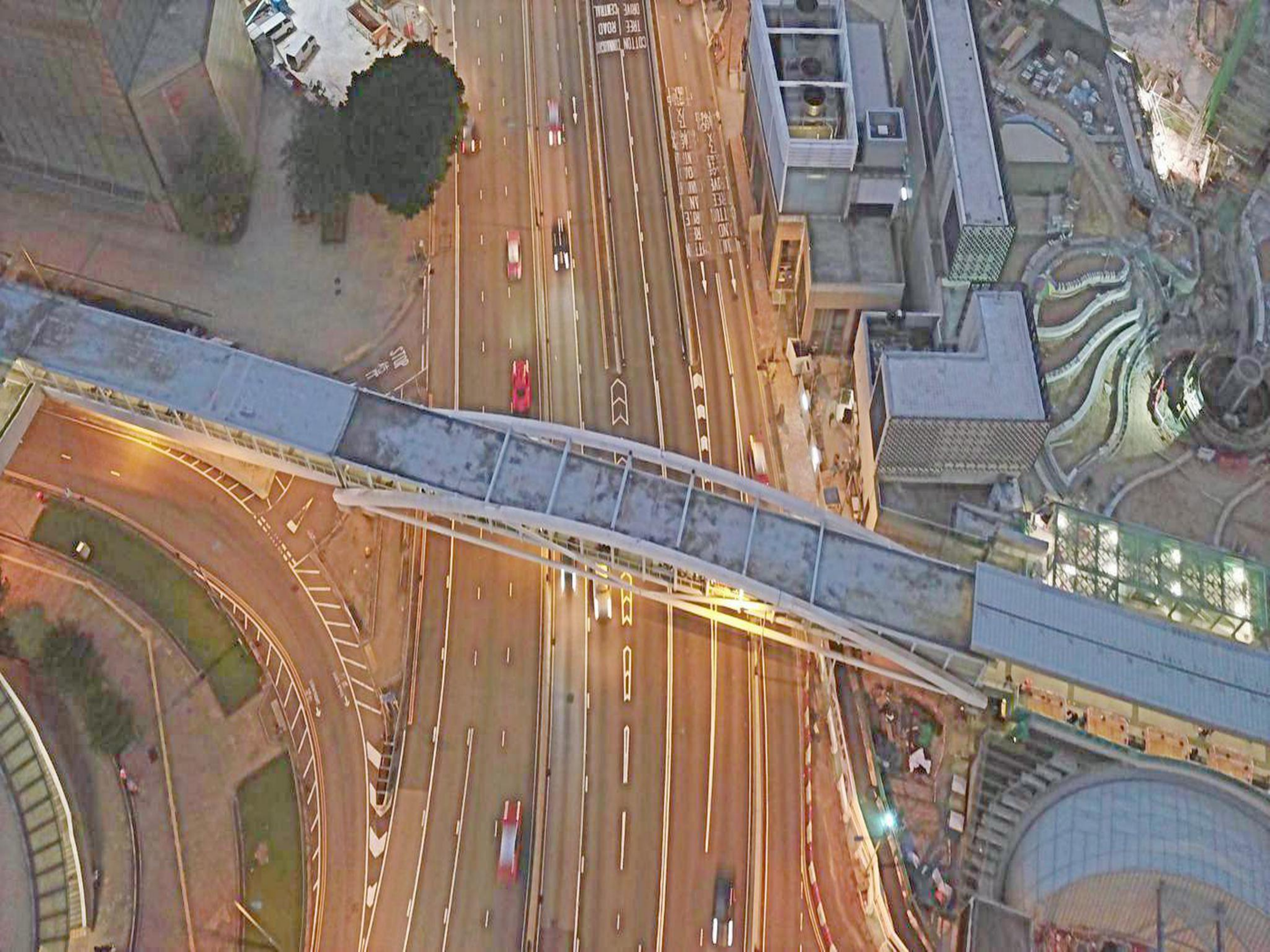}&
			\includegraphics[width=0.19\linewidth,height=0.13\linewidth]{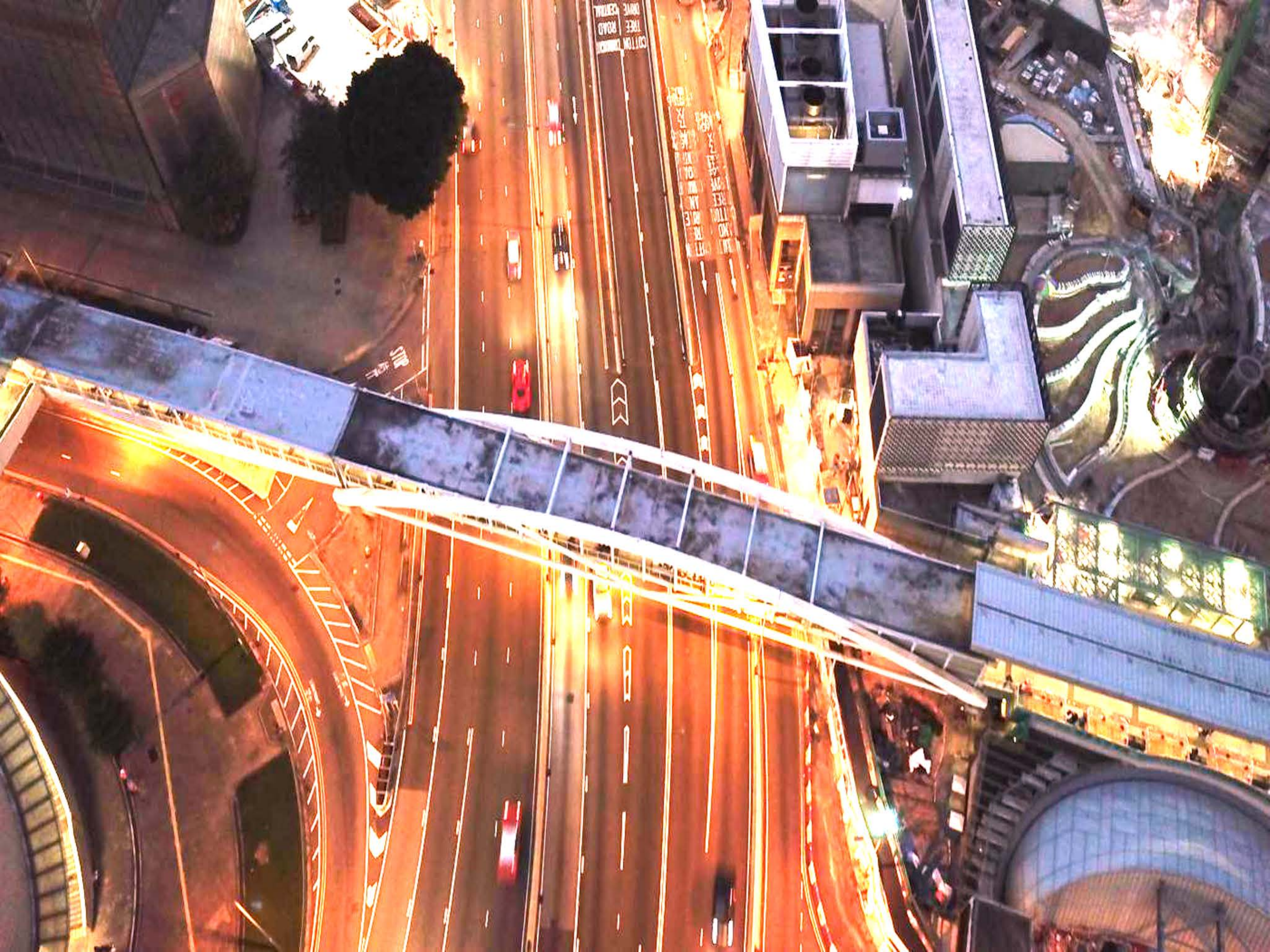}&
			\includegraphics[width=0.19\linewidth,height=0.13\linewidth]{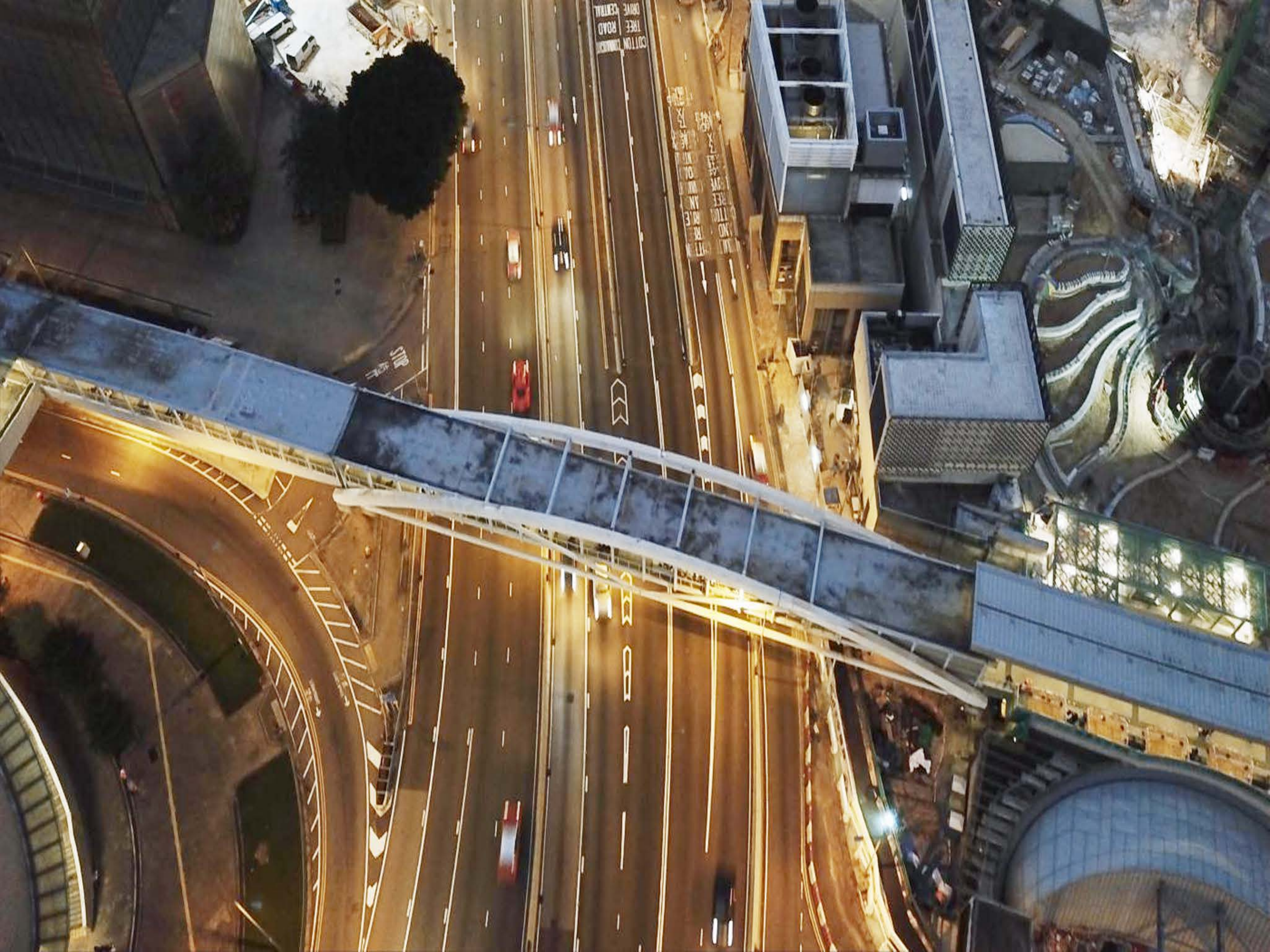}&
			\includegraphics[width=0.19\linewidth,height=0.13\linewidth]{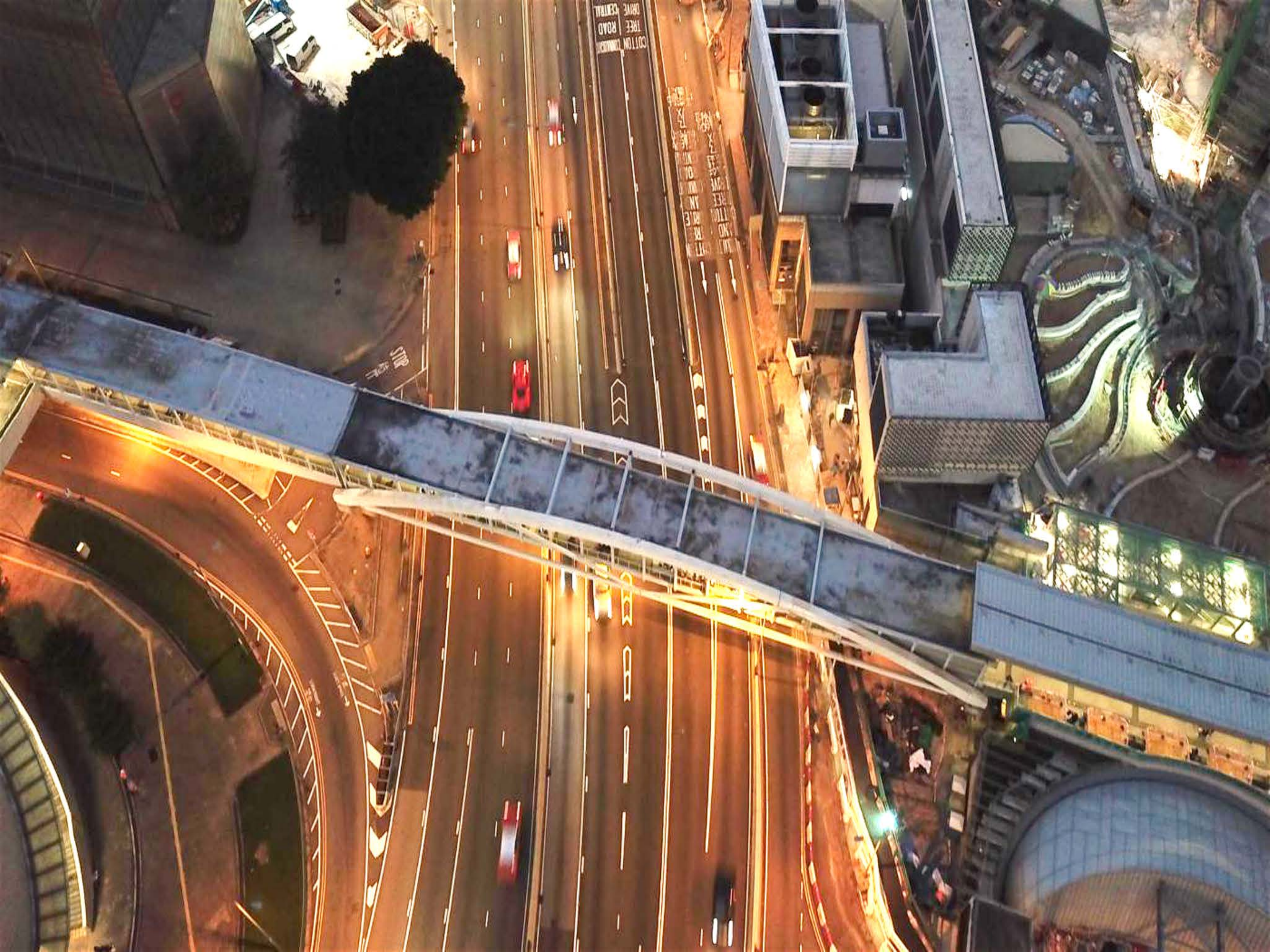}\\
			\includegraphics[width=0.19\linewidth,height=0.13\linewidth]{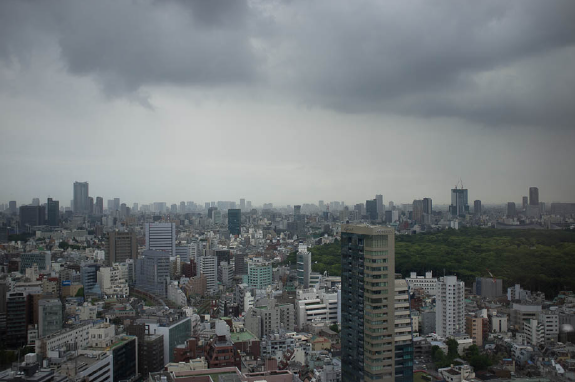}&
			\includegraphics[width=0.19\linewidth,height=0.13\linewidth]{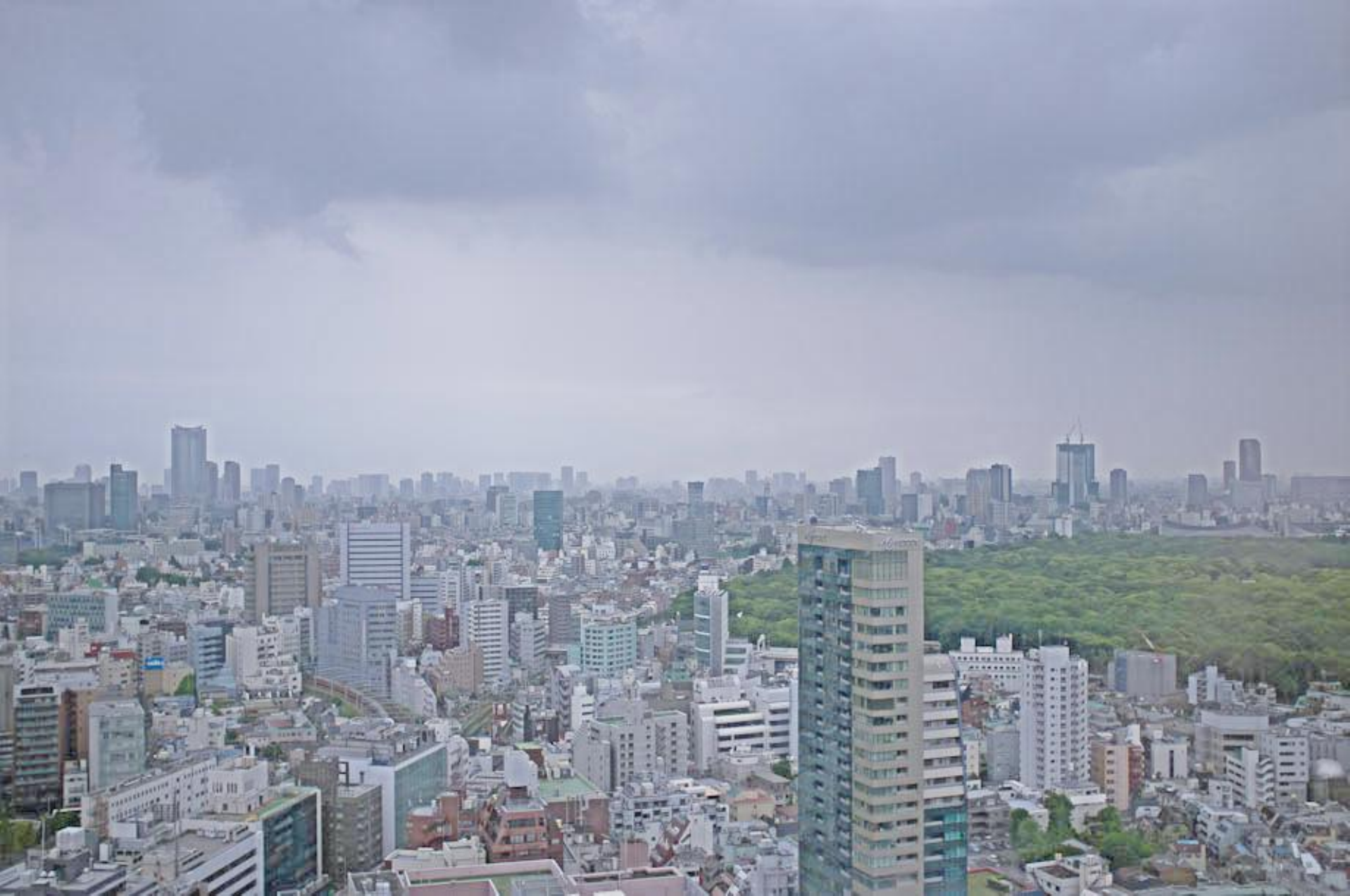}&
			\includegraphics[width=0.19\linewidth,height=0.13\linewidth]{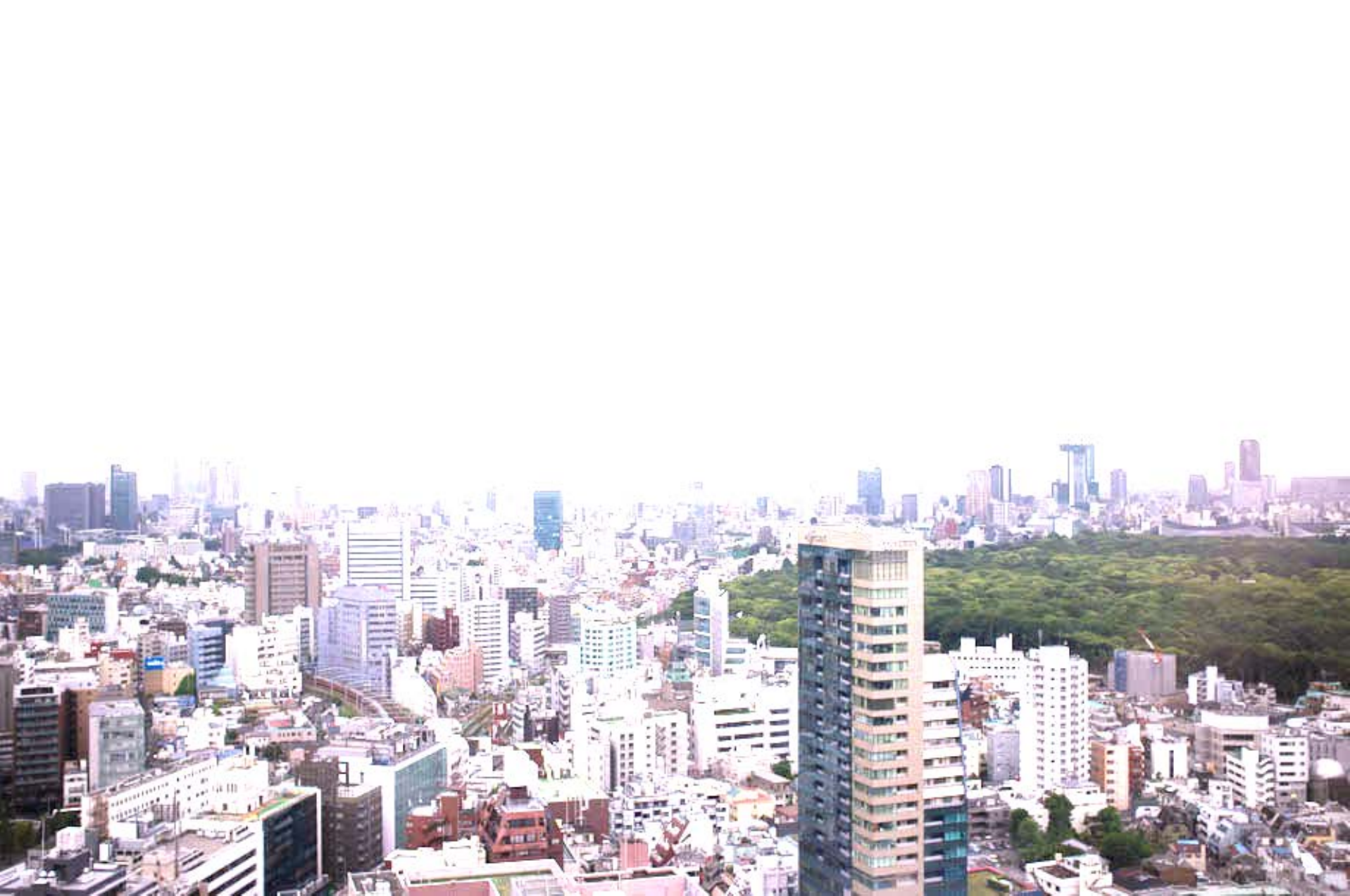}&
			\includegraphics[width=0.19\linewidth,height=0.13\linewidth]{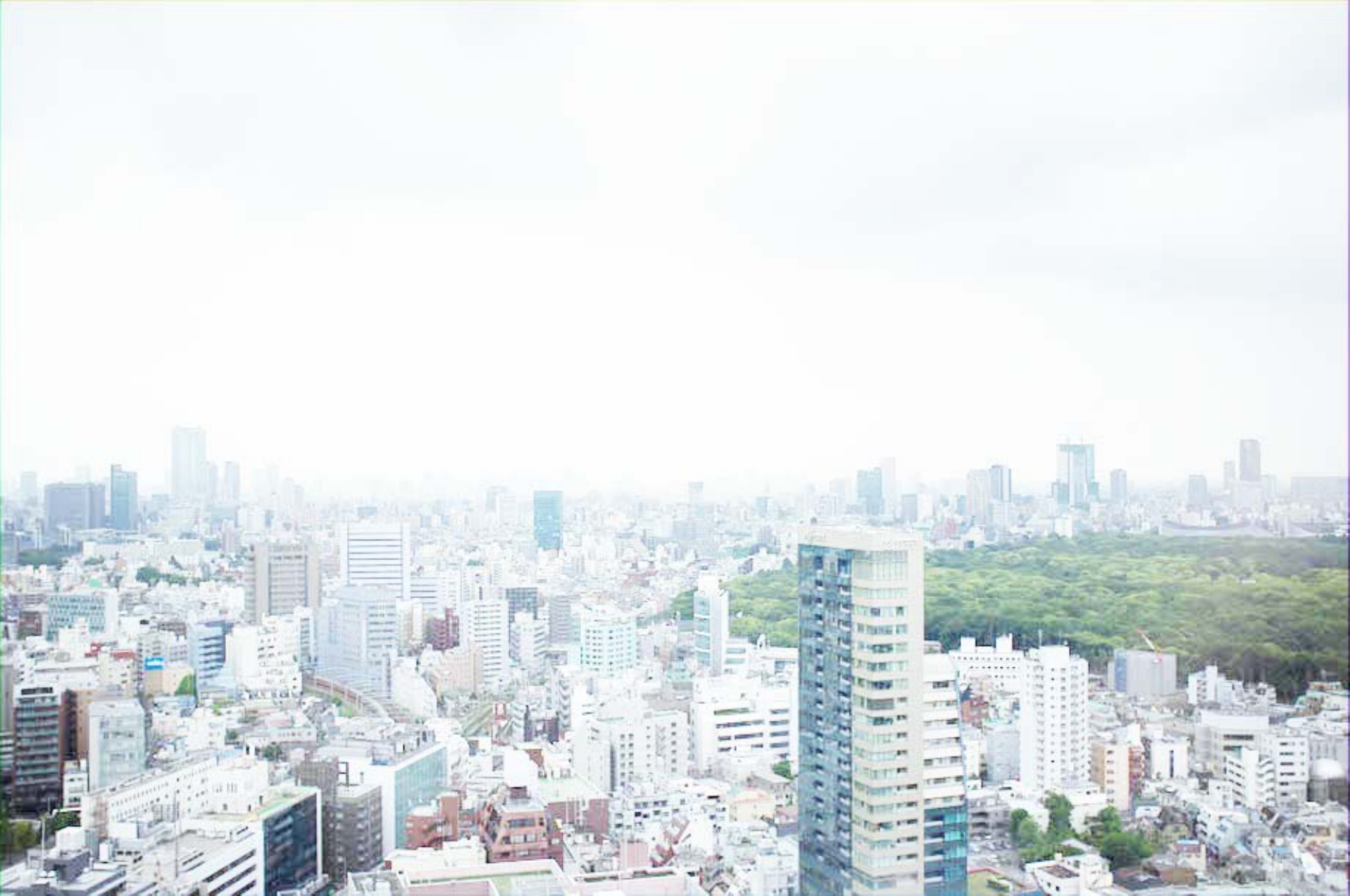}&
			\includegraphics[width=0.19\linewidth,height=0.13\linewidth]{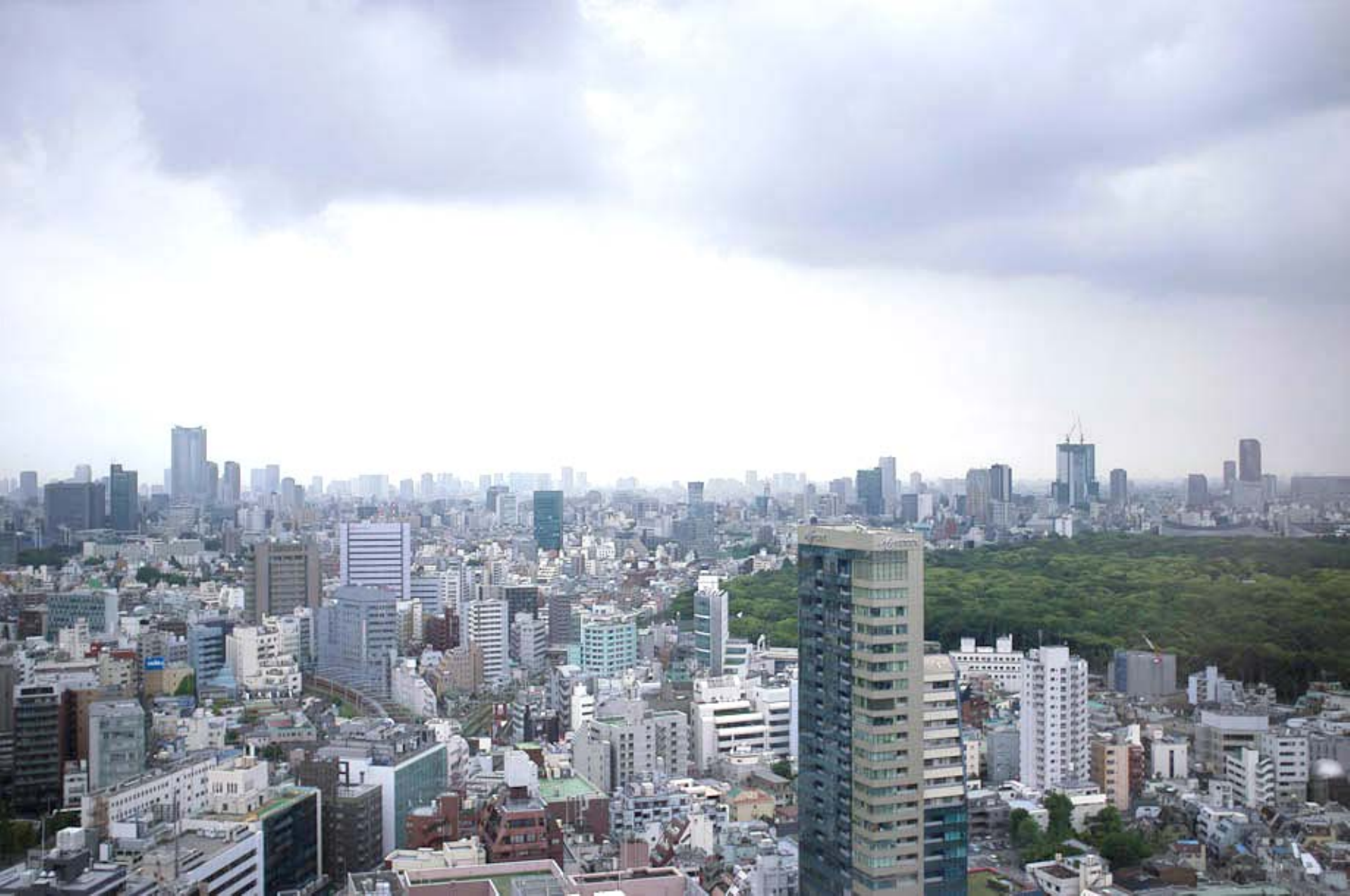}\\
			\includegraphics[width=0.19\linewidth,height=0.13\linewidth]{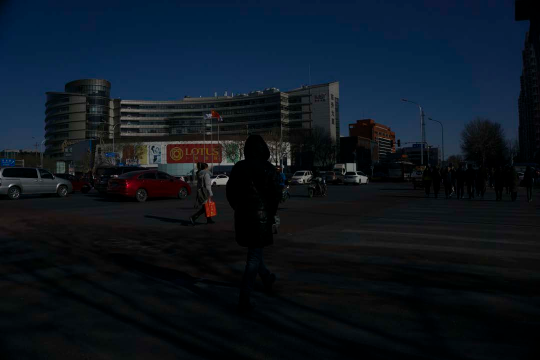}&
			\includegraphics[width=0.19\linewidth,height=0.13\linewidth]{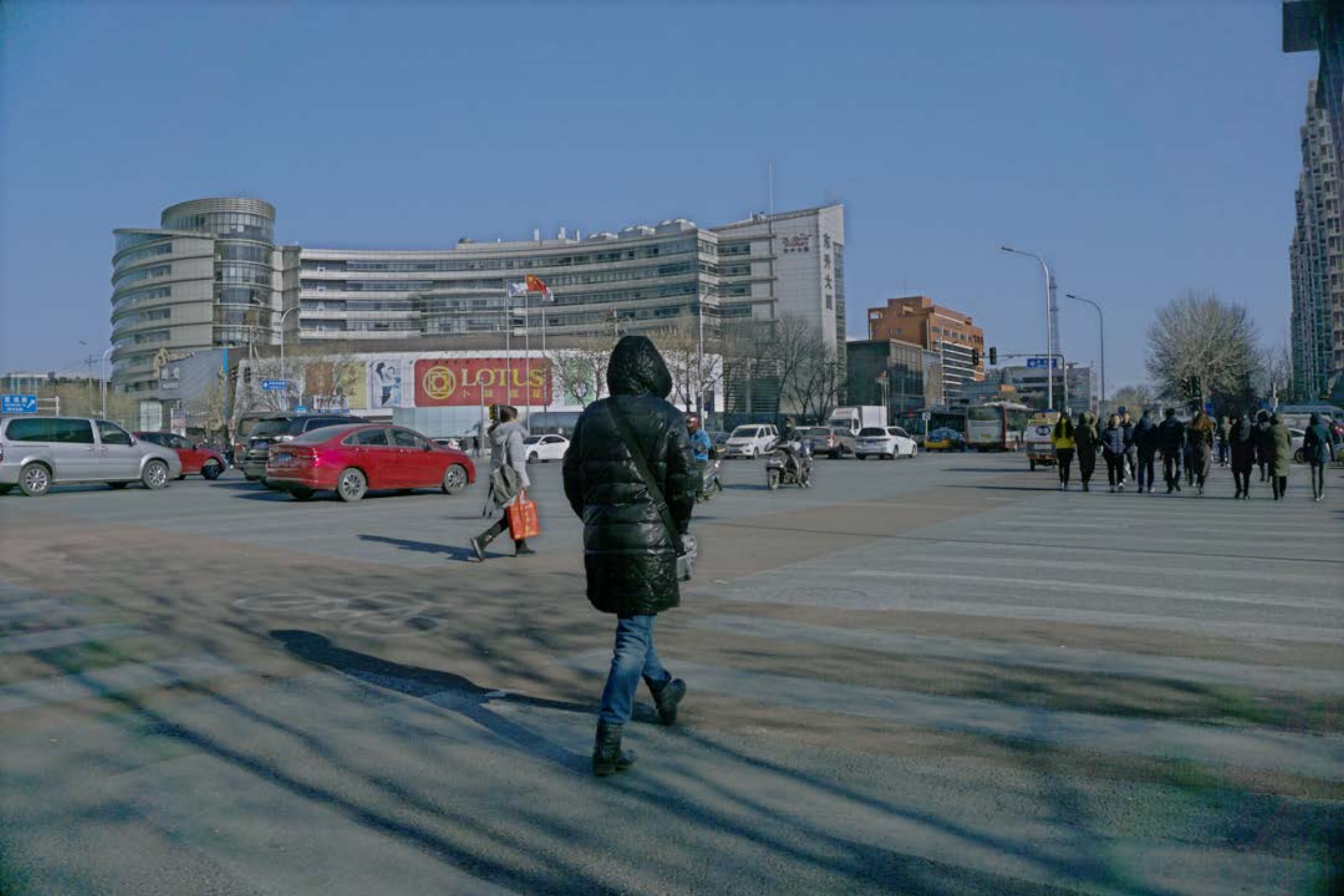}&
			\includegraphics[width=0.19\linewidth,height=0.13\linewidth]{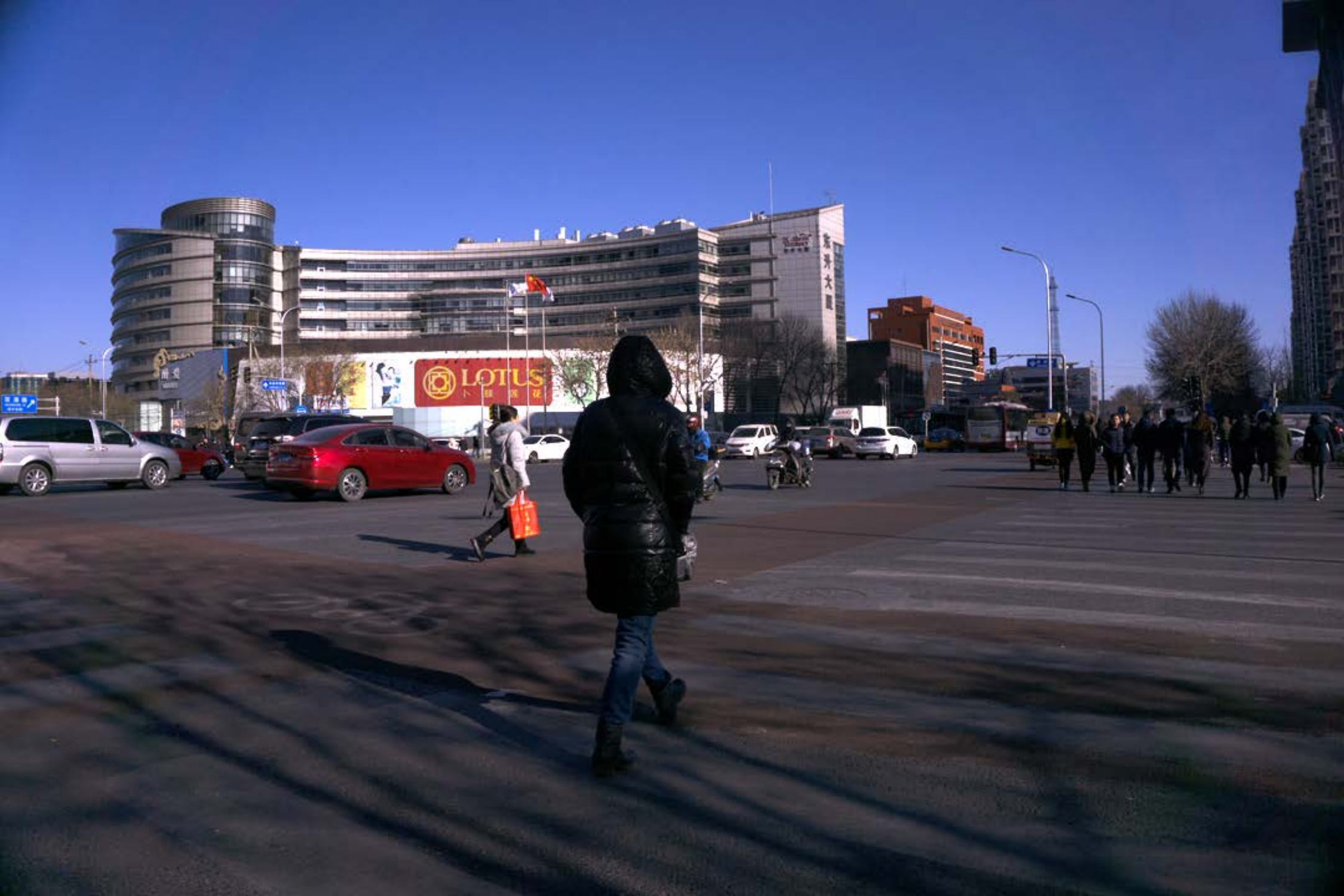}&
			\includegraphics[width=0.19\linewidth,height=0.13\linewidth]{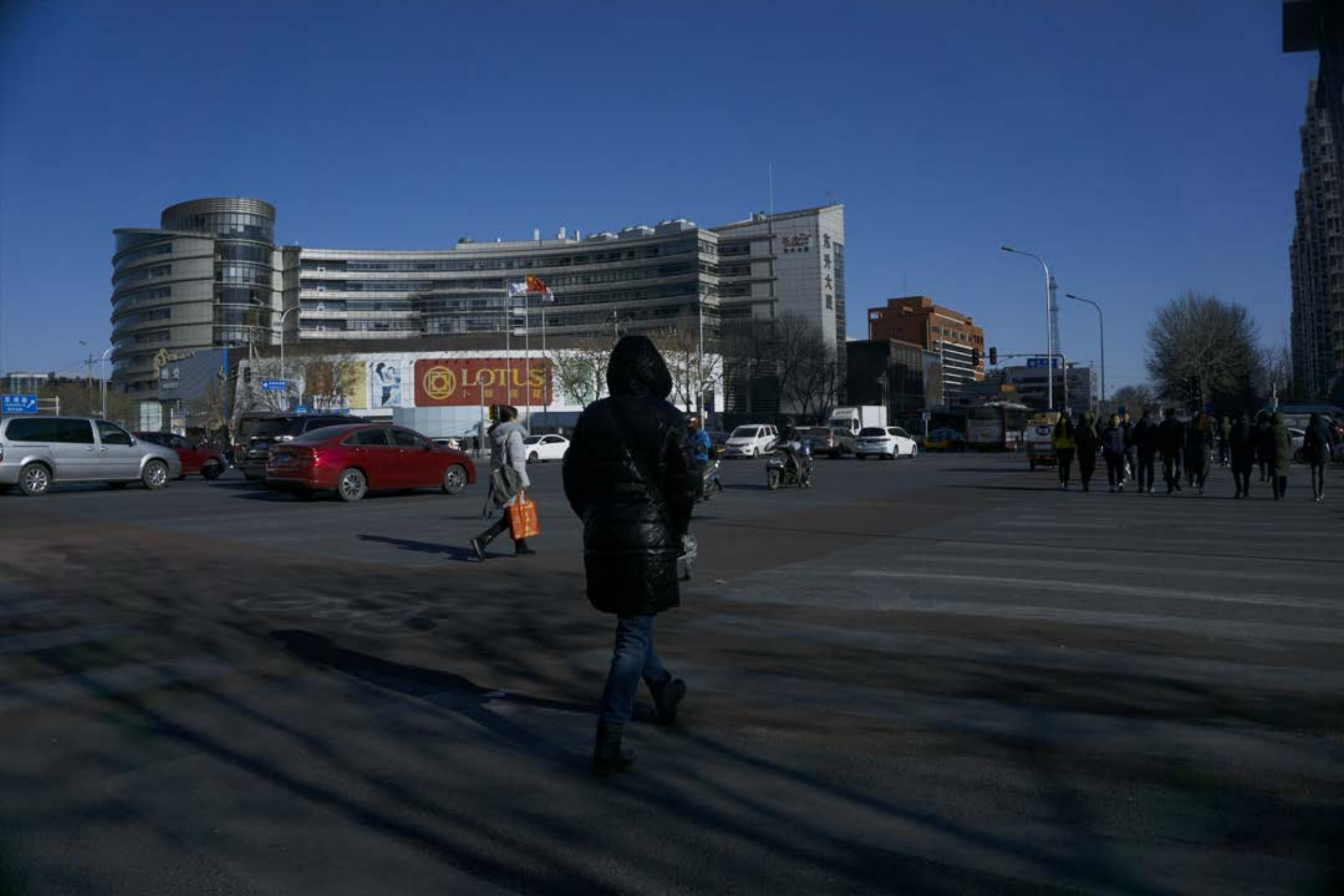}&
			\includegraphics[width=0.19\linewidth,height=0.13\linewidth]{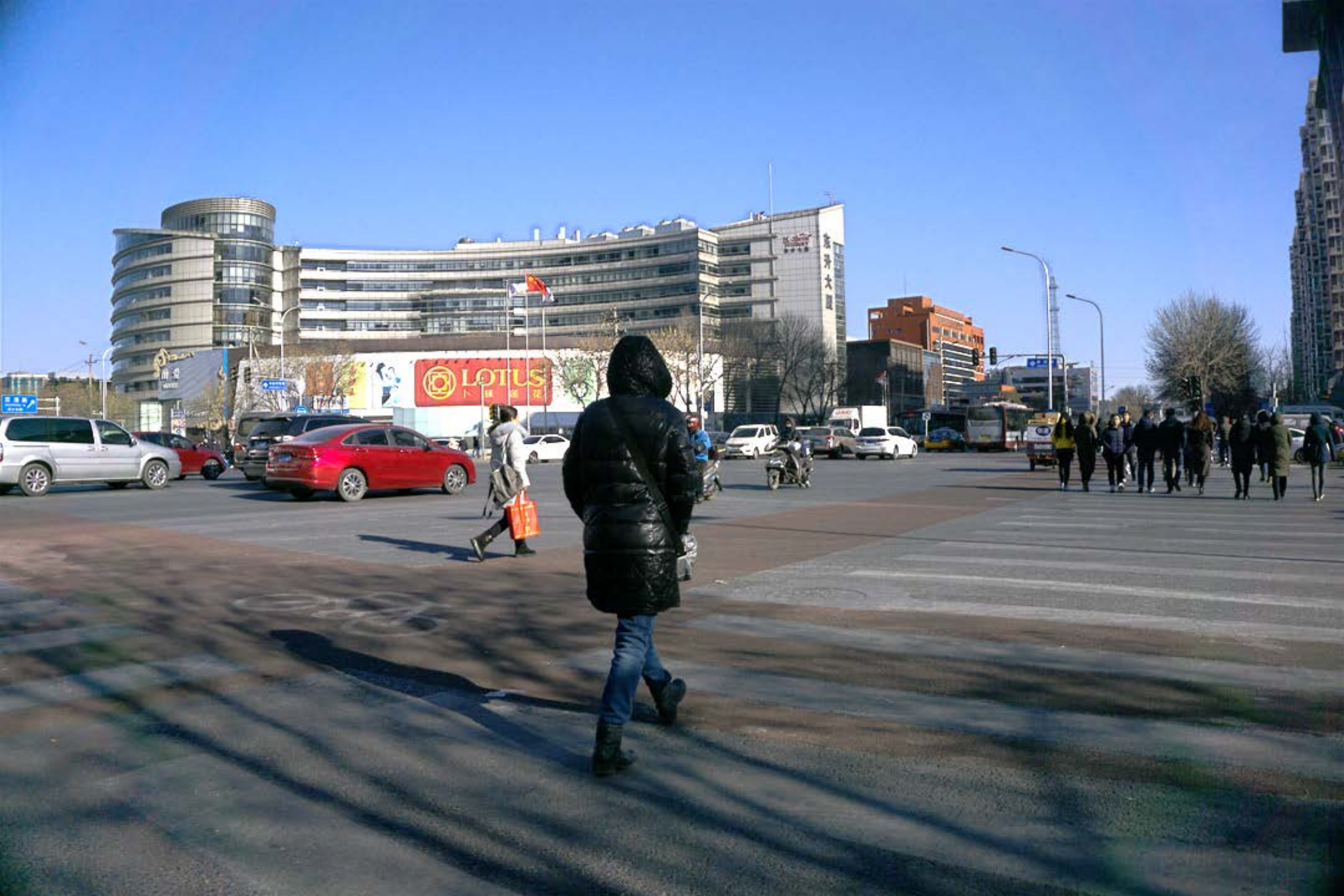}\\
			\includegraphics[width=0.19\linewidth,height=0.13\linewidth]{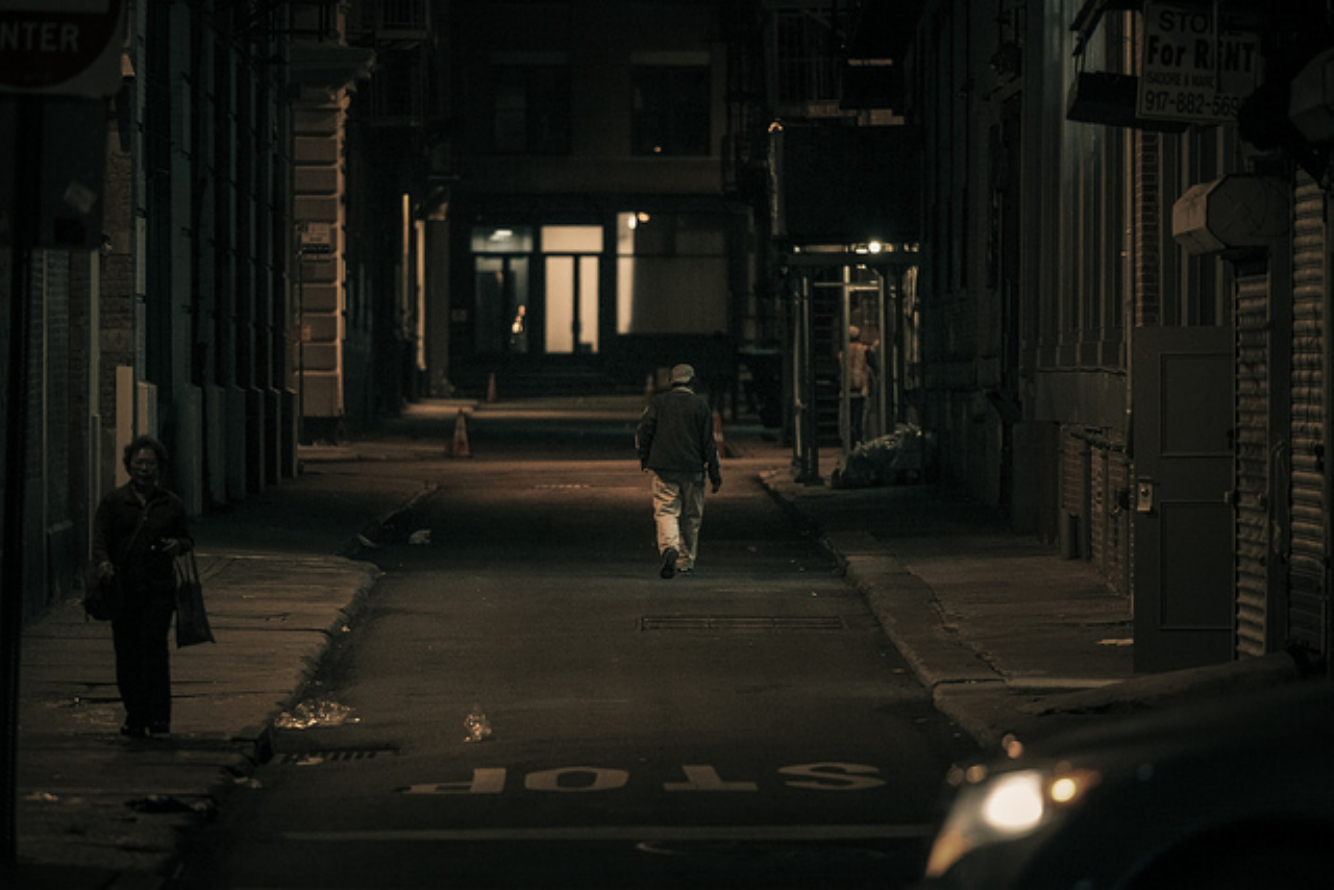}&
			\includegraphics[width=0.19\linewidth,height=0.13\linewidth]{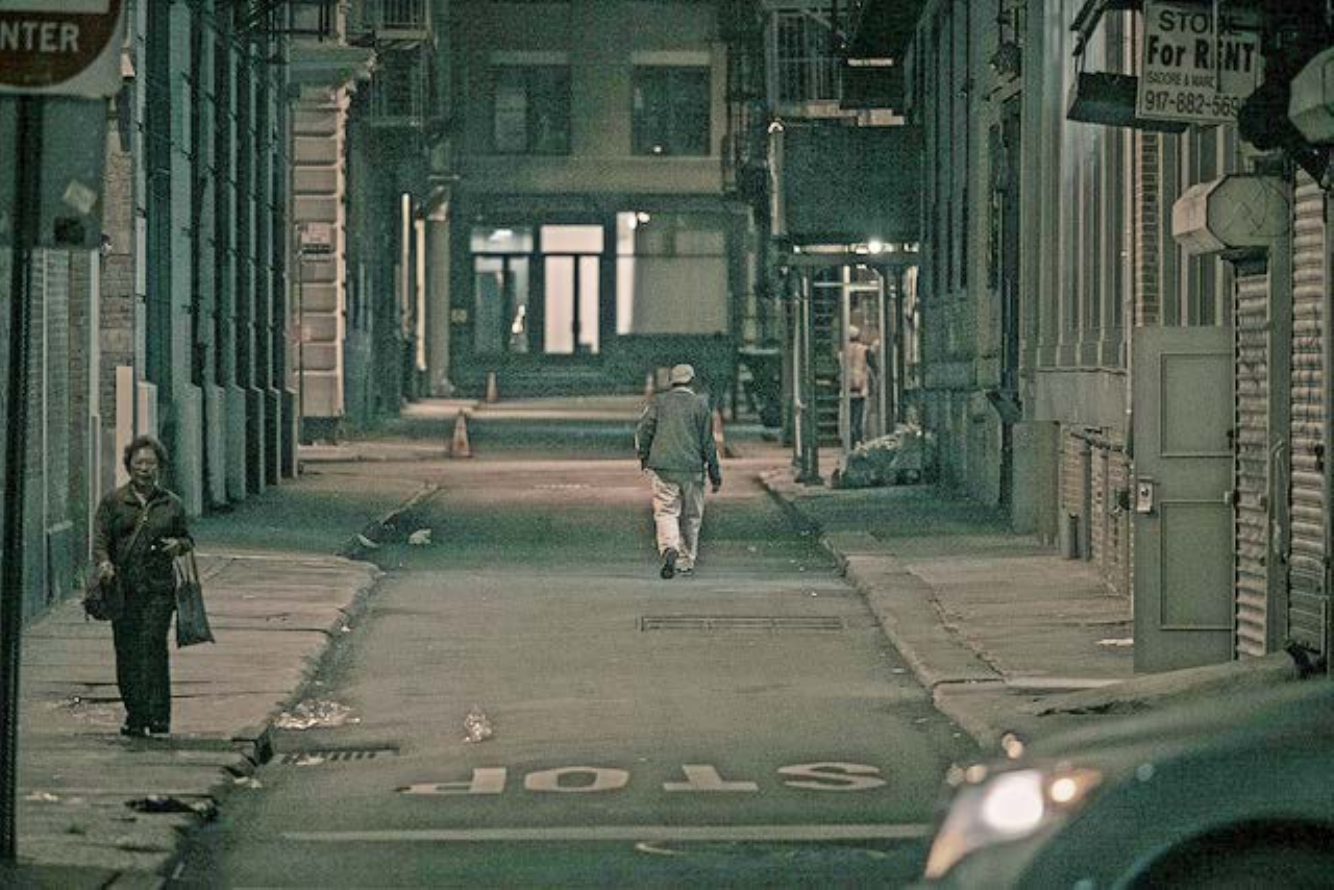}&
			\includegraphics[width=0.19\linewidth,height=0.13\linewidth]{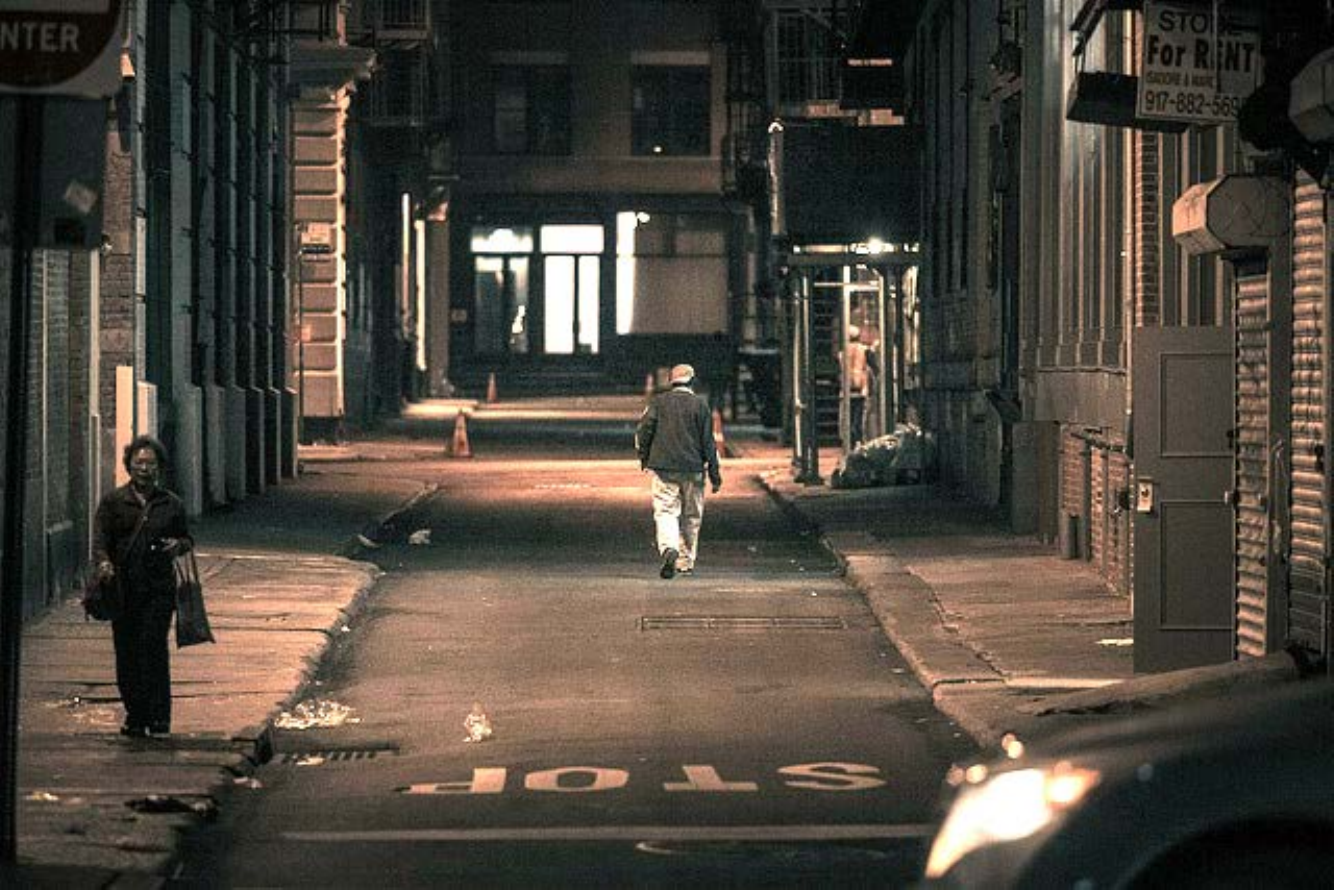}&
			\includegraphics[width=0.19\linewidth,height=0.13\linewidth]{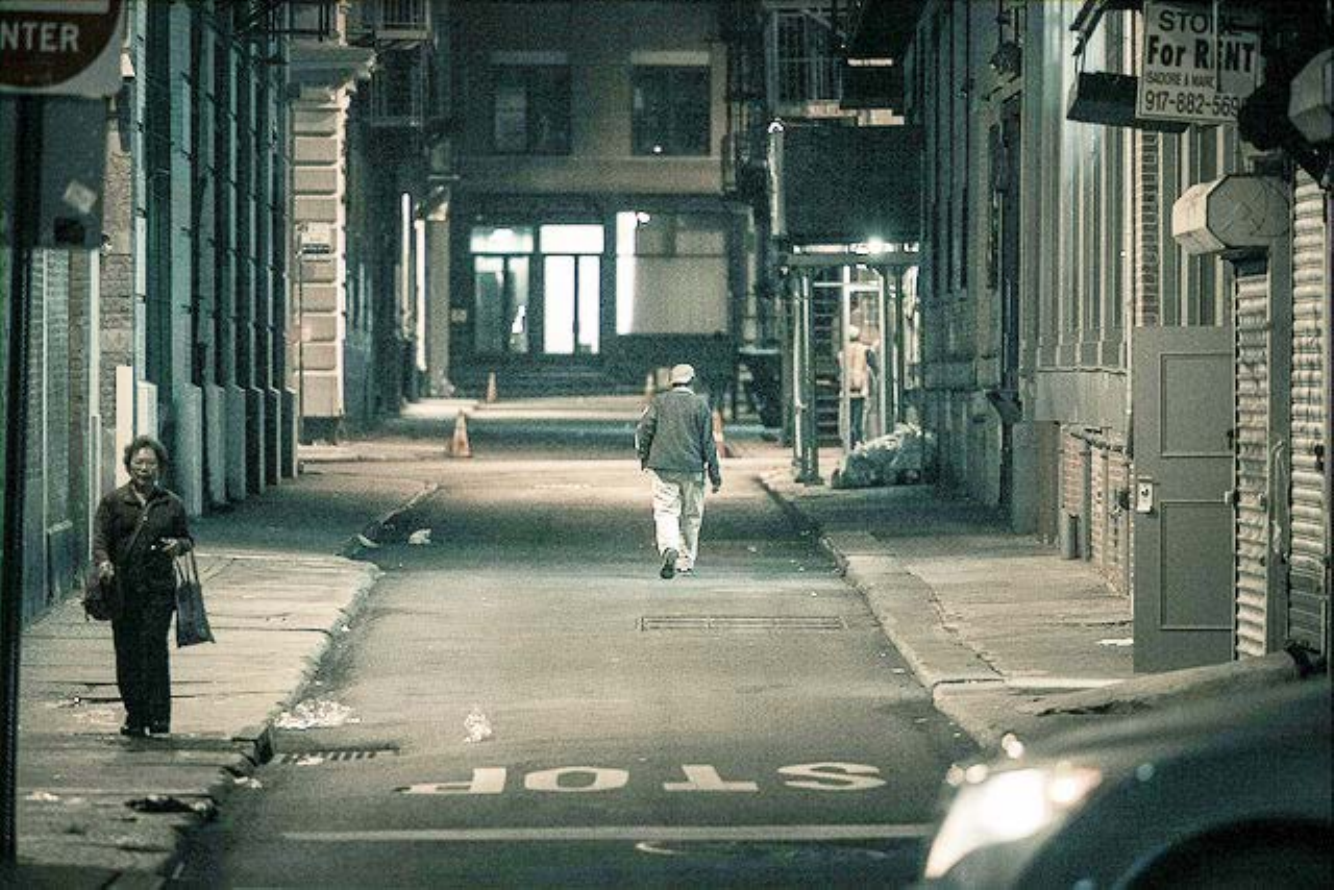}&
			\includegraphics[width=0.19\linewidth,height=0.13\linewidth]{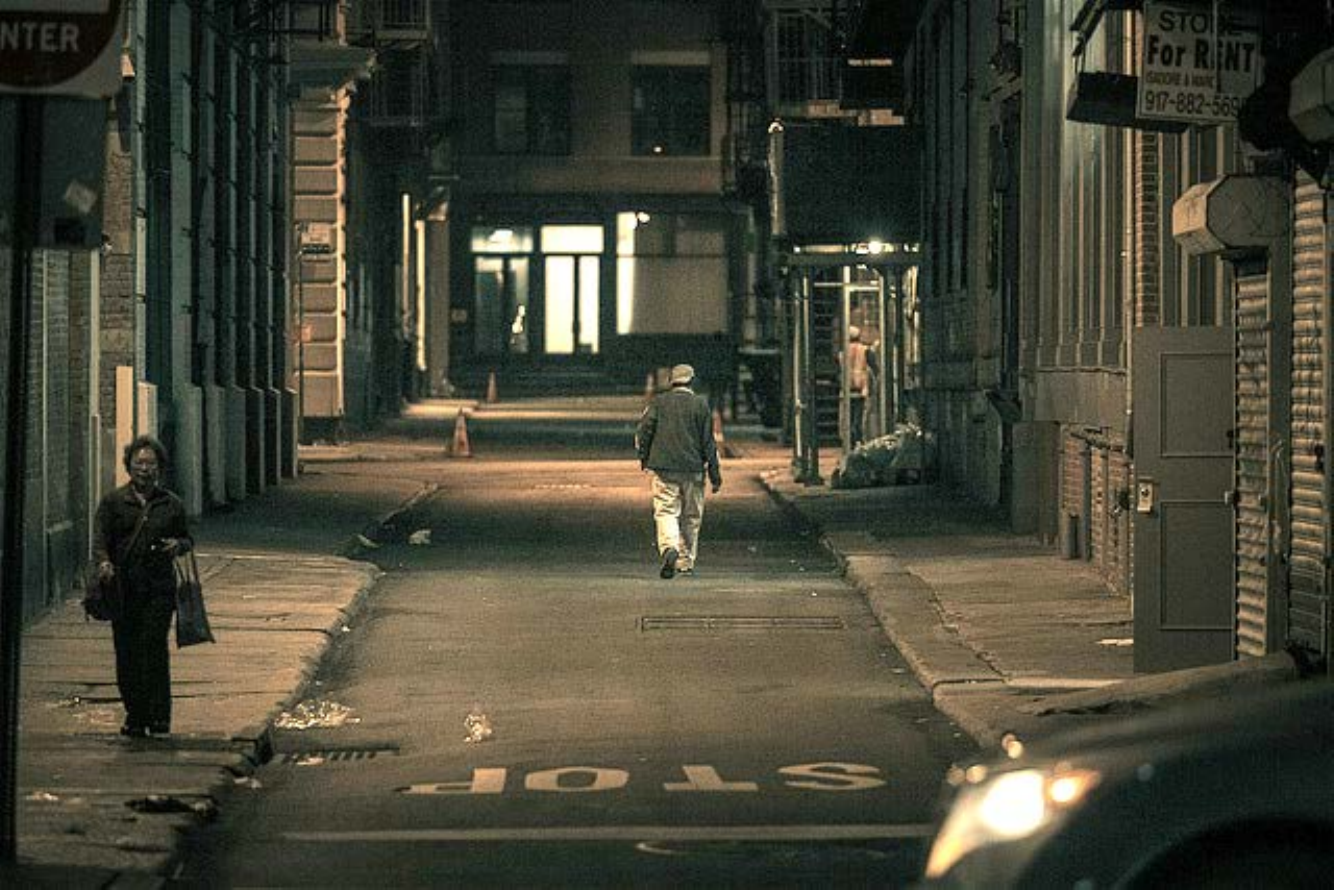}\\
			\includegraphics[width=0.19\linewidth,height=0.13\linewidth]{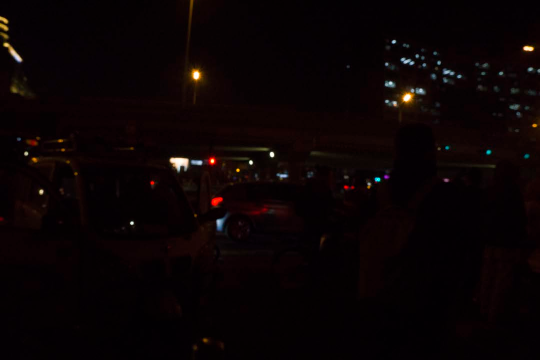}&
			\includegraphics[width=0.19\linewidth,height=0.13\linewidth]{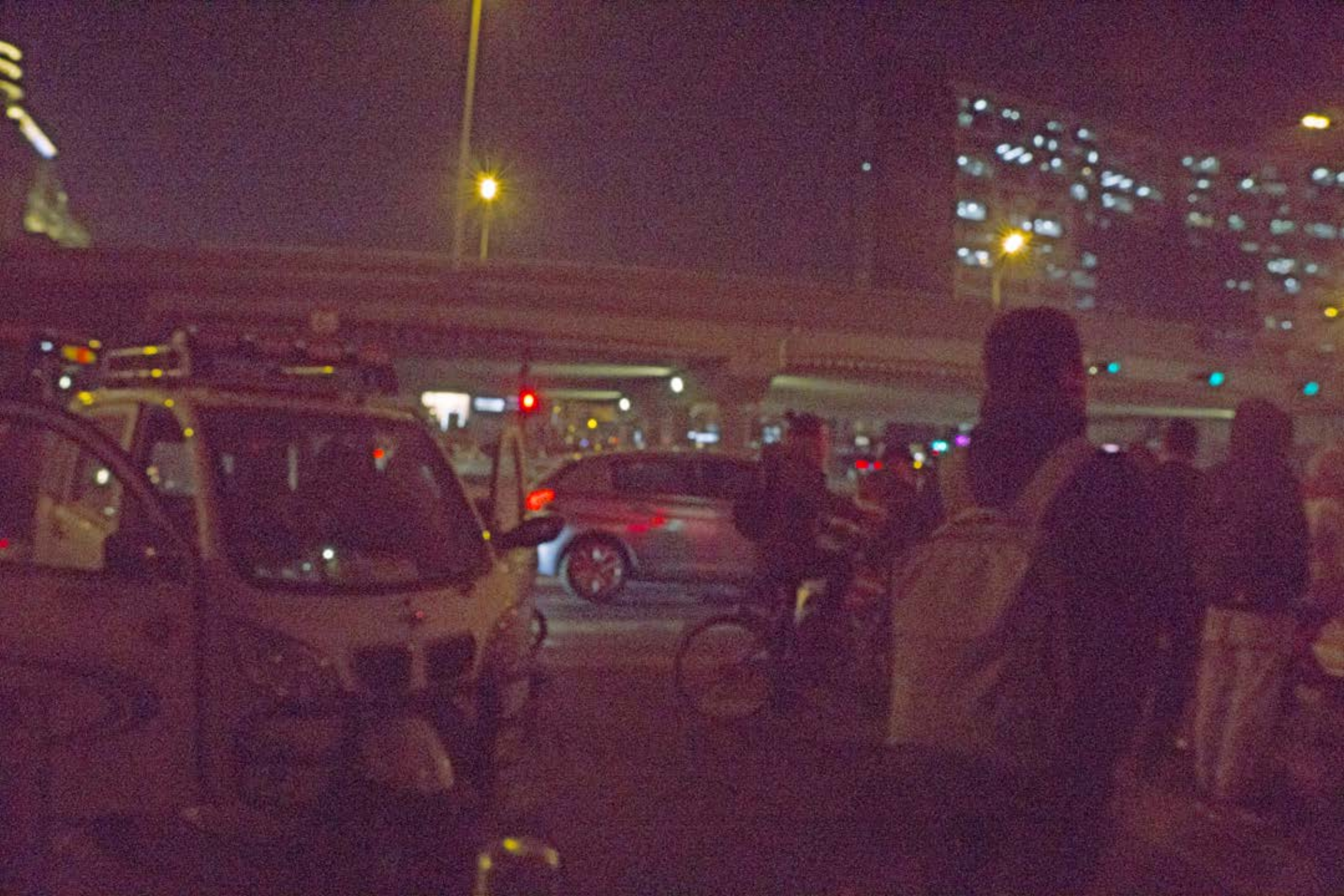}&
			\includegraphics[width=0.19\linewidth,height=0.13\linewidth]{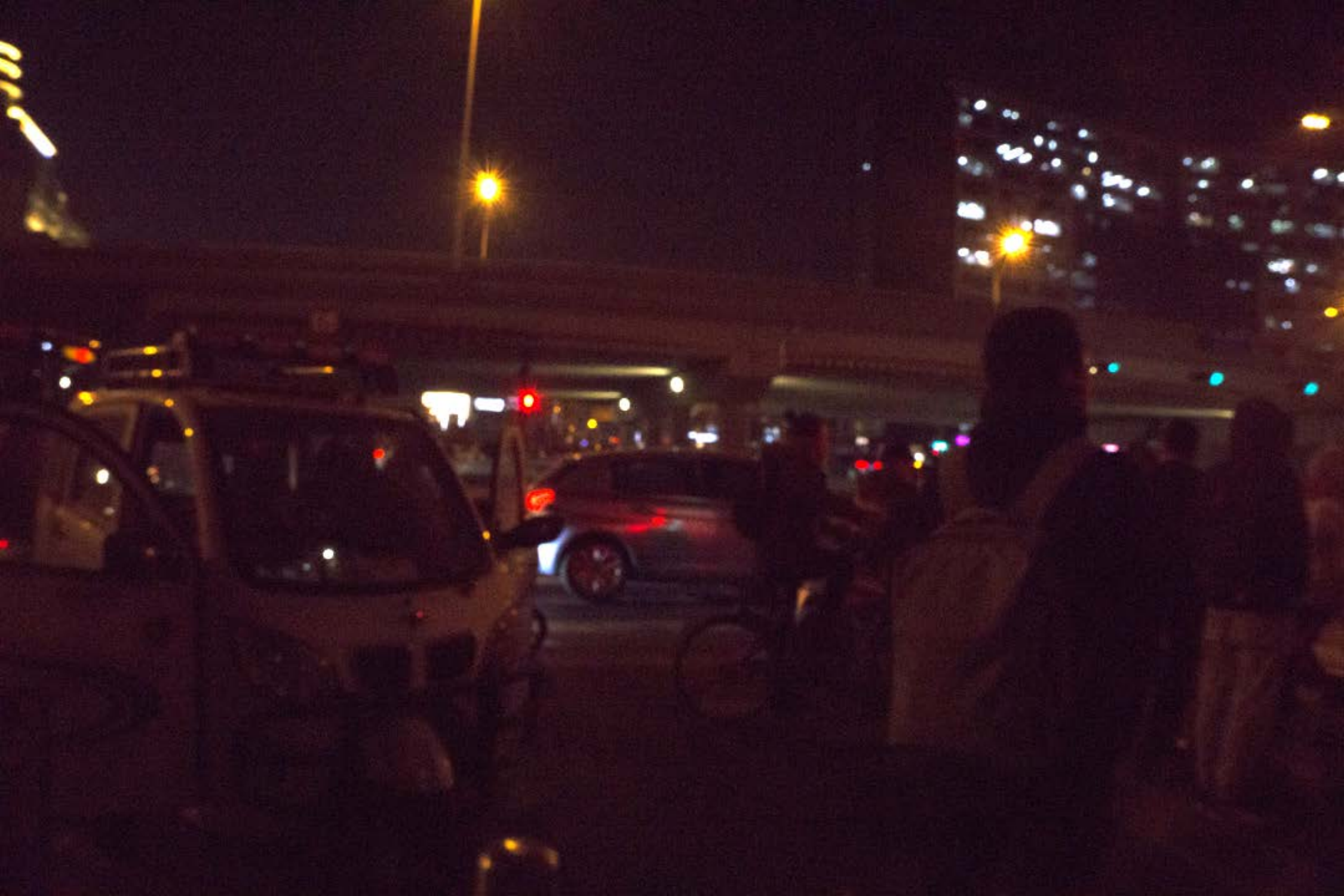}&
			\includegraphics[width=0.19\linewidth,height=0.13\linewidth]{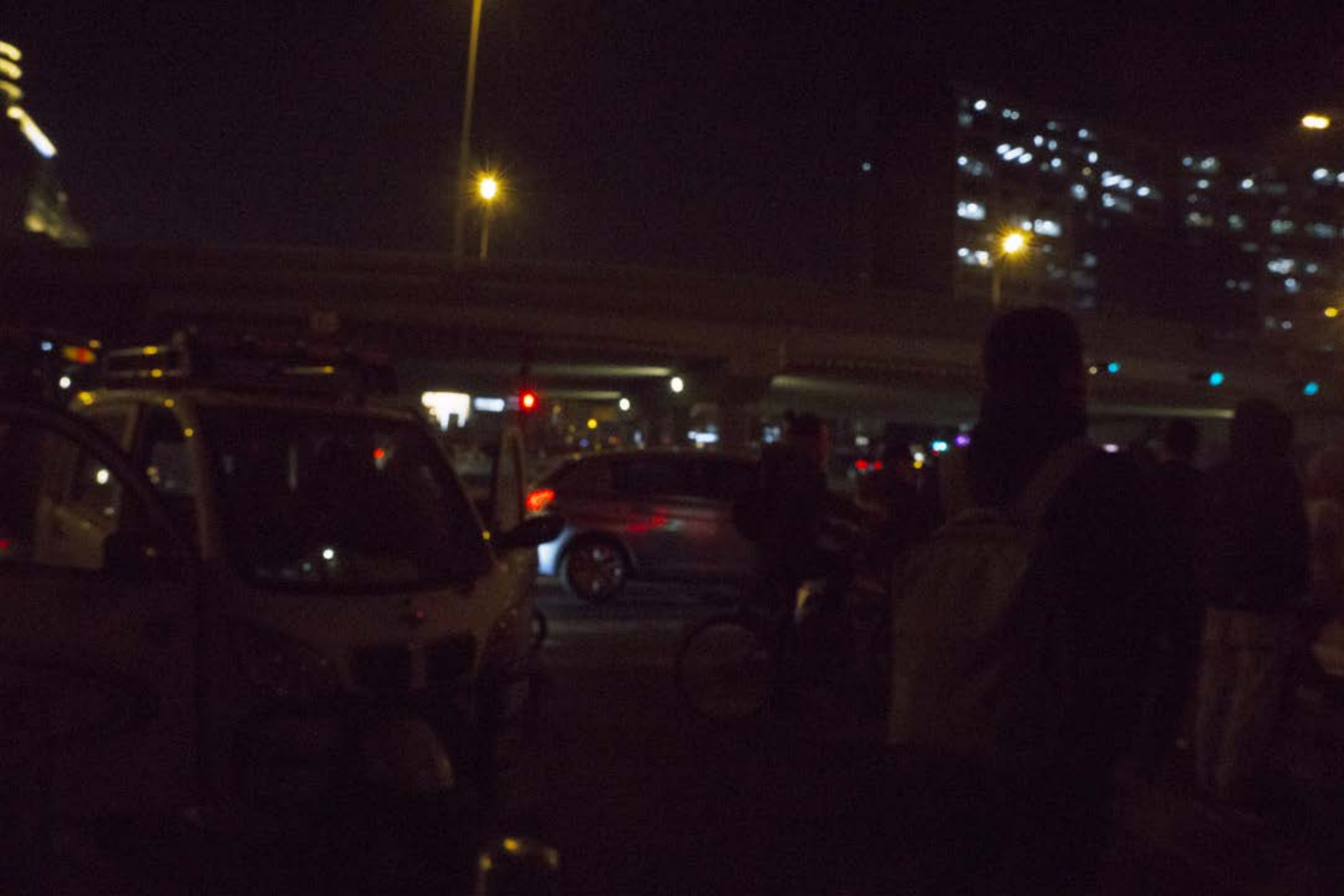}&
			\includegraphics[width=0.19\linewidth,height=0.13\linewidth]{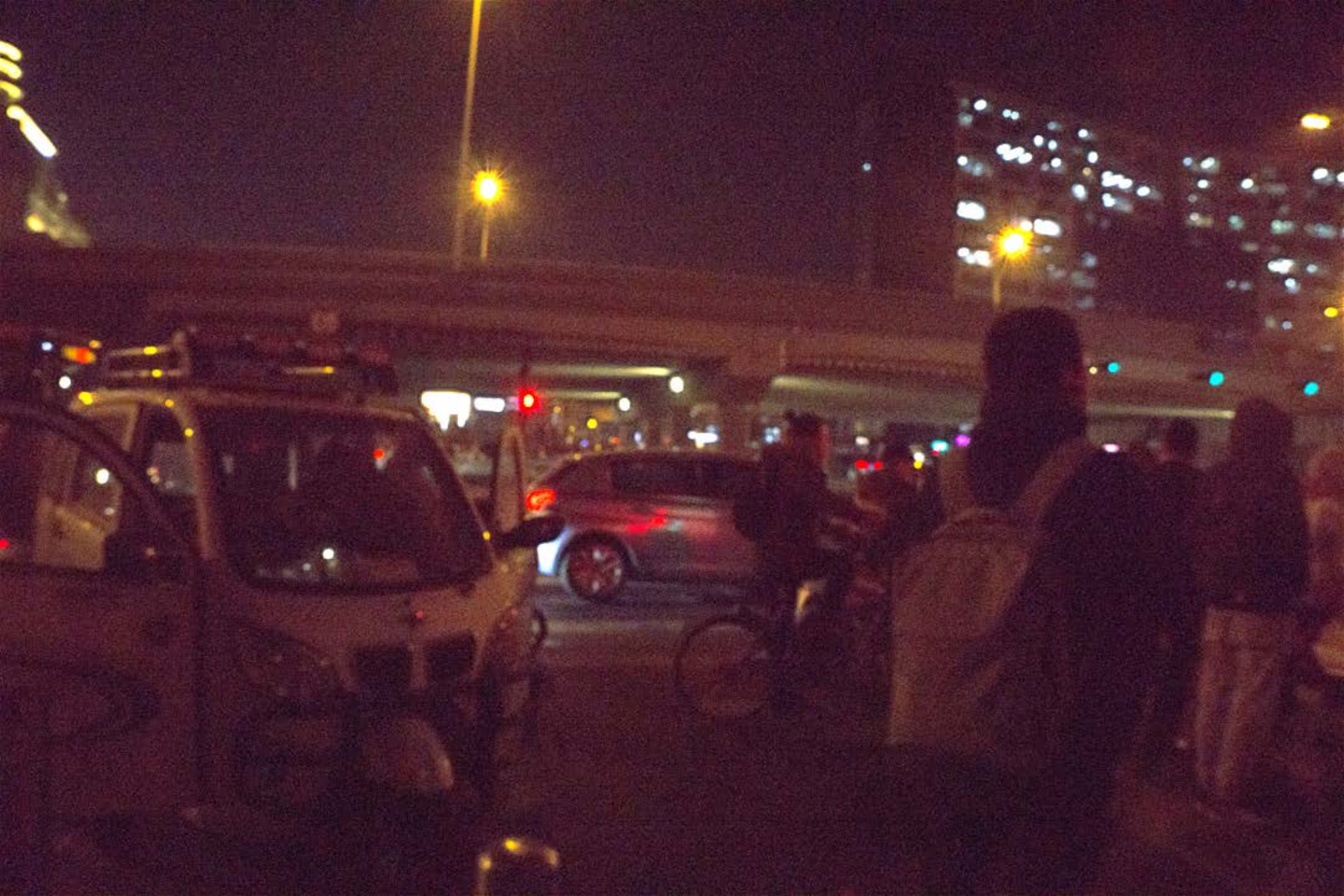}\\
			\footnotesize Input&\footnotesize DCE&\footnotesize RUAS&\footnotesize SCI&\footnotesize Ours\\\vspace{-0.4cm}
		\end{tabular}
		\caption{{Visual comparison with state-of-the-art methods on more low-light scenarios.}}
		\label{fig:in the wild}
	\end{figure*}

	\begin{figure*}[htb!]
		\centering
		\begin{tabular}{c@{\extracolsep{0.5em}}c@{\extracolsep{0.5em}}c@{\extracolsep{0.5em}}c@{\extracolsep{0.5em}}c}
			\includegraphics[width=0.18\linewidth]{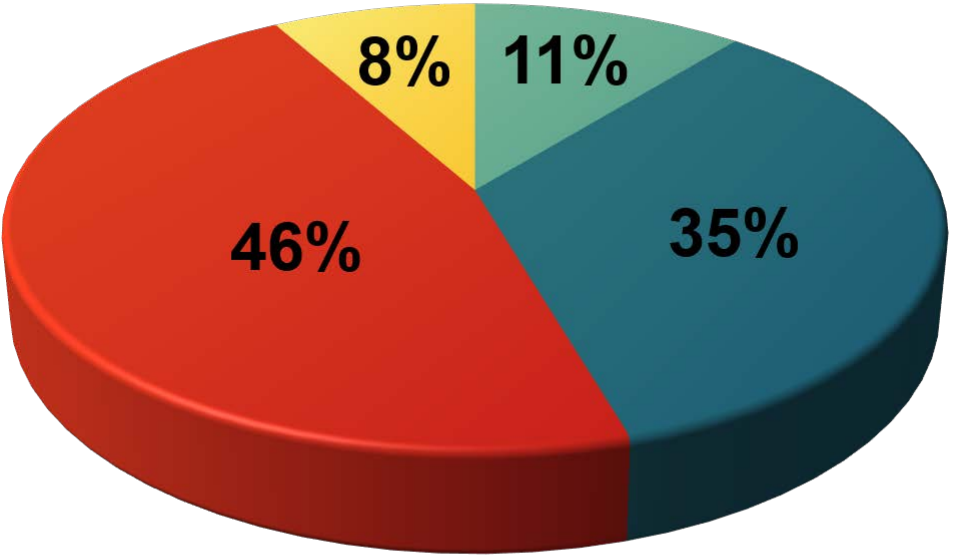}&
			\includegraphics[width=0.18\linewidth]{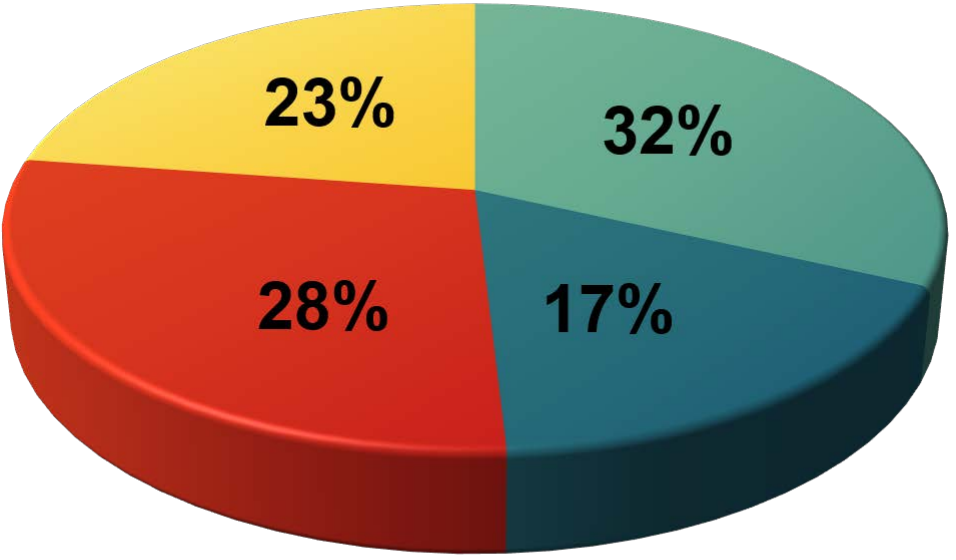}&
			\includegraphics[width=0.18\linewidth]{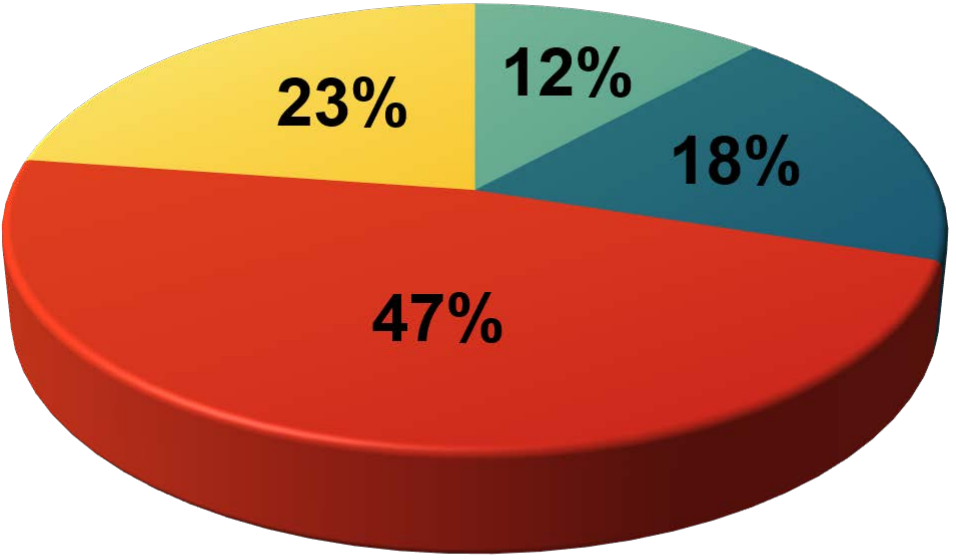}&
			\includegraphics[width=0.18\linewidth]{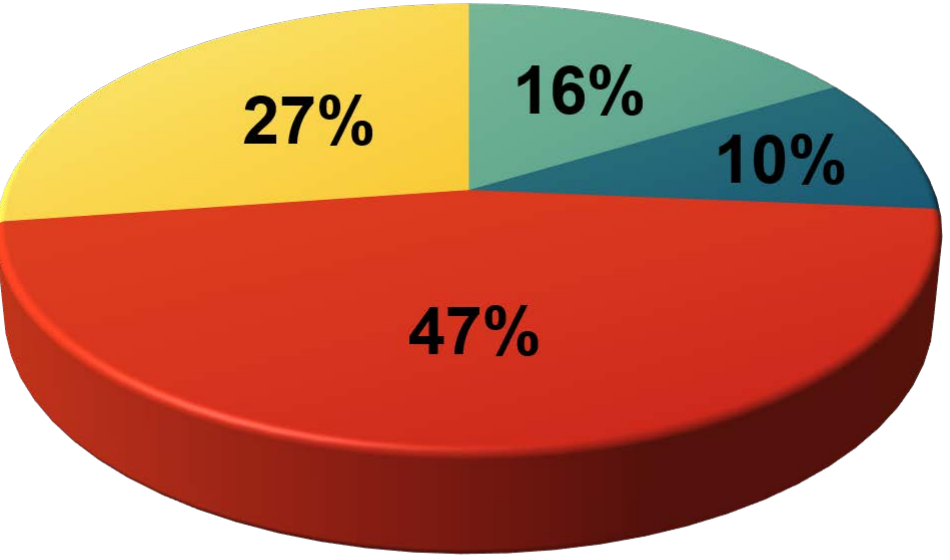}&
			\includegraphics[width=0.18\linewidth]{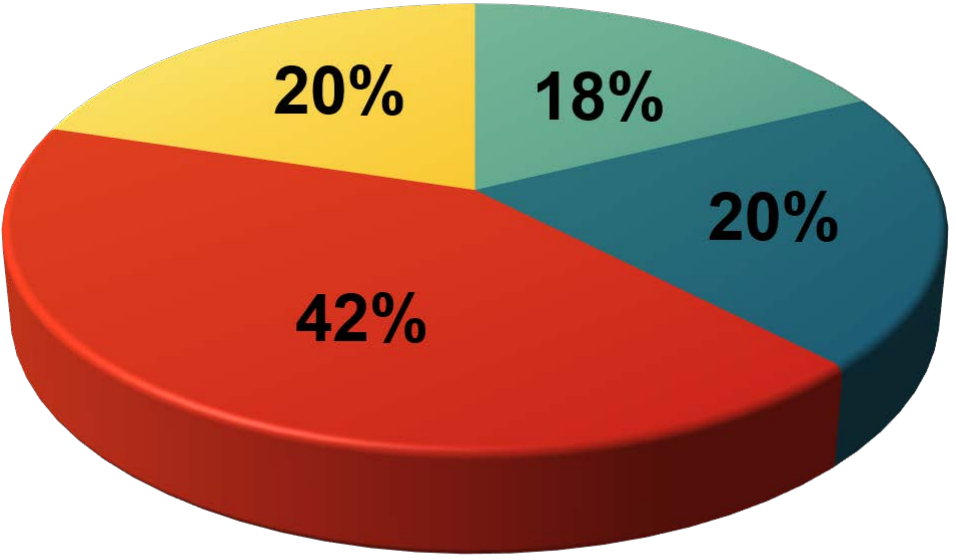}\\
			\footnotesize Brightness&\footnotesize Details&\footnotesize Color&\footnotesize Naturalness&\footnotesize Overall\\\vspace{-0.4cm}
		\end{tabular}
		
		\caption{Pie charts of average rates on different subjective score (including brightness, detail, color, natureness and overall rating). In which, DCE, RUAS, SCI and our method are denoted by green, blue, yellow, and red, respectively.}
		\label{fig:user study} 
	\end{figure*}
	
	\begin{table*}[ht]
		\footnotesize
		\setlength{\tabcolsep}{4.4mm}
		\renewcommand\arraystretch{1.5}	
		\centering
		\caption{{Quantitative comparison with advanced methods on the BAID dataset.}}
		\begin{tabular}{|c||ccccccc||c|}
			\hline
			{Method} &DCE&SCI&RUAS&RQ-LLIE&PairLIE&NeRCo&ZeroIG &{Ours}\\
			\hline
			PSNR$\uparrow$&21.2893&13.7054&14.6746&18.7379&16.8073&\textcolor{blue}{\textbf{21.3949}}&11.3921&\textcolor{red}{\textbf{22.8127}}\\ 
			SSIM$\uparrow$&\textcolor{blue}{\textbf{0.9016}}&0.7994&0.8056&0.8510&0.8387&0.8854&0.6995&\textcolor{red}{\textbf{0.9018}}\\
			LOE$\downarrow$&231.2081&266.4429&918.2778&\textcolor{red}{\textbf{31.0329}}&308.5668&171.1638&1692.79&\textcolor{blue}{\textbf{117.7952}}\\ \hline
		\end{tabular}
		\label{table:BAID}
	\end{table*}
	
	\subsubsection{Challenging Low-Light Scenes}
	In real-world low-light environments, challenges posed by unexpected noise and severe underexposure are significant and cannot be ignored. Investigating the performance of enhancers in such environments is crucial for a comprehensive performance evaluation. Here, as our method focuses on improving the visual quality of low-light images in terms of brightness and color, for a better evaluation on the LOL dataset, which includes significant noise, we introduce a recently proposed denoiser named SCUNet~\cite{zhang2023practical} as a post-processing step. Additionally, all compared methods employ the same post-processing operation as AR-LLIE to ensure fairness.  The lower section of Table~\ref{table:quan} provides quantitative comparison results on the LOL dataset, it can be observed that our method still outperforms other compared methods significantly in most numerical metrics. 
	
	In terms of visual results, Fig.~\ref{fig:lol2} provides comparative results on samples with significant noise from the LOL dataset. It can be observed that, under the fair condition of adding denoising post-processing to all methods, our method's results show more appropriate brightness and good visual effects compared to other methods. Some methods, such as KinD, still exhibit significant noise. The reason for this phenomenon may be that the enhancement process of this method might have amplified the noise to some extent, leading to unsatisfactory results even with the introduction of additional denoisers.
	

	\begin{table*}[htb!]
		\footnotesize
		\setlength{\tabcolsep}{1.9mm}
		\renewcommand\arraystretch{1.5}	
		\centering
		\caption{Efficiency on different platforms. CPU: Intel(R) Xeon(R) W-2135 CPU@3.70GHz. GPU: Geforce RTX 2080 Ti GPU.}
		\begin{tabular}{|c|c||ccccccccc||c|}
			\hline
			Platform&Size&UTVNet&RCTNet&DCE&SCI&RUAS&PairLIE&NeRCo&RQ-LLIE&ZeroIG&Ours\\
			\hline
			\multirowcell{3}{CPU\\Time}&1280$\times$720&4.18054&1.21133&33.6015&\underline{0.06641}&0.69634&7.49217&119.087&164.293&1.49291&\textbf{0.01572}\textcolor{red}{\(_{\uparrow \textbf{76.36\%}}\)}\\
			&1920$\times$1080&9.43331&1.25057&55.1763&\underline{0.14525}&1.69749&14.7476&215.811&536.284&3.43526&\textbf{0.03192}\textcolor{red}{\(_{\uparrow \textbf{78.05\%}}\)}\\
			&2560$\times$1440&16.8201&2.69246&125.331&\underline{0.34452}&3.72713&24.2392&359.129&632.791&5.94202&\textbf{0.10294}\textcolor{red}{\(_{\uparrow \textbf{70.13\%}}\)}\\
			\hline
			\multirowcell{3}{GPU\\Time}&1280$\times$720&0.74555&0.34751&0.02310&\underline{0.00026}&0.01834&0.07688&3.12394&2.82367&0.04463&\textbf{0.00017}\textcolor{red}{\(_{\uparrow \textbf{33.33\%}}\)}\\
			&1920$\times$1080&0.82506&0.39062&0.05476&\underline{0.00083}&0.04084&0.17544&---&---&0.10112&\textbf{0.00020}\textcolor{red}{\(_{\uparrow \textbf{75.00\%}}\)}\\
			&2560$\times$1440&0.94271&0.43460&0.09463&\underline{0.00217}&0.07258&0.31439&---&---&0.18170&\textbf{0.00023}\textcolor{red}{\(_{\uparrow \textbf{90.90\%}}\)}\\
			\hline
		\end{tabular}
		\label{table:efficiency}
	\end{table*}
	\begin{table*}[htb!]
		\footnotesize
		\renewcommand\arraystretch{1.5} 	
		\setlength{\tabcolsep}{2.3mm}
		\centering
		\caption{Comparing running time on the mobile devices. The computational platforms include Snapdragon 865 DSP, Kirin 990 NPU and Snapdragon 8 Gen 3 NPU. Here we choose 3 representative methods from Table~\ref{table:efficiency} for comparisons.  The best results are \textbf{bolded}.} 
		\begin{tabular}{|c||ccc|ccc|ccc|}
				\hline
				\multirow{2}*{Method} & \multicolumn{3}{c|}{Snapdragon 865 DSP}
				& \multicolumn{3}{c|}{Kirin 990 NPU} & \multicolumn{3}{c|}{Snapdragon 8 Gen 3 NPU}\\
				\cline{2-10}
				& 1280$\times$720 &1920$\times$1080 & 2560$\times$1440 & 1280$\times$720 &1920$\times$1080 & 2560$\times$1440  & 1280$\times$720 &1920$\times$1080 & 2560$\times$1440\\
				\hline
				{DCE} & 8.0520 & 23.3810 & 43.3570 & 2.2020 & 4.8150 & 10.2300 & 1.0670 & 2.4370 & 6.3800\\
				{RUAS} & 1.8600 & 4.0980 & 7.4570 & 1.7140 & 4.5780 & 9.1220 & 0.6250 & 1.8030 & 4.3480\\
				{SCI} & \underline{0.0284} & \underline{0.0730} & \underline{0.2050} & \underline{0.0489} & \underline{0.0846} & \underline{0.1970}  & \underline{0.0126} & \underline{0.0357} & \underline{0.0723}\\
				{ZeroIG} & {4.8852} & {10.1467} & {18.1752} & {8.3767} & {11.3476} & {19.6326}  & {2.1725} & {4.4142} & {11.2603}\\
				{Ours} &\textbf{0.0156}\color{red}{\tiny  \textbf{$\uparrow${45\%}}} &\textbf{0.0403}\color{red}{\tiny  \textbf{$\uparrow${45\%}}} &\textbf{0.1029}\color{red}{\tiny  \textbf{$\uparrow${50\%}}} &\textbf{0.0270}\color{red}{\tiny  \textbf{$\uparrow${45\%}}} &\textbf{0.0510}\color{red}{\tiny  \textbf{$\uparrow${40\%}}} &\textbf{0.1040}\color{red}{\tiny  \textbf{$\uparrow${47\%}}} &\textbf{0.0058}\color{red}{\tiny  \textbf{$\uparrow${54\%}}} &\textbf{0.0159}\color{red}{\tiny  \textbf{$\uparrow${55\%}}} &\textbf{0.0306}\color{red}{\tiny  \textbf{$\uparrow${57\%}}}\\
				\hline 
			\end{tabular}	
			
			\label{table:DSP/NPU time}
		\end{table*}
	 \begin{table}[h]
		\centering
		\footnotesize
		\renewcommand{\arraystretch}{1.5}
		\setlength{\tabcolsep}{2pt}
		\caption{{Comparison of computational efficiency between our method and representative efficient methods. Note that FLOPs are calculated on 2560$\times$1440 resolution images.}}
		\begin{tabular}{|c|ccccc|c|}
			\hline
			Method &DCE&SCI&RUAS&PairLIE&ZeroIG&{Ours}\\ \hline
			FLOPs(G)$\downarrow$&291.9628&\underline{1.9685}&12.0435&9116.15&1823.62&\textbf{0.2986}\\ 
			Model Size(M)$\uparrow$&0.3029&\underline{0.0013}&0.01311&1.3037&0.4716&\textbf{0.0003}\\
			\hline
		\end{tabular}
		\label{table: Modelsize}
	\end{table}
	
	{Furthermore, we conducted additional performance evaluations on the DARKFACE dataset, which consists of real night scenes captured in extremely dark conditions. In this dataset, most of the image information is often obscured by darkness, significantly increasing the difficulty of recovering clear images from low-light conditions. We present the qualitative comparison results in Fig.~\ref{fig:darkface}, where it is evident that the proposed AR-LLIE achieves more significant brightness enhancement compared to other methods. It recovers more scene information and provides richer texture details. {To provide a more comprehensive evaluation of our method, we also provided evaluation results between ours method and advanced LLIE method on the BAID dataset~\cite{lv2022backlitnet} (contains images with regions of varying brightness levels) and LOLi-Phone dataset~\cite{li2021low} (contains images with high contrast).} The corresponding results on the BAID dataset are shown in Table~\ref{table:BAID} and Fig.~\ref{fig:BAID}. It is clearly observed that our method exhibits significant advantages both in terms of numerical results and visual effects. Compared to other methods, AR-LLIE provides enhanced results with more accurate brightness control and more stable color restoration. The relevant results on the LOLi-Phone dataset are shown in Fig.~\ref{fig: LOLi}. It can be observed that other methods tend to amplify the color bias issues, resulting in unsatisfactory visual quality. In contrast, our method not only effectively enhances the brightness but also ensures some correction of the color bias, demonstrating our method's stability.}
	
	\begin{figure*}[!htb]
		\centering
		\footnotesize
		\begin{tabular}{c}
			\includegraphics[width=0.9\linewidth]{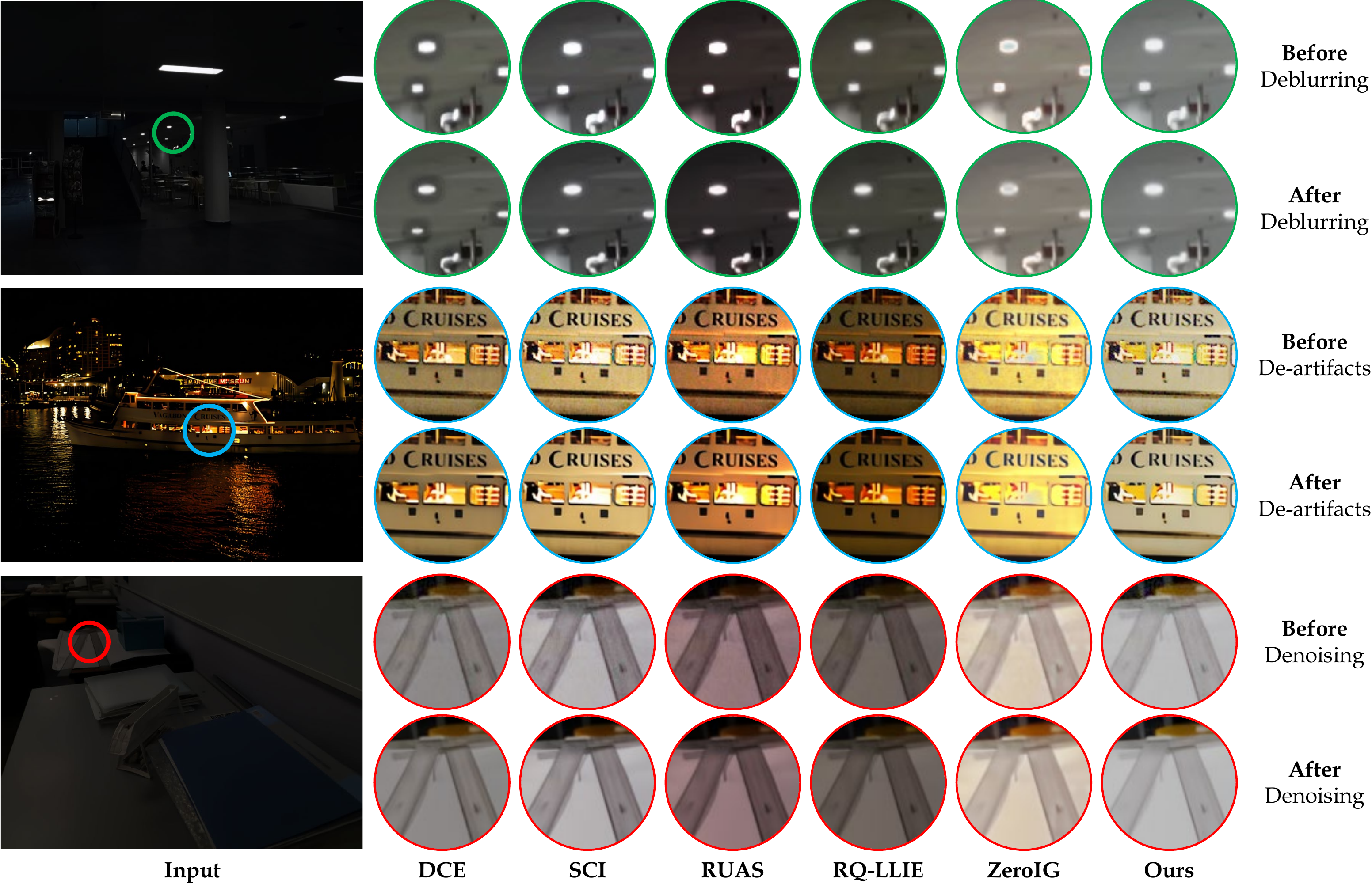}\\
		\end{tabular}
		\caption{{Exploring the impact of post-processing operations in challenging low-light scenarios.}}
		\label{fig: 3scenes}
	\end{figure*}
	
	{In fact, aside from the low-light scenarios discussed above, there are three other representative and challenging real-world scenarios: low-light scenes with motion blur, artifacts, and noise. {To conduct an in-depth analysis of these cases, we selected examples corresponding to these three scenarios from the LSRW~\cite{hai2023r2rnet} (with significant artifacts, noise, and color shift), LOL-Blur~\cite{zhou2022lednet} (with varying degrees of low light and blur), and ExDark datasets~\cite{loh2019getting} (extremely dark with significant noise) datasets for evaluation.} Fig.~\ref{fig: 3scenes} presents the results of our method compared to a range of representative approaches across these three scenarios. It can be observed that most existing representative methods also struggle to handle such conditions effectively. It is important to emphasize that the core contribution of our method lies in addressing brightness deficiency caused by environmental factors under the extreme condition of a single convolutional layer, rather than designing additional image quality enhancement schemes for specific scenarios. Moreover, attempting to tackle both brightness deficiency and other degradation factors solely with a single convolutional layer is exceedingly challenging~\cite{zhang2023learning}. To tackle above issues, we adopted the most common solution in the field, which involves introducing additional degradation-related post-processing to enhance visual quality~\cite{guo2017lime,li2021learning,zhang2023learning}. Here, we utilized three representative post-processing methods: SCUNet~\cite{zhang2023practical} for denoising, FBCNN~\cite{jiang2021towards} for de-artifacts~\cite{lin2020learning}, and MISC~\cite{liu2024motion} for deblurring. Figs.~\ref{fig: 3scenes} shows the results after applying fair post-processing to all methods. It can be observed that our method preserves more natural colors and achieves more accurate brightness control compared to other approaches, further validating its superiority.}
	
	\subsubsection{User Study}
	{Here, to evaluate the performance of the proposed method from a more comprehensive perspective, we first provide visual comparison results of the proposed AR-LLIE against three reference-free methods known for their efficiency on a large number of low-light image samples from DPED~\cite{ignatov2017dslr} (various lighting conditions and different weather conditions), VisDrone~\cite{du2019visdrone} (various weather conditions and uneven lighting), NPE~\cite{liu2021benchmarking} (uneven lighting and the impact of weather conditions), ExDark~\cite{loh2019getting}, LSRW~\cite{hai2023r2rnet}, and DARKFACE. As shown in Fig.~\ref{fig:in the wild}, it can be easily observed that the results from DCE generally suffer from low contrast and inadequate color priors, while the enhancement results from RUAS and SCI are prone to underexposure or overexposure. In contrast, the proposed method exhibits more satisfactory visual effects in exposure control and color recovery.}
	
	Subsequently, we conducted an independent user study to provide more realistic evaluation feedback. Specifically, in the survey interface we created, the enhancement results of DCE, RUAS, SCI, and AR-LLIE in four different types of real low-light scenes were displayed. We then invited 55 participants and asked them to rank the four methods based on four evaluation criteria: brightness, detail, color, and naturalness (with method names hidden). Finally, the corresponding final scores were obtained by calculating the number of votes for the top two choices. To more intuitively present the results of the user study, we quantified the survey results in a pie chart shown in Fig.~\ref{fig:user study}. It is evident that our method has significant advantages in all four evaluation criteria as well as overall performance, demonstrating more comprehensive capabilities and superior enhancement effects compared to other reference-free low-light image enhancement methods.

	\subsection{Computational Efficiency}
	As previously mentioned, a significant drawback of existing techniques is their purported high computational efficiency, which has only been validated on PC devices, overlooking the performance of the proposed algorithms on mobile devices. This oversight hinders a thorough validation of their practicality. To address the aforementioned issue, this section presents a comprehensive performance evaluation of computational efficiency, not only on PC devices but also on mobile devices, to validate the practicality of our method from a more holistic perspective. Specifically, we first compared AR-LLIE with the methods used in the enhancement performance evaluation section on PC devices. Then, we selected three enhancers known for their efficiency and decent performance on PC devices (\textit{i.e.,} DCE, RUAS, and SCI) and reported their computational efficiency on mobile terminals in comparison with our method. It should be noted that for each method, we calculated the inference time across different devices and input resolutions by randomly sampling 100 images and computing the average time to mitigate the impact of outliers.
	
	\begin{table*}[t]
		\footnotesize
		\centering
		\setlength{\tabcolsep}{6mm}
		\renewcommand\arraystretch{1.5}	
		\centering
		\caption{Comparing the quantitative results of using different re-parameterization mechanisms on the MIT dataset.}
		\begin{tabular}{|c||cccc|cccc|}
			\hline
			\multirow{2}{*}{Model} &\multicolumn{4}{c|}{$\sum_{s=1}^{S}\bm{\pi}_{s}$}& \multicolumn{4}{c|}{Metrics} \\
			\cline{2-9}
			~&ACB &ECB&DBB &AR-LLIE &PSNR$\uparrow$&SSIM$\uparrow$&DE$\uparrow$&NIQE$\downarrow$\\
			\hline
			A&\ding{55}&\ding{55}&\ding{55}&\ding{55} &15.8392&0.7055&6.4232&3.6405 \\
			B&\Checkmark&\ding{55}&\ding{55}&\ding{55} &16.0750&0.7174&6.4611&3.6403 \\
			C&\ding{55}&\Checkmark&\ding{55}&\ding{55} &16.9854&0.7312&6.5265&3.6191\\
			D&\ding{55}&\ding{55}&\Checkmark&\ding{55} &16.3962&0.7228&6.4844&3.6070\\
			E&\ding{55}&\ding{55}&\ding{55}&\Checkmark &\textbf{21.8376}&\textbf{0.7486}&\textbf{7.0297}&\textbf{3.6016}\\
			\hline
		\end{tabular}
		
		\label{table:ablation}
	\end{table*}
	\begin{figure*}[htb!]
		\centering
		\begin{tabular}{c@{\extracolsep{0.2em}}c@{\extracolsep{0.2em}}c@{\extracolsep{0.2em}}c@{\extracolsep{0.2em}}c@{\extracolsep{0.2em}}c}
			\includegraphics[width=0.16\linewidth]{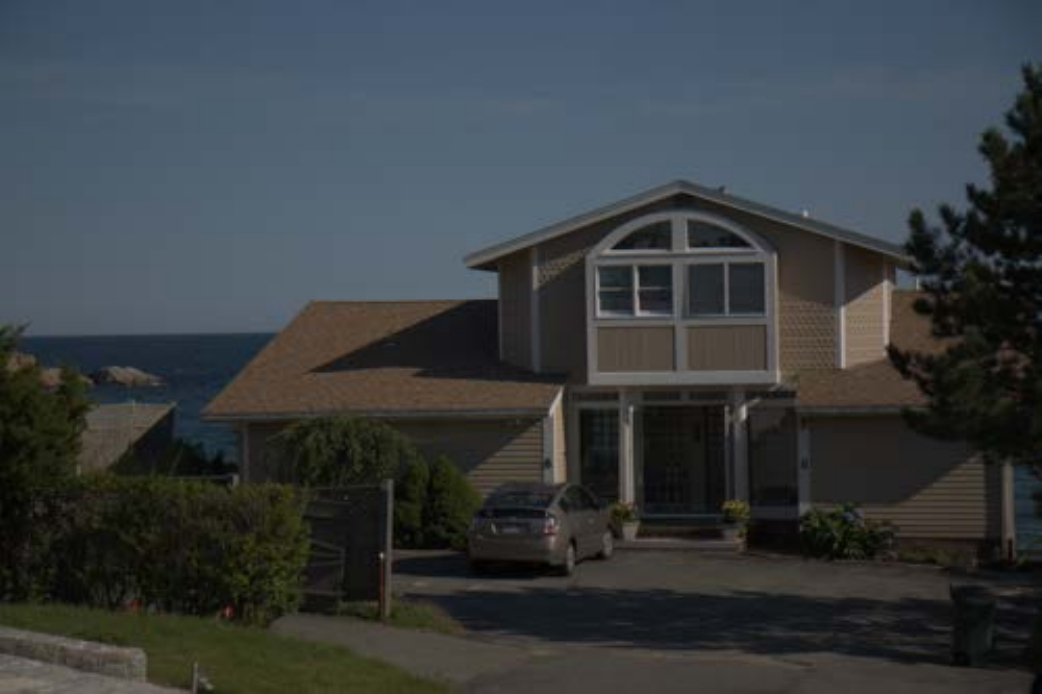}&
			\includegraphics[width=0.16\linewidth]{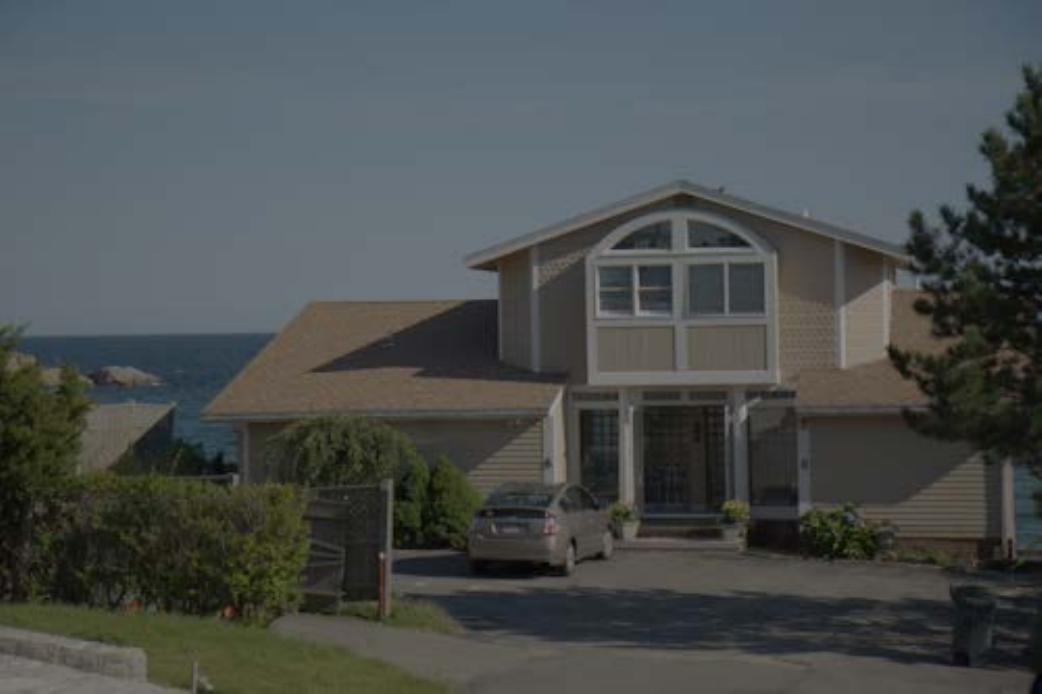}&
			\includegraphics[width=0.16\linewidth]{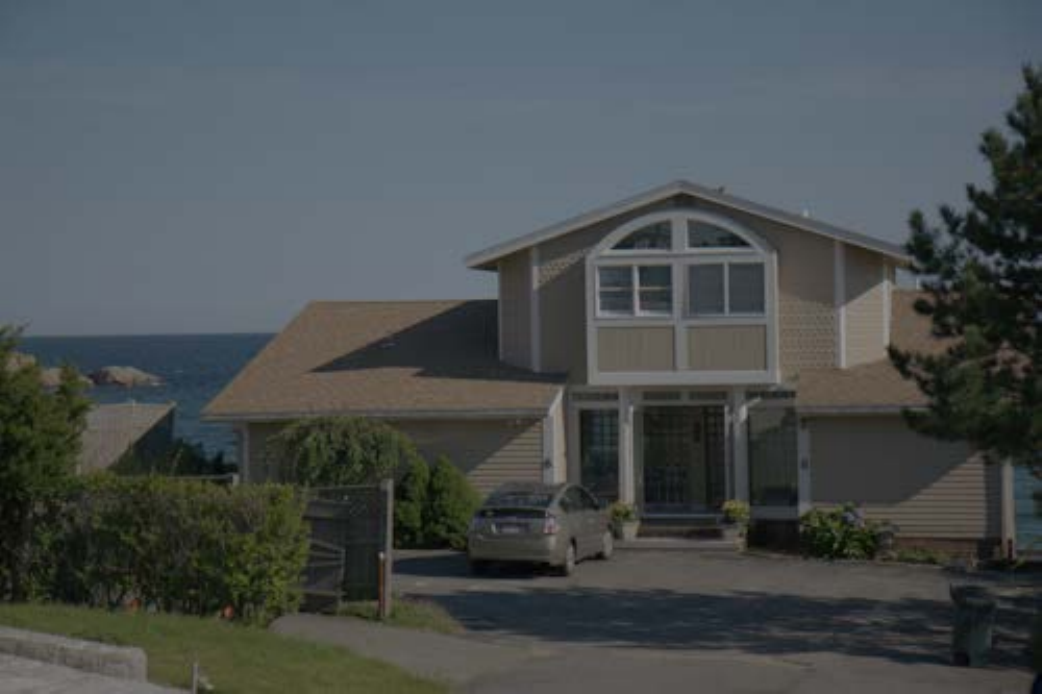}&
			\includegraphics[width=0.16\linewidth]{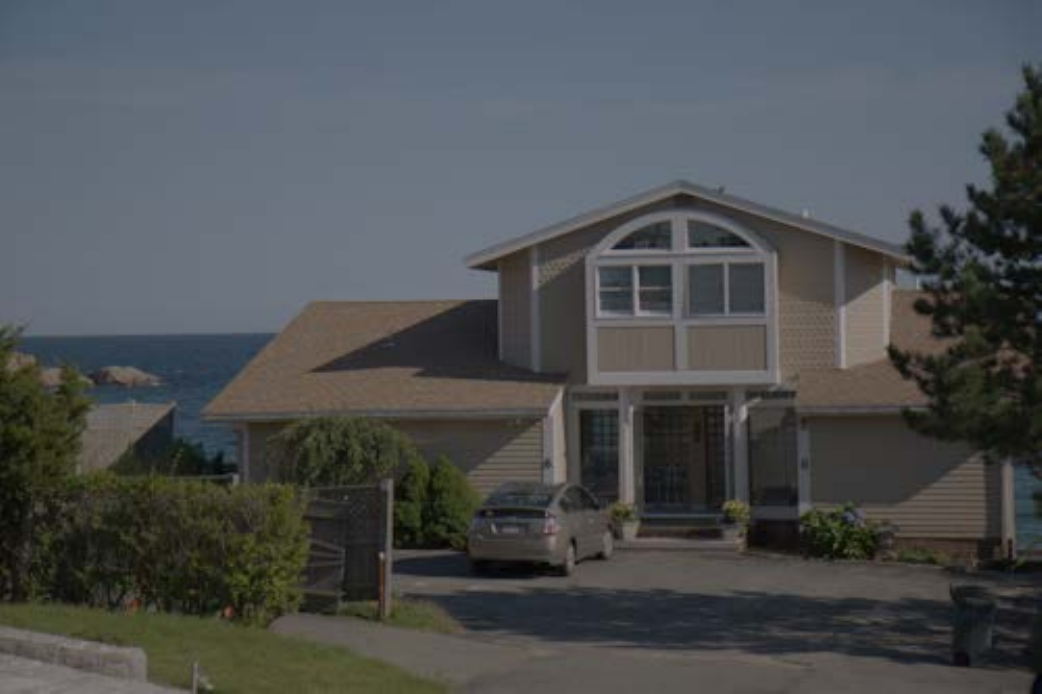}&
			\includegraphics[width=0.16\linewidth]{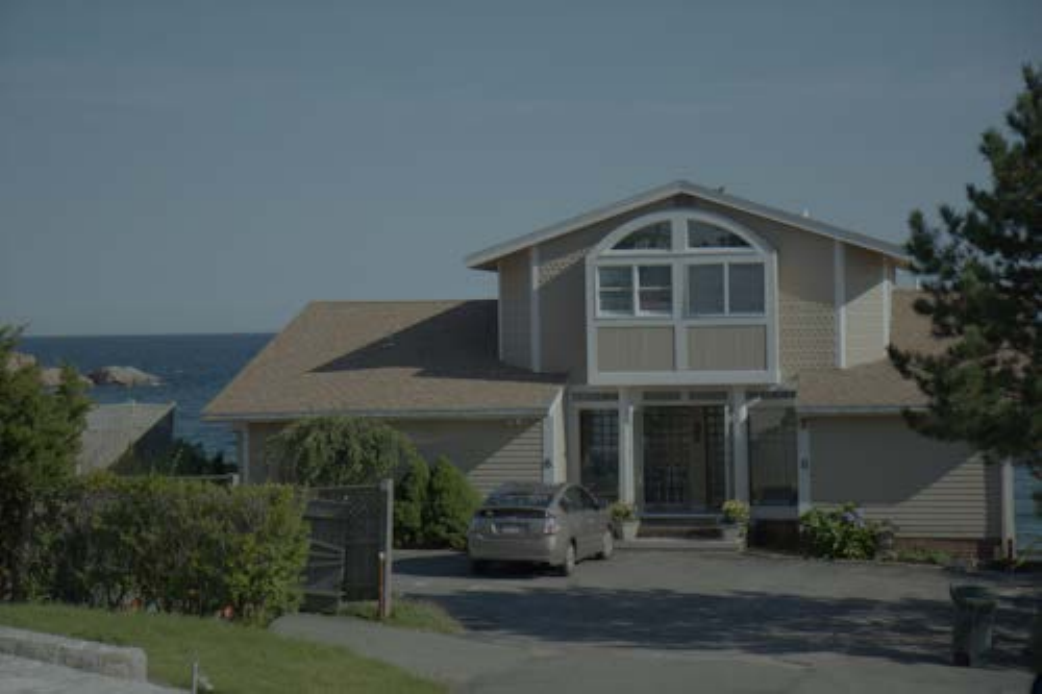}&
			\includegraphics[width=0.16\linewidth]{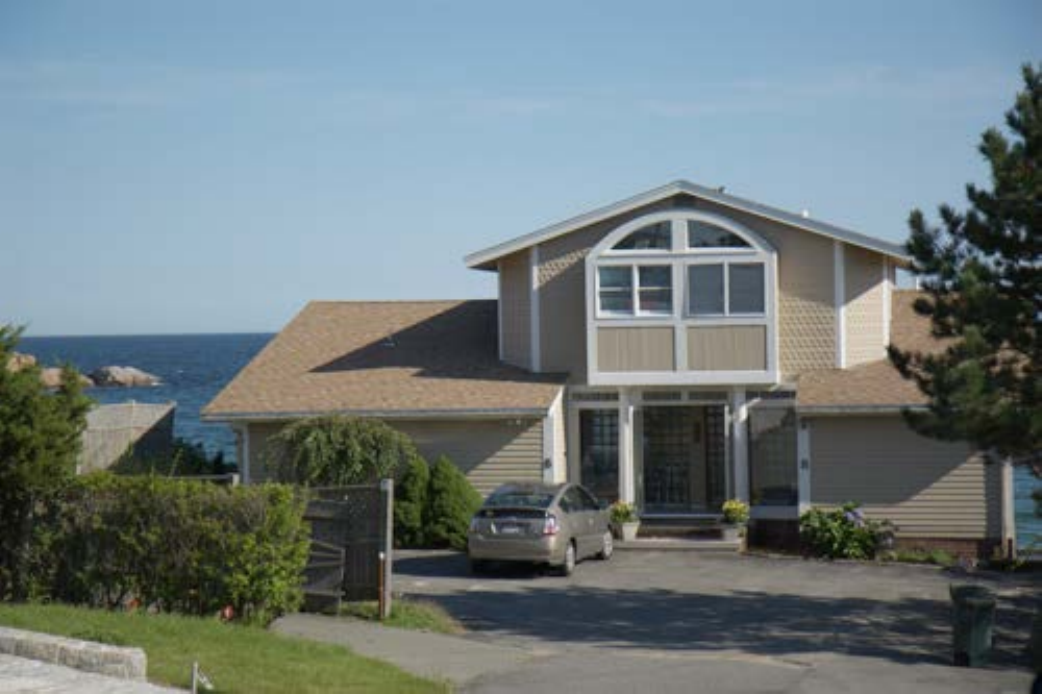}\\
			\includegraphics[width=0.16\linewidth]{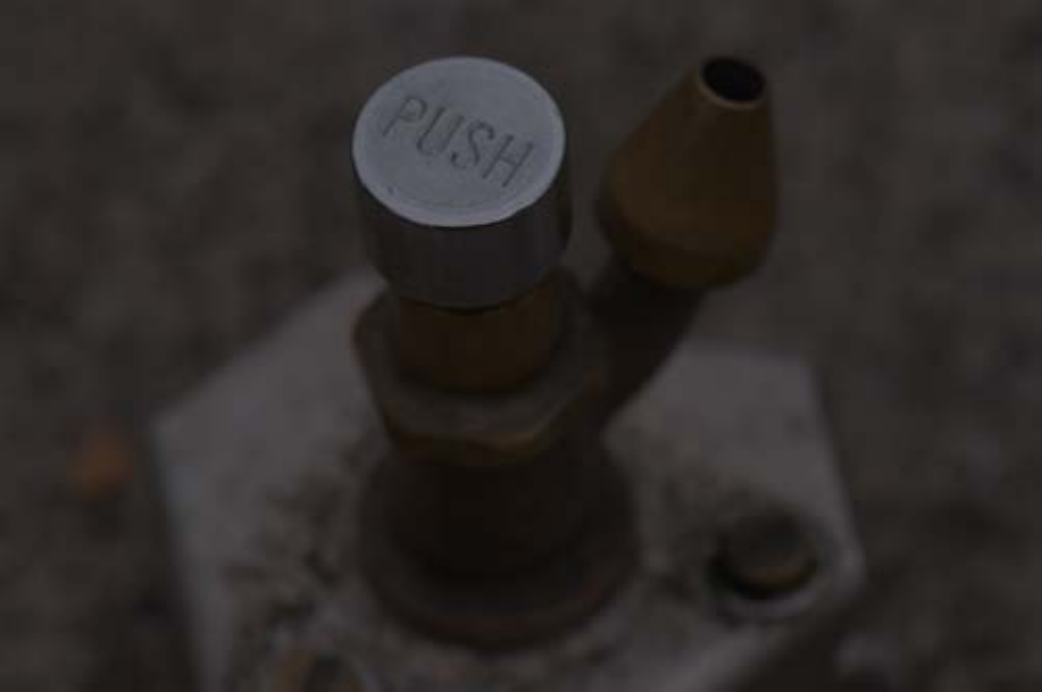}&
			\includegraphics[width=0.16\linewidth]{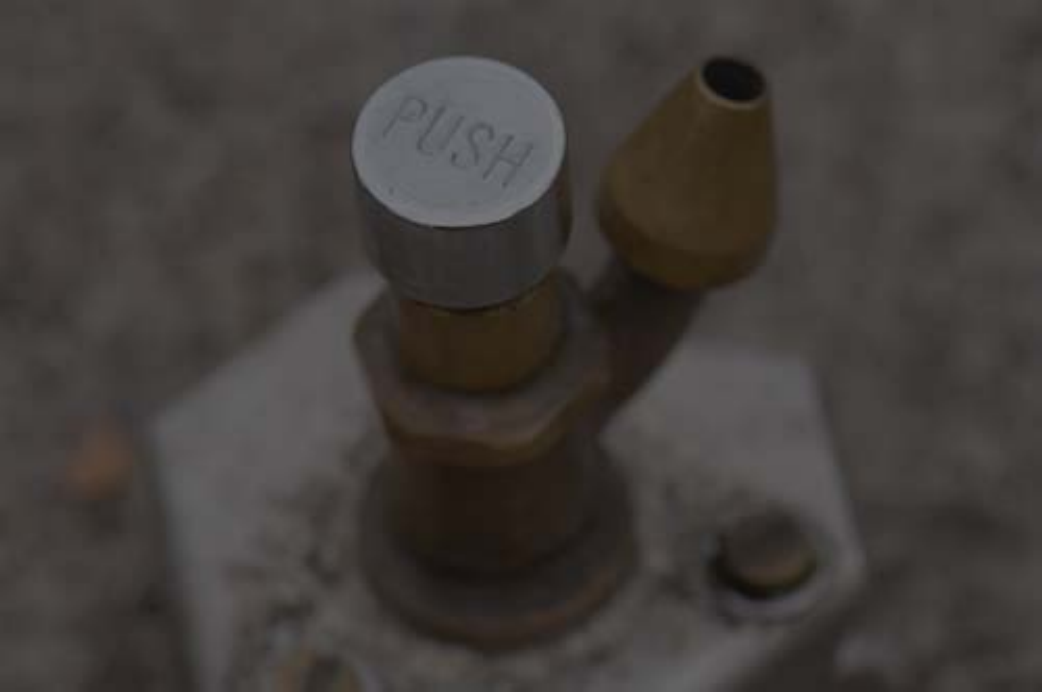}&
			\includegraphics[width=0.16\linewidth]{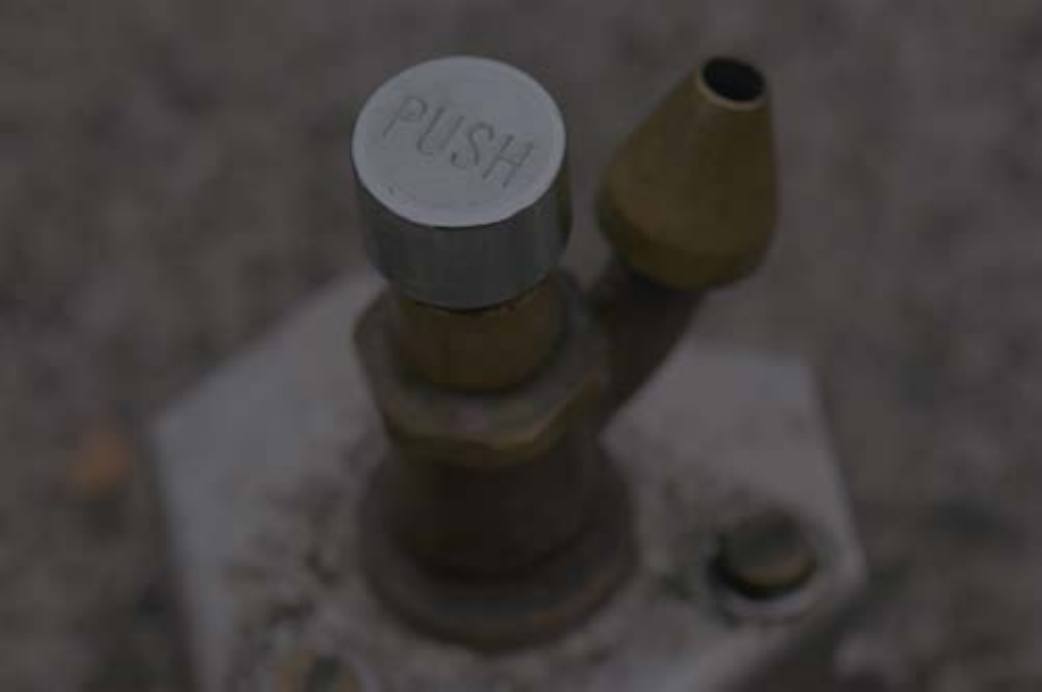}&
			\includegraphics[width=0.16\linewidth]{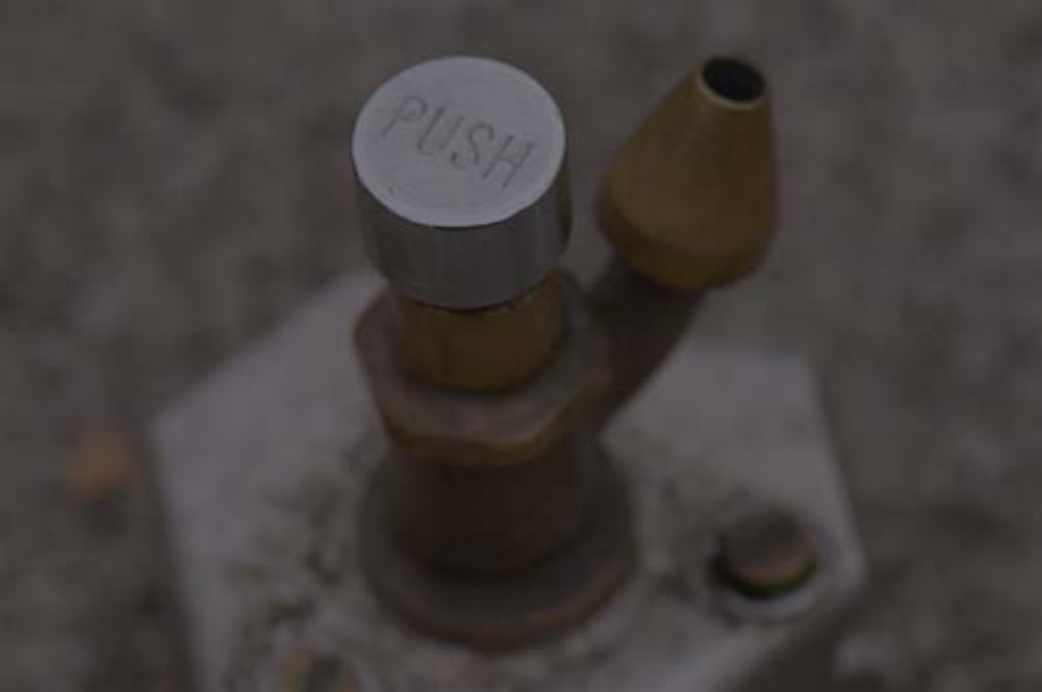}&
			\includegraphics[width=0.16\linewidth]{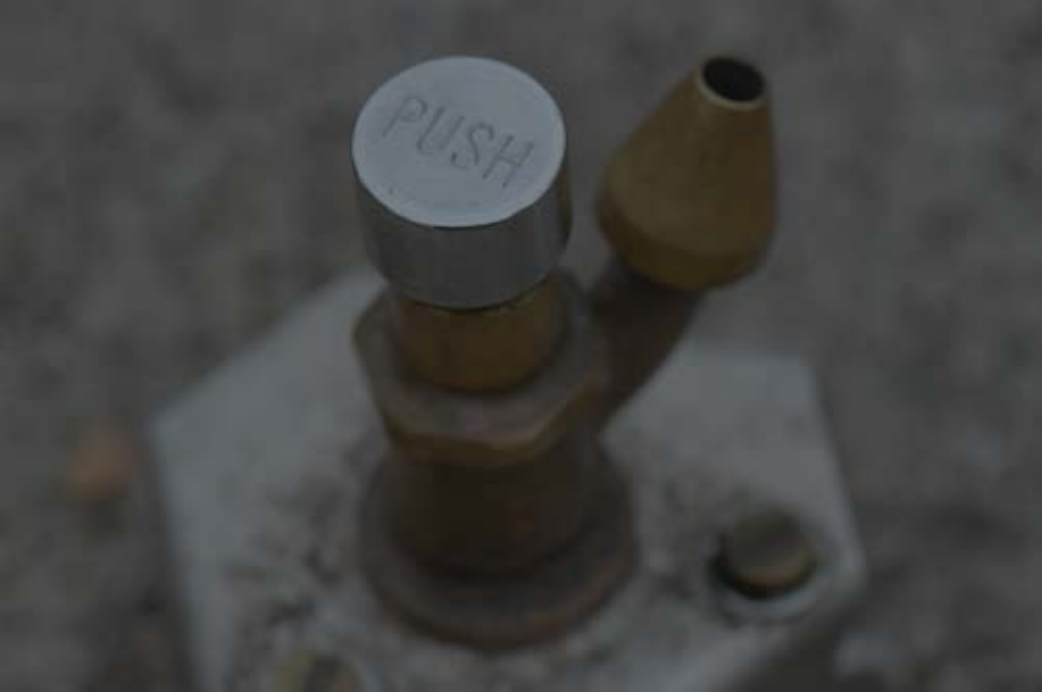}&
			\includegraphics[width=0.16\linewidth]{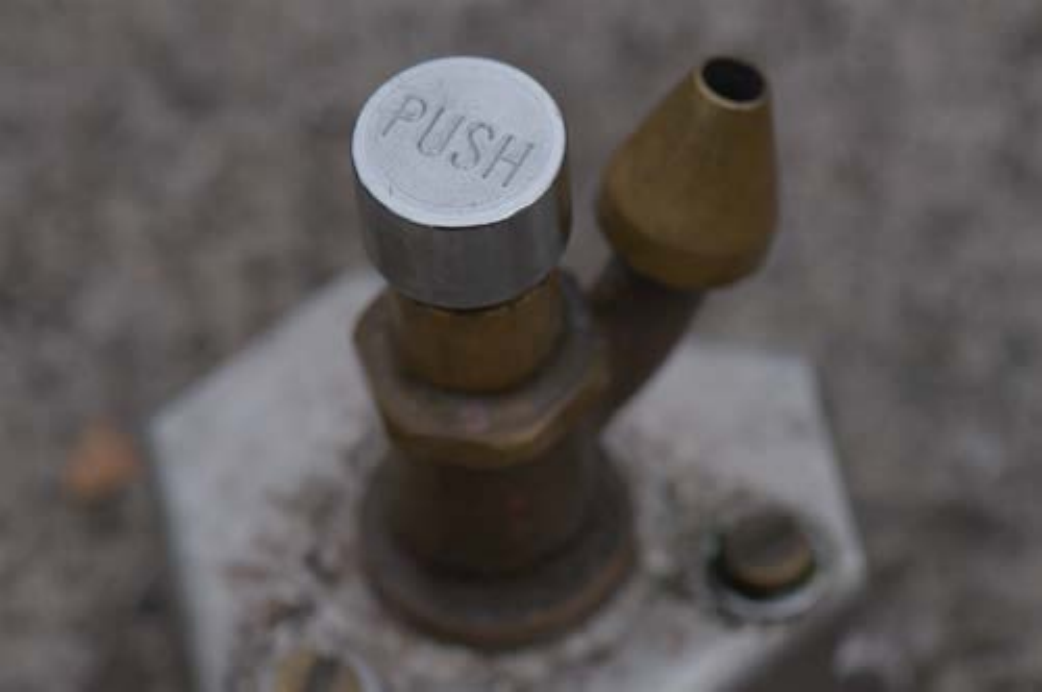}\\
			\footnotesize Input&\footnotesize w/o REP &\footnotesize w/ REP (ACB) &\footnotesize w/ REP (DBB)&\footnotesize w/ REP (ECB) &\footnotesize w/ REP (Ours)\\
		\end{tabular}
		\caption{Qualitative comparison of different re-parameterization methods.}
		\label{fig:rep}
	\end{figure*}
	
	{\subsubsection{CPU/GPU (PC devices)} We firstly conducted a performance evaluation of the inference time for recently proposed methods on a personal computer equipped with a GeForce RTX 2080 Ti GPU and an Intel(R) Xeon(R) W-2135 Processor. Specifically, to provide a more comprehensive efficiency assessment, we considered three common image resolution settings for input: 1280×720 (720P), 1920×1080 (1080P), and 2650×1440 (2K). We calculated the running time for these inputs on both the CPU and GPU. The numerical results are shown in Table~\ref{table:efficiency}, where the red numbers in the lower right corner of the cells corresponding to our AR-LLIE method indicate the percentage improvement over the second-ranking method. It is evident that our method consistently achieves the fastest inference time on both CPU and GPU, with improvements exceeding 30\% compared to the second-best method (\textit{i.e.,} SCI). Moreover, the performance gains are even more pronounced at higher resolutions, with an improvement of over 90\% for 2K resolution inputs on the GPU platform. Additionally, it is noteworthy that on the GPU, as the input image resolution increases, our method's inference time remains in the order of $1 \times 10^{-5}$ seconds, with no significant fluctuations, demonstrating both high efficiency and stability. We have also provided the numerical comparison of our proposed method with other efficiency-oriented methods in terms of FLOPs and model size in Table~\ref{table: Modelsize}. It is evident that our method shows significant advantages in both metrics, highlighting its lightweight nature and high efficiency.}

	
	\subsubsection{DSP/NPU (Mobile devices)} Although we have demonstrated the computational efficiency of the proposed method on PC devices, in practical applications, low-light enhancers are typically expected to be deployed on mobile devices. Unlike PC devices, mobile devices (\textit{e.g.,} smartphones) have significantly limited hardware resources (\textit{e.g.,} lower bandwidth and computing power). Therefore, it is crucial to evaluate the performance of enhancers under these constrained conditions. Based on the above considerations, we further assessed the inference efficiency on two typical kinds of mobile devices equipped with DSP or NPU chips. 
	For DSP evaluation, we used a Xiaomi 10 smartphone featuring the representative Snapdragon 865 DSP chip. Regarding NPU, we considered the recently launched Huawei Mate30 Pro with the Kirin 990 chip and the Xiaomi 14 with the Snapdragon 8 Gen 3 chip. Similar to the performance evaluation conducted on PC devices, we also provide results for three image resolution inputs here. Specifically, we selected three representative enhancers known for their efficiency from Table~\ref{table:efficiency} for further comparison. We used the AI Benchmark 5.0.1 version as the evaluation software. All the aforementioned mobile devices ran the software with identical settings. We quantized the inference model's parameters to the F16 algorithm and chose the TFLite GPU Delegate to achieve inference acceleration. The final results are presented in Table~\ref{table:DSP/NPU time}. As shown, our AR-LLIE method maintains a significant advantage in computational efficiency over these lightweight methods on mobile devices. Compared to the second-best method, our method's runtime is at least 40\% faster. Consistent with the evaluation results on PC devices, the advantage of our method becomes more pronounced with higher resolution inputs.
	

\begin{figure*}[t]
	\centering
	\footnotesize
	\begin{tabular}{c}
		\includegraphics[width=0.98\linewidth]{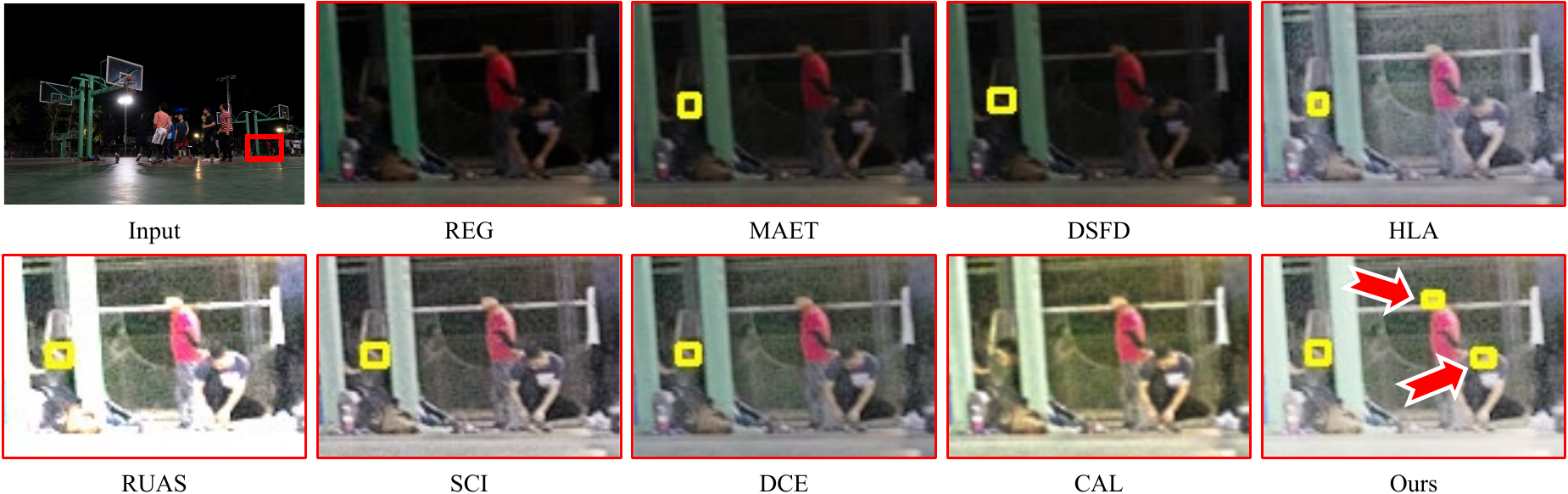}\\
	\end{tabular}
	\caption{{Visual comparison of detection results on the DARKFACE dataset.}}
	\label{fig: Det}
\end{figure*}

\begin{figure}[h]
	\centering
	\begin{tabular}{c}
		\includegraphics[width=0.95\linewidth]{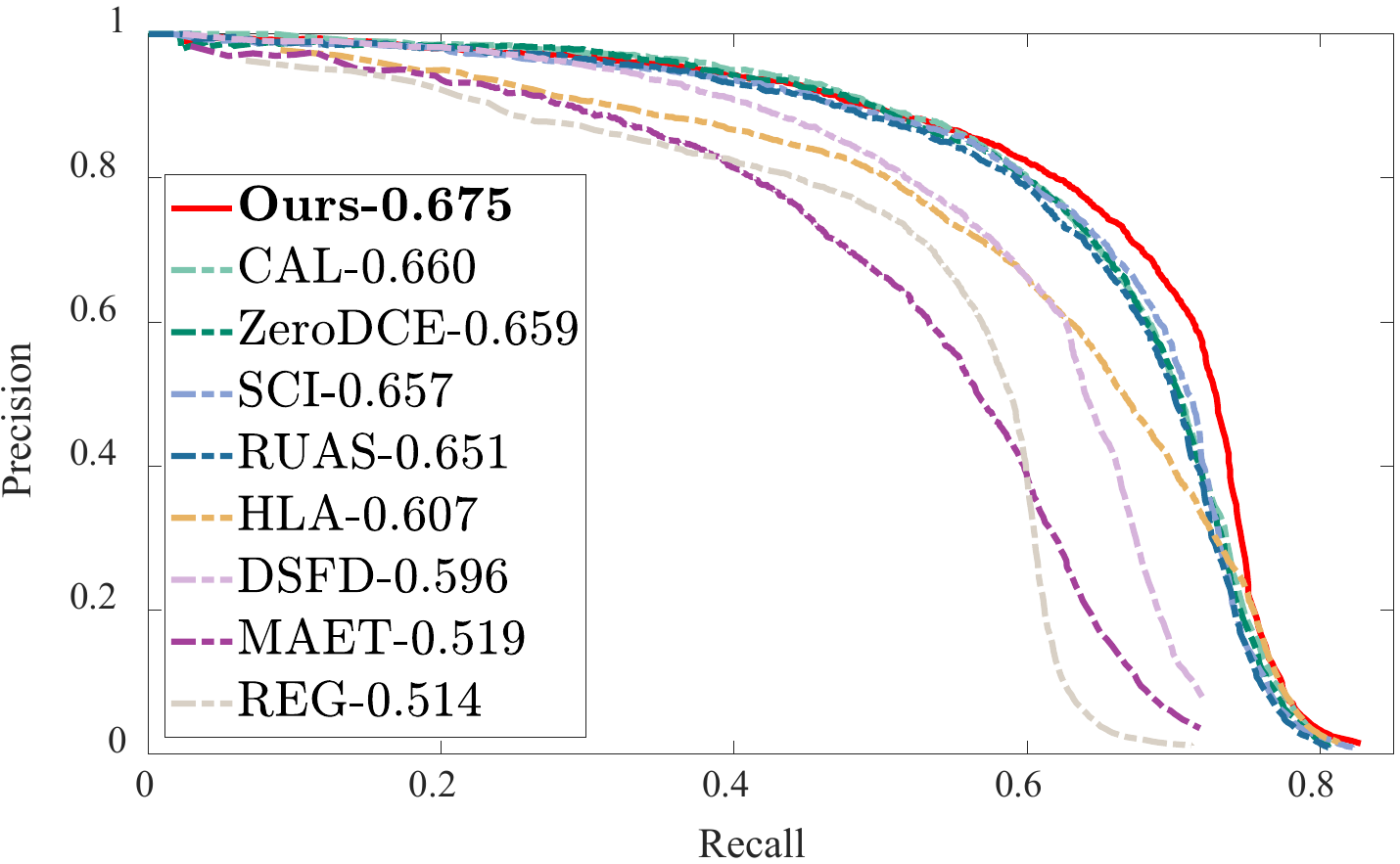}\\
	\end{tabular}
	\caption{{Precision-Recall curve on the DARKFACE dataset.}}
    \vspace{-3mm}
	\label{fig:pr}
\end{figure}

	\subsection{Application}
	Here, we use the representative face detector DSFD~\cite{li2019dsfd} as the baseline and adopt the same cascaded framework with a fine-tuning training strategy as SCI to evaluate the performance of our method in supporting downstream tasks. The related quantitative results are shown in the Precision-Recall curves in Fig.~\ref{fig:pr}. In addition to several representative unsupervised enhancers, we also introduced four detectors specifically designed for low-light scenarios (\textit{i.e.,} HLA~\cite{wang2021hla}, MAET~\cite{cui2021multitask}, REG~\cite{liang2021recurrent}, and CAL~\cite{xue2022best}) as comparative methods to provide a more comprehensive evaluation of detection performance. It is evident that our method achieved the highest detection accuracy compared to all the comparative methods, with a performance improvement of over 13.2\% compared to the baseline. This validates the potential of our method in supporting downstream tasks. We provide the relevant qualitative results in Fig.~\ref{fig: Det}, where it can be observed that our method detects more and smaller faces compared to other methods, without any false-positive or false-negative conditions, offering more suitable visual effects.

	 \begin{table}[h]
		\centering
		\footnotesize
		\renewcommand{\arraystretch}{1.5}
		\setlength{\tabcolsep}{7pt}
		\caption{{Ablation experiment results with different loss function settings on the MIT dataset.}}
		\begin{tabular}{|c|cc|ccc|}
			\hline
			Model &$\mathcal{L}_{fidelity}$ &$\mathcal{L}_{smooth}$ &PSMR$\uparrow$ &SSIM$\uparrow$ &NIQE$\downarrow$\\ \hline
			A &\checkmark&\ding{55}&18.0243&0.7488&3.5786\\ 
			B &\ding{55}&\checkmark&13.9047&0.6205&3.7946\\ 
			\hline
			Ours &\checkmark&\checkmark&24.0157&0.8571&3.5183\\ \hline
		\end{tabular}
		\vspace{-3mm}
		\label{table: abla}
	\end{table}
	\begin{figure}[htb!]
		\centering
		\footnotesize
		\begin{tabular}{c@{\extracolsep{0.2em}}c@{\extracolsep{0.2em}}c@{\extracolsep{0.2em}}c}
			\includegraphics[width=0.23\linewidth]{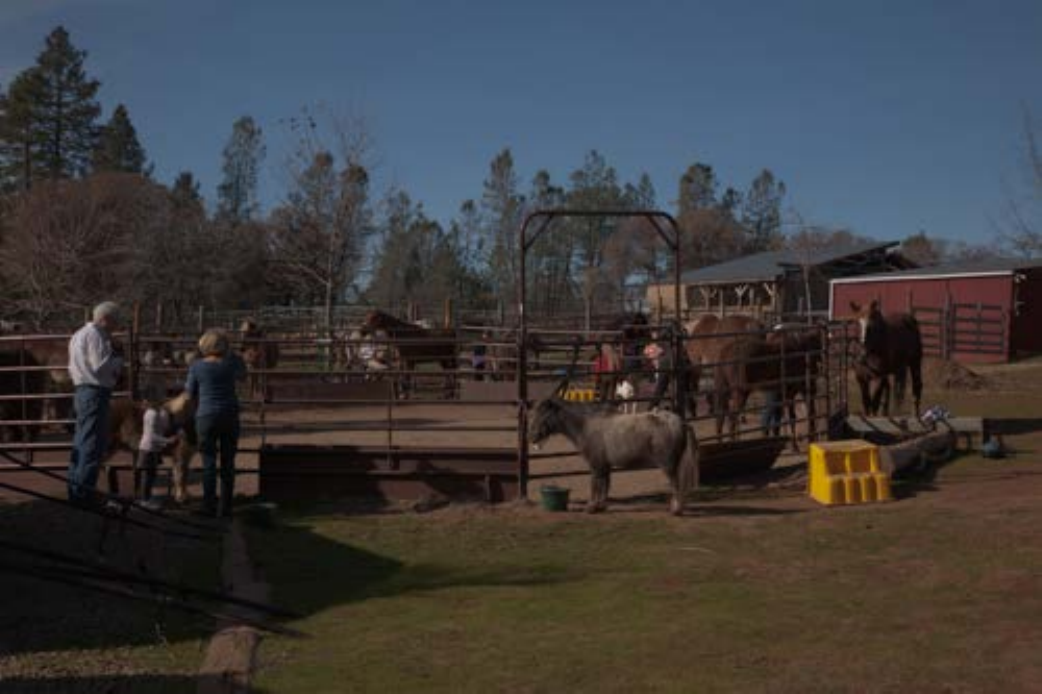}&
			\includegraphics[width=0.23\linewidth]{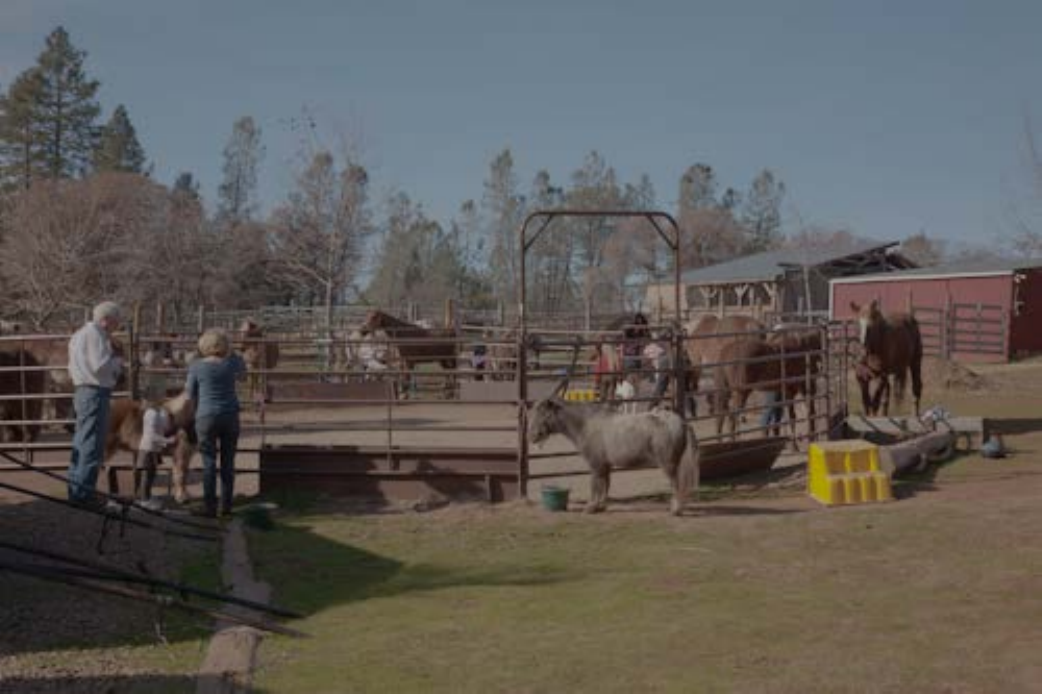}&
			\includegraphics[width=0.23\linewidth]{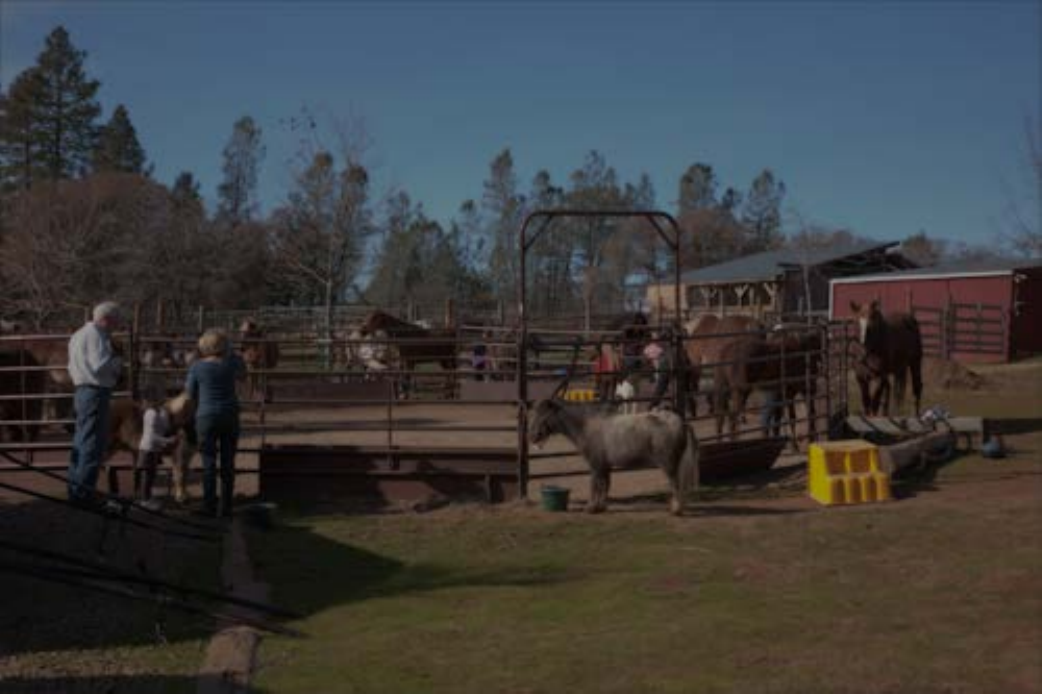}&
			\includegraphics[width=0.23\linewidth]{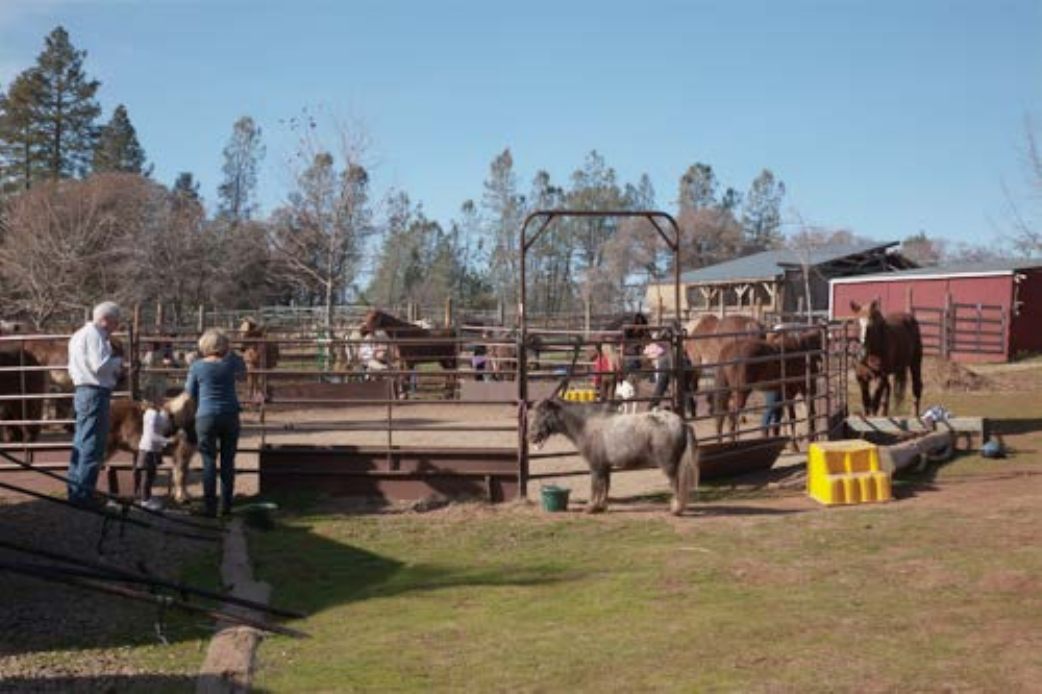}\\
			Input&\footnotesize w/o $\mathcal{L}_{smooth}$ &w/o $\mathcal{L}_{fidelity}$ &Ours\\
		\end{tabular}
		\caption{{The impact of different loss function settings on visual outcomes.}}
		\label{fig: abla}
	\end{figure}
	{\subsection{Algorithmic Analyses}
	\label{sec:ablation}
	In this section, we conduct a series of analytical experiments from multiple perspectives to validate the effectiveness of the automatic re-parameterization mechanism in our AR-LLIE. First, we consider removing the proposed automatic re-parameterization mechanism or replacing it with three commonly used re-parameterization modules (\textit{i.e.,} ACB~\cite{Ding_2019_ICCV}, DBB~\cite{ding2021diverse} and ECB~\cite{zhang2021edge}) to validate the effectiveness of our method. The relevant results are shown in Table~\ref{table:ablation} and Fig.~\ref{fig:rep}, where it can be observed that our method significantly outperforms existing re-parameterization methods. This advantage arises primarily from two key factors. First, ACB, DBB, and ECB were initially designed for tasks such as image classification, semantic segmentation, and super-resolution, respectively, which have limited relevance to image enhancement tasks. Second, the multi-branch structures employed by these methods are pre-defined, whereas our approach can flexibly and automatically construct optimal re-parameterization structures based on task requirements, fully leveraging the expanded parameter space enabled by re-parameterization.}

	\begin{figure}[htb!]
		\centering
		\begin{tabular}{c}
			\includegraphics[width=0.9\linewidth]{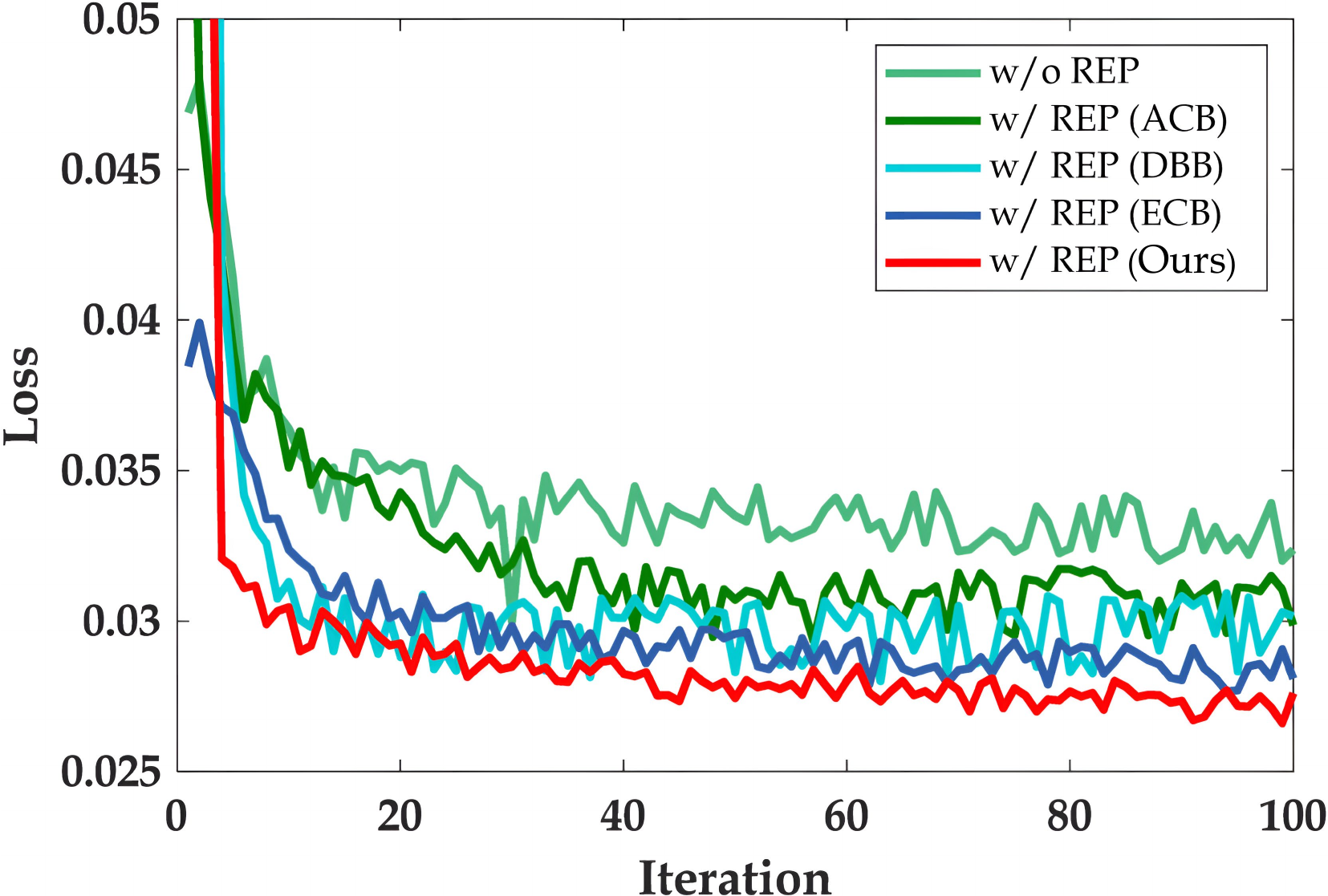}\\
		\end{tabular}
		\caption{{Convergence behaviors with different re-parameterization (REP) methods. The red line represents our AR-LLIE.}}
		\vspace{-3mm}
		\label{fig:loss}
	\end{figure}
	
	{Subsequently, we analyzed the impact of different re-parameterization mechanisms on the convergence of the algorithm. From the loss convergence curves shown in Fig.~\ref{fig:loss}, it can be seen that the case without re-parameterization has the slowest loss convergence speed. The introduction of any re-parameterization mechanism significantly improves the convergence behavior. Moreover, compared to other methods, our AR-LLIE achieves the fastest convergence speed by automatically discovering the most suitable re-parameterization structure, which avoids the potential issue of the model getting trapped in local optima when using simple structures, demonstrating the effectiveness of our approach. We have also provided ablation results on the loss functions in Table~\ref{table: abla} and Fig.~\ref{fig: abla}. It can be observed that removing any single loss function significantly impacts performance. Notably, when $\mathcal{L}_{fidelity}$ is removed, brightness control fails entirely. This demonstrates the effectiveness of the loss functions.}
	
	\begin{figure}[h]
		\centering
		\begin{tabular}{c}
			\includegraphics[width=0.95\linewidth]{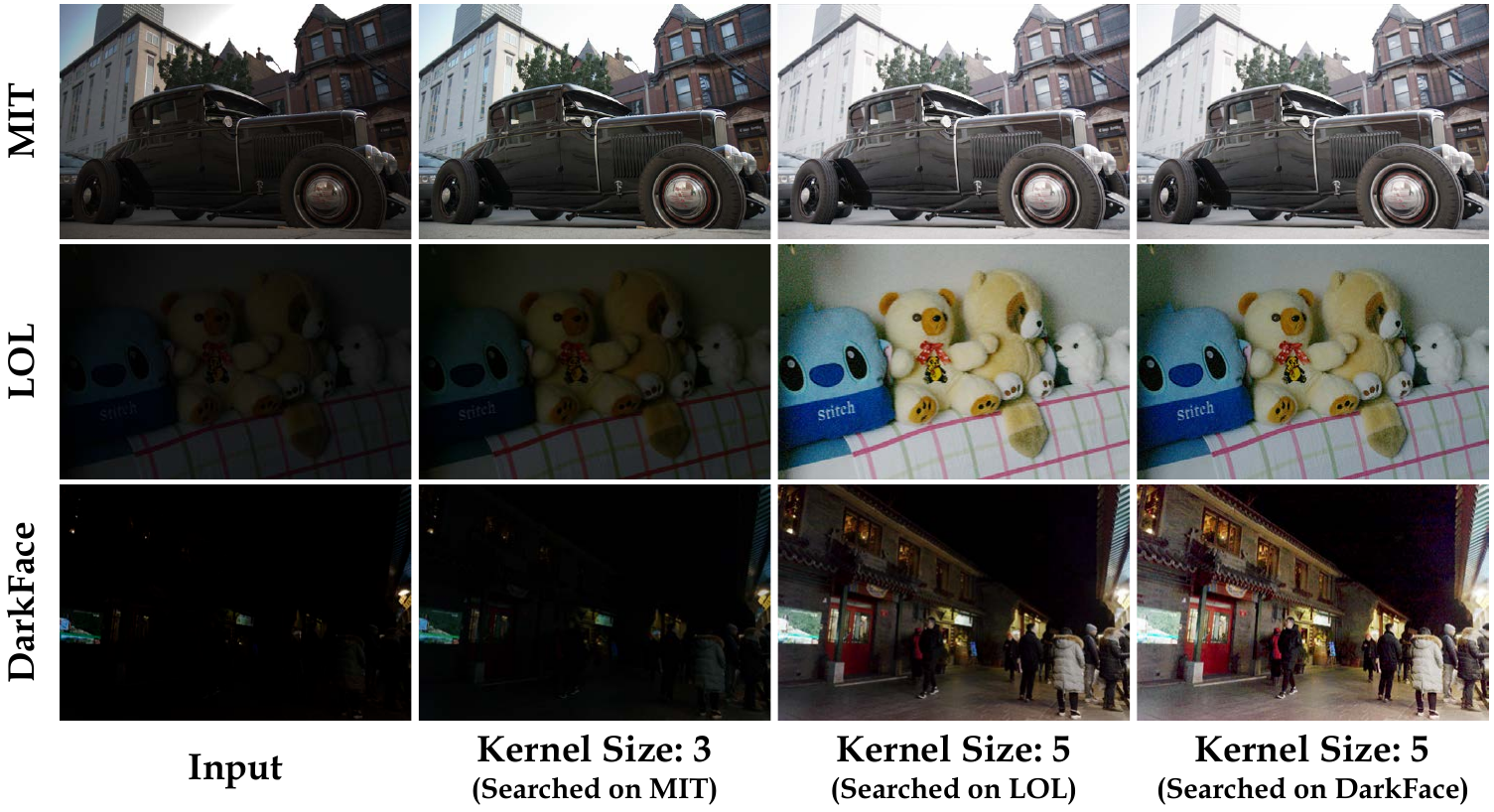}\\
		\end{tabular}
		\caption{{Visual results of structures with different convolution kernel sizes on various difficulty levels of low-light datasets.}}
		\vspace{-5mm}
		\label{fig:discuss}
	\end{figure}
	
	{\subsection{Effects of Different Kernel Size}~\label{sec:dis}
	Here we analysis the characteristics of structures obtained through the tiered NAS on different difficulty levels of low-light datasets. We consider three datasets corresponding including MIT~\cite{bychkovsky2011learning}, LOL~\cite{wei2018deep}, and DARKFACE~\cite{yang2020advancing}. Interestingly, when we re-parameterize the three obtained structures into single-layer convolution, the kernel size changes with the difficulty of the adopt data during the search phase.}
	
	{We illustrate examples of models trained on the three datasets with different structures in Fig.~\ref{fig:discuss} to provide a more intuitive analysis. From the first row, we observe that all three structures achieve satisfactory results on the MIT dataset. In the subsequent rows, we notice that the structures obtained from the search on the MIT dataset perform poorly on the more challenging datasets. In contrast, the structures with bigger kernel size exhibit significant enhancement effects on both the LOL and DARKFACE datasets. Moreover, the enhancement effect is best when the data used in the search phase is from the same dataset as the training and testing phases. Based on these results, we can infer that the limits of the enhancer vary on low-light datasets with different difficulty levels. Considering the characteristics of the re-parameterized structures, networks with larger convolution kernel sizes have the ability to handle more challenging scenarios. However, for the subsequent experiments, aiming for a balance between performance and efficiency, we use the structures obtained from the corresponding data searches on all three datasets, rather than adopting larger kernels.}
	
	\begin{figure}[!htb]
		\centering
		\footnotesize
		\begin{tabular}{c}
			\includegraphics[width=0.98\linewidth]{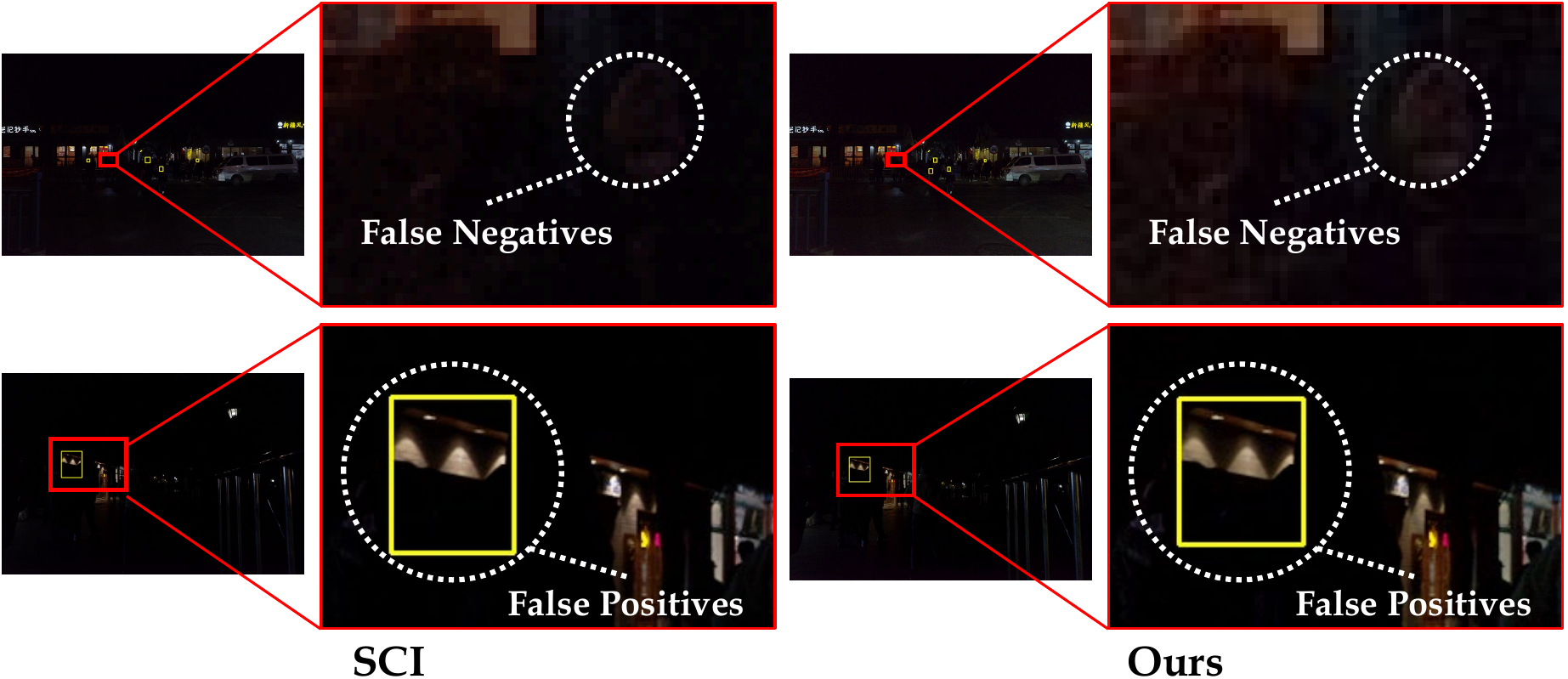}\\
		\end{tabular}
		\caption{{Limitations of assisting object detection tasks in extremely challenging low-light scenarios.}}
		\vspace{-3mm}
		\label{fig:lim}
	\end{figure}

	{\subsection{Limitations}	
	Although we have made extensive efforts through a series of experiments and analyses to demonstrate the effectiveness of the proposed AR-LLIE in various low-light scenarios, real-world low-light conditions often present even more challenging illumination degradations. In fact, when exploring the ability of the proposed method to assist downstream tasks, unexpected missed and false detections occurred in some extreme cases. Considering the importance of algorithm performance in such challenging environments, we provide a set of extreme dark condition detection visualization results to analyze the potential limitations of our method. Fig.~\ref{fig:lim} shows the results of the representative unsupervised low-light enhancer SCI and ours on extreme examples in the DARKFACE dataset. It can be observed that both methods suffer from missed and false detections. This may be due to the difficulty of low-light enhancement assistance in bridging the interference caused by background noise in the image, which makes it challenging for detection networks to accurately locate and classify targets. A potential solution is to design additional constraints at the feature detection layer, providing the network with stronger noise resistance, thereby improving detection accuracy.}
	
	
	\section{Conclusion}\label{sec:con}
	In this work, we aim to explore the limit for enhancers that can balance good visual quality with high computational efficiency. We first re-examine illumination-based enhancers, constructing a novel perspective to understand the task requirements and the connections between network design and model learning. We developed a novel model learning strategy by introducing re-parameterization techniques to expand the model's parameter space, thereby avoiding the issue of overly simple structures falling into local optima. Subsequently, a hierarchical structure modeling perspective is established for the re-parameterized structure, and a tiered search scheme is designed to simultaneously achieve high visual quality and efficiency. We conducted extensive evaluations in various low-light scenarios and on different terminal devices to validate our outstanding performance.
	

\bibliographystyle{IEEEtran}
\bibliography{AR-LLIE.bib}

\begin{IEEEbiography}[{\includegraphics[width=1in,height=1.25in,clip,keepaspectratio]{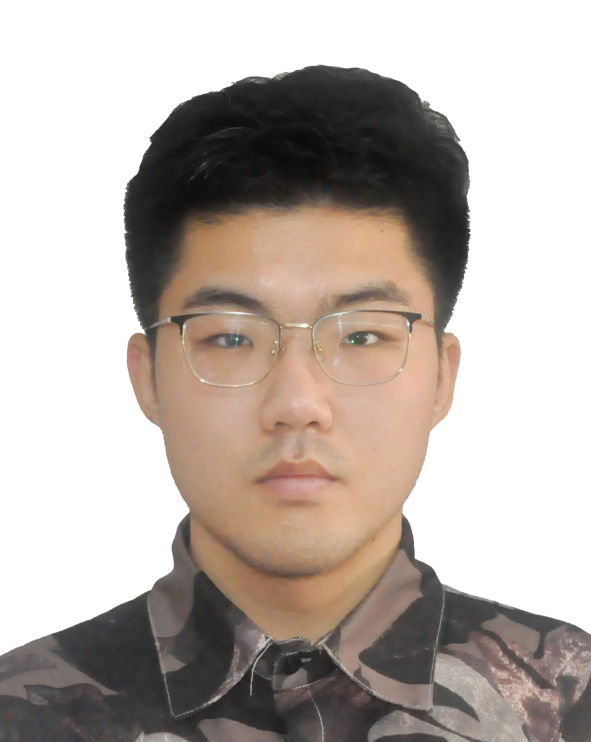}}]{Nan An}
	received the B.S. degree in computer science and technology from Liaoning University, China, in 2021. He is currently pursuing the Ph.D. degree in software engineering at Dalian University of Technology, China. His research interests include computer vision and deep learning.
\end{IEEEbiography}

\begin{IEEEbiography}[{\includegraphics[width=1in,height=1.25in,clip,keepaspectratio]{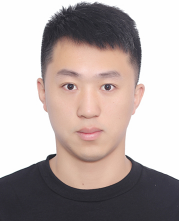}}]{Long Ma}
	received the Ph.D. degree in software engineering at Dalian University of Technology, Dalian, China, in 2023. He is currently a Postdoc the School of Software Technology, Dalian University of Technology (DUT), Dalian, China. His research interests include computer vision and deep learning.
\end{IEEEbiography}

\begin{IEEEbiography}[{\includegraphics[width=1in,height=1.25in,clip,keepaspectratio]{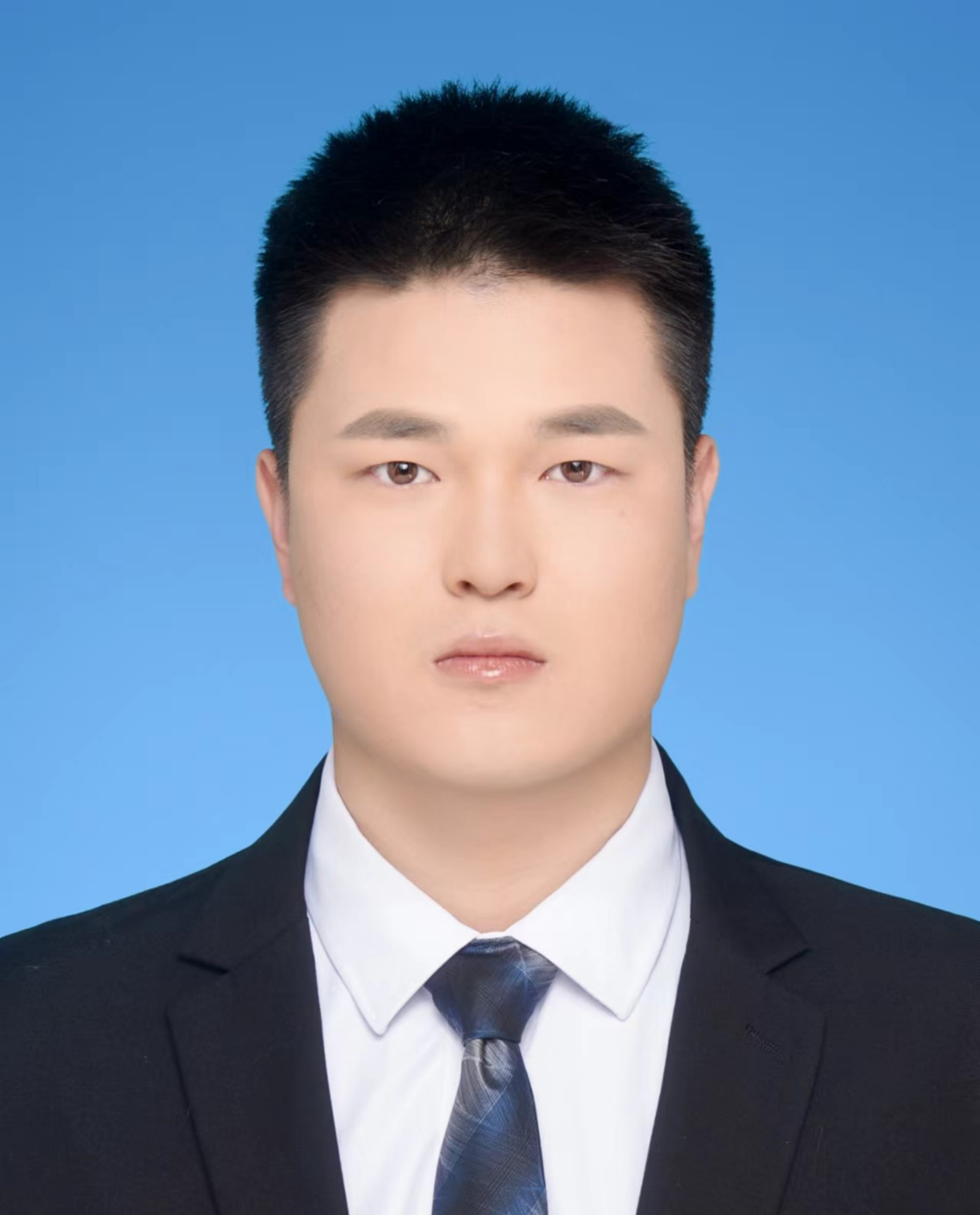}}]{Guangchao Han}
	received the B.S. in software engineering from Taiyuan University of Technology, China, in 2024. He is currently pursuing the M.S. degree in software engineering at Dalian University of Technology, China. His research interests include computer vision and deep learning.
\end{IEEEbiography}

\begin{IEEEbiography}[{\includegraphics[width=1in,height=1.25in,clip,keepaspectratio]{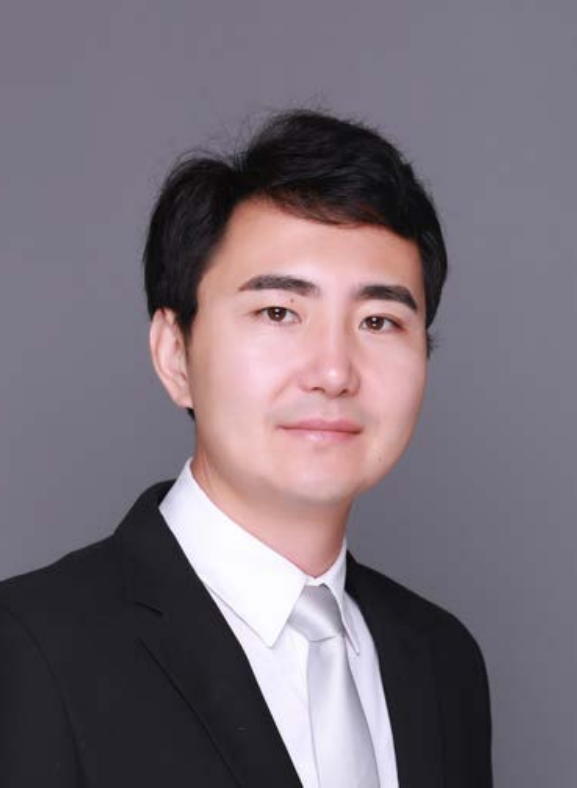}}]{Xin Fan}
	received the B.E. and Ph.D. degrees in information and communication engineering from Xian Jiaotong University, Xian, China, in 1998 and 2004, respectively. He was with Oklahoma State University, Stillwater, from 2006 to 2007, as a postdoctoral research fellow. He joined the School of Software, Dalian University of Technology, Dalian, China, in 2009. His current research interests include computational geometry and machine learning, and their applications to low-level image processing and DTI-MR image analysis.
\end{IEEEbiography}

\begin{IEEEbiography}[{\includegraphics[width=1in,height=1.25in,clip,keepaspectratio]{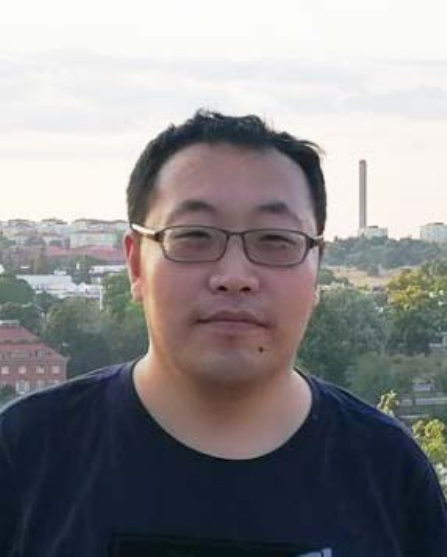}}]{Risheng Liu} 
	received his B.Sc. (2007) and Ph.D. (2012) from Dalian University of Technology, China. From 2010 to 2012, he was doing research as joint Ph.D. in robotics institute at Carnegie Mellon University. From 2016 to 2018, He was doing research as Hong Kong Scholar at the Hong Kong Polytechnic University. He is currently a full professor with the Digital Media Department at International School of Information Science \& Engineering, Dalian University of Technology (DUT). He was awarded the “Outstanding Youth Science Foundation” of the National Natural Science Foundation of China. He serves as associate editor for Pattern Recognition, the Journal of Electronic Imaging (Senior Editor), The Visual Computer, and IET Image Processing. His research interests include optimization, computer vision and multimedia.
\end{IEEEbiography}

\end{document}